%% file: 5331.tex
\let\mypdfximage\pdfximage
\def\pdfximage{\immediate\mypdfximage}

\documentclass[runningheads]{llncs}
\usepackage{graphicx}
\usepackage{comment}
\usepackage{amsmath,amssymb} 
\usepackage{color}
\usepackage{ dsfont }		
\usepackage[export]{adjustbox}
\usepackage{multicol}
\usepackage{rotating}		
 \usepackage{array}
\usepackage{subfig}
\usepackage{float}

\usepackage[table]{xcolor}   

\definecolor{tablegray}{rgb}{1.,0.85,0.85}

\usepackage{xspace}
\newcommand*{\eg}{e.g.\@\xspace}
\newcommand*{\ie}{i.e.\@\xspace}

\usepackage{tabu}
\usepackage{adjustbox}
\usepackage{graphicx}

\newcolumntype{x}[1]{>{\centering\arraybackslash\hspace{0pt}}p{#1}}
\newlength{\breite}
\newlength{\breiteimg}
\input{math_commands.tex}

\begin{document}
\pagestyle{headings}
\mainmatter
\def\ECCVSubNumber{5331}  
\title{Adversarial Robustness on In- and Out-Distribution Improves Explainability}

\titlerunning{Adversarial Robustness on In- and Out-Distribution Improves Explainability}

\author{Maximilian Augustin \and
Alexander Meinke \and
Matthias Hein}
\authorrunning{M. Augustin \and A. Meinke \and M. Hein}

\institute{University of T\"ubingen, Germany }

\maketitle

\begin{abstract}
Neural networks have led to major improvements in image classification but suffer
from being non-robust to adversarial changes, unreliable uncertainty estimates on out-distribution samples and their
inscrutable black-box decisions. In this work we propose RATIO, a training procedure for Robustness via Adversarial Training on In- and Out-distribution,
which leads to robust models with 
reliable and robust confidence estimates on the out-distribution. 
RATIO has similar generative properties to adversarial training so that visual counterfactuals produce class specific features.
While adversarial training comes at the price of lower clean accuracy, RATIO
achieves state-of-the-art $l_2$-adversarial robustness on CIFAR10 and maintains better clean accuracy.
\end{abstract}

\section{Introduction}
Deep neural networks have shown phenomenal success in achieving high accuracy on challenging classification tasks \cite{CunBenHin2015}. However, they are lacking in terms of robustness against adversarial attacks \cite{SzeEtAl2014}, make overconfident predictions \cite{GuoEtAl2017,HeiAndBit2019} especially on out-of-distribution (OOD) data \cite{NguYosClu2015,HenGim2017} and their black box decisions are inscrutable \cite{WacMitRus2018}. Progress has been made with respect to all these aspects but there is currently no approach which is accurate, robust, has good confidence estimates and is explainable. Adversarial training (AT) \cite{MadEtAl2018} leads to models robust against adversarial attacks in a defined threat model and has recently been shown to produce classifiers with generative capabilities \cite{santurkar2019computer}. However, AT typically suffers from a significant drop in accuracy and is over-confident on OOD data as we show in this paper. Adversarial confidence enhanced training (ACET) \cite{HeiAndBit2019} enforces low confidence in a neighborhood around OOD samples and can be seen as adversarial training on the out-distribution. ACET leads to models with good OOD detection performance even in an adversarial setting and suffers from a smaller loss in clean accuracy compared to AT. However, ACET models typically are significantly less robust than adversarially trained models. 

In this paper we show that combining AT and ACET into RATIO, Robustness via Adversarial Training on In- and Out-distribution, inherits the good properties of adversarial training and ACET without, or at least with significantly reduced, negative effects, e.g. we get SOTA $l_2$-robustness on CIFAR10 and have better clean accuracy than AT. On top of this we get reliable confidence estimates on the out-distribution even in a worst case scenario. In particular AT yields highly overconfident predictions on out-distribution images in the absence of class specific features whereas RATIO only yields high confident predictions if recognizable features are present. In summary, RATIO achieves high clean accuracy, is robust, calibrated and has
generative properties which can be used to produce high-quality visual counterfactual explanations: 
see Table~\ref{Tab:Summary_cifar10} for a summary of our results for CIFAR10 and SVHN and Table~\ref{Tab:Summary_cifar100} for CIFAR100 and restricted ImageNet~\cite{tsipras2018robustness}.

\input{res/summary.tex}

\section{Related Work}
\textbf{Adversarial Robustness.} 
Adversarial attacks are small changes of an image with respect to some distance measure, which change the decision of a classifier \cite{SzeEtAl2014}.
Many defenses have been proposed but with more powerful or adapted attacks most of them could be defeated \cite{CroHei2020,CarWag2017,AthEtAl2018,MosEtAl18}. Adversarial training (AT) \cite{MadEtAl2018} is the most widely used approach that has not been broken. However, adversarial robustness comes at the price of a drop in accuracy \cite{SchEtAl2018,StuHeiSch2019}. Recent variations are using other losses \cite{ZhaEtAl2019} and boost robustness via generation of additional training data \cite{CarEtAl19,AlaEtAl19} or pre-training \cite{pmlr-v97-hendrycks19a}. Another line of work 
are provable defenses, either deterministic \cite{WonEtAl18,CroEtAl2018,MirGehVec2018,GowEtAl18} or based on randomized smoothing \cite{li2018certified,lecuyer2018certified,CohenARXIV2019}. However, provable defenses are still not competitive with the
empirical robustness of adversarial training for datasets like CIFAR10 and have even worse accuracy. We show that using AT
on the in-distribution and out-distribution leads to a smaller drop in clean accuracy and similar or better robustness.\\
\textbf{Confidence on In- and Out-distribution.}
Neural networks have been shown to yield overly confident predictions far away from the training data \cite{NguYosClu2015,HenGim2017,LeiEtAl2017} and this is
even provably the case for ReLU networks~\cite{HeiAndBit2019}. Moreover, large neural networks are not calibrated on the in-distribution and have a bias to be overconfident~\cite{GuoEtAl2017}. The overconfidence on the out-distribution has been tackled in \cite{LeeEtAl2018,HeiAndBit2019,HenMazDie2019} by enforcing low-confidence predictions on a large out-distribution dataset e.g. using 
the 80 million tiny images dataset\cite{HenMazDie2019} leads to state-of-the-art results. However, if one maximizes the confidence in 
a ball around out-distribution-samples, most OOD methods are again overconfident \cite{SchEtAl2018,HeiAndBit2019,sehwag2019better,meinke2020towards} and only AT on the out-distribution as in ACET \cite{HeiAndBit2019} or methods providing guaranteed worst case OOD performance \cite{meinke2020towards,bitterwolf20provable} work in this worst-case setting. We show that RATIO leads to better worst case
OOD performance than ACET.\\
\textbf{Counterfactual Explanations.}
Counterfactual explanations have been proposed in \cite{WacMitRus2018} as a tool
for making classifier decisions plausible, since humans also justify decisions via
counterfactuals ``I would have decided for X, if Y had been true'' \cite{Mil2017}. 
Other forms are explanations based on image features \cite{HenEtAl2016,HenEtAl2018}. 
However, changing the decision for image classification
in \emph{image space} for non-robust models leads to adversarial samples \cite{DongEtAl2017} with changes that are visually not meaningful.
Thus visual counterfactuals are often based on generative models or restrictions on the space of image manipulation \cite{SamEtAl2018,ParVit2019,ChaEtAl2019,GoyEtAl2019,zhu2016generative,wang2018high}. 
Robust models wrt $l_2$-adversarial attacks \cite{tsipras2018robustness,santurkar2019computer}
have been shown to change their decision when class-specific features appear in the image, which is a prerequisite for meaningful counterfactuals \cite{BarSelRag2020}.
RATIO generates better counterfactuals, i.e. the confidence of the counterfactual images obtained by an $l_2$-adversarial attack tends to be high only after features of the alternative class have appeared. Especially for out-distribution images the difference to AT is pronounced.\\
\textbf{Robust, reliable and explainable classifiers.} This is the holy grail of machine learning. A model which is accurate and calibrated \cite{GuoEtAl2017} on the in-distribution, reliably has low confidence on out-distribution inputs, is robust to adversarial manipulation and has explainable decisions. Up to our knowledge there is no model which claims to have all these properties. The closest one we are aware of is the JEM-0 of \cite{grathwohl2019your} which is supposed to be robust, detects out-of-distribution samples and has generative properties. They state ``JEM does not confidently classify nonsensical
images, so instead, ... natural image properties visibly emerge''. We show that RATIO gets us closer to this ultimate goal and outperforms JEM-0 in all aspects: accuracy, robustness, (worst-case) out-of-distribution detection, and visual counterfactual explanations. 

\section{RATIO: Robust, Reliable and Explainable Classifier}
In the following we are considering multi-class (image) classification. We have the logits of a classifier $f:[0,1]^d \rightarrow \R^K$ where $d$ is the input dimension
and $K$ the number of classes. With $\Delta=\{ p \in [0,1]^K \,|\, \sum_{i=1}^K p_i=1\}$ we denote the 
predicted probability distribution of $f$ over the labels by $\hat{p}:\R^d \rightarrow \Delta$ which is obtained using the softmax function: $\hat{p}_{f,s}(x)=\frac{e^{f_s(x)}}{\sum_{j=1}^K e^{f_j(x)}}$, $s=1,\ldots,K$. We further denote the training set by $(x_i,y_i)_{i=1}^N$ with $x_i \in [0,1]^d$ and $y_i \in \{1,\ldots,K\}$. As loss we always use the cross-entropy loss defined as 
\begin{equation}
L(p,\hat{p}_f) = \sum_{j=1}^K p_j \log(\hat{p}_{f,j}),
\end{equation} 
where $p \in \Delta$ is the true distribution and $\hat{p}_f$ the predicted distribution.

\subsection{Robustness via Adversarial Training}

An adversarial sample of $x$ with respect to some threat model $T(x) \subset \R^d$ is a point 
$z \in T(x) \cap [0,1]^d$ such that the decision of the classifier $f$ changes for $z$ while an oracle would unambiguously associate $z$ with the class of $x$. In particular this implies that $z$ shows no meaningful class-associated features of any other class. 
Formally, let $y$ be the correct label of $x$, then $z$ is an adversarial sample if
\begin{equation}
 \argmax_{k \neq y} f_k(z) > f_y(x), \qquad z \in [0,1]^d \cap T(x),
\end{equation}
assuming that the threat model is small enough such that no real class change occurs.
Typical threat models are $l_p$-balls of a given radius $\epsilon$, that is 
\begin{equation}
T(x)=B_p(x,\epsilon)=\{z \in \R^d \,|\, \norm{z - x}_p \leq \epsilon\}.
\end{equation}
The robust test accuracy is then defined as the lowest possible accuracy when every test image $x$ is allowed to be changed to some $z \in T(x) \cap [0,1]^d$.
Plain models have a robust test accuracy close to zero, even for ``small'' threat models.

Several strategies for adversarial robustness have been proposed, but adversarial training (AT)~\cite{MadEtAl2018} has proven to produce robust classifiers across datasets and network architectures without adding significant computational overhead during inference  (compared to randomized
smoothing \cite{li2018certified,lecuyer2018certified,CohenARXIV2019}).

The objective of adversarial training for a threat model $T(x) \subset \R^d$ is:
\begin{equation}\label{Eq:DefRobust}
\minop_{f} \Exp_{(x,y) \sim p_{\mathrm{in}}}\Big[\maxop_{z \in T(x)} L(\textbf{e}_y,\hat{p}_f(z))\Big],
\end{equation}
where $\textbf{e}_y$ is a one-hot encoding of label $y$ and $p_{\mathrm{in}}(x,y)$ is the training distribution. During training one approximately solves the inner maximization problem in \eqref{Eq:DefRobust} via projected gradient descent (PGD) and then computes the gradient wrt $f$ at the approximate solution of the inner problem. The community has put emphasis on robustness wrt $l_\infty$ but recently there is more interest in other threat models e.g. $l_2$-balls \cite{TraBon2019,RonEtAl2019,santurkar2019computer}. In particular, it has
been noted \cite{tsipras2018robustness,santurkar2019computer} that robust models wrt an $l_2$-ball have the property that ``adversarial'' samples generated within a sufficiently large $l_2$-ball tend to have image features of the predicted class. Thus they are not ``adversarial'' samples
in the sense defined above as the true class has changed or is at least ambiguous.

The main problem of AT is that robust classifiers suffer from  a significant drop in accuracy compared to normal training \cite{tsipras2018robustness}. This trade-off \cite{schmidt2018adversarially,StuHeiSch2019} can be mitigated e.g. via training $50\%$ on clean samples and $50\%$ on adversarial samples at the price of reduced robustness \cite{StuHeiSch2019} or via semi-supervised learning \cite{stanforth2019labels,najafi2019robustness,CarEtAl19}. 

\subsection{Worst-case OOD detection via Adversarial Training on the Out-distribution}

While adversarial training yields robust classifiers, similarly to plain models it suffers from overconfident predictions on out-of-distribution samples. 
Overconfident predictions are a problem for safety-critical systems as the classifier is not
reliably flagging when it operates ``out of its specification'' and thus its confidence in the prediction cannot be used to trigger human intervention.

In order to mitigate over-confident predictions \cite{HeiAndBit2019,HenMazDie2019} proposed to enforce low confidence on images from a chosen out-distribution $p_{\mathrm{out}}(x)$. A generic out-distribution would be all natural images and thus \cite{HenMazDie2019} suggest the 80 million tiny images dataset \cite{80mtiny} as a proxy for this. While \cite{HenMazDie2019} consistently reduce confidence on different out-of-distribution datasets, similar to plain training for the 
in distribution one can again get overconfident predictions by maximizing the confidence in a small ball around a given out-distribution image (adversarial
attacks on the out-distribution \cite{HeiAndBit2019,meinke2020towards}).

Thus \cite{HeiAndBit2019} proposed Adversarial Confidence Enhanced Training (ACET) which enforces low confidence in an entire neighborhood around the out-distribution samples which can be seen as a form of AT on the out-distribution:
\begin{equation}\label{Eq:ACET_objective}
\minop_{f}   \Exp_{(x,y) \sim p_{\mathrm{in}}}\Big[L(\vec{e}_y,\hat{p}_f(x))\Big] + \lambda  \,\Exp_{(x,y) \sim p_{\mathrm{out}}}\Big[\maxop_{\norm{z-x}_2\leq \epsilon} L(\ones/K,\hat{p}_f(z))\Big],
\end{equation}
where $\ones$ is the vector of all ones (outlier exposure \cite{HenMazDie2019} has the same objective without the inner maximization
for the out-distribution). Different from \cite{HeiAndBit2019} we use the same loss for in-and out-distribution, whereas they used the maximal log-confidence over
all classes as loss for the out-distribution. In our experience the maximal log-confidence is more difficult to optimize, but both losses are minimized by the
uniform distribution over the labels. Thus the difference is rather small and we also denote this version as ACET.

\subsection{RATIO: Robustness via Adversarial Training on In-and Out-distribution}

We propose RATIO: adversarial training on in-and out-distribution. This combination
leads to synergy effects where most positive attributes of AT and ACET are fused without having larger drawbacks. 
The objective of RATIO is given by: 
\begin{equation}\label{Eq:RATIO}
\minop_{f}   \Exp_{(x,y) \sim p_{\mathrm{in}}}\Big[\maxop_{\norm{z-x}_2\leq \epsilon_i} L(\vec{e}_y,\hat{p}_f(z))\Big] + \lambda  \,\Exp_{(x,y) \sim p_{\mathrm{out}}}\Big[\maxop_{\norm{z-x}_2\leq \epsilon_o} L(\ones/K,\hat{p}_f(z))\Big],
\end{equation}
where $\lambda$ has the interpretation of $\frac{p_o}{p_i}$, the probability to see out-distribution $p_o$ and in-distribution $p_i$ samples at test time. Here we have specified an
$l_2$-threat model for in-and out-distribution but the objective can be adapted to different threat models which could be different for in-
and out-distribution. The surprising part of RATIO is that the addition of the out-distribution part can improve the results even on the in-distribution in
terms of (robust) accuracy. 
The reason is that adversarial training on the out-distribution ensures that spurious features do not change the confidence of the classifier. This behavior generalizes to the in-distribution and thus ACET (adversarial training on the out-distribution) is also robust on the in-distribution ($52.3\%$ robust accuracy for $l_2$ with $\epsilon=0.5$ on CIFAR10). One problem of adversarial training is overfitting on the training set \cite{rice2020overfitting}. Our RATIO has seen more images at training time and while the direct goal is distinct (keeping one-hot prediction on the in-distribution and uniform prediction on out-distribution) both aim at constant behavior of the classifier over the $l_2$-ball and thus the effectively increased training size improves generalization (in contrast to AT, RATIO has its peak robustness at the end of the training). 
Moreover, RATIO typically only shows high confidence if class-specific features have appeared which we use in the generative process described next.
\section{Visual Counterfactual Explanations}
\begin{figure}[ht!]
\begin{tabular}{p{1cm}x{\breite}x{\breite}x{\breite}x{\breite}x{\breite}x{\breite}x{\breite}x{\breite}}
Model  & Orig. & $\epsilon=0.5$ & $\epsilon=1.0$ & $\epsilon=1.5$ & $\epsilon=2.0$ & $\epsilon=2.5$ & $\epsilon=3.0$\\
\begin{turn}{90} \hspace{-.4cm} Plain \end{turn} & \multicolumn{7}{c}{\includegraphics[width=0.91\textwidth,valign=c]{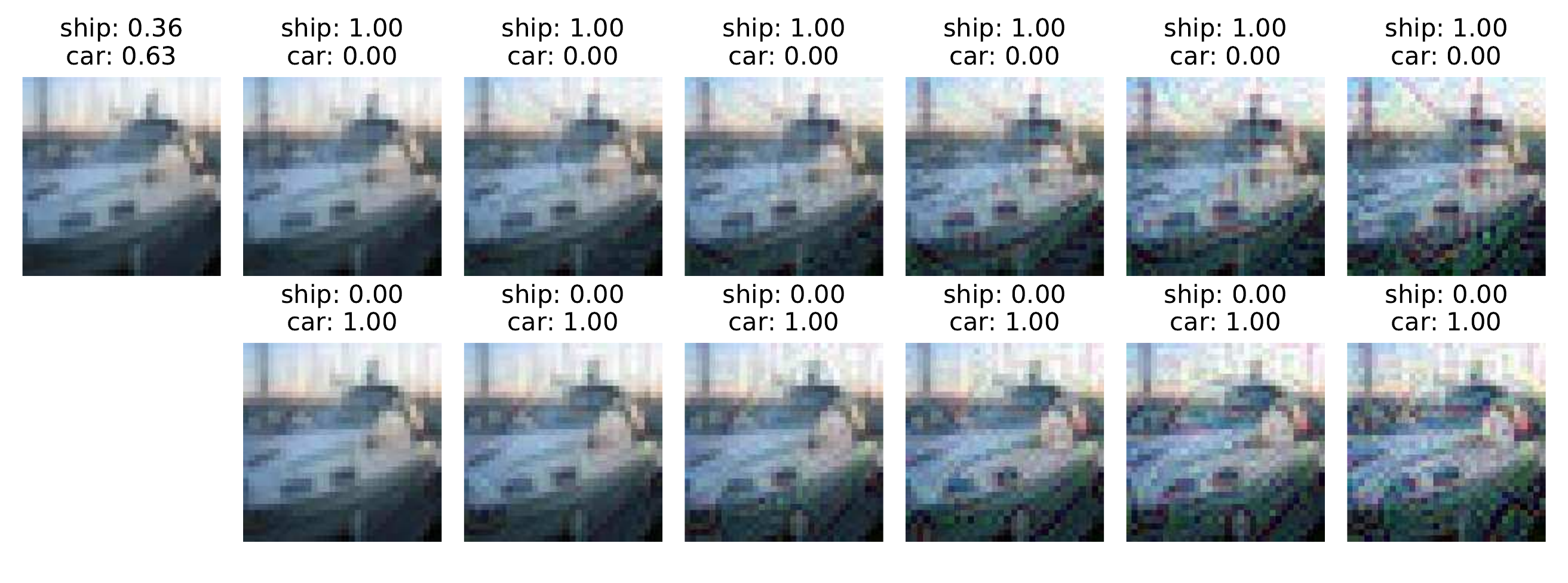}} \\
\end{tabular}
\caption{\label{fig:plain-cf} Failure of a visual counterfactual for a plain model. The targeted attack immediately produces very high confidence in both classes
but instead of class features only high-frequency noise appears because plain models are not robust.}
\end{figure}

The idea of a counterfactual explanation \cite{WacMitRus2018} is to provide the smallest change of a given input such that the decision changes into a desired
target class e.g. how would this X-ray image need to look in order to change the diagnosis from X to Y.
Compared to sensitivity based explanations \cite{BaeEtAl2010,ZeiFer2014} or explanations based on feature attributions \cite{BacEtAl2015} counterfactual explanations have the advantage that they have an ``operational meaning'' which couples the explanation directly to the decision of the classifier. On the other hand the
counterfactual explanation requires us to specify a metric or a budget for the allowed change of the image which can be done directly in image space or in 
the latent space of a generative model. However, our goal is that the classifier directly learns what meaningful changes are and we do not want to impose that
via a generative model. Thus we aim at visual counterfactual explanations directly in image space with a fixed budget for changing the image. As the decision
changes, features of this class should appear in the image (see Figure \ref{fig:cf}). Normally trained models will not achieve this
since non-robust models change their prediction for non-perceptible perturbations \cite{SzeEtAl2014}, see Figure \ref{fig:plain-cf}. Thus robustness against ($l_2$-)adversarial perturbations is a necessary requirement for visual counterfactuals
and indeed \cite{tsipras2018robustness,santurkar2019computer} have shown ``generative properties'' of $l_2$-robust models.

A \textit{visual counterfactual} for the original point $x$ classified as $c=\argmax_{k=1,\ldots,K} f_k(x)$, a target class $t \in \{1,\ldots,K\}$ and a budget $\epsilon$ is defined as
\begin{equation}\label{eq:cf} x^{(t)} = \argmax_{z \in [0,1]^d,\,\norm{x-z}_2 \leq \epsilon} \hat{p}_{f,t}(z),
\end{equation}
where $\hat{p}_{f,t}(z)$ is the confidence for class $t$ of our classifier for the image $z$. If $t \neq c$ it answers the counterfactual question 
of how to use the given budget to change the original input $x$ so that the classifier is most confident in class $t$. Note that in our definition we include the case where $t=c$, that is we ask how to change the input $x$ classified as $c$ to get even more confident in class $c$. In Figure \ref{fig:cf} we illustrate
both directions and show how for robust models class specific image features appear when optimizing the confidence of that class. This shows that the optimization
of visual counterfactuals can be done directly in image space.

\input{res/summary2.tex}

\section{Experiments}\label{Sec:Exp}
\begin{figure}[ht]
     \centering
     \subfloat[][ID worst-case confidence]{\includegraphics[width=0.45\textwidth]{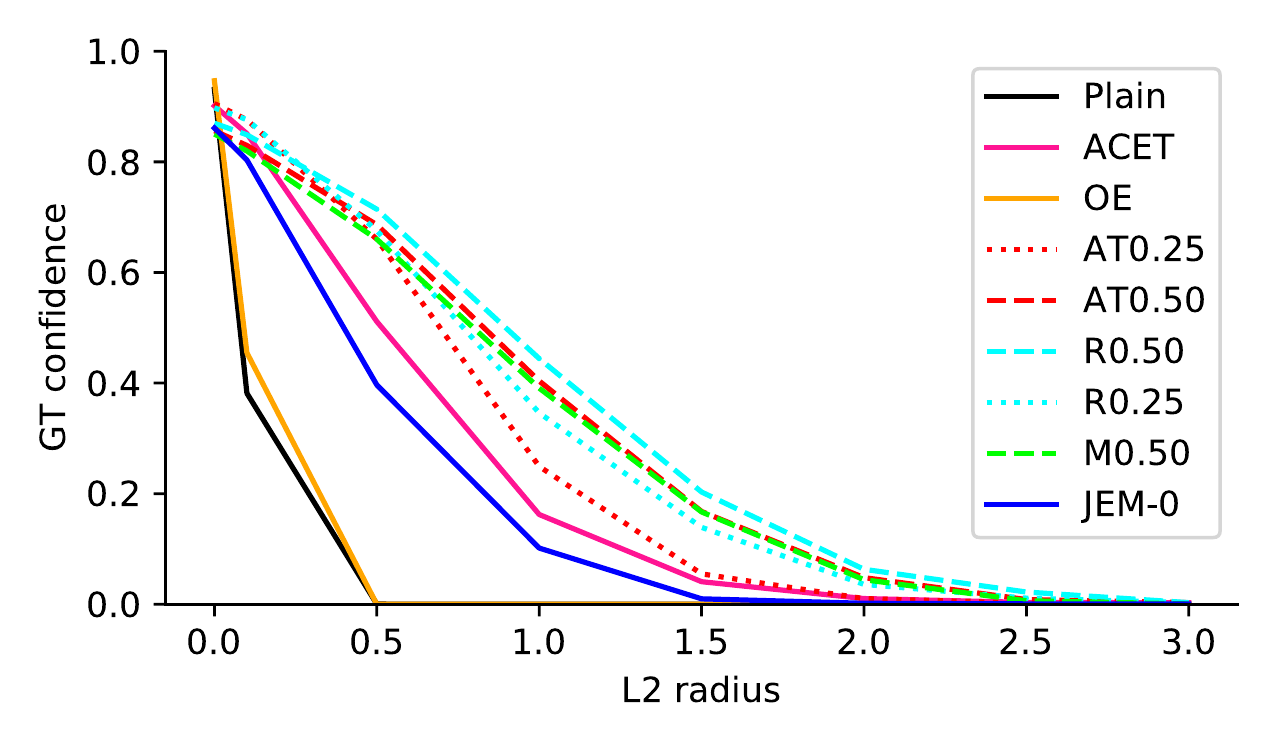}\label{Fig:CIFAR_ID_CONF}}
		\quad \;
     \subfloat[OD worst-case confidence][OD worst-case confidence]{ \includegraphics[width=0.45\textwidth]{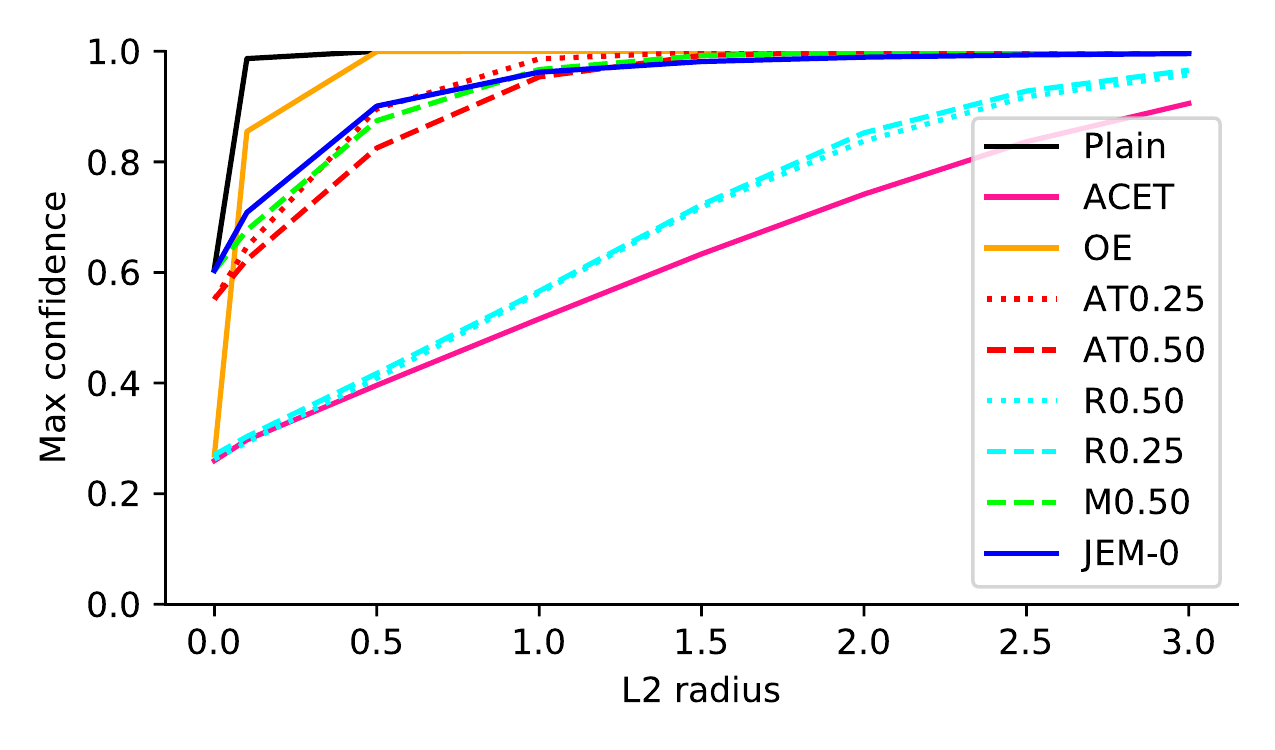}\label{Fig:CIFAR_OD_CONF}}
     \caption{(a) Mean confidence in true label as a function of the attack $l_2$-radius around CIFAR10 test images. RATIO and AT0.5 have a reasonable decay of the confidence. 
		          (b) Mean of maximal confidence around OD-data (tiny images) over the attack $l_2$-radius. All methods except
							RATIO and ACET are overconfident.}
     \label{Fig:CIFAR_ID_OD_CONF}
\end{figure}

\textbf{Comparison, Training and Attacks.} We validate our approach on SVHN \cite{SVHN}, CIFAR10/100 \cite{cifar10} and restricted ImageNet \cite{santurkar2019computer}.
On CIFAR10 we compare RATIO to a pretrained JEM-0~\cite{grathwohl2019your} and the AT model \cite{robustness} with $l_2=0.5$ (M$_{0.5}$)
(both not available on the other datasets). As an ablation study of RATIO we train a plain model, outlier exposure (OE) \cite{HenMazDie2019}, ACET \cite{HeiAndBit2019} and AT with $l_2=0.5$ (AT$_{0.5}$) and $l_2=0.25$ (AT$_{0.25}$), using the same hyperparameters as for our RATIO training. On SVHN we use a ResNet18 architecture for all methods and on the other datasets we use ResNet50, both with standard input normalization. For ACET on CIFAR10 we use ResNet18 since for ResNet50 we could not obtain a model with good worst case OOD performance as the attack seemed to fail at some point during training (on CIFAR100 this was even the case for ResNet18 and thus we omit it from comparison). In general ACET is difficult
to train. For RATIO the additional adversarial training on the in-distribution seems to stabilize the training and we did not encounter any problems.
As out-distribution for SVHN and CIFAR we use 80 million tiny images \cite{80mtiny} as suggested in \cite{HenMazDie2019} and for restricted ImageNet the remaining ImageNet classes. 
For the out-distribution we always use 
$l_2$-attacks with radius $\epsilon_o=1$ for SVHN/CIFAR and $\epsilon_o=7$ on restricted ImageNet (both ACET and RATIO) whereas on the in-distribution we use $\epsilon_i=0.25$ and $\epsilon_i=0.5$ and $\epsilon_i=1.75$ and $\epsilon_i=3.5$, respectively (both AT and RATIO). Therefore RATIO/AT models are labeled by $\epsilon_i$. For further training details see the Appendix. 
For the adversarial attacks on in- and out-distribution we use the recent Auto-Attack \cite{CroHei2020} which is an ensemble of four attacks, including the
black-box Square Attack \cite{andriushchenko2019square} and three white-box attacks (FAB-attack \cite{croce2019minimally} and AUTO-PGD with different losses). For each of the white-box attacks, a budget of $100$ iterations and $5$ restarts is used and a query limit of  $5000$ for Square attack. In \cite{CroHei2020} they
show that Auto-Attack consistently improves the robustness evaluation for a large number of models (including JEM-x). 
\\
\textbf{Calibration on the in-distribution.}
With RATIO we aim for reliable confidence estimates, in particular no overconfident predictions. In order to have comparable confidences
for the different models we train, especially when we check visual counterfactuals or feature generation, we first need to ``align'' their confidences.
We do this by minimizing the expected calibration error (ECE) via temperature rescaling \cite{GuoEtAl2017}. 
Note that this rescaling does not change the classification and thus has no impact on (robust) accuracy and only a minor influence
on the (worst case) AUC values for OOD-detection. For details see the Appendix.\\ 
\textbf{(Robust) Accuracy on the in-distribution.}
Using Auto-Attack \cite{CroHei2020} we evaluate robustness on the full test set for both CIFAR and r. Imagenet and 10000 test samples for SVHN. 
Tables \ref{Tab:Summary_cifar10} and \ref{Tab:Summary_cifar100} contain (robust $l_2$) accuracy, detailed results, including $l_\infty$ attacks, can be found in the Appendix.  
On CIFAR10, RATIO achieves significantly higher robust accuracy than AT for $l_2$-and $l_\infty$-attacks. Thus the additional adversarial training on the out-distribution with radius $\epsilon_o=1$ boosts the robustness on the in-distribution. In particular, RATIO$_{0.25}$ achieves better $l_2$-robustness than AT$_{0.5}$
and M$_{0.5}$ at $\approx 2.7\%$ higher clean accuracy. In addition, R$_{0.5}$ yields new state-of-the-art $l_2$-robust accuracy at radius $0.5$ (see \cite{CroHei2020} for a benchmark) while having higher test accuracy than AT$_{0.5}$, M$_{0.5}$. Moreover, the $l_2$-robustness at radius $1.0$ and the $l_\infty$-robustness at $8/255$ is significantly better. Interestingly, although ACET is not designed to yield adversarial robustness on the in-distribution, it achieves more than $50\%$ robust accuracy for $l_2=0.5$ and outperforms JEM-0 in all benchmarks. However, as our goal is to have a model which is both robust and accurate, we recommend to use $R_{0.25}$ for CIFAR10 which has a drop of only $2.6\%$ in test accuracy compared to a plain model while having similar robustness to M$_{0.5}$ and AT$_{0.5}$. Similar observations as for CIFAR10 hold for CIFAR100 and for Restricted ImageNet, see Table \ref{Tab:Summary_cifar100}, even though
for CIFAR100 AT and RATIO suffer a higher loss in accuracy.
On SVHN, RATIO outperforms AT in terms of robust accuracy trained with the same $l_2$-radius but the effect is less than for CIFAR10.
We believe that this is due to the fact that that the images obtained from the 80 million tiny image dataset (out distribution) do not reflect the specific structure of SVHN numbers which makes (worst case) outlier detection an easier task. This is supported by the fact that ACET achieves better clean accuracy on SVHN than both OE and the plain model while it has worse clean accuracy on CIFAR10.\\
\textbf{Visual Counterfactual Generation.}
We use 500 step Auto-PGD \cite{CroHei2020} for a targeted attack with the objective in \eqref{eq:cf}. However, note that this non-convex optimization problem has been shown to be NP-hard \cite{KatzEtAl2017}. In Figure \ref{fig:cf}, \ref{fig:cf_svhn} and \ref{fig:cf_imagenet} and in the Appendix we show generated counterfactuals for all datasets. For CIFAR10 AT$_{0.5}$ performs very similar to RATIO$_{0.25}$ in terms of the emergence of class specific image features. In particular, we often see the appearance of characteristic features such as pointed ears for cats, wheels for cars and trucks, large eyes for both cats and dogs
and the antlers for deers. JEM-0 and ACET perform worse but for both of them one observes the appearance of image features. However, particularly the images of 
JEM-0 have a lot of artefacts. For SVHN RATIO$_{0.25}$ on average performs  better than AT$_{0.25}$ and ACET. 
It is interesting to note that for both datasets class-specific features emerge already for an $l_2$-radius of $1.0$. Thus it seems questionable if $l_2$-adversarial robustness beyond a radius of $1.0$
should be enforced. Due to the larger number of classes, CIFAR100 counterfactuals are of slightly lower quality. For Restricted ImageNet the visual counterfactuals show class-specific features but can often be identified as synthetic due to misaligned features.\\
\textbf{Reliable Detection of (Worst-case) Out-of-Distribution Images.}
A reliable classifier should assign low confidence to OOD images. This is not the case for plain models and AT. 
As the 80 million tiny image dataset has been used for training for ACET and RATIO (respectively other ImageNet classes for Restricted ImageNet), we evaluate the discrimination of in-distribution versus out-distribution on other datasets as in \cite{meinke2020towards}, see the Appendix for details. We use $\max_k \hat{p}_{f,k}(x)$ as feature to discriminate in-and out-distribution (binary classification) and compute the AUC.
However, it has been shown that even state-of-the-art methods like outlier exposure (OE)
suffer from overconfident predictions if one searches for the most confident prediction in a small neighborhood 
around the
the out-distribution image \cite{meinke2020towards}. Thus we also report the worst-case AUC by maximizing the confidence in an $l_2$-ball of radius $1.0$ (resp. $7.0$ for R. ImageNet)
around OOD images via Auto-PGD \cite{CroHei2020} with $100$ steps and $5$ random restarts. Figure \ref{Fig:CIFAR_ID_OD_CONF} further shows that while RATIO behaves similar to AT around samples from the data distribution, which explains similar counterfactuals, it has a flatter confidence profile around out-distribution samples. 
\input{res/counterfactuals_plot_cifar}

\input{res/counterfactuals_plot_svhn}

\noindent on $1024$ points from each out-distribution ($300$ points for LSUN\_CR).
Using the worst case confidences of these points we find 
empirical upper bounds on the worst-case AUC under our threat model. We report both the average-case AUCs as well as the worst-case AUCs in the Appendix.
The average AUC over all OOD datasets is reported in Tables \ref{Tab:Summary_cifar10} and \ref{Tab:Summary_cifar100}.
The AT-model of Madry et. al (M$_{0.5}$) perform worse than the plain model even on the average case task. However, we see that with our more aggressive data augmentation this problem is somewhat alleviated (AT$_{0.5}$ and AT$_{0.25}$). As expected ACET, has good worst-case OOD performance but is similar to the plain model for the average case. JEM-0 has bad worst-case AUCs and we cannot confirm the claim  that ``JEM does not confidently classify nonsensical images''\cite{grathwohl2019your}. As expected, OE has state-of-the-art performance on the clean task but has no robustness on the out-distribution, so it fails completely in this regime. Our RATIO models show strong performance on all tasks and even outperform the ACET model which shows that adversarial robustness wrt the in-distribution also helps with adversarial robustness on the out-distribution. 
On SVHN the average case OOD task is simple enough that several models achieve near perfect AUCs, but again only ACET and our RATIO models manage to retain strong performance in the worst case setting. The worst-case AUC of AT models is significantly worse than that of ACET and RATIO.\\
\input{res/imagenet_plots_in.tex}
\input{res/imagenet_plots_out.tex}
\textbf{Feature Generation on OOD images.}
Finally, we test the abilities to generate image features with a targeted attack on OOD images (taken from 80m tiny image dataset resp. ImageNet classes not belonging to R. ImageNet). The setting is similar to the visual counterfactuals. We take
some OOD image and then optimize the confidence in the class which is predicted on the OOD image. The results
can be found in Figure \ref{Fig:CIFAR_OD} and \ref{Fig:CIFAR100_OD} and additional samples are attached in the Appendix. For CIFAR10 all methods are able to generate image features of the class but the predicted confidences are only reasonable for ACET and RATIO$_{0.25}$ whereas AT$_{0.5}$ and JEM-0 are overconfident when no strong class features are visible. 
This observation generalizes to SVHN and mostly CIFAR100 and r. Imagenet, \ie RATIO generally has the best OOD-confidence profile.\\
\textbf{Summary.}
In summary, in Table \ref{Tab:Summary_cifar10} and \ref{Tab:Summary_cifar100} we can see that RATIO$_{0.25}$ resp. RATIO$_{1.75}$ is except for CIFAR100 the only model which
has no clear failure case. Here the subjective definition of a failure case (highlighted in red) is an entry which is ``significantly worse'' than the best possible in this metric. Thus we think that RATIO succeeds in being state-of-the-art in generating a model which is accurate, robust, has reliable confidence and is able to produce meaningful visual counterfactuals.  
Nevertheless RATIO is not perfect and we discuss failure cases of all models in the Appendix.

\section{Conclusion and Outlook}
We have shown that adversarial robustness on in-distribution and out-distribution (as a proxy of all natural images) gets us closer to a classifier
which is accurate, robust, has reliable confidence estimates and is able to produce visual counterfactual explanations with strong class specific image features. For the usage in safety-critical in systems it would be ideal if these properties can be achieved in a provable way which remains an open problem. 

\input{res/od_plots.tex}

\section*{Acknowledgements}
M.H and A.M. acknowledge support by the BMBF T\"ubingen AI Center (FKZ: 01IS18039A) and by DFG TRR 248, project number 389792660 and the DFG Excellence Cluster “Machine Learning -New Perspectives for Science”, EXC 2064/1, project number 390727645. A.M. thanks the IMPRS for Intelligent Systems.

%
%
\bibliographystyle{splncs04}
\bibliography{biblio,Literatur}

\appendix
\newpage
\section{Additional Results}
\subsection{Detailed results on robustness evaluation on the in-distribution}
The full adversarial robustness evaluation for all datasets can be found in Table~\ref{Tab:IDRobustness_cifar10}.
\input{res/cifar10_id.tex}

\subsection{Comparison to TRADES on CIFAR10}
TRADES \cite{ZhaEtAl2019} has shown competitive performance regarding adversarial robustness and thus we compare to TRADES on CIFAR10.
We trained two ResNet50 models with the popular TRADES training scheme \cite{ZhaEtAl2019}. The first one is based on the official PyTorch implementation with $l_2$ radius 0.5 and uses the same hyperparameters that were used to train the reference models.  As those hyperparameters are likely better suited for $L_\infty$ training, we further implemented TRADES training in our framework and trained an additional model with the same hyperparameters that were used for the AT and RATIO models. While our own implementation of TRADES is slightly better than standard adversarial training, our RATIO-0.5 model outperforms both TRADES models in terms of accuracy and robustness .

\input{res/appendix_trades}

\subsection{Detailed results on out-of-distribution detection}\label{App:OOD}

For the evaluation of OOD performance we use datasets not seen during training time. In detail, for CIFAR10 we use: SVHN, CIFAR100, downscaled ImagetNet with the CIFAR10 classes removed, the classroom set of
 LSUN \cite{LSUN}, as well as uniform noise images and smoothed noise \cite{HeiAndBit2019}, for SVHN: same, SVHN exchanged with CIFAR10, for CIFAR100: same, CIFAR100 exchanged with CIFAR10. For Restricted ImageNet we use datasets with higher image resolution: Flowers\cite{Nilsback08}, Food101\cite{lee2017cleannet}, FGVC Aircraft\cite{maji13fine-grained}, Cars\cite{KrauseStarkDengFei-Fei_3DRR2013} and uniform noise.
The detailed results for the out-of-distribution detection performance of all models can be found
in Table \ref{Tab:OOD_Cifar10} for CIFAR10, SVHN and CIFAR100 and in Table \ref{Tab:OOD_Cifar10-b}
for restricted ImageNet. The summary tables in the main paper have provided the average AUC and average worst case
AUC over the respective out-of-distribution datasets.

\section{Experimental Setup}\label{App:Exp}
\subsection{CIFAR10}
For our experiments on CIFAR10 \cite{krizhevsky2009learning} we use a standard ResNet50 architecture and SGD with Nesterov momentum ($\beta = 0.9$) and a base learning rate of 0.1 and weight decay of $5e-4$. Our training schedule spans 220 epochs and we decrease the learning rate by a factor of 10 in epochs 100, 150 and 200. As data augmentation for all our trained CIFAR10 models we use the recommended AutoAugment policy from \cite{cubuk18autoaugment}, including Cutout \cite{devries17cutout}. 

For adversarial and RATIO training on CIFAR10, we use $100\%$ adversarial training on the train distribution, \ie the model only sees perturbed samples during training. Instead of solving the robust min-max formulation (\eqref{Eq:DefRobust}) directly, we use the logits-based loss from \cite{carlini16towards} in the inner maximization problem, \ie for a training sample $(x,y)$ we approximately solve:
\begin{equation}
\maxop_{z \in T(x)} \maxop_{i \neq y} f_i(z) - f_y(z).
\end{equation}
To compute $z$, we use a 7-step PGD with the $l_2$-normalized gradient with step size 0.1 and momentum weight 0.9 which returns the point with the highest loss across its trajectory. 
This deviation from the standard adversarial training scheme of \cite{MadEtAl2018} is justified by our empirical experience that for this small number of steps the optimization of the logits-based loss even leads to higher cross-entropy loss than optimizing the cross-entropy loss directly. Note that for the actual update of
the model we use the gradient of the cross-entropy loss.

The same scheme applies to the inner maximization problem for the adversarial training on the out-distribution (cross-entropy loss to uniform distribution, see also \eqref{Eq:ACET_objective}) 
in ACET and RATIO training, where we again use PGD with momentum and a step size of $0.1$. We again emphasize that unlike \cite{HeiAnd2017} who used a smoothed form of noise as out-distribution, we use 80 Million Tiny Images which makes ACET resp. the adversarial training on the out-distribution a substantially harder task. As 
the radius of the $l_2$-threat model on the out-distribution is significantly larger than on the in-distribution we increase the initial number of iterations to 20. For pure ACET training we noticed that even a 20-step attack is often too weak to find an approximate maximum of the inner maximization problem which results in the model gradually becoming less robust. We therefore incrementally increase the number of ACET iterations to 40 by adding 5 steps for each update of the learning rate. However even with those adjustments, pure ACET training on CIFAR10 remains very unstable and reproducibly ends up with a maximum-mean confidence close to 0.1 for CIFAR10 test samples. We therefore use a smaller ResNet18 with a 100 epoch schedule for all ACET experiments where this training scheme can be used without problems. We note that RATIO does not suffer from ACET's stability problems and in this setting the training reliably works.

The threat models are $l_2$-balls of radius 0.25 resp. 0.5 on the in-distribution and a $l_2$-ball of radius 1.0  n the out-distribution. The number of iterations and the radii of the threat model for the different methods are summarized in Table~\ref{Tab:STEPS}. We use a batch size of 128 for plain and adversarial training and a total batch size of 256 for OE, ACET and RATIO training, \ie 128 samples from the in- and 128 samples from the out-distribution. 

As adversarial training is prone to overfitting on the training set \cite{rice2020overfitting}, resulting in a loss in robust accuracy on the test set in the last epochs of training, we use the robust accuracy on the test set under the 7 step PGD attack as early stopping criterion (note that the $7$-step PGD attack is significantly weaker than what we use later on for evaluation of robustness).
\begin{table}[t]
\begin{center}
\caption{\label{Tab:STEPS} Radii and number of inner optimization steps for the different methods when applicable.  }
\begin{tabular}{l|ccccccccc}
\hline
CIFAR10 & Plain &  OE  & ACET  & M$_{0.5}$ &  AT$_{0.5}$ & AT$_{0.25}$ & JEM-0 &  R$_{0.5}$ & R$_{0.25}$ \\
\hline
AT radius &  -  & - & - & 0.5 & 0.5 & 0.25 & - &  0.5  &  0.25  \\
AT steps  &  -  & - & - & 7   & 7   & 7    & - &  7    & 7  \\
ACET radius &  - & - & 1.0 & - & - & - & - & 1.0 & 1.0 \\
ACET steps &  - & - & 20-40 & - & - & - & - & 20-40 & 20-40 \\
\hline
\end{tabular}
\end{center}
\end{table}

\input{res/cifar10_ood.tex}

\subsection{SVHN}
Our SVHN training scheme is similar to the CIFAR10 schedule, in particular the chosen radii and number of steps are identical for all training methods. The only differences are the model architecture and data augmentation. For SVHN, we use a ResNet18 architecture with a 100 epochs schedule which decreases the learning rate in epochs 50, 75 and 90. The data augmentation scheme consists of input normalization, random cropping and Cutout. 

\subsection{CIFAR100}
Our CIFAR100 schedule is an exact replicate of the CIFAR10 schedule. However we were not able to train an OD-robust ResNet18 or ResNet50 ACET model. Throughout multiple training runs with different
PGD schedules and different data augmentation schemes, the training failed as the the attack was not able to approximate the inner maximization problem, resulting in a model with unexpected worst-case out-distribution performance. We therefore decided to not include any ACET Cifar100 results. We further note that RATIO trained perfectly stable despite its use of the same ACET loss. 

\subsection{Restricted ImageNet}
To demonstrate the feasibility of our method on higher-resolution images, we use restricted ImageNet\cite{santurkar2019computer}, a subset of the ILSVRC2012 dataset\cite{russakovsky2015}. It contains 9 super-classes, consisting of 118 ("dog") to 3 ("frog") individual ILSVRC2012 classes. 
We adopt the overall training scheme from \cite{santurkar2019computer}, including the ResNet50 architecture and data augmentation with a slightly shorter 75 epoch LR schedule which decays the initial LR of $0.1$ at epochs $30$, $60$ and $75$ by a factor of 10.
We also tested the AutoAugment ImageNet policy, however, found that it performed worse than the simpler transform based on random crops, flips, color jitter and a lighting transformation.  

As 80 million tiny images only contains images with a resolution of 32x32 and we want to make use of the full 224x224 native resolution of the ImageNet ResNet50 models, we use all the remaining classes from ILSVRC2012 as out-distribution for RATIO and ACET training. 

As a result of the larger image resolution, we scale up the $l_2$ radii and use $1.75$ and $3.5$ in place of $0.25$ and $0.5$ on the in-distribution and $7.0$ instead of $1.0$ for out-distribution worst-case training. We continue using the 7 step PGD during training on the in-distribution with a stepsize of $0.7$. Due to computational complexity, we use a simple 10 step PGD with a stepsize of $1.0$ on the out-distribution instead of increasing the number of iterations during training. On restricted ImageNet, we noticed an unexpected drop in clean accuracy on the larger RATIO models and therefore add an additional clean in-distribution loss to the RATIO and AT models. In detail, adversarial training uses a 50/50 scheme with 128 standard and 128 perturbed samples per batch while RATIO uses 128 clean and 128 perturbed samples from the in-distribution and 128 perturbed samples from the out distribution, resulting in a total batch-size of 384. Such a scheme typically improves clean accuracy while reducing the robustness, however we note that our 50/50 AT models are able to compete with Madry's standard AT model ($100\%$ adversarial training) and are thus a fair baseline for RATIO.

\section{\label{App:Temp} Calibration on the In-Distribution by Temperature Rescaling}

We follow \cite{GuoEtAl2017} and compute the expected calibration error (ECE) by grouping the confidences in $M$ equally sized bins ${B_m \subset [1/K,1]}$ and then estimate the ECE via:
\begin{equation}
\mathrm{ECE} = \sum_{m=1}^M \frac{|B_m|}{n} \left|\mathrm{acc}(B_m)-\mathrm{conf}(B_m)\right|.
\end{equation}
On the validation set we use $10$ bins (note that validation sets are small than the test set) and for the evaluation of the final calibration on the test set we use $15$ bins. For the finding the temperature we pick 500 geometrically spaced temperatures on the interval $T\in[0.05,2.71]$ and choose the minimizer for each model. Since M$_{0.5}$ and JEM-0 have used the entire training set and removing data from the test set would make the accuracy values harder to compare, we use the CIFAR10.1 dataset \cite{recht2018cifar10.1} for calibration on CIFAR10. On SVHN we use 2000 points from the unused additional data, on CIFAR100 the first 2000 test points and on R.Imagenet we use a random subset of 2000 test points as our validation set. For CIFAR100 and R.Imagenet we omit these 2000 points when testing the OOD performance. 

The results on the validation sets are given in Table \ref{Tab:Cal}. Note that all models that rely on an out-distribution (OE, ACET, RATIO) are initially highly uncalibrated on the in-distribution. The reason is simply that they were trained on the distribution ${p_{\mathrm{joint}}(x,y)=\frac{1}{2}p_{\mathrm{in}}(x,y) + \frac{1}{2}p_{\mathrm{out}}(x,y)}$ so they produce underconfident predictions on $p_{\mathrm{in}}$. This effect is far weaker for SVHN because TinyImages can be separated from SVHN much more easily than from CIFAR10 and thus the effect on the in-distribution confidences is much lower. On CIFAR10, if we compute the ECE with respect to the distribution where half the samples are drawn from the in-distribution and half the samples from the out-distribution with a randomly selected label, then the out-distribution trained models achieve drastically lower ECEs, whereas the models that are only trained on the in-distribution receive much higher scores, e.g. on the CIFAR10 test set the (non-temperature rescaled) AT$_{0.5}$ model goes from an ECE of $1.1$ on $p_{\mathrm{in}}$ to $24.2$ on $p_{\mathrm{joint}}$, while RATIO$_{0.25}$ improves from $20.0$ to $6.6$. This is why minimizing the ECE on the validation set by temperature rescaling is necessary in order to directly compare the confidence values of differently trained models. 

\input{res/table_cal.tex}

\section{Additional Samples}
We provide additional samples of the visual counterfactuals and feature generation from OOD images. Moreover, we have a separate section where we discuss
failure cases of our and other methods.

\subsection{Visual Counterfactuals}
In Figure \ref{fig:cf} we generated visual counterfactual explanations on test images that were misclassified by all models. For space reasons we omitted some models which we present now in Figures \ref{fig:vc_cifar_ext1} and \ref{fig:vc_cifar_ext2} as well as Figures \ref{fig:vc_svhn_ext1} and \ref{fig:vc_svhn_ext2} for CIFAR10 and SVHN respectively. For the reader's convenience we also repeat the samples from the main paper in these figures for better comparison. As already demonstrated in Figure \ref{fig:plain-cf}, the non-robust models (Plain and OE) have no interpretable visual counterfactuals. JEM-0 produces some visible class features but introduces severe artefacts so that JEM-0 is inferior in its ``generative'' properties compared to ACET, AT and RATIO. The other models are qualitatively similar to each other since they all exhibit adversarial robustness on the in-distribution. For the models shown in the main text, we present additional visual counterfactuals in Figures \ref{fig:vc_cifar_new1} through \ref{fig:vc_cifar_new3} for CIFAR10 and Figures \ref{fig:vc_svhn_new1} through \ref{fig:vc_svhn_new3} for SVHN. On CIFAR10 ACET, AT$_{0.5}$ and RATIO$_{0.25}$ all show good generative performance. In order to avoid cherry picking we also demonstrate this on a random selection of visual counterfactuals (we show 48 randomly selected test samples from the set of 70 test points that all models classify wrongly) at radius $\epsilon=3.0$ in Figure \ref{fig:cifar_overview} for CIFAR10. Interestingly ACET achieves good generative performance around the in-distribution with smaller in-distribution robustness than both AT and RATIO models. 

On SVHN we also observe that ACET, AT$_{0.5}$ and RATIO$_{0.25}$ perform very well at generating plausible counterfactuals. Surprisingly, AT$_{0.25}$ performs much worse as it introduces many visual artefacts. This is in so far interesting as the in-distribution robustness of AT$_{0.25}$ is very similar to RATIO$_{0.25}$. This suggests that robustness to the out-distribution improves the generative properties. Again, we show a panel of 48 randomly chosen visual counterfactuals in Figure \ref{fig:svhn_overview} (out of the 199 images which are misclassified by all methods). Note that the reason for the common misclassification is often that the labels of these cases are either completely wrong or the label is correct for some digit shown but not for the more central digit which should be predicted (already in the first row of 8 images there are 4 obviously wrong cases). In particular for high contrast images generating a meaningful visual counterfactual with an $l_2$-budget of $3.0$ is very difficult.

Further visual counterfactuals for both AT and RATIO models on CIFAR100 can be found in Figures \ref{fig:vc_cifar100_new1} to \ref{fig:vc_cifar100_new3} and restricted ImageNet in Figures \ref{fig:vc_imagenet_new1} to \ref{fig:vc_imagenet_new3}. Additionally, we present an randomly generated overview for CIFAR100 in Figure  \ref{fig:cifar100_overview} and restricted ImageNet in Figure \ref{fig:restricted_overview}. Despite the increased complexity of CIFAR100, the models are mostly able to produce meaningful counterfactuals and RATIO often yields a higher image quality than standard adversarial training. The overview plot for CIFAR100 however shows that the overall counterfactual quality for CIFAR100 is worse than for CIFAR10 or SVHN. This is to be expected as the number of samples per class is ten times smaller in comparison to CIFAR10 which makes it hard to learn class specific features, especially when the model has to distinguish fine grained classes like "oak", "maple" and other tree species. 

Even though the generation of high quality images in 224x224 directly in image space is a very challenging task, the samples show that robust models are able to produce high quality counterfactuals. Often the changes integrate seamlessly into the reference image by only adding specific textures or details and thereby removing the ground-truth subject. However, it can also be the case that the new class features are painted on top of the existing image and often remain slightly transparent due to the fixed $l_2$ budget. We further discuss this in Section \ref{subsec:failure}. 
 Throughout our experiments, the average image quality for the different restricted ImageNet models appears to be about equal and there is no model that clearly outperforms the others. 

\subsection{Feature Generation on OOD images}
In Figure \ref{Fig:CIFAR_OD} we showed examples of class specific feature generation where we started from samples of the 80 million tiny images dataset not seen during training and maximized the confidence in the predicted class on the original images within $l_2$-balls of different sizes. We omitted some models there because their results are either trivial, in the sense that they do not generate visible features (Plain and OE), or are qualitatively similar to other models that we presented (M$_{0.5}$, AT$_{0.25}$, RATIO$_{0.5}$). For completeness we show the missing models on the same starting points as in the main text in Figure \ref{fig:cifar_od_ext} and Figure \ref{fig:svhn_od_ext} respectively. We also give additional examples of feature generation in Figures \ref{fig:od_cifar_new1} through \ref{fig:od_cifar_new2} for CIFAR10 as well as Figures \ref{fig:od_svhn_new1} through \ref{fig:od_svhn_new2} for SVHN.

Note that as more class specific features appear, the confidences of RATIO and ACET increase much more gradually than for the other models. In particular on the original images they have very low confidence if class-specific features are not already present. AT$_{0.5}$ and JEM-0 have both very little robustness on the out-distribution and therefore make overconfident predictions long before visible class features emerge. To make this observation more quantitative we provided Figure \ref{Fig:CIFAR_ID_OD_CONF} in the main paper. It shows that the mean confidence on worst-case OOD samples is only high at relatively large radii for RATIO and ACET, where clearly real features can already be present, but for Plain, OE, AT and JEM-0 it is already high for a much
smaller radius where class-specific features on average have not appeared yet.

The qualitiative appearance for CIFAR10 is similar to the visual counterfactuals. JEM-0 produces some class-specific features but also a lot of artefacts and in general the generated images look worse than the ones of ACET, AT and RATIO. For SVHN the picture is similar even though JEM-0 is even more overconfident on 
the original images. However, here the generated images for RATIO$_{0.25}$ seem qualitatively better than all other methods as they have fewer artefacts and 
stronger class-specific features. Note that in two cases the targeted attack on the ACET model does not find any images of high confidence. This is fine if indeed
no class-specific features can be generated within the threat model for the class predicted for the original image.

Additional OD samples for CIFAR100 can be found in Figures \ref{fig:od_cifar100_new1} and \ref{fig:od_cifar100_new2} and Figures \ref{fig:od_imagenet_new1} and \ref{fig:od_imagenet_new2} contain restricted ImageNet samples for $l_2$ radii from 3.5 to 21. 
For CIFAR100, all models are mostly able to generate class specific features for the larger radii, however even after calibration, AT remains overconfident and assigns high confidences for images without recognizable features. Both RATIO models on the other hand mostly assign a low confidence to OD images and smoothly increase their confidence with the appearance of class structures. 

The OD behaviour of our restricted ImageNet models is also similar to that of the other datasets. While the quality of the final images is rather consistent across models, the AT models are again overconfident whereas the RATIO models assign noticeable lower confidences to the out-distribution images. 

\subsection{Failure Cases}\label{subsec:failure}
While RATIO manages to improve on the state-of-the-art in several ways, RATIO is not yet a perfect model and we want to illustrate some failure cases. We present several failures in Figures \ref{fig:vc_cifar_failure1} through \ref{fig:od_svhn_fail2}. First, it can happen that the counterfactual explanations are not very informative to a human observer. On CIFAR10 we observed this problem in distinguishing the ``car'' and ``truck'' classes. When performing feature generation even with RATIO it is still possible that the confidence in the targeted class rises much more sharply than is desirable or worse, high confidence is achieved even when nearly no visible class-specific features are present. However, we stress that the failures are common to all the models that we tested and thus present challenges for future work.

On SVHN many of our failures to generate plausible counterfactuals have to do with the test data being either clearly mislabeled or unrecognizable. Even though it is impossible to properly inpaint the target number with the given budget, RATIO sometimes still makes high confidence predictions even though these cases are rare.

On Cifar100 most failures occur for one of the four classes "girl", "woman", "boy" or "man", as can be seen in the failure examples in Figures \ref{fig:vc_cifar100_failure1} to \ref{fig:od_cifar100_fail1}. Despite the fact that human anatomy is complex and our visual system being especially observant towards images containing humans, we believe that for RATIO the failure cases are also caused by the fact that our out-distribution dataset during training contains a significant number of humans. Thus, the enforcement of uniform confidence in  a neighbourhood around those datapoints could result in worse generative performance for the related classes.

Figures \ref{fig:vc_imagenet_failure} to \ref{fig:od_imagenet_failure} show some failures for the restricted ImageNet models. While the models are mostly able to generate class specific features, they are often layered on top of the original image instead of cleverly manipulating existing image structures. This is likely caused by the higher dimensional image space in combination with our simple $l_2$ threat model, which promotes many small changes instead of few sparse ones. This can also cause images that show multiple individual structures, for example different dog heads, instead of one connected structure. Overall we could, however, barely find any failure cases that do not show any class features, even though all models are clearly overconfident in certain situations. They assign extremely high confidence values to transparent class structures layered on top of an existing image or other barely visible class features. We believe that to achieve an even better quality on high resolution images, it might be necessary to replace the simple $l_2$ based threat model with a more localized one.
  
\input{res/appendix_vc_cifar_main_paper_extended}
\input{res/appendix_vc_cifar_new}
\input{res/appendix_cifar_overview}

\input{res/appendix_od_cifar_main_paper_extended}

\input{res/appendix_od_cifar_new}

\input{res/appendix_vc_svhn_main_paper_extended}
\input{res/appendix_vc_svhn_new}
\input{res/appendix_svhn_overview}

\input{res/appendix_od_svhn_main_paper_extended}
\input{res/appendix_od_svhn_new}

\input{res/appendix_vc_cifar100_new}
\input{res/appendix_cifar100_overview}
\input{res/appendix_od_cifar100_new}

\input{res/appendix_vc_restricted_new}
\input{res/appendix_restricted_overview}
\input{res/appendix_od_restricted_new}

\input{res/appendix_failure_vc_cifar10.tex}
\input{res/appendix_failure_od_cifar10.tex}
\input{res/appendix_failure_vc_svhn.tex}
\input{res/appendix_failure_od_svhn.tex}

\input{res/appendix_failure_vc_cifar100.tex}
\input{res/appendix_failure_od_cifar100.tex}

\input{res/appendix_failure_vc_restricted.tex}
\input{res/appendix_failure_od_restricted.tex}

\end{document}

%% file: math_commands.tex
\usepackage{amsmath,amsfonts,bm}

\def\eqref#1{equation~\ref{#1}}

\def\1{\bm{1}}

\newcommand{\test}{\mathcal{D_{\mathrm{test}}}}

\DeclareMathAlphabet{\mathsfit}{\encodingdefault}{\sfdefault}{m}{sl}
\SetMathAlphabet{\mathsfit}{bold}{\encodingdefault}{\sfdefault}{bx}{n}

\newcommand{\R}{\mathbb{R}}

\DeclareMathOperator*{\argmax}{arg\,max}

\def\R{\mathbb{R}}

\newcommand{\norm}[1]{\left\|#1\right\|}

\def\Exp{\mathbb{E}}

\def\ones{\mathop{\rm e}\nolimits}

\def\argmax{\mathop{\rm arg\,max}\limits}
\def\minop{\mathop{\rm min}\limits}
\def\maxop{\mathop{\rm max}\limits}

\def\max{\mathop{\rm max}\nolimits}
\def\ones{\mathbf{1}}

%% file: res/summary.tex
\newcommand{\firstcolumnS}{20mm}
\newcommand{\cifarS}{8mm}
\newcommand{\svhnS}{11.116mm}
\newcolumntype{C}[1]{>{\centering\arraybackslash}p{#1}}

\begin{table}[t]
\setlength{\tabcolsep}{4.3pt}
\caption{\label{Tab:Summary_cifar10} \textit{Summary:} 
We show clean and robust accuracy in an $l_2$-threat model with $\epsilon=0.5$ and the expected calibration error (ECE). For OOD detection we report the mean of clean and worst case AUC over several out-distributions in an $l_2$-threat model with $\epsilon=1.0$ as well as the mean maximal confidence (MMC) on the out-distributions. In light red we highlight failure cases for certain metrics. Only RATIO-0.25 (R$_{0.25}$) has good performance across all metrics.}
\begin{tabular}{p{\firstcolumnS}|ccccccccc}
\hline
\textbf{CIFAR10} & Plain &  OE & ACET  & M$_{0.5}$ &  AT$_{0.5}$ & AT$_{0.25}$ & JEM-0   &  R$_{0.5}$ & \textbf{R$_{0.25}$} \\
\hline
Acc. $\uparrow$                &  96.2 & \textbf{96.4}  & 94.1 &  \cellcolor{tablegray} 90.8 &   \cellcolor{tablegray} 90.8 &   94.0 & 92.8 & \cellcolor{tablegray} 91.1  & 93.5   \\
R. Acc.$_{0.5}$ $\uparrow$     &      \cellcolor{tablegray}   0.0   & \cellcolor{tablegray} 0.0   & \cellcolor{tablegray} 52.3 &  69.3 & 70.4 &   65.0 & \cellcolor{tablegray} 40.5 &  \textbf{73.3}  &  70.5  \\
ECE  (in \%)  $\downarrow$            &  \textbf{1.0} &  2.9 &    2.8 &  2.6 &    2.2 &  2.2 &             3.9 &             2.8 &   2.7 \\
\hline
AUC $\uparrow$                 &  94.2 &   \textbf{96.5} &  94.7 & \cellcolor{tablegray} 81.8 & \cellcolor{tablegray} 88.9 &  92.7 & \cellcolor{tablegray} 75.0 &  95.6 &  95.0 \\
WC AUC$_{1.0}$ $\uparrow$      & \cellcolor{tablegray} 1.6 &  \cellcolor{tablegray}  8.7 &  81.9 & \cellcolor{tablegray} 48.5 & \cellcolor{tablegray} 57.4 & \cellcolor{tablegray} 42.0 & \cellcolor{tablegray} 14.6 &  83.6 &  \textbf{84.3} \\
MMC $\downarrow$               & \cellcolor{tablegray} 62.0 &  \textbf{31.9} &  39.1 & \cellcolor{tablegray} 62.7 & \cellcolor{tablegray} 55.8 & \cellcolor{tablegray}  55.2 & \cellcolor{tablegray} 69.7 &  \textbf{31.9} &  33.9 \\
\hline
\end{tabular}

\begin{tabular}{p{\firstcolumnS}|C{\svhnS}C{\svhnS}C{\svhnS}C{\svhnS}C{\svhnS}C{\svhnS}C{\svhnS}C{\svhnS}}
\hline
\textbf{SVHN} & Plain & OE & ACET & AT$_{0.5}$ & AT$_{0.25}$ &  R$_{0.5}$ & \textbf{R$_{0.25}$} \\
\hline
Acc. $\uparrow$                & 97.3 & 97.6 & \textbf{97.8} & \cellcolor{tablegray} 94.4 & 96.7 & \cellcolor{tablegray} 94.3 & 96.8 \\
R. Acc.$_{0.5}$ $\uparrow$     & \cellcolor{tablegray} 0.9 & \cellcolor{tablegray} 0.3 & \cellcolor{tablegray} 28.8 &  68.1 & 63.0 & \textbf{68.4} & 64.8  \\
ECE    $\downarrow$            &  0.9 & 0.9  & 1.6  & 1.6   & \textbf{0.8}  & 2.0  & 1.8 \\
\hline
AUC $\uparrow$                 & 96.9 & 99.6 & 99.8 & \cellcolor{tablegray} 91.0 & 97.0 & 99.8 & \textbf{99.9} \\
WC AUC$_{1.0}$ $\uparrow$      & \cellcolor{tablegray} 8.5  & \cellcolor{tablegray} 18.2 & 96.0 & \cellcolor{tablegray} 51.1 & \cellcolor{tablegray} 48.3 & \textbf{97.5} & \textbf{97.5} \\
MMC $\downarrow$               & \cellcolor{tablegray} 61.5 & 16.3 & 11.8 & \cellcolor{tablegray} 67.1 & \cellcolor{tablegray} 49.1 & 12.1 & \textbf{11.1} \\
\hline
\end{tabular}

\end{table}

%% file: res/summary2.tex
\newcommand{\imagenetS}{9.35mm}
\begin{table}[t]
\setlength{\tabcolsep}{4.3pt}
\renewcommand{\arraystretch}{1.0}
\caption{\label{Tab:Summary_cifar100} Summary for CIFAR100 and R. ImageNet (see Table \ref{Tab:Summary_cifar10} for details).}

\begin{tabular}{p{\firstcolumnS}|C{\svhnS}C{\svhnS}C{\svhnS}C{\svhnS}C{\svhnS}C{\svhnS}C{\svhnS}C{\svhnS}}
\hline
\textbf{CIFAR100}              & Plain & OE   & ACET & AT$_{0.5}$ & AT$_{0.25}$ &  R$_{0.5}$ & \textbf{R$_{0.25}$} \\
\hline
Acc. $\uparrow$                &  \textbf{81.5} & 81.4 &   -   &  \cellcolor{tablegray}70.6      & \cellcolor{tablegray}75.8       & \cellcolor{tablegray}69.2      & \cellcolor{tablegray}74.4  \\
R. Acc.$_{0.5}$ $\uparrow$     &  \cellcolor{tablegray}0.0  & \cellcolor{tablegray}0.0  &   -   &  43.2      & 37.3       & \textbf{45.6}      & 42.4  \\
ECE    $\downarrow$            &  \textbf{1.2}  & \cellcolor{tablegray} 7.2  &  -   &  1.3       & 1.5        & 3.2       & 2.0   \\
\hline
AUC $\uparrow$                 & 84.0   & \textbf{91.9}  &   -  & \cellcolor{tablegray} 75.6        & \cellcolor{tablegray} 79.4       & 87.0      & 86.9   \\
WC AUC$_{1.0}$ $\uparrow$      & \cellcolor{tablegray} 0.4    & \cellcolor{tablegray} 14.6  & -   & \cellcolor{tablegray} 29.9       & \cellcolor{tablegray} 24.8        & \textbf{55.5}       & 54.5   \\
MMC $\downarrow$               & \cellcolor{tablegray} 51.1  & 21.8  &  -  & \cellcolor{tablegray} 45.8       & \cellcolor{tablegray} 47.1        & 24.4       & 31.0   \\
\hline
\end{tabular}

\begin{tabular}{p{\firstcolumnS}|C{\imagenetS}C{\imagenetS}C{\imagenetS}C{\imagenetS}C{\imagenetS}C{\imagenetS}C{\imagenetS}C{\imagenetS}C{\imagenetS}}
\hline
\textbf{R.Imagenet}              & Plain & OE    & ACET &  M$_{3.5}$ & AT$_{3.5}$ & AT$_{1.75}$ &  R$_{3.5}$ & \textbf{R$_{1.75}$} \\
\hline
Acc. $\uparrow$                &  96.6   & \textbf{97.2}  & 96.2 & \cellcolor{tablegray}  90.3     & \cellcolor{tablegray} 93.5       & 95.5        & \cellcolor{tablegray} 93.9       & 95.5  \\
R. Acc.$_{3.5}$ $\uparrow$     & \cellcolor{tablegray} 0.0    & \cellcolor{tablegray} 0.0   & \cellcolor{tablegray} 6.2  &   47.7     & 47.7       & 36.7        & \textbf{49.2}       & 43.0  \\
ECE    $\downarrow$            &  0.6    & 1.8   & 0.9  &   0.7      & 0.9        & 0.5         & \textbf{0.3}        & 0.7   \\
\hline
AUC $\uparrow$                 &  92.7   & \textbf{98.9}  & 97.74 &  \cellcolor{tablegray}  83.6    & \cellcolor{tablegray} 84.3       & \cellcolor{tablegray} 86.5        & 97.2   & 97.8 \\
WC AUC$_{7.0}$ $\uparrow$      & \cellcolor{tablegray} 0.0    & \cellcolor{tablegray} 1.8   & 87.54 &  \cellcolor{tablegray}  44.2    & \cellcolor{tablegray} 37.5       & \cellcolor{tablegray} 16.3        & \textbf{90.9}       & 90.6  \\
MMC $\downarrow$               & \cellcolor{tablegray} 67.9   & \textbf{20.6}  & 34.85 &   \cellcolor{tablegray} 69.2    &  \cellcolor{tablegray} 75.2       & \cellcolor{tablegray} 81.8        & 33.6       & 32.3  \\
\hline
\end{tabular}
\end{table}

%% file: res/counterfactuals_plot_cifar.tex
\begin{figure}[ht]
\begin{tabular}{p{1cm}x{\breite}x{\breite}x{\breite}x{\breite}x{\breite}x{\breite}x{\breite}x{\breite}}
Model  & Orig. & $\epsilon=0.5$ & $\epsilon=1.0$ & $\epsilon=1.5$ & $\epsilon=2.0$ & $\epsilon=2.5$ & $\epsilon=3.0$\\
\begin{turn}{90} \hspace{-.4cm} ACET \end{turn} & \multicolumn{7}{c}{\includegraphics[width=0.91\textwidth,valign=c]{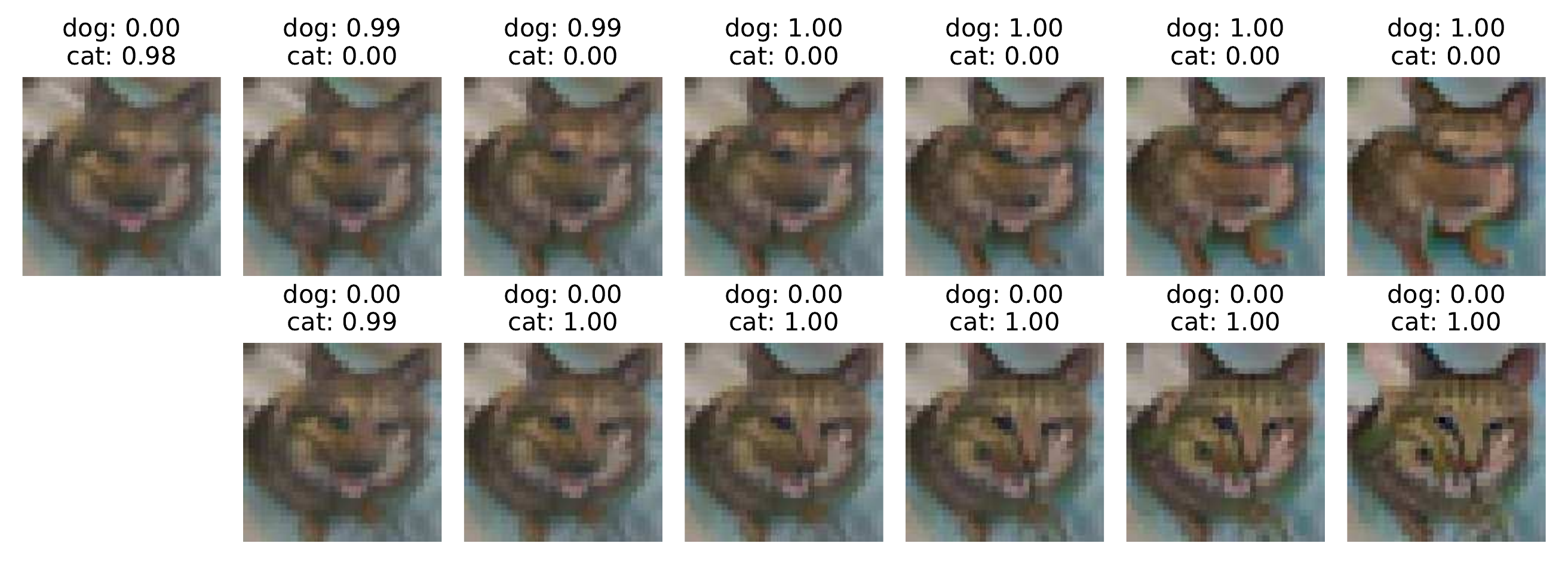}} \\
\hline
\begin{turn}{90} \hspace{-.35cm}  JEM-0 \end{turn}  &  \multicolumn{7}{c}{\includegraphics[width=0.91\textwidth,valign=c]{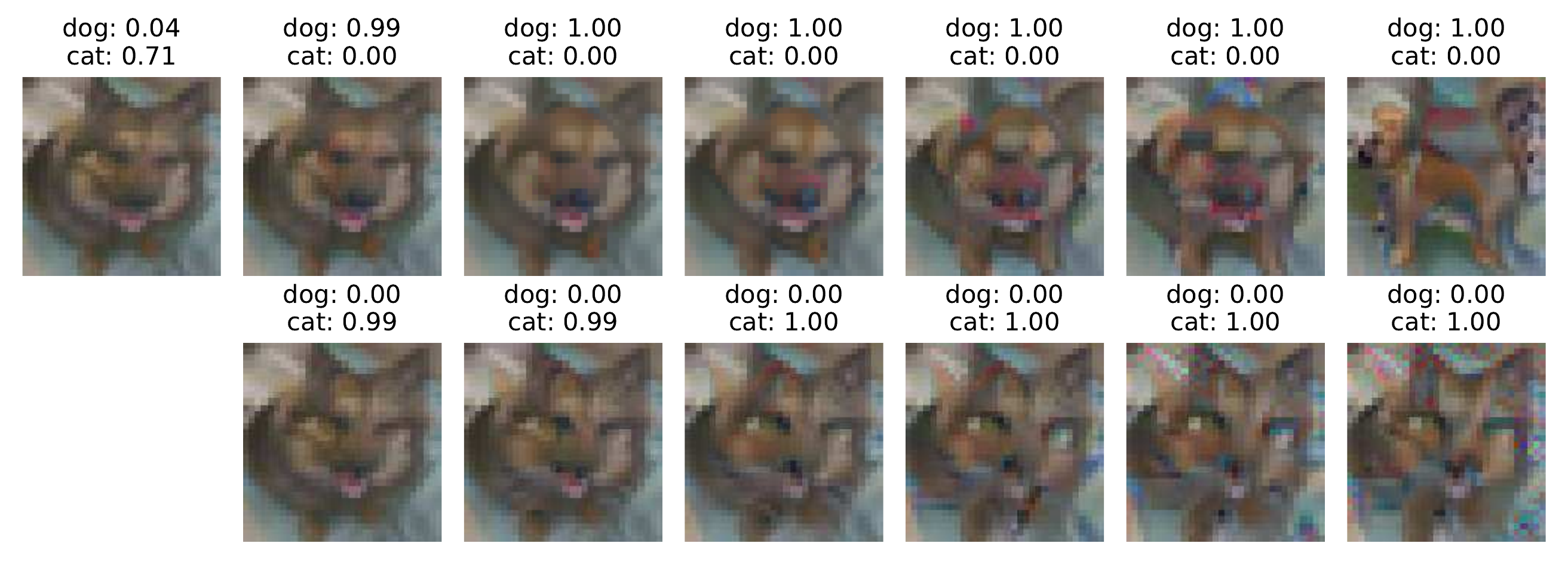}} \\
\hline
\begin{turn}{90} \hspace{-.5cm} AT-0.50 \end{turn}  &  \multicolumn{7}{c}{\includegraphics[width=0.91\textwidth,valign=c]{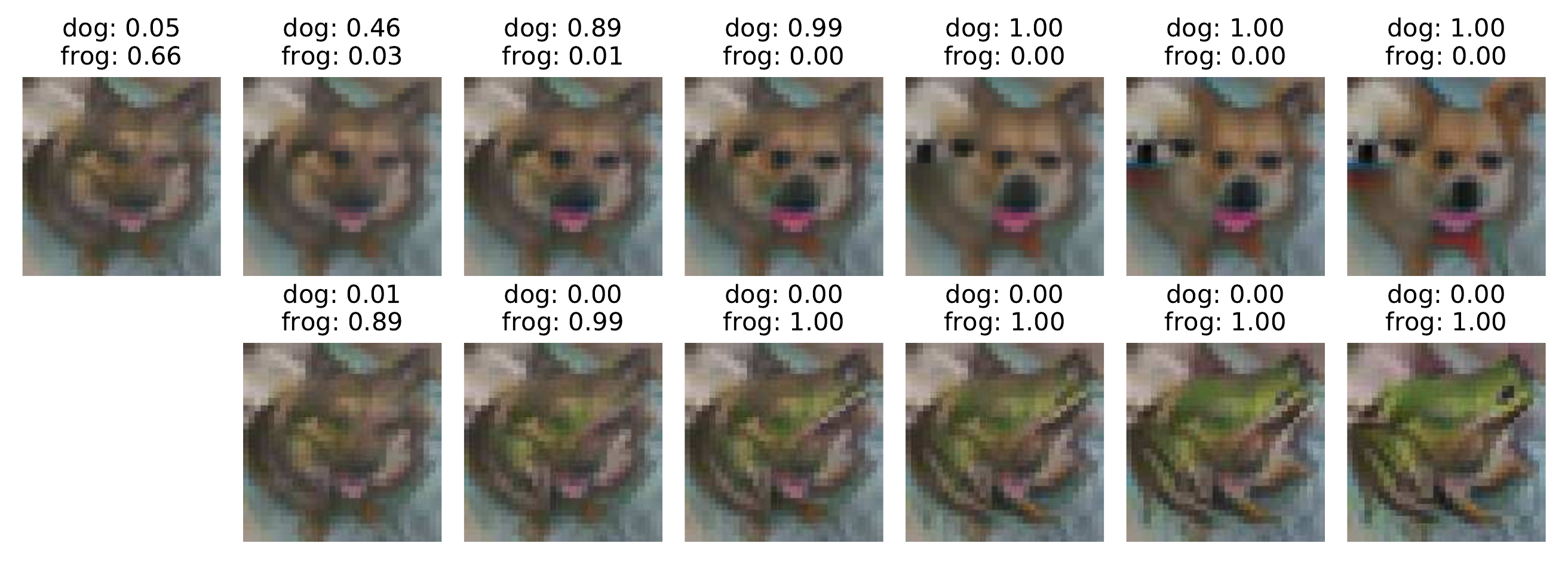}} \\
\hline
\begin{turn}{90} \hspace{-.8cm} RATIO-0.25 \end{turn}  &  \multicolumn{7}{c}{\includegraphics[width=0.91\textwidth,valign=c]{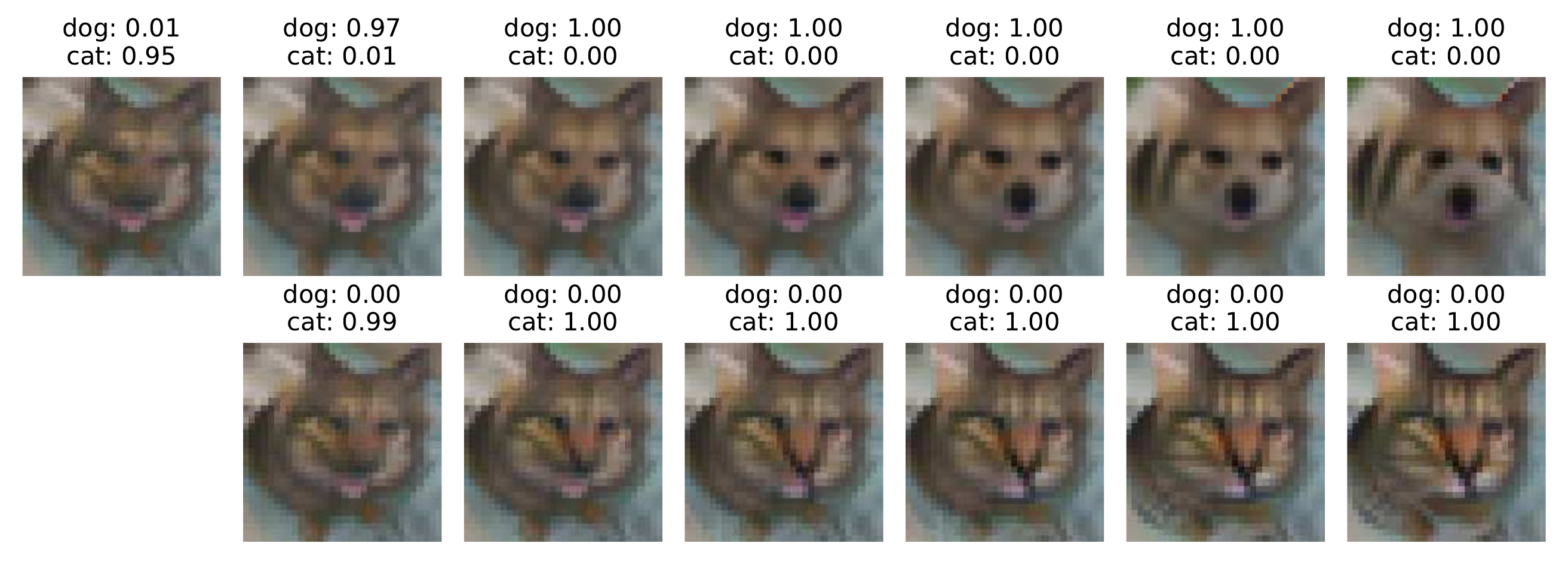}} \\
\end{tabular}	
\caption{\label{fig:cf}\textbf{Visual Counterfactuals (CIFAR10):} The dog image on the left is misclassified by all models 
(confidence for true and predicted class are shown). The top row shows visual counterfactuals for the correct class (how to change the image so that it is classified as dog) and the bottom row shows how to change the image in order to increase the confidence in the wrong prediction for different budgets of the $l_2$-radius ($\epsilon=0.5$ to $\epsilon=3$). More examples are in the appendix.}
\end{figure}
\clearpage

%% file: res/counterfactuals_plot_svhn.tex
\begin{figure}[t]
\begin{tabular}{p{1cm}x{\breite}x{\breite}x{\breite}x{\breite}x{\breite}x{\breite}x{\breite}x{\breite}}
Model  & Orig. & $\epsilon=0.5$ & $\epsilon=1.0$ & $\epsilon=1.5$ & $\epsilon=2.0$ & $\epsilon=2.5$ & $\epsilon=3.0$\\
\begin{turn}{90} \hspace{-.4cm} AT-0.25 \end{turn}  &  \multicolumn{7}{c}{\includegraphics[width=0.91\textwidth,valign=c]{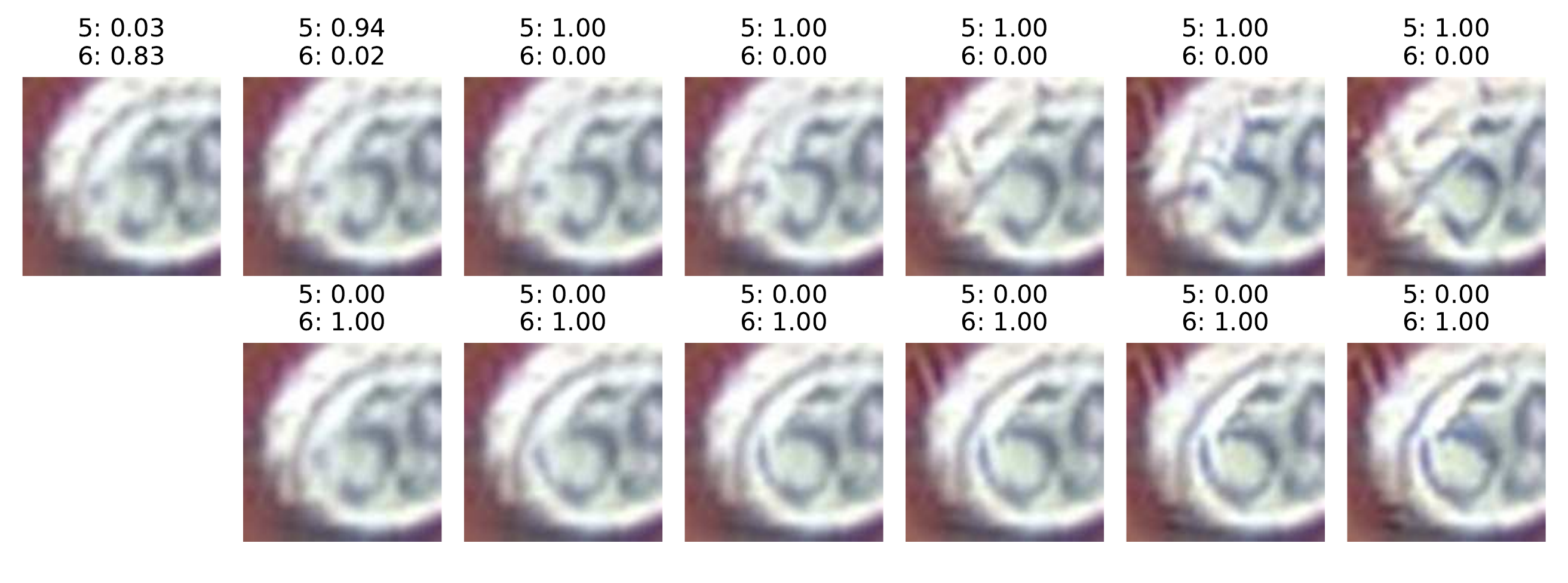}} \\
\hline
\begin{turn}{90} \hspace{-.8cm} RATIO-0.25 \end{turn} & \multicolumn{7}{c}{\includegraphics[width=0.91\textwidth,valign=c]{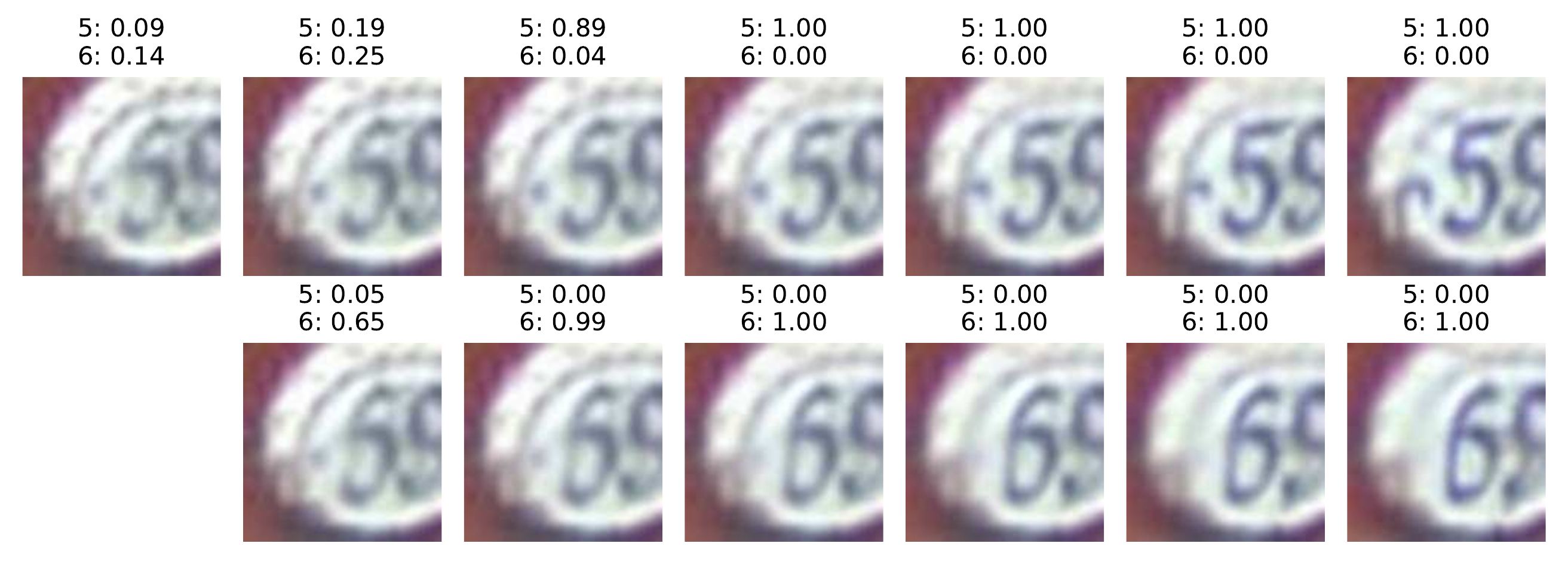}} \\
\end{tabular}	
\caption{\label{fig:cf_svhn}\textbf{Visual Counterfactuals (SVHN):} The 5 on the left is misclassified by all models. We show counterfactuals for the true class the predicted class (see Figure \ref{fig:cf}). RATIO consistently produces samples with fewer artefacts than AT.}
\end{figure}

%% file: res/imagenet_plots_in.tex
\begin{figure}[t]
\begin{tabular}{p{1cm}x{\breite}x{\breite}x{\breite}x{\breite}x{\breite}x{\breite}x{\breite}x{\breite}}
Model  & Orig. & $\epsilon=0.5$ & $\epsilon=1.0$ & $\epsilon=1.5$ & $\epsilon=2.0$ & $\epsilon=2.5$ & $\epsilon=3.0$\\
\begin{turn}{90} \hspace{-.8cm} RATIO-0.25 \end{turn}  &  \multicolumn{7}{c}{\includegraphics[width=0.91\textwidth,valign=c]{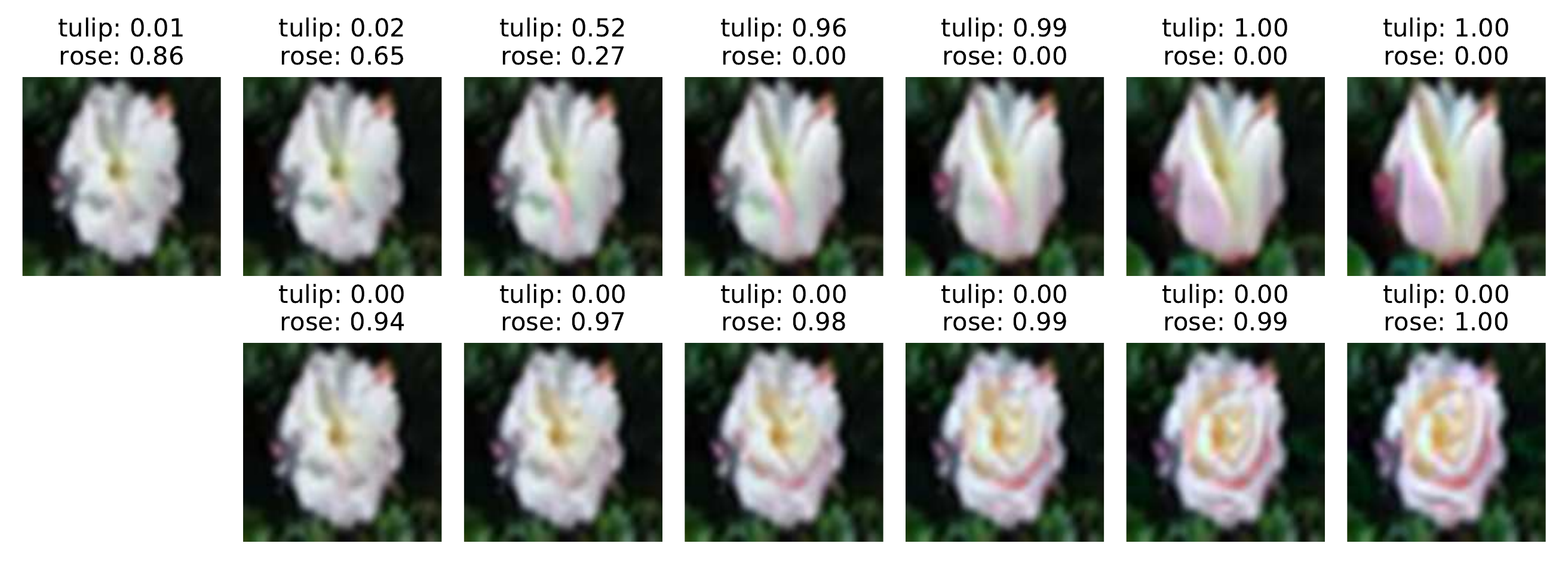}} \\
\hline
\end{tabular}
\begin{tabular}{p{1cm}x{\breite}x{\breite}x{\breite}x{\breite}x{\breite}x{\breite}x{\breite}x{\breite}}
Model  & Orig. & $\epsilon=3.5$ & $\epsilon=7.0$ & $\epsilon=10.5$ & $\epsilon=14.0$ & $\epsilon=17.5$ & $\epsilon=21.0$\\
\begin{turn}{90} \hspace{-.8cm} RATIO-1.75 \end{turn} & \multicolumn{7}{c}{\includegraphics[width=0.91\textwidth,valign=c]{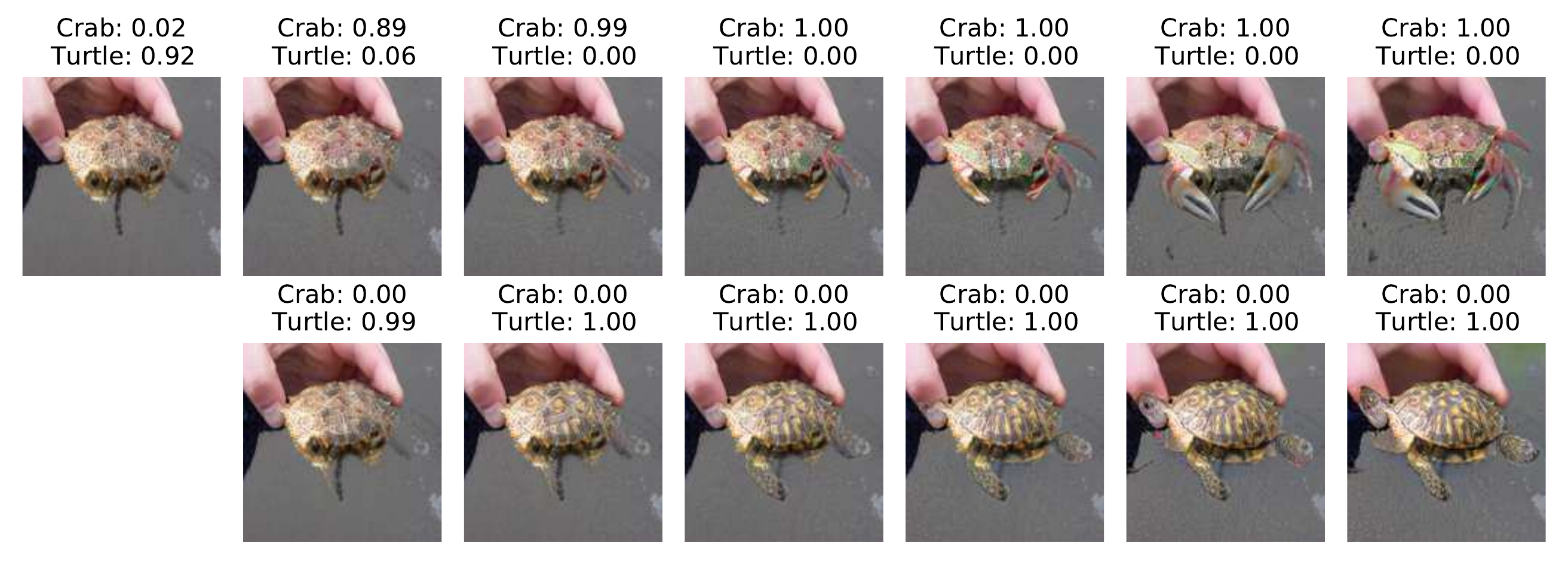}} \\
\end{tabular}	
\caption{\label{fig:cf_imagenet}\textbf{Visual Counterfactuals} top: RATIO-0.25 for CIFAR100 and bottom: RATIO-1.75 for RestrictedImageNet.}
\end{figure}

%% file: res/imagenet_plots_out.tex
\begin{figure}[h]
\begin{tabular}
{p{1cm}x{\breite}x{\breite}x{\breite}x{\breite}x{\breite}x{\breite}x{\breite}x{\breite}}
Model  & Orig. & $\epsilon=0.5$ & $\epsilon=1.0$ & $\epsilon=1.5$ & $\epsilon=2.0$ & $\epsilon=2.5$ & $\epsilon=3.0$\\
\begin{turn}{90} \hspace{-.4cm} R-0.25 \end{turn} & \multicolumn{7}{c}{\includegraphics[width=0.91\textwidth,valign=c]{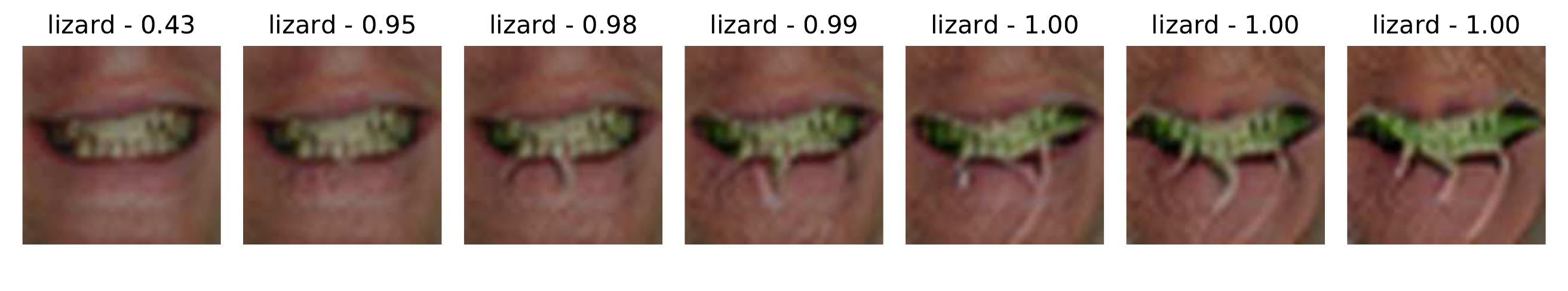}} \\
\hline
Model  & Orig. & $\epsilon=3.5$ & $\epsilon=7.0$ & $\epsilon=10.5$ & $\epsilon=14.0$ & $\epsilon=17.5$ & $\epsilon=21.0$\\
\begin{turn}{90} \hspace{-.4cm} R-1.75 \end{turn} & \multicolumn{7}{c}{\includegraphics[width=0.91\textwidth,valign=c]{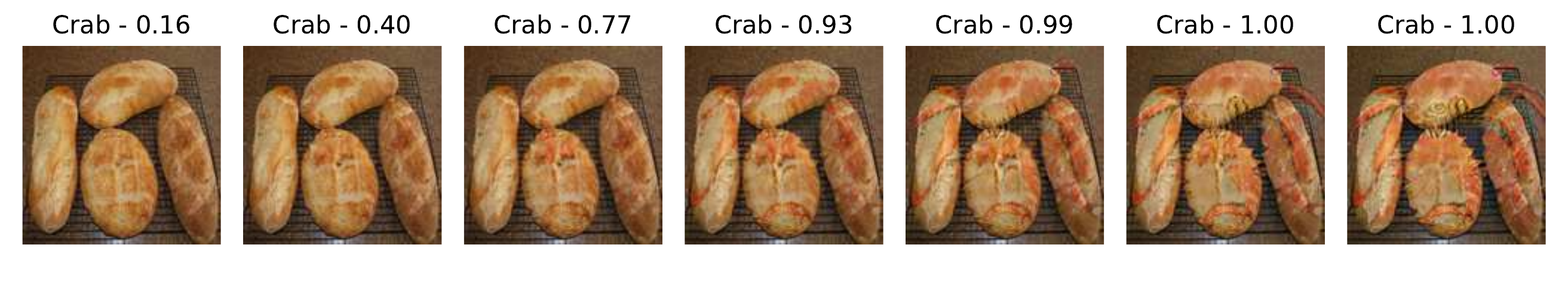}} \\
\end{tabular}	
\caption{\label{Fig:CIFAR100_OD}\textbf{Feature Generation for out-distribution images} top: RATIO-0.25 for CIFAR100 and bottom: RATIO-1.75 for R.ImageNet}
\end{figure}

%% file: res/od_plots.tex
\begin{figure}
\begin{tabular}{p{1cm}x{\breite}x{\breite}x{\breite}x{\breite}x{\breite}x{\breite}x{\breite}x{\breite}}
Model  & Orig. & $\epsilon=0.5$ & $\epsilon=1.0$ & $\epsilon=1.5$ & $\epsilon=2.0$ & $\epsilon=2.5$ & $\epsilon=3.0$\\
\begin{turn}{90} \hspace{-.4cm} ACET \end{turn} & \multicolumn{7}{c}{
\includegraphics[width=0.91\textwidth,valign=c]{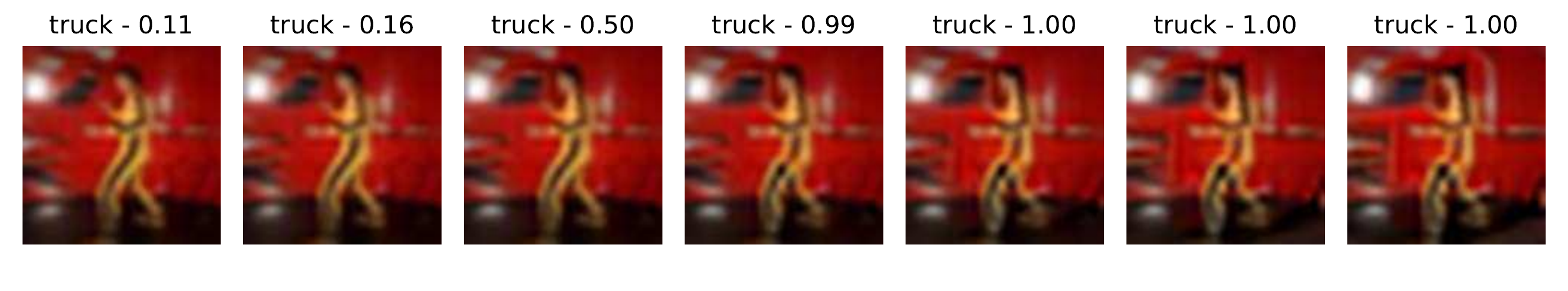}
} \\
\hline
\begin{turn}{90} \hspace{-.4cm} JEM-0 \end{turn} & \multicolumn{7}{c}{
\includegraphics[width=0.91\textwidth,valign=c]{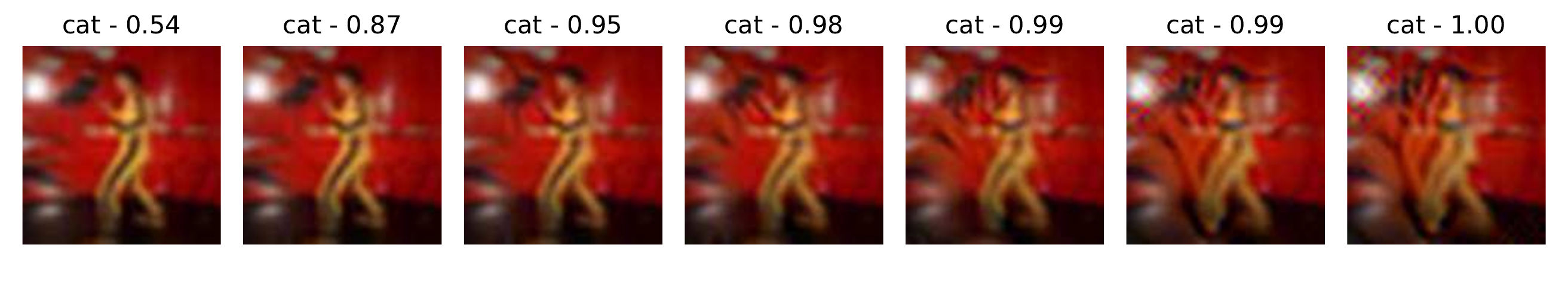}
} \\
\hline
\begin{turn}{90} \hspace{-.4cm} AT-0.50 \end{turn} & \multicolumn{7}{c}{
\includegraphics[width=0.91\textwidth,valign=c]{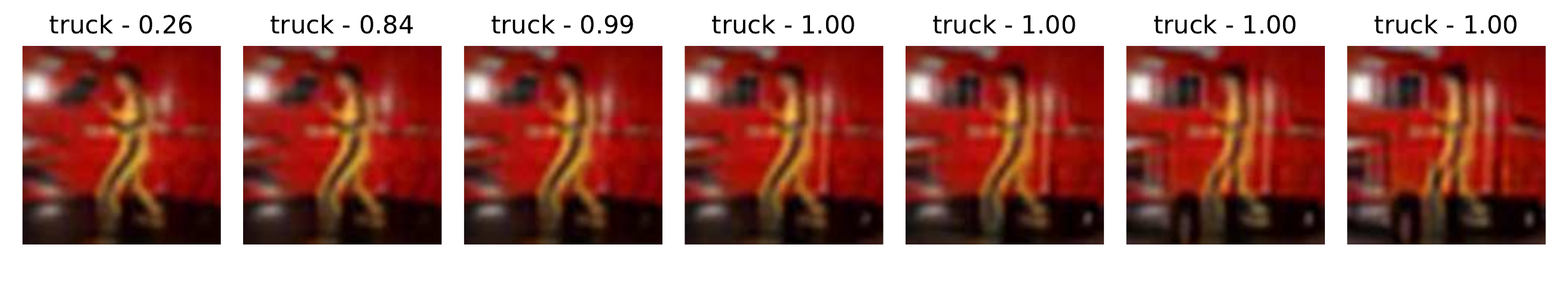}
} \\
\hline
\begin{turn}{90} \hspace{-.4cm} R-0.25 \end{turn} & \multicolumn{7}{c}{
\includegraphics[width=0.91\textwidth,valign=c]{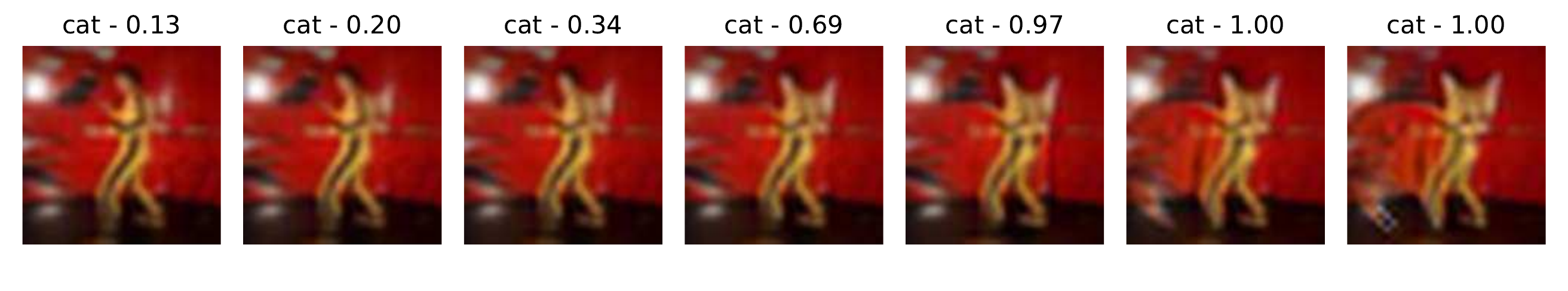}
} \\
\hline
\hline
\begin{turn}{90} \hspace{-.4cm} ACET \end{turn} & \multicolumn{7}{c}{
\includegraphics[width=0.91\textwidth,valign=c]{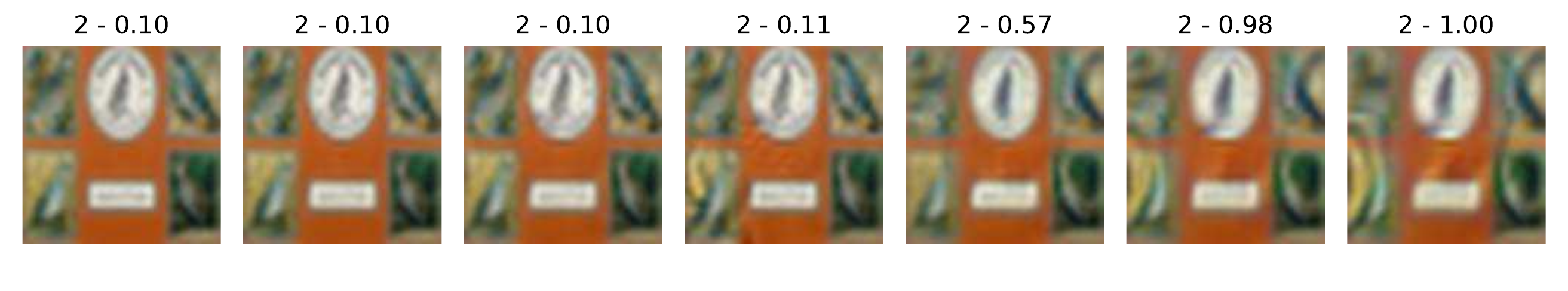}
} \\
\hline
\begin{turn}{90} \hspace{-.4cm} AT-0.25 \end{turn} & \multicolumn{7}{c}{
\includegraphics[width=0.91\textwidth,valign=c]{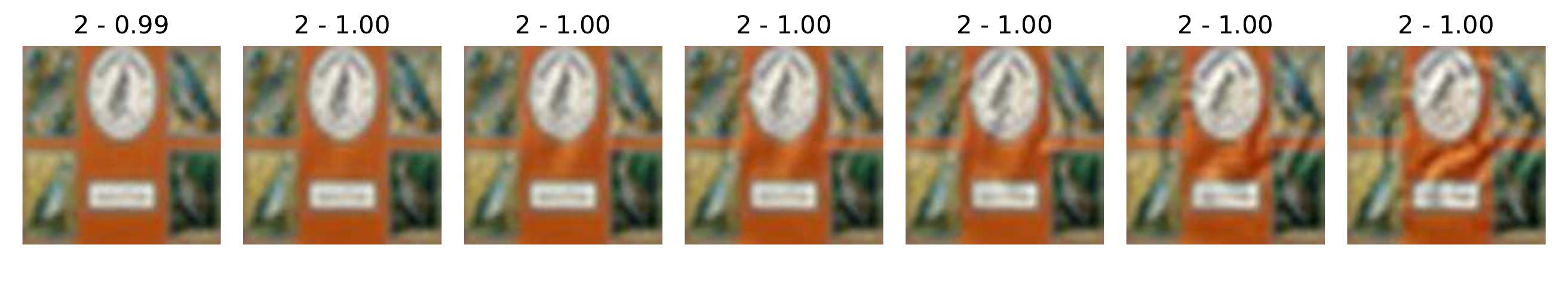}
} \\
\hline
\begin{turn}{90} \hspace{-.4cm} R-0.25 \end{turn} & \multicolumn{7}{c}{
\includegraphics[width=0.91\textwidth,valign=c]{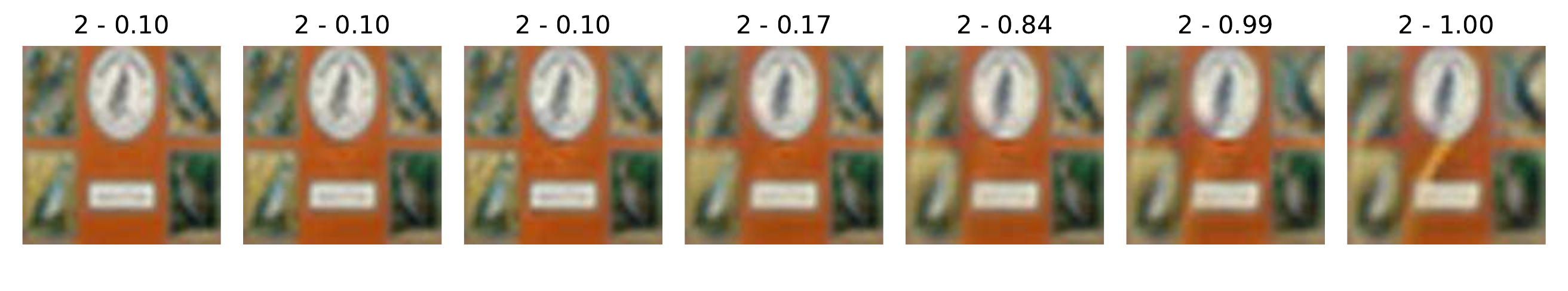}
} \\
\hline
\end{tabular}	
\caption{\textbf{Feature Generation for out-distribution images (CIFAR10 (top), SVHN (bottom)):} targeted attacks towards the class achieving highest
confidence on original image for different budgets of the $l_2$-radius ranging from $\epsilon=0.5$ to $\epsilon=3$. RATIO-0.25 generates the visually
best images and in particular has reasonable confidence values for its decision. While AT-0.5/AT-0.25 generates also good images it is overconfident into
the target class.
\label{Fig:CIFAR_OD}}
\end{figure}

%% file: res/cifar10_id.tex
\newcommand{\firstcolumnID}{14mm}
\newcommand{\cifarID}{7.95mm}
\newcommand{\svhnID}{11.517mm}
\newcommand{\imagenetID}{9.641mm}
\newcolumntype{C}[1]{>{\centering\arraybackslash}p{#1}}
\begin{table}[t]
\begin{center}
\setlength{\tabcolsep}{5pt}
\renewcommand{\arraystretch}{1.0}
\caption{\label{Tab:IDRobustness_cifar10} \textit{ID performance:} Evaluation of the classification accuracy and adversarial robustness under various threat models. }
\begin{tabular}{p{\firstcolumnID}|C{\cifarID}C{\cifarID}C{\cifarID}C{\cifarID}C{\cifarID}C{\cifarID}C{10mm}C{\cifarID}C{\cifarID}}
\hline
CIFAR10 & Plain &  OE  & ACET  & M$_{0.5}$ &  AT$_{0.5}$ & AT$_{0.25}$ & JEM-0 &  R$_{0.5}$ & R$_{0.25}$ \\
\hline
Clean                   &  96.15 & \textbf{96.41}  & 94.06  & 90.83 &   90.79 & 94.00 & 92.84  &  91.08  & 93.53   \\
$l_2$ $0.5$             &  0.00   & 0.01   & 52.66 &  69.32 & 70.42 &   65.02 & 40.67 &  \textbf{73.27}  &  70.49  \\
$l_2$ $1.0$             &  0.00   & 0.00   & 14.95 &  36.86 & 38.20 &   20.83 &  6.77 &  \textbf{44.09}  &  34.15  \\
$l_{\infty}$ $8/255$    &  0.00   & 0.00   & 12.66 &  29.59 & 30.58 &   14.30 &  7.19 &  \textbf{37.71}  &  29.18  \\
\hline
\end{tabular}

\renewcommand{\arraystretch}{1.0}
\begin{tabular}{p{\firstcolumnID}|C{\svhnID}C{\svhnID}C{\svhnID}C{\svhnID}C{\svhnID}C{\svhnID}C{\svhnID}C{\svhnID}}
\hline
SVHN & Plain & OE & ACET & AT$_{0.5}$ & AT$_{0.25}$ &  R$_{0.5}$ & R$_{0.25}$ \\
\hline
Clean                &  97.33 & 97.56 & \textbf{97.77} &  94.38 & 96.70 & 94.30 & 96.80  \\
$l_2$ $0.5$          &   0.85 &  0.33 & 28.82 &  68.58 & 63.18 & \textbf{68.85} & 65.02  \\
$l_2$ $1.0$          &   0.01 &  0.00 & 4.20  &  26.40 & 17.19 & \textbf{27.06} & 20.30  \\
$l_{\infty}$ $8/255$ &   0.00 &  0.00 & 4.03  &  27.56 & 19.05 & \textbf{28.20} & 22.14  \\
\hline
\end{tabular}

\renewcommand{\arraystretch}{1.0}
\begin{tabular}{p{\firstcolumnID}|C{\svhnID}C{\svhnID}C{\svhnID}C{\svhnID}C{\svhnID}C{\svhnID}C{\svhnID}C{\svhnID}}
\hline
CIFAR100          & Plain  & OE    & ACET & AT$_{0.5}$ & AT$_{0.25}$ &  R$_{0.5}$ & R$_{0.25}$ \\
\hline
Clean                &  \textbf{81.49} & 81.43 &  -    &  70.60     & 75.84       & 69.17      & 74.38  \\
$l_2$ $0.5$          &  0.00  & 0.01  &   -   &  43.23     & 37.26       & \textbf{45.55}      & 42.41  \\
$l_2$ $1.0$          &  0.00  & 0.00  &   -   &  19.14     & 9.45        & \textbf{23.87}      & 16.30  \\
$l_{\infty}$ $8/255$ &  0.00  & 0.00  &  -    &  13.74     & 5.96        & \textbf{18.49}      & 12.39  \\
\hline
\end{tabular}

\renewcommand{\arraystretch}{1.0}
\begin{tabular}{p{\firstcolumnID}|C{\imagenetID}C{\imagenetID}C{\imagenetID}C{\imagenetID}C{\imagenetID}C{\imagenetID}C{\imagenetID}C{\imagenetID}C{\imagenetID}}
\hline
RImagenet              & Plain  & OE    & ACET &  M$_{3.5}$ & AT$_{3.5}$ & AT$_{1.75}$ &  R$_{3.5}$ & R$_{1.75}$ \\
\hline
Clean                &  96.55 & \textbf{97.22} & 96.24 &   90.25    & 93.49      & 95.45       & 93.94      & 95.46      \\
$l_2$ $3.5$          &  00.00 & 00.00          & 6.25 &   47.66    & 47.66      & 36.72       & \textbf{49.22}      & 42.97      \\
$l_2$ $7.0$          &  00.00 & 00.00          & 0.78 & 14.06  & 11.72 & 2.54  &  \textbf{15.43} &      12.70  \\
$l_{\infty}$ $8/255$ &  00.00 & 00.00          & 1.17 &   11.72    & 10.55      &  1.95       & \textbf{17.58}      & 10.55      \\
\hline
\end{tabular}
\end{center}
\end{table}

%% file: res/appendix_trades.tex
\newcommand{\tradesID}{15mm}
\newcolumntype{C}[1]{>{\centering\arraybackslash}p{#1}}
\begin{table}[H]
\begin{center}
\setlength{\tabcolsep}{5pt}
\renewcommand{\arraystretch}{1.0}
\caption{\label{Tab:TRADES} Evaluation of the CIFAR10 TRADES model.  }
\begin{tabular}{p{\firstcolumnID}|C{\svhnID}C{\svhnID}C{\svhnID}C{\svhnID}C{\tradesID}}
\hline
 CIFAR10 & Clean &  $l_2$ 0.5  & $l_2$ 1.0  & AUC &  WC AUC\\
\hline
Reference & 86.22 & 68.81 & 43.80 & 81.81 & 49.86\\
Ours & 90.87 & 71.02 & 39.53 & 88.80 & 56.59\\
\hline
\end{tabular}
\end{center}
\end{table}

%% file: res/cifar10_ood.tex
\newcommand{\firstcolumnOD}{16mm}
\newcommand{\cifarOD}{8mm}
\newcommand{\svhnOD}{11.1385mm}
\newcommand{\imagenetOD}{9.3mm}
\newcolumntype{C}[1]{>{\centering\arraybackslash}p{#1}}
\begin{table}[H]
\setlength{\tabcolsep}{5.1pt}
\renewcommand{\arraystretch}{.96}
\caption{\label{Tab:OOD_Cifar10} \textit{OOD performance (CIFAR10, SVHN, CIFAR100):} The area under the ROC curve (AUC) on the binary task of separating the in- from the out-distribution based on the confidence. For each dataset the first table shows the average-case AUC and the second ones show the worst-case AUC with a threat model $l_2=1.0$ around the out-distribution samples.}
\begin{tabular}{p{\firstcolumnOD}|ccccccccc}
\hline
& \multicolumn{8}{C{8cm}}{\textbf{CIFAR10}} \\
Av. Case & Plain &  OE & ACET & M$_{0.5}$ &  AT$_{0.5}$ & AT$_{0.25}$ & JEM-0 & R$_{0.5}$ & R$_{0.25}$ \\
\hline
SVHN           &              96.8 &  \textbf{99.4} &              93.6 &              91.9 &              93.5 &              95.3 &              89.3 &               96.5 &              96.4 \\
CIFAR100       &     \textbf{91.6} &     91.4 &              90.4 &              84.3 &              85.3 &              89.1 &              87.6 &               90.8 &    \textbf{91.6} \\
LSUN\_CR       &              95.6 &     \textbf{99.6} &              98.2 &              89.7 &              90.7 &              92.5 &              91.6 &               98.0 &              98.3 \\
Imagenet-      &     \textbf{91.6} &     89.8 &              91.0 &              84.8 &              85.9 &              89.0 &              86.7 &               90.5 &              91.3 \\
Noise          &              94.3 &     99.3 &              95.0 &              93.7 &              94.9 &              95.5 &              83.1 &      \textbf{97.8} &              97.6 \\
Uni. Noise     &              95.0 &     99.5 &     \textbf{99.9} &              46.1 &              83.2 &              94.6 &              11.8 &      \textbf{99.9} &     \textbf{99.9} \\
\hline
Worst Case & Plain & OE &  ACET & M$_{0.5}$ &  AT$_{0.5}$ & AT$_{0.25}$ & JEM-0 & R$_{0.5}$ & R$_{0.25}$ \\
\hline
SVHN           &               0.0 &   0.6  &              76.1 &              57.1 &              62.0 &              40.1 &               7.3 &      \textbf{81.3} &              81.1 \\
CIFAR100       &               0.0 &   2.7  &              69.9 &              47.9 &              48.5 &              31.8 &              19.2 &               71.9 &     \textbf{73.0} \\
LSUN\_CR       &               0.0 &   4.0  &              87.9 &              52.0 &              52.8 &              36.5 &              20.6 &               87.3 &     \textbf{89.1} \\
Imagenet-      &               0.0 &   1.5  &              72.8 &              50.6 &              51.1 &              36.8 &              21.2 &               72.4 &     \textbf{73.5} \\
Noise          &               0.0 &   0.0  &              84.5 &              62.9 &              67.9 &              38.8 &              16.5 &               88.9 &     \textbf{89.4} \\
Uni. Noise     &               9.4 &  43.1  &     \textbf{99.9} &              20.7 &              62.0 &              67.8 &               2.5 &               99.8 &             99.8  \\
\hline
\end{tabular}

\begin{tabular}{p{\firstcolumnOD}|C{\svhnOD}C{\svhnOD}C{\svhnOD}C{\svhnOD}C{\svhnOD}C{\svhnOD}C{\svhnOD}C{\svhnOD}}
\hline
& \multicolumn{6}{C{8cm}}{\textbf{SVHN}} \\
Av. Case & Plain & OE & ACET & AT$_{0.5}$ & AT$_{0.25}$ &  R$_{0.5}$ & R$_{0.25}$ \\
\hline
CIFAR10      &  95.8 &  \textbf{100.0} &  \textbf{100.0} &  88.5 &   95.7 &  \textbf{100.0} &  \textbf{100.0} \\
CIFAR100     &  95.6 &  \textbf{100.0} &  \textbf{100.0} &  87.8 &   95.5 &  \textbf{100.0} &  \textbf{100.0} \\
LSUN\_CR     &  97.1 &  \textbf{100.0} &  \textbf{100.0} &  87.3 &   95.8 &  \textbf{100.0} &  \textbf{100.0} \\
Imagenet-    &  96.2 &  \textbf{100.0} &  \textbf{100.0} &  87.9 &   95.9 &  \textbf{100.0} &  \textbf{100.0} \\
Noise        &  97.2 &  97.8 &   99.2 &  97.5 &   99.2 &   99.1 &   \textbf{99.5} \\
Uni. Noise   &  99.9 &  \textbf{100.0} &  \textbf{100.0} &  97.2 &   99.7 &  \textbf{100.0} &   99.9 \\
\hline
Worst Case & Plain & OE & ACET & AT$_{0.5}$ & AT$_{0.25}$ &  R$_{0.5}$ & R$_{0.25}$ \\
\hline
CIFAR10      &   0.0 &   1.3 &  \textbf{99.8} &  43.4 &   43.7 &  \textbf{99.8} &   \textbf{99.8} \\
CIFAR100     &   0.0 &   2.5 &  \textbf{99.8} &  42.7 &   39.5 &  99.7 &   \textbf{99.8}\\
LSUN\_CR     &   0.0 &   1.0 &  99.8 &  36.6 &   42.4 &  \textbf{99.9} &   \textbf{99.9} \\
Imagenet-    &   0.0 &   4.3 &  99.8 &  39.4 &   43.7 &  \textbf{99.9} &   \textbf{99.9} \\
Noise        &   0.0 &   0.0 &  76.8 &  71.2 &   52.7 &  85.6 &  \textbf{85.7} \\
Uni. Noise   &  51.1 &  \textbf{99.9} &  \textbf{99.9} &  73.0 &   67.7 &  \textbf{99.9} &   \textbf{99.9} \\
\hline
\end{tabular}

\begin{tabular}{p{\firstcolumnOD}|C{\svhnOD}C{\svhnOD}C{\svhnOD}C{\svhnOD}C{\svhnOD}C{\svhnOD}C{\svhnOD}C{\svhnOD}}
\hline
& \multicolumn{6}{C{8cm}}{\textbf{CIFAR100}} \\
Av. Case     & Plain & OE      & ACET & AT$_{0.5}$ & AT$_{0.25}$ &  R$_{0.5}$ & R$_{0.25}$ \\
\hline
SVHN         &  86.8 &  \textbf{95.8} &  - &  82.2 &   81.0 &  83.8 &   84.5 \\
CIFAR10      &  81.1 &  \textbf{84.3} &  - &  73.0 &   76.0 &  71.9 &   73.2 \\
LSUN\_CR     &  83.1 &  \textbf{97.5} &  - &  81.0 &   80.4 &  93.6 &   91.7 \\
Imagenet-    &  83.9 &  86.2 &  - &  74.3 &   77.7 &  79.6 &   81.0 \\
Noise        &  85.9 &  87.6 &  - &  84.8 &   82.3 &  \textbf{93.0} &   91.1 \\
Uni. Noise   &  73.2 &  99.7 &  - &  58.5 &   78.7 &  \textbf{99.8} &   99.6 \\
\hline
Worst Case   & Plain & OE     & ACET & AT$_{0.5}$ & AT$_{0.25}$ &  R$_{0.5}$ & R$_{0.25}$ \\
\hline
SVHN         &  0.0 &  5.6 &   - &  30.2      &  20.6       &  41.4      &   \textbf{42.6} \\
CIFAR10      &  0.0 &  5.0 &   - &  27.3      &  18.9       &  \textbf{31.3}      &   28.9 \\
LSUN\_CR     &  0.0 &  5.0 &   - &  30.0      &  21.3       &  58.9      &   \textbf{59.2} \\
Imagenet-    &  0.0 &  4.9 &   - &  31.3      &  23.3       &  \textbf{34.1}      &   31.3 \\
Noise        &  0.0 &  6.2 &   - &  32.6      &  22.2       &  \textbf{68.3}      &   67.5 \\
Uni. Noise   &  2.5 & 60.6 &  - &  27.9      &  42.2       &  \textbf{99.2}      &   97.5 \\
\hline
\end{tabular}
\end{table}

\begin{table}[ht!]
\setlength{\tabcolsep}{5.1pt}
\renewcommand{\arraystretch}{1.}
\caption{\label{Tab:OOD_Cifar10-b} \textit{OOD performance (R. ImageNet):} The area under the ROC curve (AUC) on the binary task of separating the in- from the out-distribution based on the confidence. For each dataset the first table shows the average-case AUC and the second ones show the worst-case AUC with a threat model $l_2=7.0$ around the out-distribution samples.}
\begin{tabular}{p{\firstcolumnOD}|C{\imagenetOD}C{\imagenetOD}C{\imagenetOD}C{\imagenetOD}C{\imagenetOD}C{\imagenetOD}C{\imagenetOD}C{\imagenetOD}}
\hline
& \multicolumn{7}{C{8cm}}{\textbf{R.ImageNet}} \\
Av. Case     & Plain & OE      & ACET & M$_{3.5}$ & AT$_{3.5}$ & AT$_{1.75}$ &  R$_{3.5}$ & R$_{1.75}$ \\
\hline
Flowers      & 90.6  & \textbf{96.2}    & 94.1  & 74.4      & 79.8       & 83.2        & 91.3       & 92.8 \\
Food101      & 91.6  & \textbf{99.3}    & 98.3  & 79.9      & 83.8       & 86.9        & 98.1       & 98.7 \\
FGVC         & 89.6  & \textbf{99.7}    & 98.8  & 79.8      & 80.8       & 81.5        & 98.8       & 99.1 \\
Cars         & 93.9  & \textbf{99.9}    & \textbf{99.9} & 83.5      & 86.3       & 90.0        & 99.8       & \textbf{99.9}  \\
Uni. Noise   & 98.7  & 99.6    & \textbf{100.0} & 95.9      & 88.2       & 87.4        & \textbf{100.0}      & \textbf{100.0}  \\
\hline
Worst Case   & Plain & OE     & ACET & M$_{3.5}$  & AT$_{3.5}$ & AT$_{1.75}$ &  R$_{3.5}$ & R$_{1.75}$ \\
\hline
SVHN         & 0.0   & 1.8    & 83.4 & 32.2       & 35.9       & 16.1        & \textbf{86.6}       & 86.5  \\
CIFAR10      & 0.0   & 1.8    & 87.7 & 33.3       & 32.5       & 11.9        & \textbf{90.6}       & 90.5  \\
LSUN\_CR     & 0.0   & 1.8    & 92.4 & 39.5       & 33.0       & 13.6        & \textbf{95.8}       & 95.7  \\
Imagenet-    & 0.0   & 1.8    & 94.0 & 44.7       & 50.0       & 31.8        & \textbf{97.9}       & \textbf{97.9}  \\
Uni. Noise   & 0.0   & 1.8    & \textbf{100.0} & 83.6       & 43.3       & 16.1        & \textbf{100.0}      & 99.9 \\ 
\hline
\end{tabular}

\end{table}

%% file: res/table_cal.tex
\newcommand{\firstcolumnCAL}{17.9mm}
\newcommand{\cifarCAL}{8mm}
\newcommand{\svhnCAL}{10.89mm}
\newcommand{\imagenetCAL}{9.09mm}

\newcolumntype{C}[1]{>{\centering\arraybackslash}p{#1}}
\begin{table}[t]
\begin{center}
\setlength{\tabcolsep}{5pt}
\renewcommand{\arraystretch}{1.0}
\caption{\label{Tab:Cal} \textit{Calibration:} The ECE on the 2000 point validation set before (ECE$_b$) and after (ECE$_a$) temperature rescaling, as well as the temperature minimizing the ECE on the validation set. }
\begin{tabular}{p{\firstcolumnCAL}|ccccccccc}
\hline
CIFAR10 & Plain &  OE  & ACET  & M$_{0.5}$ &  AT$_{0.5}$ & AT$_{0.25}$ & JEM-0 &  R$_{0.5}$ & R$_{0.25}$ \\
\hline
ECE$_b$    (in \%)    &  4.8   & 14.1  & 31.5 & 4.0   & 2.0  &  4.3   & 10.6  &  25.0  &  26.1  \\
$T$                   &  1.61  & 0.10  & 0.35 & 1.20  & 1.11 &  1.39  & 1.88  &  0.41  &  0.43  \\
ECE$_a$    (in \%)    &  0.7   & 6.5   & 6.7  & 1.8   & 1.0  &  1.0   & 2.9   &  5.7   &  4.6   \\
\hline
\end{tabular}

\renewcommand{\arraystretch}{1.0}
\begin{tabular}{p{\firstcolumnCAL}|C{\svhnCAL}C{\svhnCAL}C{\svhnCAL}C{\svhnCAL}C{\svhnCAL}C{\svhnCAL}C{\svhnCAL}C{\svhnCAL}}
\hline
SVHN & Plain & OE & ACET & AT$_{0.5}$ & AT$_{0.25}$ &  R$_{0.5}$ & R$_{0.25}$ \\
\hline
ECE$_b$    (in \%)    &  0.5   & 0.5   &  0.9  &  3.0  & 0.8   & 3.3   &  1.2 \\
$T$                   &  1.01  & 0.97  & 0.49  & 0.86  & 0.57  & 1.10  &  0.85  \\
ECE$_a$    (in \%)    &  0.5   & 0.5   &  0.5  &  0.9  &  0.6  & 1.1   &  0.8 \\
\hline
\end{tabular}

\renewcommand{\arraystretch}{1.0}
\begin{tabular}{p{\firstcolumnCAL}|C{\svhnCAL}C{\svhnCAL}C{\svhnCAL}C{\svhnCAL}C{\svhnCAL}C{\svhnCAL}C{\svhnCAL}C{\svhnCAL}}
\hline
CIFAR100 & Plain & OE & ACET & AT$_{0.5}$ & AT$_{0.25}$ &  R$_{0.5}$ & R$_{0.25}$ \\
\hline
ECE$_b$    (in \%)    &  7.0   & 12.6  &  -  &  3.6  & 1.3   & 24.3  &  22.8 \\
$T$                   &  1.43  & 0.63  & -  & 0.86  & 1.02  & 0.60  &  0.67  \\
ECE$_a$    (in \%)    &  1.2   & 7.5   &   -  &  1.5  &  1.1  & 2.4   &  2.1  \\
\hline
\end{tabular}

\renewcommand{\arraystretch}{1.0}
\begin{tabular}{p{\firstcolumnCAL}|C{\imagenetCAL}C{\imagenetCAL}C{\imagenetCAL}C{\imagenetCAL}C{\imagenetCAL}C{\imagenetCAL}C{\imagenetCAL}C{\imagenetCAL}C{\imagenetCAL}}
\hline
R.ImageNet            & Plain & OE    & ACET &  M$_{3.5}$ & AT$_{3.5}$ & AT$_{1.75}$ &  R$_{3.5}$ & R$_{1.75}$ \\
\hline
ECE$_b$    (in \%)    & 0.7   & 4.5   & 26.9 &   2.8      &  1.6       & 0.9         & 15.3       &  15.4  \\
$T$                   & 1.17  & 0.55  & 0.34 &   0.88     &  1.01      & 0.96        & 0.49       &  0.49  \\
ECE$_a$    (in \%)    & 0.3   & 1.3   & 1.08 &   1.6      &  1.5       & 0.7         & 1.2        &  1.1   \\
\hline
\end{tabular}
\end{center}

\end{table}

%% file: res/appendix_vc_cifar_main_paper_extended.tex
\newcommand{\breiteb}{12.1mm}
\begin{figure}
\begin{tabular}{p{1cm}p{\breite}p{\breite}p{\breite}p{\breite}p{\breite}p{\breite}p{\breite}p{\breite}}
Model  & Orig. & $\epsilon=0.5$ & $\epsilon=1.0$ & $\epsilon=1.5$ & $\epsilon=2.0$ & $\epsilon=2.5$ & $\epsilon=3.0$\\
\begin{turn}{90} \hspace{-.4cm} Plain \end{turn} & \multicolumn{7}{c}{\includegraphics[width=0.91\textwidth,valign=c]{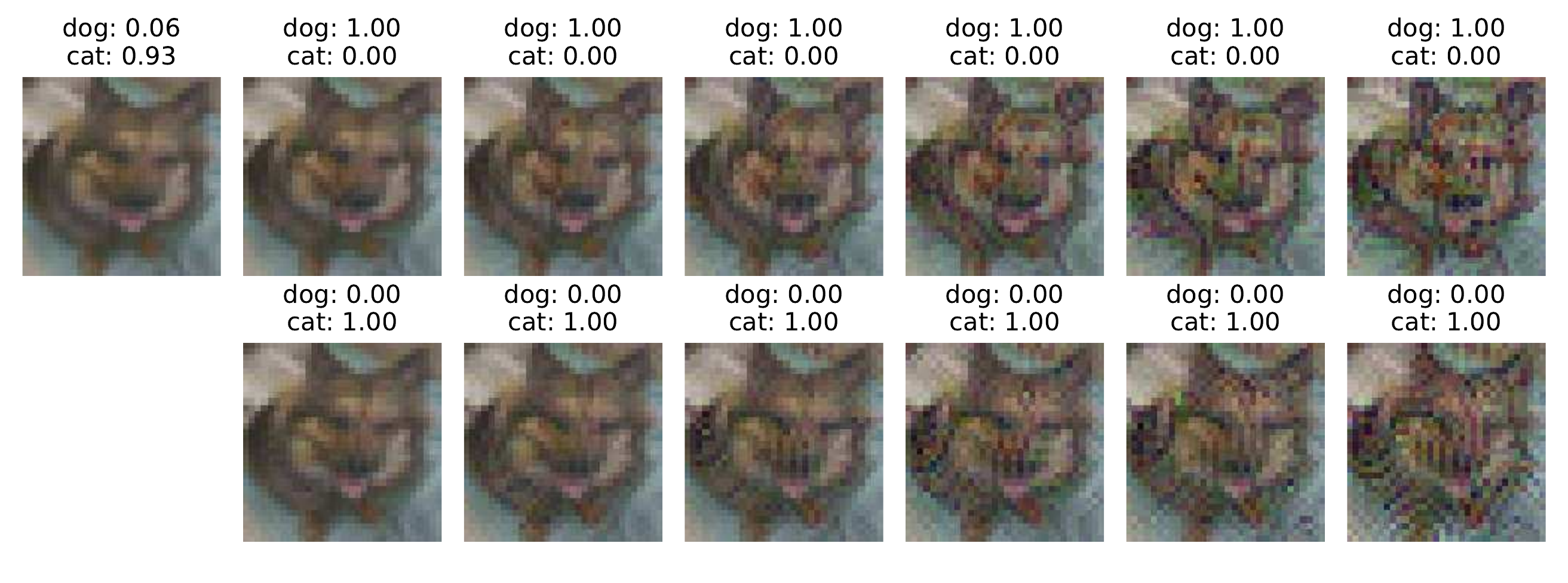}} \\
\hline
\begin{turn}{90} \hspace{-.33cm}  OE \end{turn}  &  \multicolumn{7}{c}{\includegraphics[width=0.91\textwidth,valign=c]{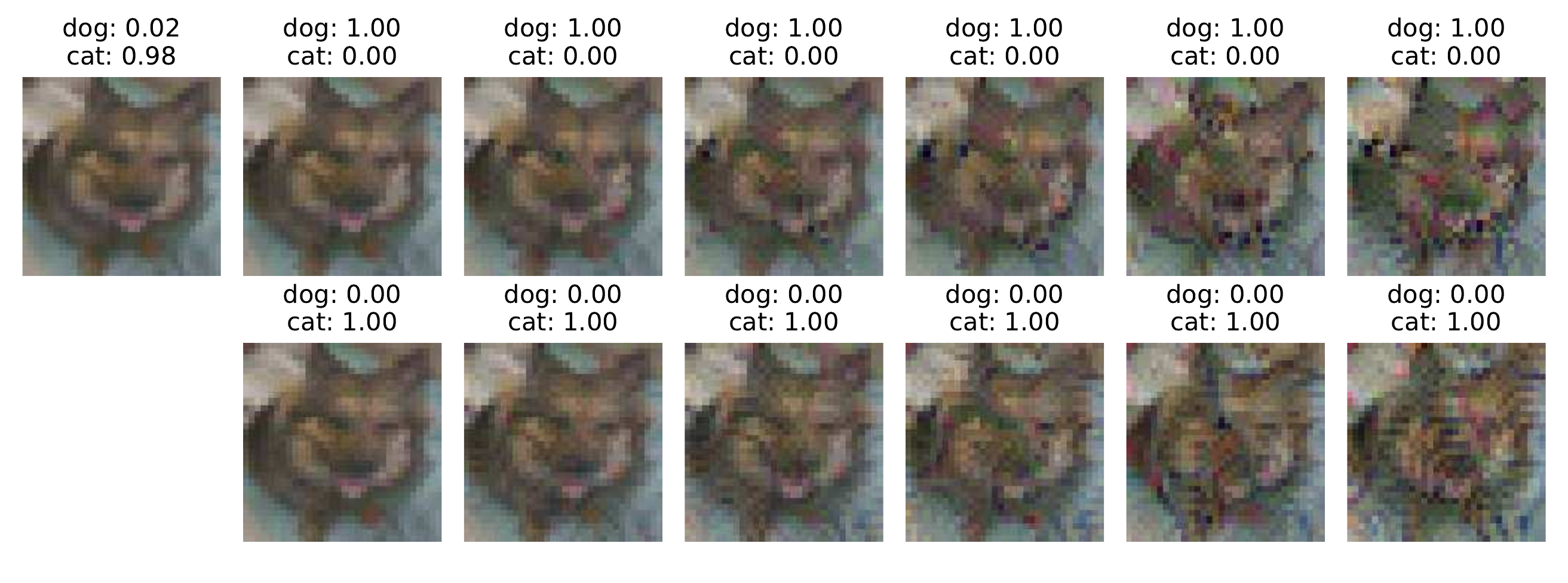}} \\
\hline
\begin{turn}{90} \hspace{-.4cm} ACET \end{turn}  &  \multicolumn{7}{c}{\includegraphics[width=0.91\textwidth,valign=c]{pics/CIFAR10/ACET/VC/11.pdf}} \\
\hline
\begin{turn}{90} \hspace{-.5cm} M-0.50 \end{turn}  &  \multicolumn{7}{c}{\includegraphics[width=0.91\textwidth,valign=c]{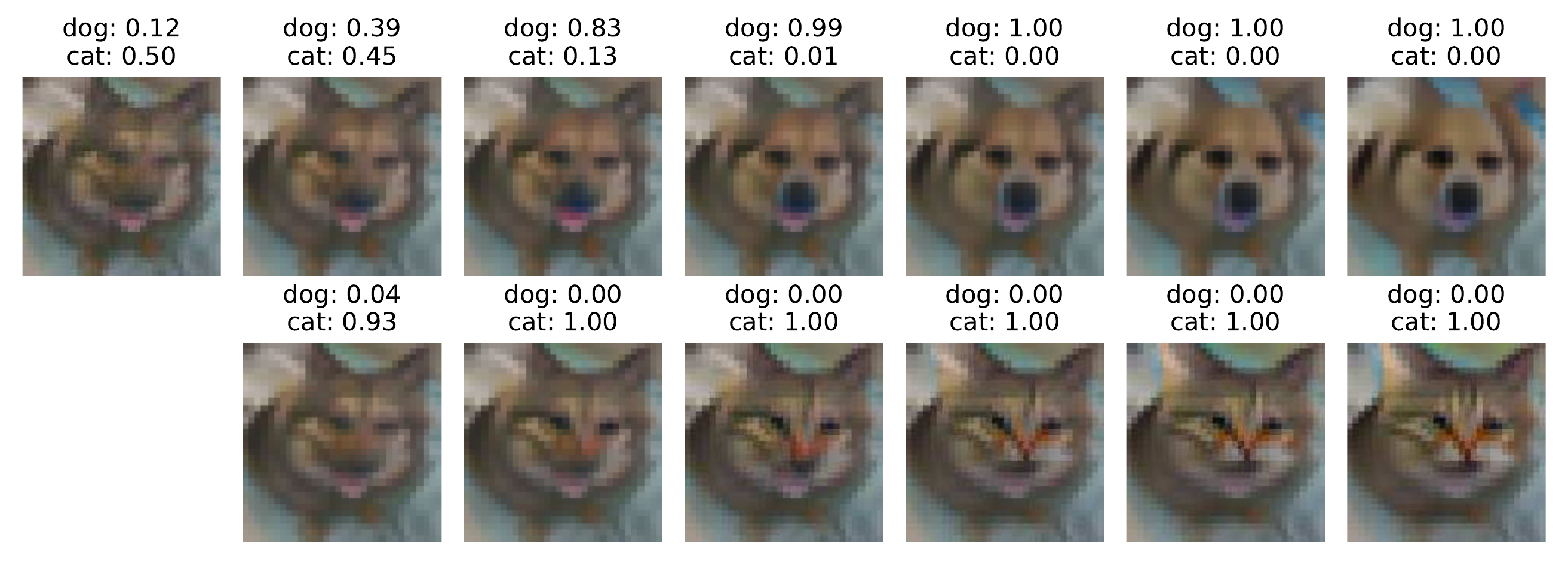}} \\
\end{tabular}	
\caption{\label{fig:vc_cifar_ext1} Visual Counterfactuals for all 9 CIFAR10 models (extension of Figure~\ref{fig:cf}). The original test image (top left) has been misclassified by all models. Per model, we generate images which increase the confidence in the grounth truth class (upper row) and falsely predicted class (bottom row) for varying $l_2$ budgets. Continues on next page. }
\end{figure}

\begin{figure}
\begin{center}
\begin{tabular}{p{1cm}x{\breiteb}x{\breiteb}x{\breiteb}x{\breiteb}x{\breiteb}x{\breiteb}x{\breiteb}x{\breiteb}}
Model  & Orig. & $\epsilon=0.5$ & $\epsilon=1.0$ & $\epsilon=1.5$ & $\epsilon=2.0$ & $\epsilon=2.5$ & $\epsilon=3.0$\\
\begin{turn}{90} \hspace{-.4cm} AT-0.50 \end{turn}  &  \multicolumn{7}{c}{\includegraphics[width=0.75\textwidth,valign=c]{pics/CIFAR10/AT05/VC/11.pdf}} \\
\hline
\begin{turn}{90} \hspace{-.4cm} AT-0.25 \end{turn}  &  \multicolumn{7}{c}{\includegraphics[width=0.75\textwidth,valign=c]{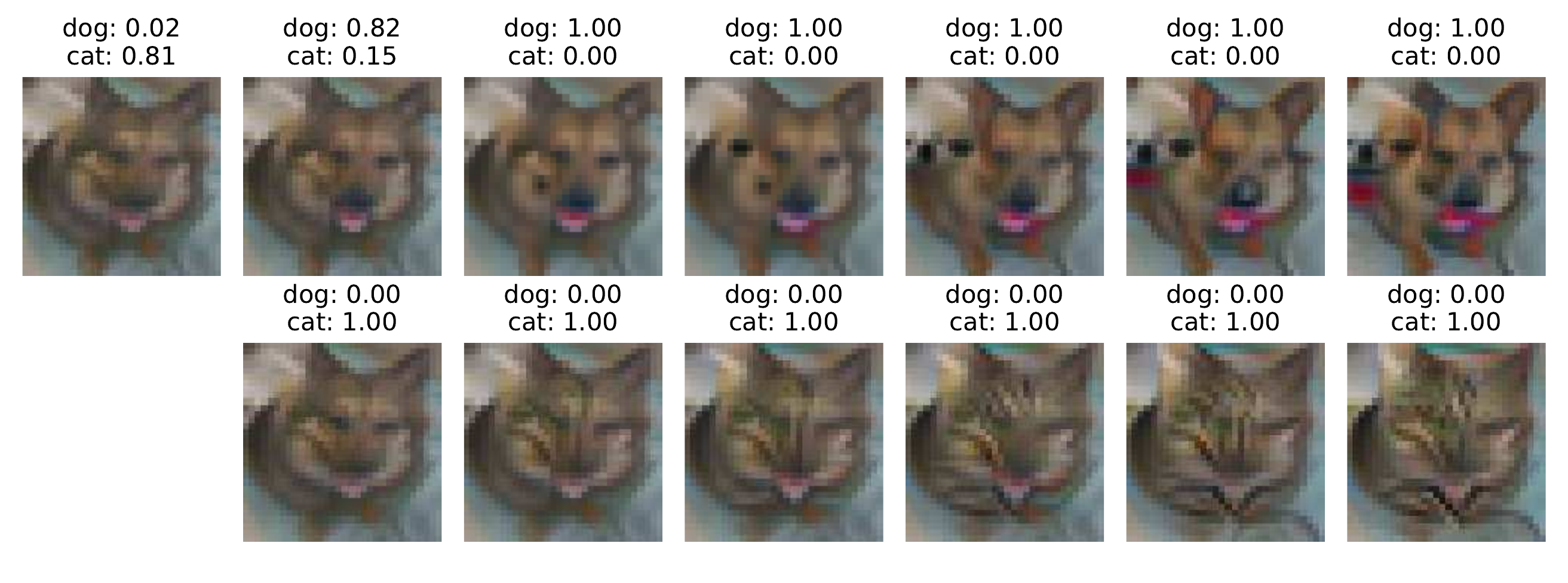}} \\
\hline
\begin{turn}{90} \hspace{-.33cm}  JEM-0 \end{turn}  &  \multicolumn{7}{c}{\includegraphics[width=0.75\textwidth,valign=c]{pics/CIFAR10/EBM/VC/11.pdf}} \\
\hline
\begin{turn}{90} \hspace{-.5cm} RATIO-0.50 \end{turn}  &  \multicolumn{7}{c}{\includegraphics[width=0.75\textwidth,valign=c]{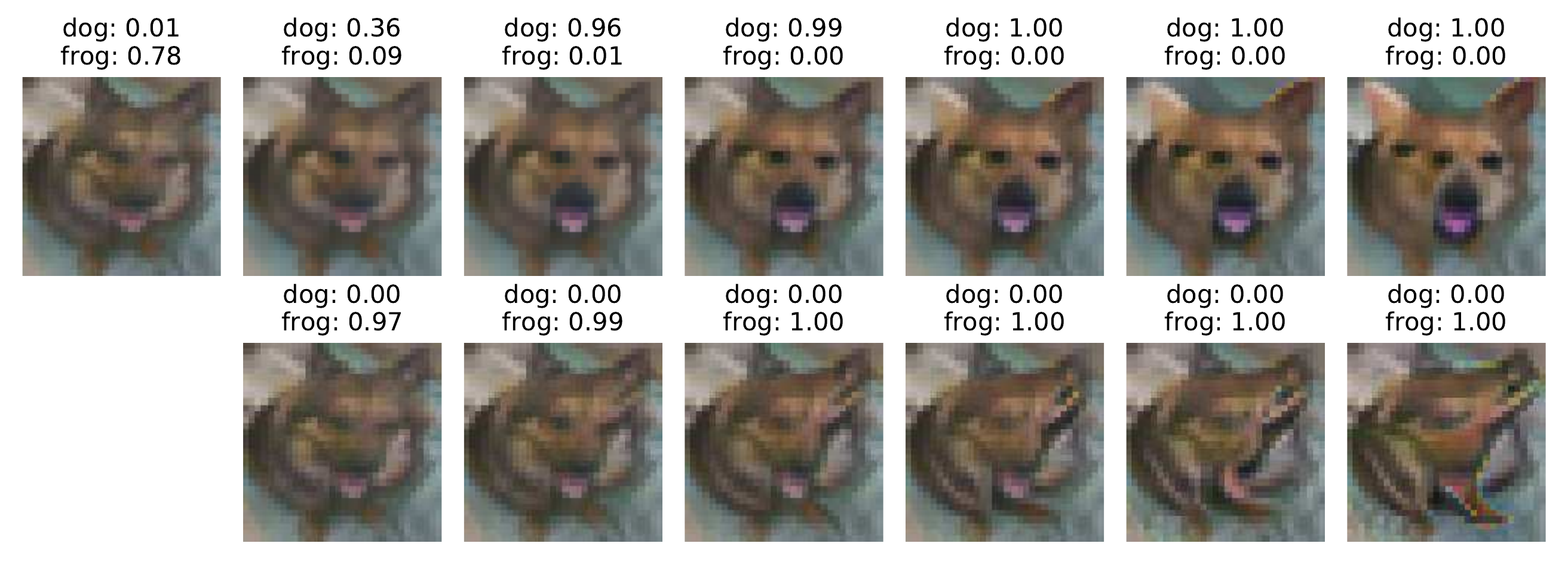}} \\
\hline
\begin{turn}{90} \hspace{-.5cm} RATIO-0.25 \end{turn}  &  \multicolumn{7}{c}{\includegraphics[width=0.75\textwidth,valign=c]{pics/CIFAR10/Ratio025/VC/11.pdf}} \\
\end{tabular}	
\caption{\label{fig:vc_cifar_ext2} Visual Counterfactuals for all 9 CIFAR10 models (extension of Figure \ref{fig:cf}) (continued).}
\end{center}
\end{figure}

%% file: res/appendix_vc_cifar_new.tex
\begin{figure}[ht!]
\begin{tabular}{p{1cm}x{\breite}x{\breite}x{\breite}x{\breite}x{\breite}x{\breite}x{\breite}x{\breite}}
Model  & Orig. & $\epsilon=0.5$ & $\epsilon=1.0$ & $\epsilon=1.5$ & $\epsilon=2.0$ & $\epsilon=2.5$ & $\epsilon=3.0$\\
\begin{turn}{90} \hspace{-.4cm} ACET \end{turn} & \multicolumn{7}{c}{\includegraphics[width=0.91\textwidth,valign=c]{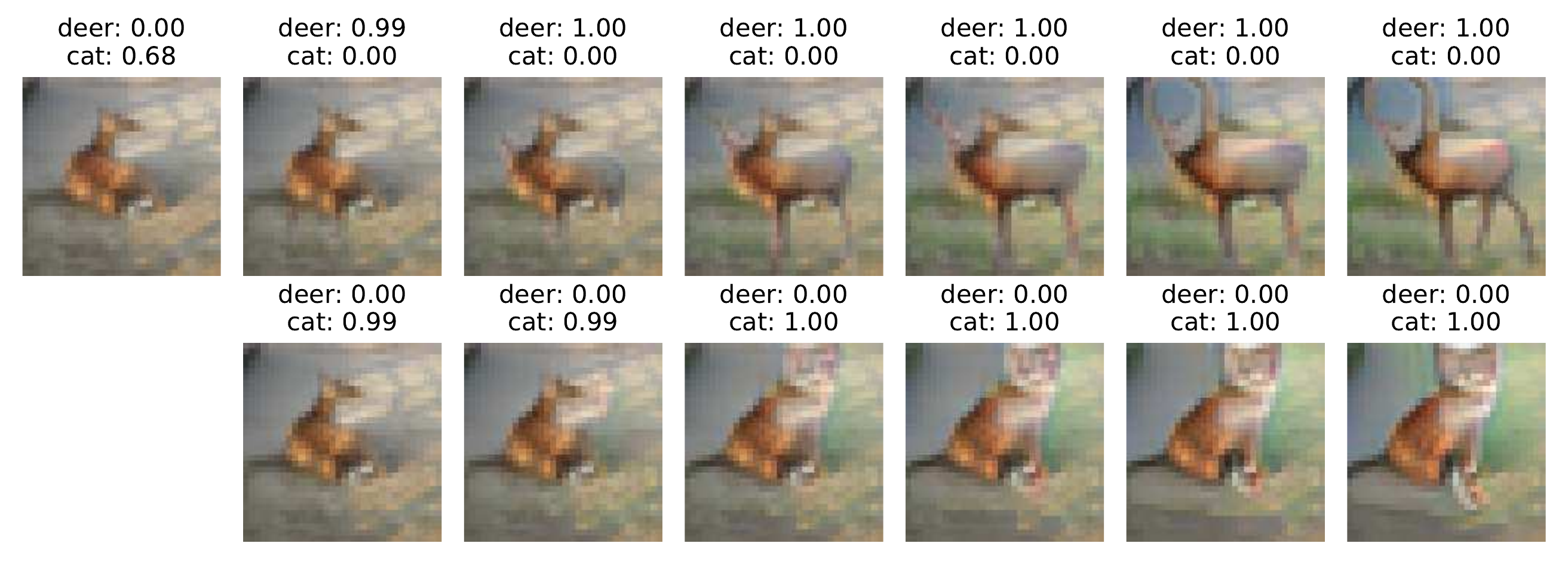}} \\
\hline
\begin{turn}{90} \hspace{-.33cm}  JEM-0 \end{turn}  &  \multicolumn{7}{c}{\includegraphics[width=0.91\textwidth,valign=c]{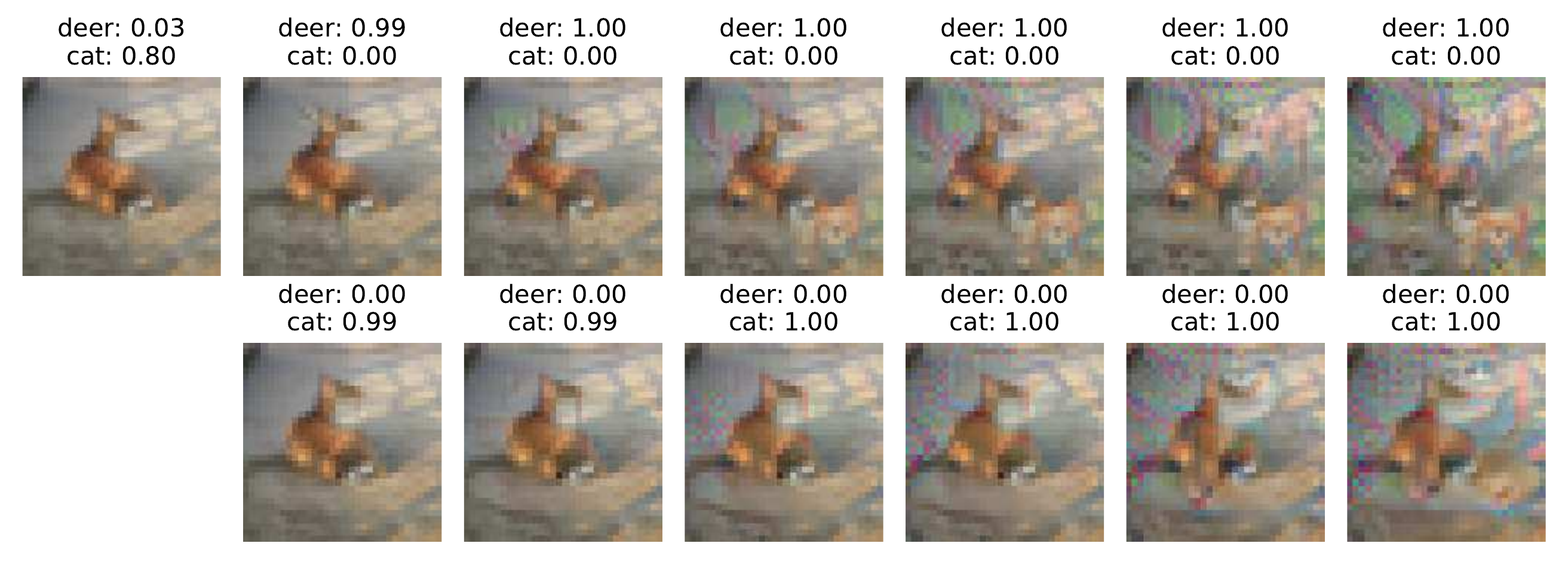}} \\
\hline
\begin{turn}{90} \hspace{-.4cm} AT-0.50 \end{turn}  &  \multicolumn{7}{c}{\includegraphics[width=0.91\textwidth,valign=c]{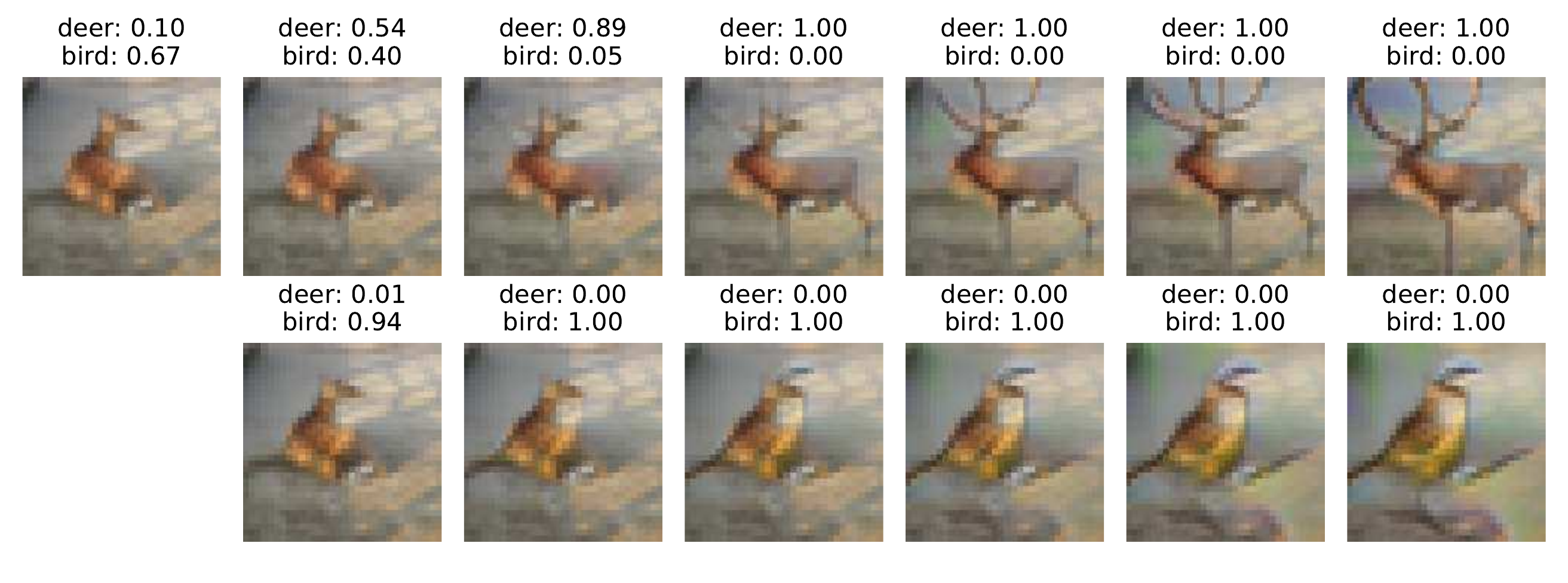}} \\
\hline
\begin{turn}{90} \hspace{-.9cm} RATIO-0.25 \end{turn}  &  \multicolumn{7}{c}{\includegraphics[width=0.91\textwidth,valign=c]{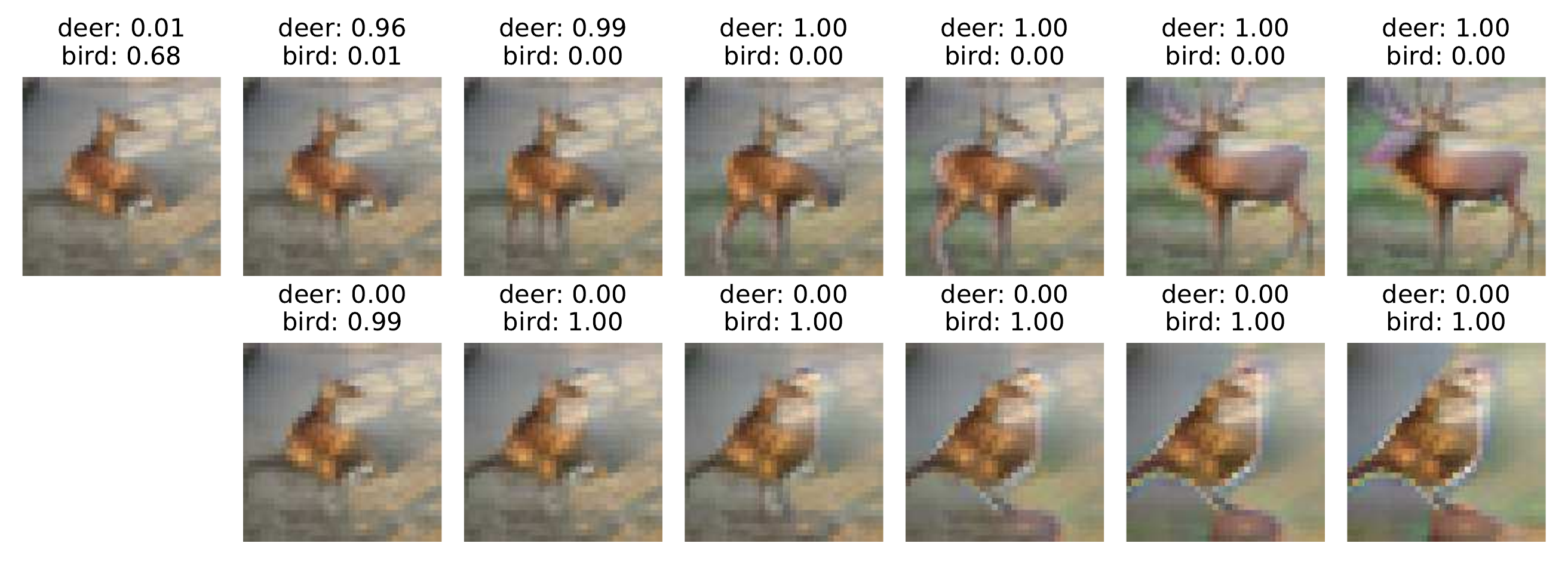}} \\
\end{tabular}	
\vspace{-.3cm}
\caption{\label{fig:vc_cifar_new1} Additional Visual Counterfactuals for CIFAR10 test samples misclassified by all methods. RATIO$_{0.25}$ and AT$_{0.5}$ produce images of similar high quality whereas JEM-0 generates strong noise patterns and less visible class specific features. ACET produces identifiable features, but with a lower quality than RATIO$_{0.25}$ and AT$_{0.5}$. }
\end{figure}

\begin{figure}[ht!]
\begin{tabular}{p{1cm}x{\breite}x{\breite}x{\breite}x{\breite}x{\breite}x{\breite}x{\breite}x{\breite}}
Model  & Orig. & $\epsilon=0.5$ & $\epsilon=1.0$ & $\epsilon=1.5$ & $\epsilon=2.0$ & $\epsilon=2.5$ & $\epsilon=3.0$\\
\begin{turn}{90} \hspace{-.4cm} ACET \end{turn} & \multicolumn{7}{c}{\includegraphics[width=0.91\textwidth,valign=c]{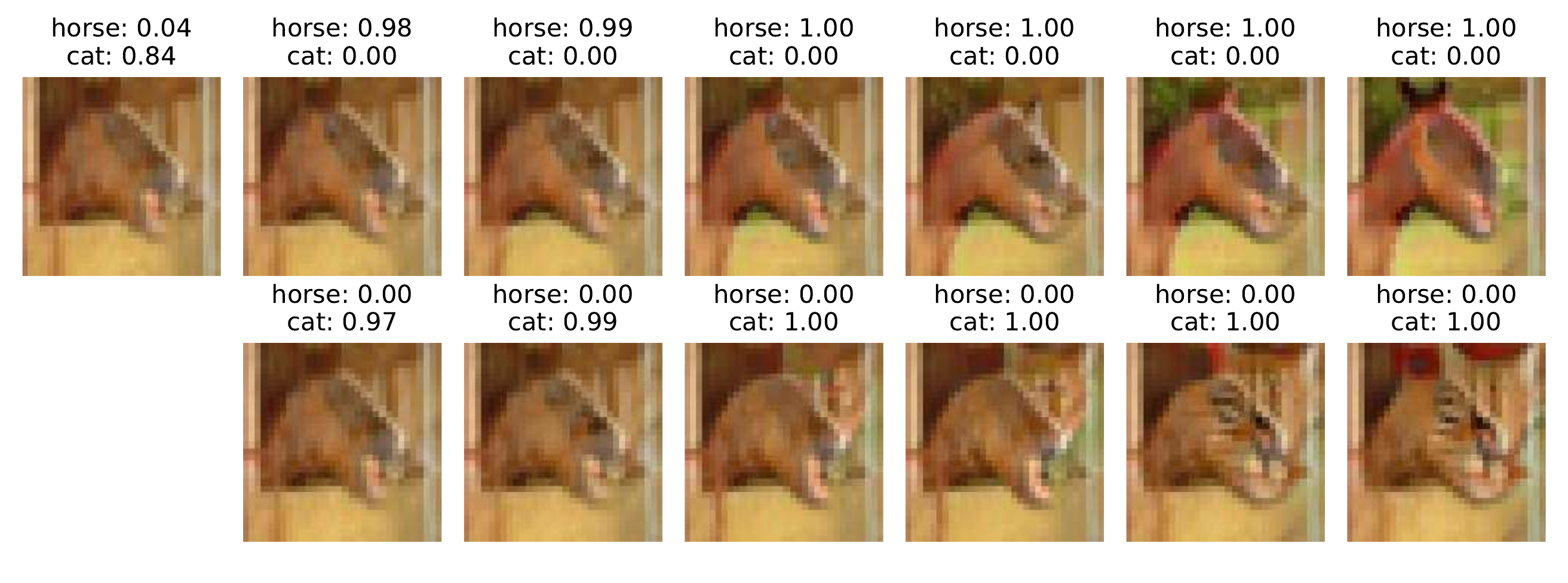}} \\
\hline
\begin{turn}{90} \hspace{-.33cm}  JEM-0 \end{turn}  &  \multicolumn{7}{c}{\includegraphics[width=0.91\textwidth,valign=c]{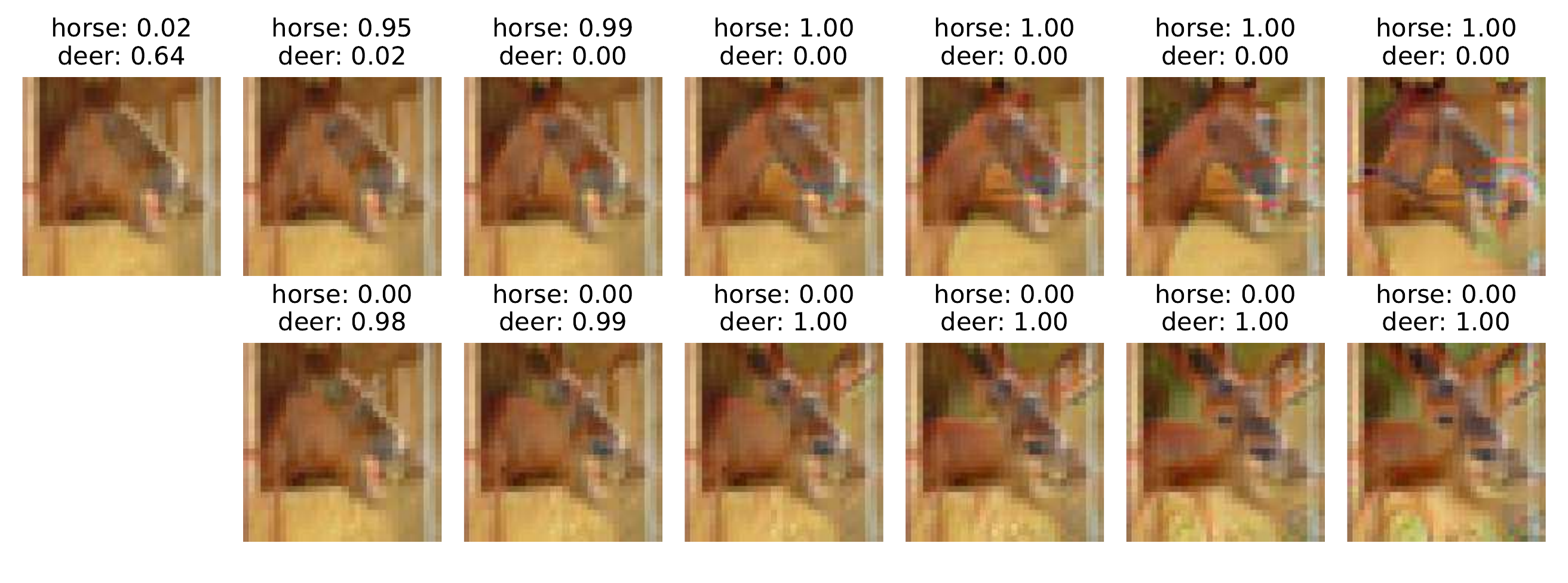}} \\
\hline
\begin{turn}{90} \hspace{-.4cm} AT-0.50 \end{turn}  &  \multicolumn{7}{c}{\includegraphics[width=0.91\textwidth,valign=c]{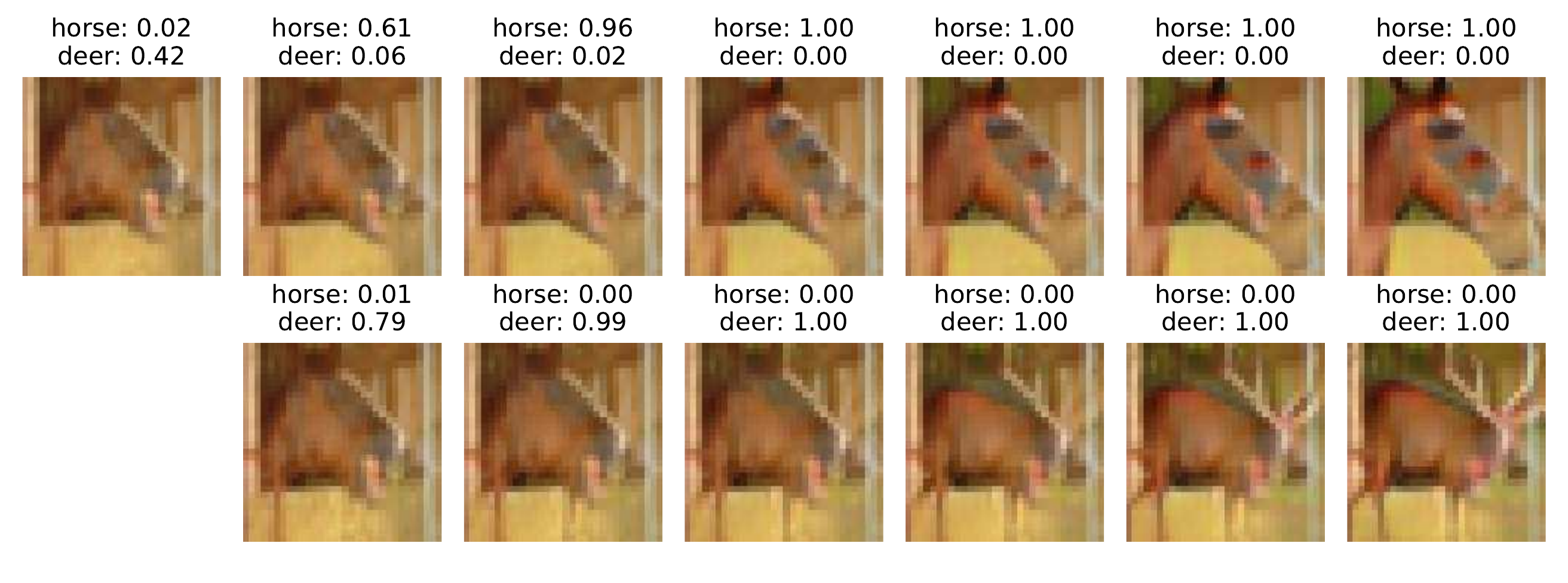}} \\
\hline
\begin{turn}{90} \hspace{-.9cm} RATIO-0.25 \end{turn}  &  \multicolumn{7}{c}{\includegraphics[width=0.91\textwidth,valign=c]{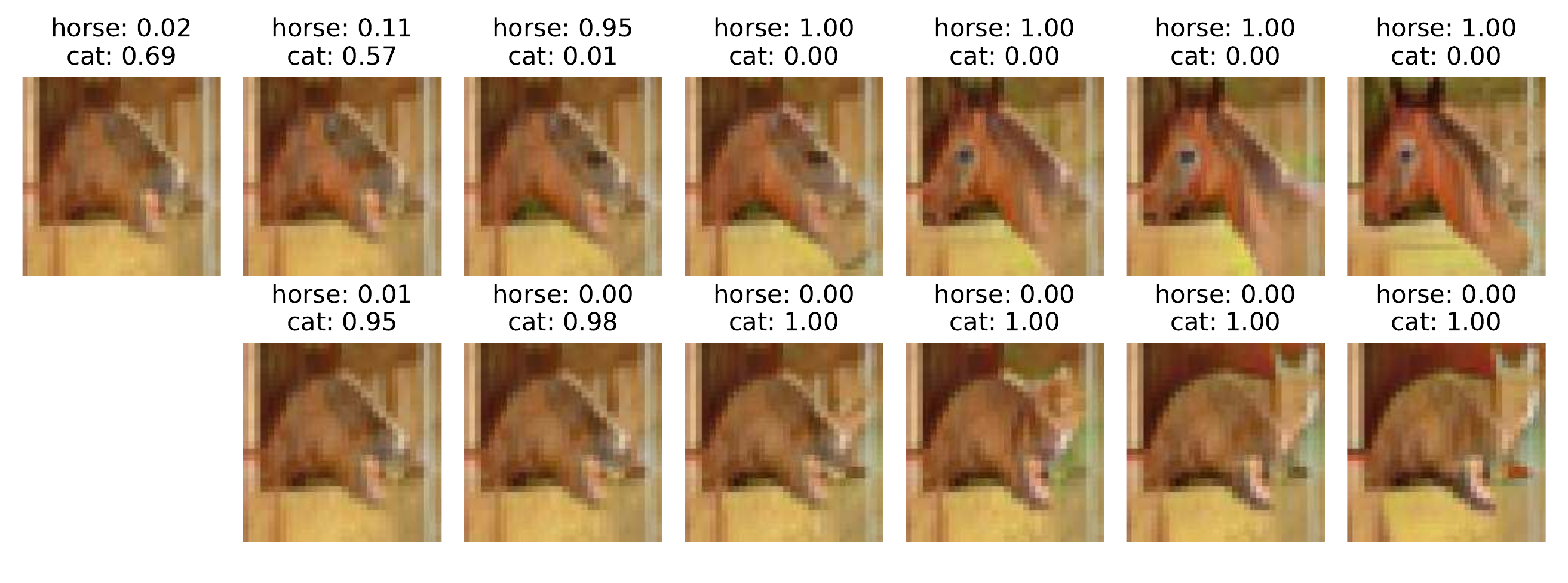}} \\
\end{tabular}	
\vspace{-.3cm}
\caption{\label{fig:vc_cifar_new2} Additional Visual Counterfactuals for misclassified samples from the CIFAR10 test set. Even though RATIO$_{0.25}$ was trained with a smaller radius on in-distribution samples than AT$_{0.5}$, it is the only model capable of generating a realistic horse head for larger radii.}
\end{figure}

\begin{figure}[ht!]
\begin{tabular}{p{1cm}x{\breite}x{\breite}x{\breite}x{\breite}x{\breite}x{\breite}x{\breite}x{\breite}}
Model  & Orig. & $\epsilon=0.5$ & $\epsilon=1.0$ & $\epsilon=1.5$ & $\epsilon=2.0$ & $\epsilon=2.5$ & $\epsilon=3.0$\\
\begin{turn}{90} \hspace{-.4cm} ACET \end{turn} & \multicolumn{7}{c}{\includegraphics[width=0.91\textwidth,valign=c]{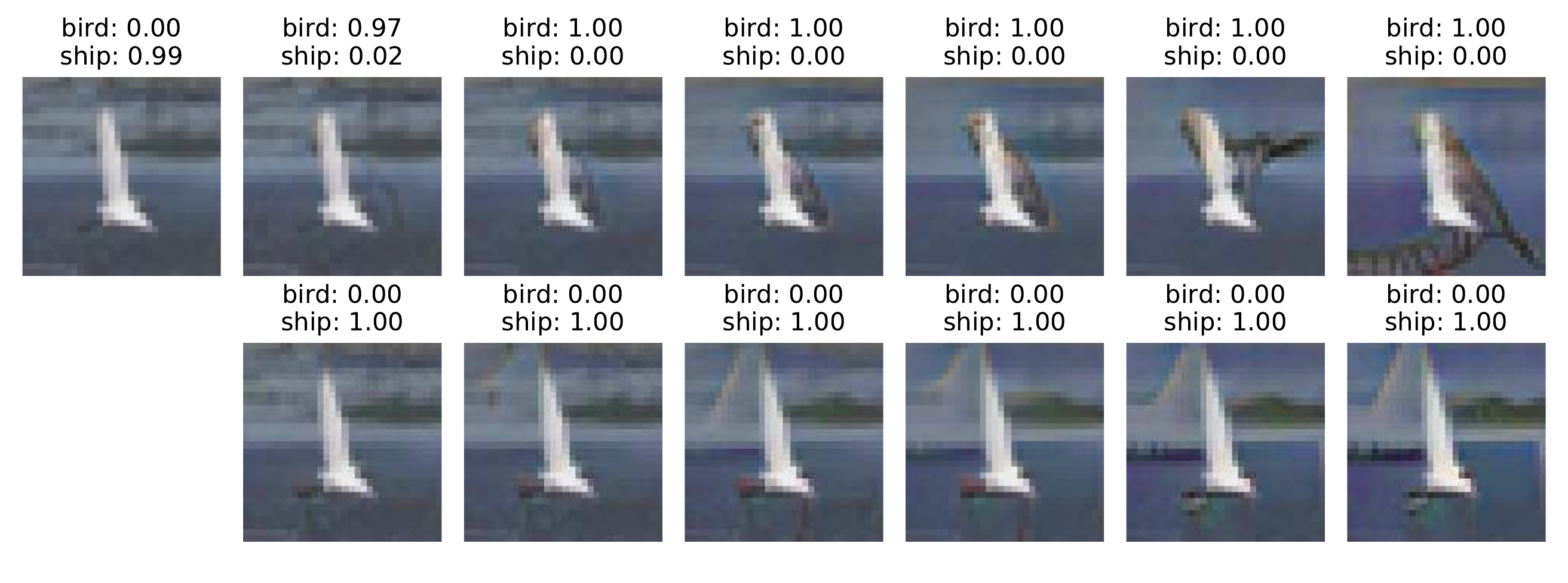}} \\
\hline
\begin{turn}{90} \hspace{-.33cm}  JEM-0 \end{turn}  &  \multicolumn{7}{c}{\includegraphics[width=0.91\textwidth,valign=c]{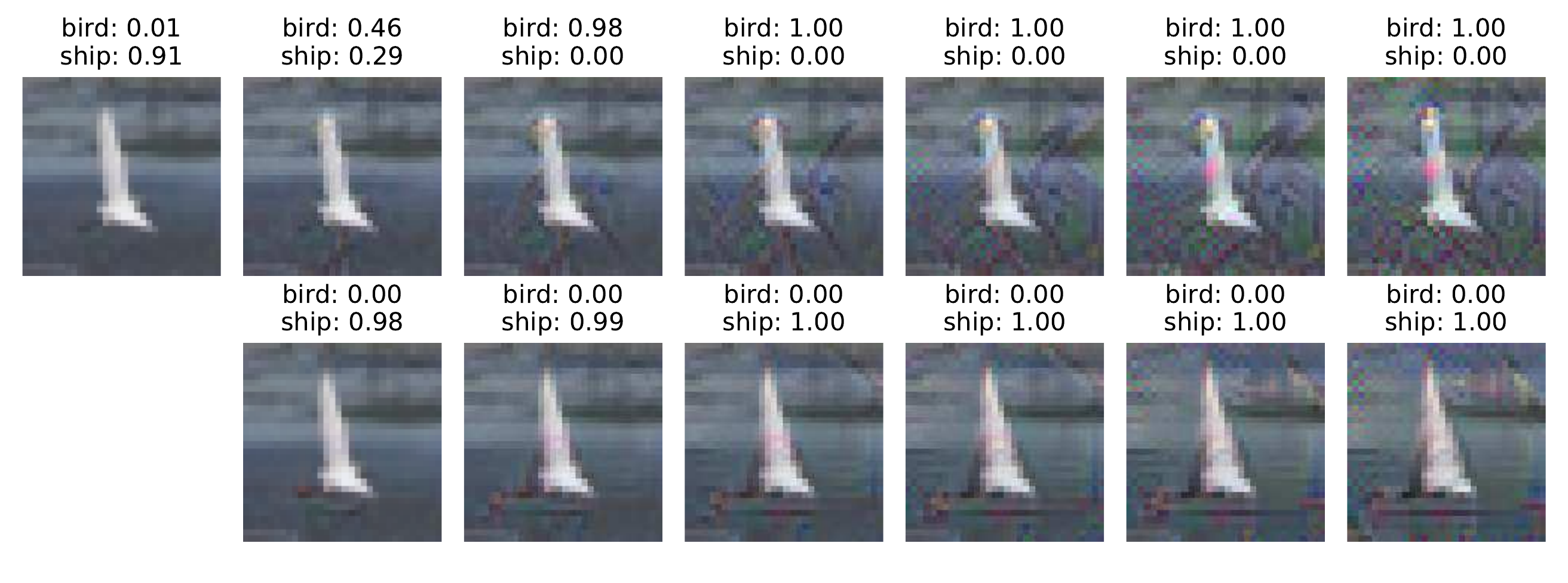}} \\
\hline
\begin{turn}{90} \hspace{-.4cm} AT-0.50 \end{turn}  &  \multicolumn{7}{c}{\includegraphics[width=0.91\textwidth,valign=c]{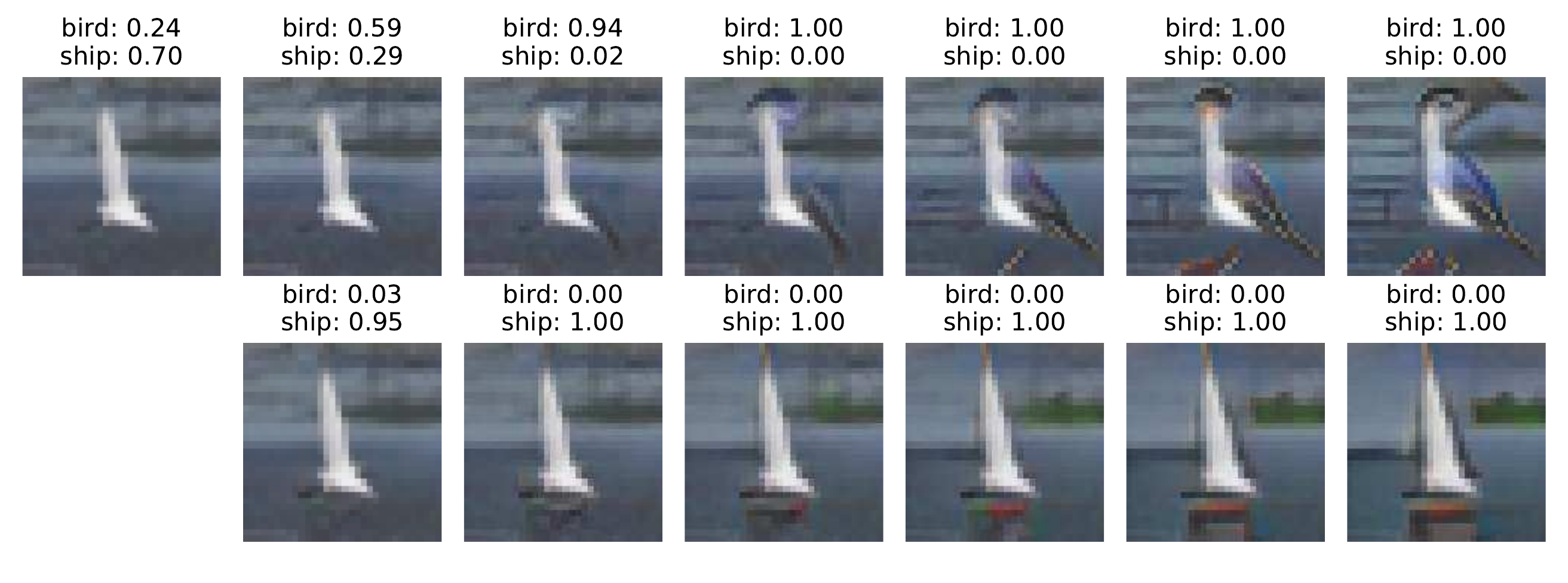}} \\
\hline
\begin{turn}{90} \hspace{-.9cm} RATIO-0.25 \end{turn}  &  \multicolumn{7}{c}{\includegraphics[width=0.91\textwidth,valign=c]{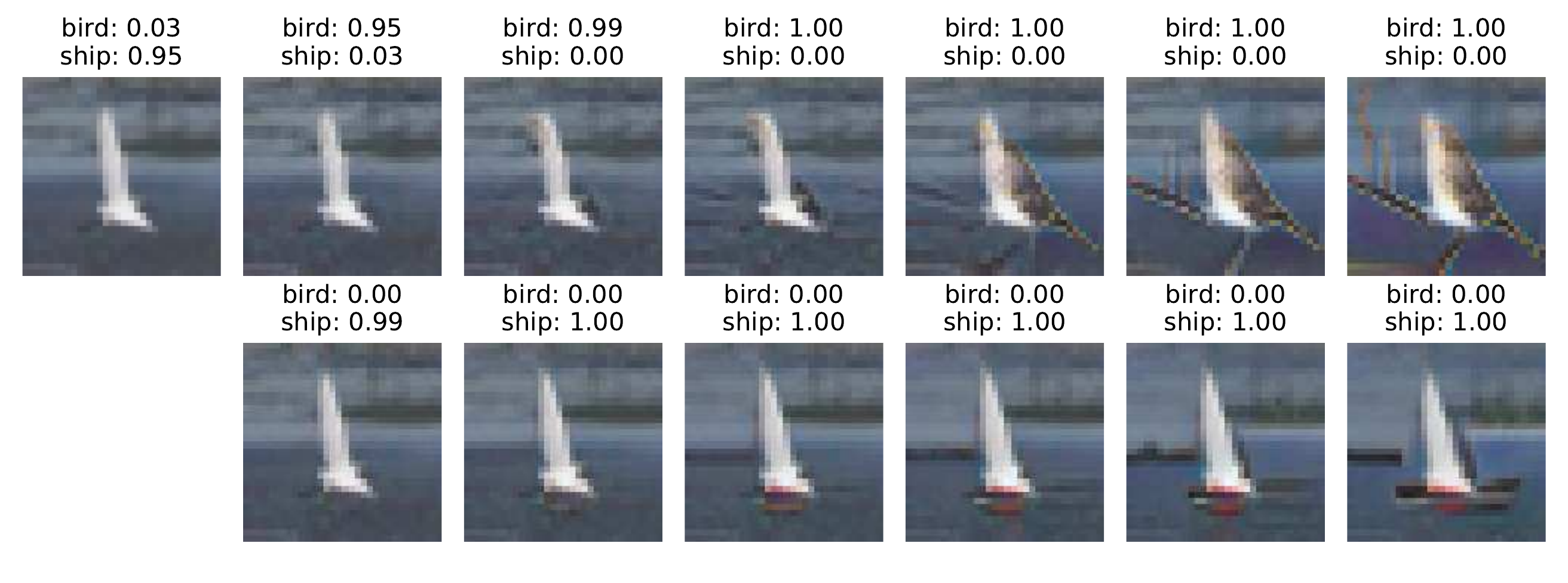}} \\
\end{tabular}	
\vspace{-.3cm}
\caption{\label{fig:vc_cifar_new3} Additional Visual Counterfactuals for misclassified samples from the CIFAR10 test set. All models except JEM-0 generate realistic images for both target classes.}
\end{figure}

%% file: res/appendix_cifar_overview.tex
\begin{figure}
\begin{adjustbox}{max width=\textwidth}
\begin{tabu}{ccccccccccccccccc}
\multicolumn{8}{c}{Original} & & \multicolumn{8}{c}{ACET}\\
\rowfont{\tiny}
cat & frog & dog & dog & dog & bird & dog & horse & & cat & frog & dog & dog & dog & bird & dog & horse\\
\includegraphics[width=0.056\textwidth,valign=c]{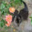}
 & 
\includegraphics[width=0.056\textwidth,valign=c]{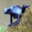}
 & 
\includegraphics[width=0.056\textwidth,valign=c]{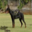}
 & 
\includegraphics[width=0.056\textwidth,valign=c]{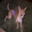}
 & 
\includegraphics[width=0.056\textwidth,valign=c]{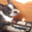}
 & 
\includegraphics[width=0.056\textwidth,valign=c]{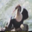}
 & 
\includegraphics[width=0.056\textwidth,valign=c]{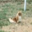}
 & 
\includegraphics[width=0.056\textwidth,valign=c]{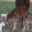}
 & 
\hspace{ 0.028 \textwidth} & 
\includegraphics[width=0.056\textwidth,valign=c]{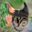}
 & 
\includegraphics[width=0.056\textwidth,valign=c]{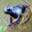}
 & 
\includegraphics[width=0.056\textwidth,valign=c]{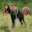}
 & 
\includegraphics[width=0.056\textwidth,valign=c]{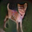}
 & 
\includegraphics[width=0.056\textwidth,valign=c]{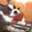}
 & 
\includegraphics[width=0.056\textwidth,valign=c]{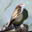}
 & 
\includegraphics[width=0.056\textwidth,valign=c]{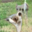}
 & 
\includegraphics[width=0.056\textwidth,valign=c]{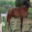}
\\
\rowfont{\tiny}
bird & ship & ship & bird & bird & dog & cat & deer & & bird & ship & ship & bird & bird & dog & cat & deer\\
\includegraphics[width=0.056\textwidth,valign=c]{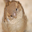}
 & 
\includegraphics[width=0.056\textwidth,valign=c]{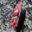}
 & 
\includegraphics[width=0.056\textwidth,valign=c]{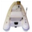}
 & 
\includegraphics[width=0.056\textwidth,valign=c]{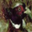}
 & 
\includegraphics[width=0.056\textwidth,valign=c]{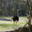}
 & 
\includegraphics[width=0.056\textwidth,valign=c]{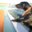}
 & 
\includegraphics[width=0.056\textwidth,valign=c]{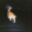}
 & 
\includegraphics[width=0.056\textwidth,valign=c]{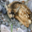}
 & 
\hspace{ 0.028 \textwidth} & 
\includegraphics[width=0.056\textwidth,valign=c]{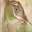}
 & 
\includegraphics[width=0.056\textwidth,valign=c]{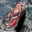}
 & 
\includegraphics[width=0.056\textwidth,valign=c]{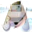}
 & 
\includegraphics[width=0.056\textwidth,valign=c]{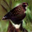}
 & 
\includegraphics[width=0.056\textwidth,valign=c]{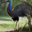}
 & 
\includegraphics[width=0.056\textwidth,valign=c]{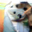}
 & 
\includegraphics[width=0.056\textwidth,valign=c]{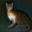}
 & 
\includegraphics[width=0.056\textwidth,valign=c]{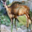}
\\
\rowfont{\tiny}
bird & deer & bird & frog & frog & deer & truck & dog & & bird & deer & bird & frog & frog & deer & truck & dog\\
\includegraphics[width=0.056\textwidth,valign=c]{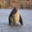}
 & 
\includegraphics[width=0.056\textwidth,valign=c]{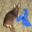}
 & 
\includegraphics[width=0.056\textwidth,valign=c]{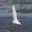}
 & 
\includegraphics[width=0.056\textwidth,valign=c]{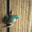}
 & 
\includegraphics[width=0.056\textwidth,valign=c]{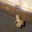}
 & 
\includegraphics[width=0.056\textwidth,valign=c]{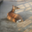}
 & 
\includegraphics[width=0.056\textwidth,valign=c]{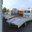}
 & 
\includegraphics[width=0.056\textwidth,valign=c]{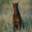}
 & 
\hspace{ 0.028 \textwidth} & 
\includegraphics[width=0.056\textwidth,valign=c]{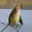}
 & 
\includegraphics[width=0.056\textwidth,valign=c]{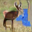}
 & 
\includegraphics[width=0.056\textwidth,valign=c]{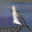}
 & 
\includegraphics[width=0.056\textwidth,valign=c]{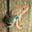}
 & 
\includegraphics[width=0.056\textwidth,valign=c]{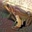}
 & 
\includegraphics[width=0.056\textwidth,valign=c]{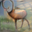}
 & 
\includegraphics[width=0.056\textwidth,valign=c]{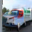}
 & 
\includegraphics[width=0.056\textwidth,valign=c]{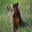}
\\
\rowfont{\tiny}
truck & horse & bird & bird & bird & cat & dog & truck & & truck & horse & bird & bird & bird & cat & dog & truck\\
\includegraphics[width=0.056\textwidth,valign=c]{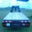}
 & 
\includegraphics[width=0.056\textwidth,valign=c]{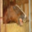}
 & 
\includegraphics[width=0.056\textwidth,valign=c]{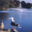}
 & 
\includegraphics[width=0.056\textwidth,valign=c]{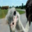}
 & 
\includegraphics[width=0.056\textwidth,valign=c]{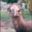}
 & 
\includegraphics[width=0.056\textwidth,valign=c]{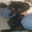}
 & 
\includegraphics[width=0.056\textwidth,valign=c]{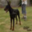}
 & 
\includegraphics[width=0.056\textwidth,valign=c]{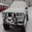}
 & 
\hspace{ 0.028 \textwidth} & 
\includegraphics[width=0.056\textwidth,valign=c]{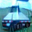}
 & 
\includegraphics[width=0.056\textwidth,valign=c]{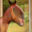}
 & 
\includegraphics[width=0.056\textwidth,valign=c]{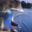}
 & 
\includegraphics[width=0.056\textwidth,valign=c]{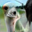}
 & 
\includegraphics[width=0.056\textwidth,valign=c]{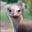}
 & 
\includegraphics[width=0.056\textwidth,valign=c]{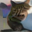}
 & 
\includegraphics[width=0.056\textwidth,valign=c]{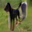}
 & 
\includegraphics[width=0.056\textwidth,valign=c]{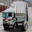}
\\
\rowfont{\tiny}
deer & plane & car & car & dog & horse & cat & cat & & deer & plane & car & car & dog & horse & cat & cat\\
\includegraphics[width=0.056\textwidth,valign=c]{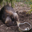}
 & 
\includegraphics[width=0.056\textwidth,valign=c]{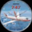}
 & 
\includegraphics[width=0.056\textwidth,valign=c]{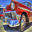}
 & 
\includegraphics[width=0.056\textwidth,valign=c]{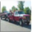}
 & 
\includegraphics[width=0.056\textwidth,valign=c]{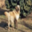}
 & 
\includegraphics[width=0.056\textwidth,valign=c]{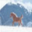}
 & 
\includegraphics[width=0.056\textwidth,valign=c]{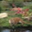}
 & 
\includegraphics[width=0.056\textwidth,valign=c]{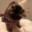}
 & 
\hspace{ 0.028 \textwidth} & 
\includegraphics[width=0.056\textwidth,valign=c]{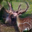}
 & 
\includegraphics[width=0.056\textwidth,valign=c]{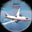}
 & 
\includegraphics[width=0.056\textwidth,valign=c]{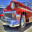}
 & 
\includegraphics[width=0.056\textwidth,valign=c]{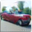}
 & 
\includegraphics[width=0.056\textwidth,valign=c]{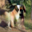}
 & 
\includegraphics[width=0.056\textwidth,valign=c]{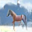}
 & 
\includegraphics[width=0.056\textwidth,valign=c]{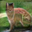}
 & 
\includegraphics[width=0.056\textwidth,valign=c]{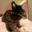}
\\
\rowfont{\tiny}
cat & dog & cat & deer & cat & car & plane & dog & & cat & dog & cat & deer & cat & car & plane & dog\\
\includegraphics[width=0.056\textwidth,valign=c]{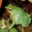}
 & 
\includegraphics[width=0.056\textwidth,valign=c]{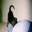}
 & 
\includegraphics[width=0.056\textwidth,valign=c]{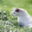}
 & 
\includegraphics[width=0.056\textwidth,valign=c]{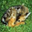}
 & 
\includegraphics[width=0.056\textwidth,valign=c]{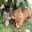}
 & 
\includegraphics[width=0.056\textwidth,valign=c]{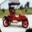}
 & 
\includegraphics[width=0.056\textwidth,valign=c]{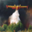}
 & 
\includegraphics[width=0.056\textwidth,valign=c]{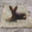}
 & 
\hspace{ 0.028 \textwidth} & 
\includegraphics[width=0.056\textwidth,valign=c]{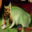}
 & 
\includegraphics[width=0.056\textwidth,valign=c]{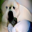}
 & 
\includegraphics[width=0.056\textwidth,valign=c]{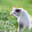}
 & 
\includegraphics[width=0.056\textwidth,valign=c]{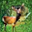}
 & 
\includegraphics[width=0.056\textwidth,valign=c]{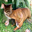}
 & 
\includegraphics[width=0.056\textwidth,valign=c]{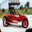}
 & 
\includegraphics[width=0.056\textwidth,valign=c]{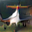}
 & 
\includegraphics[width=0.056\textwidth,valign=c]{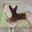}
\vspace{2mm}\\
\multicolumn{8}{c}{JEM-0} & & \multicolumn{8}{c}{AT-0.5}\\
\includegraphics[width=0.056\textwidth,valign=c]{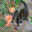}
 & 
\includegraphics[width=0.056\textwidth,valign=c]{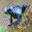}
 & 
\includegraphics[width=0.056\textwidth,valign=c]{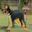}
 & 
\includegraphics[width=0.056\textwidth,valign=c]{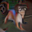}
 & 
\includegraphics[width=0.056\textwidth,valign=c]{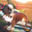}
 & 
\includegraphics[width=0.056\textwidth,valign=c]{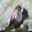}
 & 
\includegraphics[width=0.056\textwidth,valign=c]{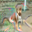}
 & 
\includegraphics[width=0.056\textwidth,valign=c]{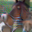}
 & 
\hspace{ 0.028 \textwidth} & 
\includegraphics[width=0.056\textwidth,valign=c]{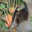}
 & 
\includegraphics[width=0.056\textwidth,valign=c]{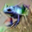}
 & 
\includegraphics[width=0.056\textwidth,valign=c]{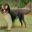}
 & 
\includegraphics[width=0.056\textwidth,valign=c]{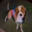}
 & 
\includegraphics[width=0.056\textwidth,valign=c]{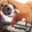}
 & 
\includegraphics[width=0.056\textwidth,valign=c]{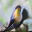}
 & 
\includegraphics[width=0.056\textwidth,valign=c]{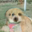}
 & 
\includegraphics[width=0.056\textwidth,valign=c]{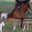}
\vspace{1mm}\\
\includegraphics[width=0.056\textwidth,valign=c]{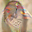}
 & 
\includegraphics[width=0.056\textwidth,valign=c]{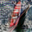}
 & 
\includegraphics[width=0.056\textwidth,valign=c]{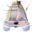}
 & 
\includegraphics[width=0.056\textwidth,valign=c]{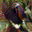}
 & 
\includegraphics[width=0.056\textwidth,valign=c]{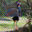}
 & 
\includegraphics[width=0.056\textwidth,valign=c]{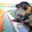}
 & 
\includegraphics[width=0.056\textwidth,valign=c]{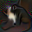}
 & 
\includegraphics[width=0.056\textwidth,valign=c]{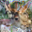}
 & 
\hspace{ 0.028 \textwidth} & 
\includegraphics[width=0.056\textwidth,valign=c]{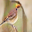}
 & 
\includegraphics[width=0.056\textwidth,valign=c]{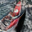}
 & 
\includegraphics[width=0.056\textwidth,valign=c]{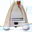}
 & 
\includegraphics[width=0.056\textwidth,valign=c]{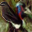}
 & 
\includegraphics[width=0.056\textwidth,valign=c]{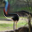}
 & 
\includegraphics[width=0.056\textwidth,valign=c]{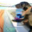}
 & 
\includegraphics[width=0.056\textwidth,valign=c]{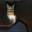}
 & 
\includegraphics[width=0.056\textwidth,valign=c]{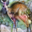}
\vspace{1mm}\\
\includegraphics[width=0.056\textwidth,valign=c]{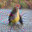}
 & 
\includegraphics[width=0.056\textwidth,valign=c]{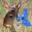}
 & 
\includegraphics[width=0.056\textwidth,valign=c]{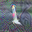}
 & 
\includegraphics[width=0.056\textwidth,valign=c]{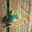}
 & 
\includegraphics[width=0.056\textwidth,valign=c]{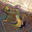}
 & 
\includegraphics[width=0.056\textwidth,valign=c]{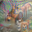}
 & 
\includegraphics[width=0.056\textwidth,valign=c]{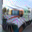}
 & 
\includegraphics[width=0.056\textwidth,valign=c]{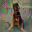}
 & 
\hspace{ 0.028 \textwidth} & 
\includegraphics[width=0.056\textwidth,valign=c]{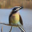}
 & 
\includegraphics[width=0.056\textwidth,valign=c]{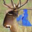}
 & 
\includegraphics[width=0.056\textwidth,valign=c]{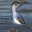}
 & 
\includegraphics[width=0.056\textwidth,valign=c]{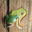}
 & 
\includegraphics[width=0.056\textwidth,valign=c]{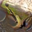}
 & 
\includegraphics[width=0.056\textwidth,valign=c]{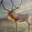}
 & 
\includegraphics[width=0.056\textwidth,valign=c]{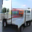}
 & 
\includegraphics[width=0.056\textwidth,valign=c]{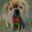}
\vspace{1mm}\\
\includegraphics[width=0.056\textwidth,valign=c]{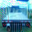}
 & 
\includegraphics[width=0.056\textwidth,valign=c]{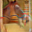}
 & 
\includegraphics[width=0.056\textwidth,valign=c]{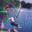}
 & 
\includegraphics[width=0.056\textwidth,valign=c]{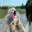}
 & 
\includegraphics[width=0.056\textwidth,valign=c]{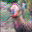}
 & 
\includegraphics[width=0.056\textwidth,valign=c]{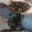}
 & 
\includegraphics[width=0.056\textwidth,valign=c]{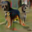}
 & 
\includegraphics[width=0.056\textwidth,valign=c]{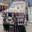}
 & 
\hspace{ 0.028 \textwidth} & 
\includegraphics[width=0.056\textwidth,valign=c]{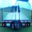}
 & 
\includegraphics[width=0.056\textwidth,valign=c]{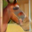}
 & 
\includegraphics[width=0.056\textwidth,valign=c]{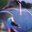}
 & 
\includegraphics[width=0.056\textwidth,valign=c]{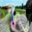}
 & 
\includegraphics[width=0.056\textwidth,valign=c]{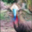}
 & 
\includegraphics[width=0.056\textwidth,valign=c]{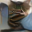}
 & 
\includegraphics[width=0.056\textwidth,valign=c]{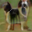}
 & 
\includegraphics[width=0.056\textwidth,valign=c]{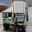}
\vspace{1mm}\\
\includegraphics[width=0.056\textwidth,valign=c]{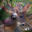}
 & 
\includegraphics[width=0.056\textwidth,valign=c]{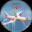}
 & 
\includegraphics[width=0.056\textwidth,valign=c]{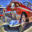}
 & 
\includegraphics[width=0.056\textwidth,valign=c]{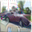}
 & 
\includegraphics[width=0.056\textwidth,valign=c]{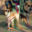}
 & 
\includegraphics[width=0.056\textwidth,valign=c]{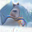}
 & 
\includegraphics[width=0.056\textwidth,valign=c]{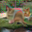}
 & 
\includegraphics[width=0.056\textwidth,valign=c]{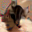}
 & 
\hspace{ 0.028 \textwidth} & 
\includegraphics[width=0.056\textwidth,valign=c]{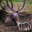}
 & 
\includegraphics[width=0.056\textwidth,valign=c]{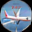}
 & 
\includegraphics[width=0.056\textwidth,valign=c]{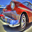}
 & 
\includegraphics[width=0.056\textwidth,valign=c]{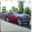}
 & 
\includegraphics[width=0.056\textwidth,valign=c]{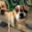}
 & 
\includegraphics[width=0.056\textwidth,valign=c]{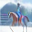}
 & 
\includegraphics[width=0.056\textwidth,valign=c]{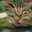}
 & 
\includegraphics[width=0.056\textwidth,valign=c]{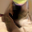}
\vspace{1mm}\\
\includegraphics[width=0.056\textwidth,valign=c]{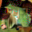}
 & 
\includegraphics[width=0.056\textwidth,valign=c]{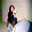}
 & 
\includegraphics[width=0.056\textwidth,valign=c]{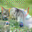}
 & 
\includegraphics[width=0.056\textwidth,valign=c]{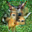}
 & 
\includegraphics[width=0.056\textwidth,valign=c]{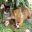}
 & 
\includegraphics[width=0.056\textwidth,valign=c]{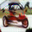}
 & 
\includegraphics[width=0.056\textwidth,valign=c]{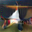}
 & 
\includegraphics[width=0.056\textwidth,valign=c]{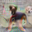}
 & 
\hspace{ 0.028 \textwidth} & 
\includegraphics[width=0.056\textwidth,valign=c]{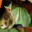}
 & 
\includegraphics[width=0.056\textwidth,valign=c]{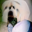}
 & 
\includegraphics[width=0.056\textwidth,valign=c]{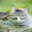}
 & 
\includegraphics[width=0.056\textwidth,valign=c]{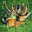}
 & 
\includegraphics[width=0.056\textwidth,valign=c]{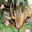}
 & 
\includegraphics[width=0.056\textwidth,valign=c]{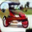}
 & 
\includegraphics[width=0.056\textwidth,valign=c]{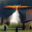}
 & 
\includegraphics[width=0.056\textwidth,valign=c]{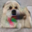}
\vspace{2mm}\\
\multicolumn{8}{c}{RATIO-0.5} & & \multicolumn{8}{c}{RATIO-0.25}\\
\includegraphics[width=0.056\textwidth,valign=c]{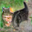}
 & 
\includegraphics[width=0.056\textwidth,valign=c]{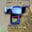}
 & 
\includegraphics[width=0.056\textwidth,valign=c]{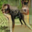}
 & 
\includegraphics[width=0.056\textwidth,valign=c]{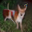}
 & 
\includegraphics[width=0.056\textwidth,valign=c]{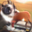}
 & 
\includegraphics[width=0.056\textwidth,valign=c]{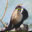}
 & 
\includegraphics[width=0.056\textwidth,valign=c]{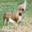}
 & 
\includegraphics[width=0.056\textwidth,valign=c]{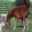}
 & 
\hspace{ 0.028 \textwidth} & 
\includegraphics[width=0.056\textwidth,valign=c]{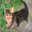}
 & 
\includegraphics[width=0.056\textwidth,valign=c]{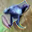}
 & 
\includegraphics[width=0.056\textwidth,valign=c]{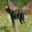}
 & 
\includegraphics[width=0.056\textwidth,valign=c]{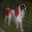}
 & 
\includegraphics[width=0.056\textwidth,valign=c]{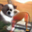}
 & 
\includegraphics[width=0.056\textwidth,valign=c]{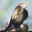}
 & 
\includegraphics[width=0.056\textwidth,valign=c]{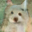}
 & 
\includegraphics[width=0.056\textwidth,valign=c]{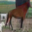}
\vspace{1mm}\\
\includegraphics[width=0.056\textwidth,valign=c]{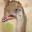}
 & 
\includegraphics[width=0.056\textwidth,valign=c]{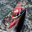}
 & 
\includegraphics[width=0.056\textwidth,valign=c]{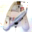}
 & 
\includegraphics[width=0.056\textwidth,valign=c]{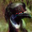}
 & 
\includegraphics[width=0.056\textwidth,valign=c]{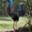}
 & 
\includegraphics[width=0.056\textwidth,valign=c]{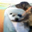}
 & 
\includegraphics[width=0.056\textwidth,valign=c]{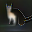}
 & 
\includegraphics[width=0.056\textwidth,valign=c]{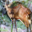}
 & 
\hspace{ 0.028 \textwidth} & 
\includegraphics[width=0.056\textwidth,valign=c]{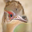}
 & 
\includegraphics[width=0.056\textwidth,valign=c]{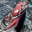}
 & 
\includegraphics[width=0.056\textwidth,valign=c]{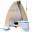}
 & 
\includegraphics[width=0.056\textwidth,valign=c]{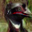}
 & 
\includegraphics[width=0.056\textwidth,valign=c]{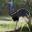}
 & 
\includegraphics[width=0.056\textwidth,valign=c]{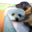}
 & 
\includegraphics[width=0.056\textwidth,valign=c]{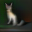}
 & 
\includegraphics[width=0.056\textwidth,valign=c]{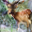}
\vspace{1mm}\\
\includegraphics[width=0.056\textwidth,valign=c]{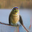}
 & 
\includegraphics[width=0.056\textwidth,valign=c]{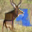}
 & 
\includegraphics[width=0.056\textwidth,valign=c]{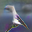}
 & 
\includegraphics[width=0.056\textwidth,valign=c]{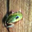}
 & 
\includegraphics[width=0.056\textwidth,valign=c]{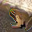}
 & 
\includegraphics[width=0.056\textwidth,valign=c]{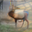}
 & 
\includegraphics[width=0.056\textwidth,valign=c]{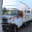}
 & 
\includegraphics[width=0.056\textwidth,valign=c]{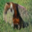}
 & 
\hspace{ 0.028 \textwidth} & 
\includegraphics[width=0.056\textwidth,valign=c]{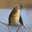}
 & 
\includegraphics[width=0.056\textwidth,valign=c]{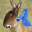}
 & 
\includegraphics[width=0.056\textwidth,valign=c]{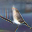}
 & 
\includegraphics[width=0.056\textwidth,valign=c]{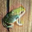}
 & 
\includegraphics[width=0.056\textwidth,valign=c]{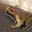}
 & 
\includegraphics[width=0.056\textwidth,valign=c]{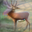}
 & 
\includegraphics[width=0.056\textwidth,valign=c]{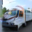}
 & 
\includegraphics[width=0.056\textwidth,valign=c]{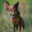}
\vspace{1mm}\\
\includegraphics[width=0.056\textwidth,valign=c]{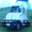}
 & 
\includegraphics[width=0.056\textwidth,valign=c]{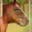}
 & 
\includegraphics[width=0.056\textwidth,valign=c]{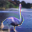}
 & 
\includegraphics[width=0.056\textwidth,valign=c]{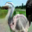}
 & 
\includegraphics[width=0.056\textwidth,valign=c]{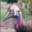}
 & 
\includegraphics[width=0.056\textwidth,valign=c]{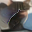}
 & 
\includegraphics[width=0.056\textwidth,valign=c]{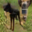}
 & 
\includegraphics[width=0.056\textwidth,valign=c]{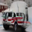}
 & 
\hspace{ 0.028 \textwidth} & 
\includegraphics[width=0.056\textwidth,valign=c]{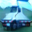}
 & 
\includegraphics[width=0.056\textwidth,valign=c]{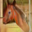}
 & 
\includegraphics[width=0.056\textwidth,valign=c]{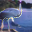}
 & 
\includegraphics[width=0.056\textwidth,valign=c]{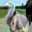}
 & 
\includegraphics[width=0.056\textwidth,valign=c]{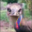}
 & 
\includegraphics[width=0.056\textwidth,valign=c]{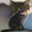}
 & 
\includegraphics[width=0.056\textwidth,valign=c]{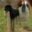}
 & 
\includegraphics[width=0.056\textwidth,valign=c]{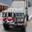}
\vspace{1mm}\\
\includegraphics[width=0.056\textwidth,valign=c]{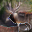}
 & 
\includegraphics[width=0.056\textwidth,valign=c]{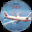}
 & 
\includegraphics[width=0.056\textwidth,valign=c]{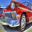}
 & 
\includegraphics[width=0.056\textwidth,valign=c]{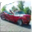}
 & 
\includegraphics[width=0.056\textwidth,valign=c]{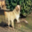}
 & 
\includegraphics[width=0.056\textwidth,valign=c]{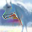}
 & 
\includegraphics[width=0.056\textwidth,valign=c]{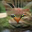}
 & 
\includegraphics[width=0.056\textwidth,valign=c]{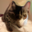}
 & 
\hspace{ 0.028 \textwidth} & 
\includegraphics[width=0.056\textwidth,valign=c]{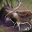}
 & 
\includegraphics[width=0.056\textwidth,valign=c]{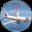}
 & 
\includegraphics[width=0.056\textwidth,valign=c]{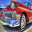}
 & 
\includegraphics[width=0.056\textwidth,valign=c]{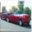}
 & 
\includegraphics[width=0.056\textwidth,valign=c]{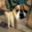}
 & 
\includegraphics[width=0.056\textwidth,valign=c]{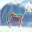}
 & 
\includegraphics[width=0.056\textwidth,valign=c]{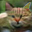}
 & 
\includegraphics[width=0.056\textwidth,valign=c]{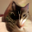}
\vspace{1mm}\\
\includegraphics[width=0.056\textwidth,valign=c]{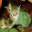}
 & 
\includegraphics[width=0.056\textwidth,valign=c]{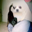}
 & 
\includegraphics[width=0.056\textwidth,valign=c]{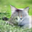}
 & 
\includegraphics[width=0.056\textwidth,valign=c]{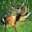}
 & 
\includegraphics[width=0.056\textwidth,valign=c]{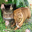}
 & 
\includegraphics[width=0.056\textwidth,valign=c]{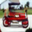}
 & 
\includegraphics[width=0.056\textwidth,valign=c]{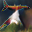}
 & 
\includegraphics[width=0.056\textwidth,valign=c]{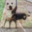}
 & 
\hspace{ 0.028 \textwidth} & 
\includegraphics[width=0.056\textwidth,valign=c]{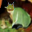}
 & 
\includegraphics[width=0.056\textwidth,valign=c]{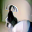}
 & 
\includegraphics[width=0.056\textwidth,valign=c]{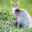}
 & 
\includegraphics[width=0.056\textwidth,valign=c]{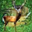}
 & 
\includegraphics[width=0.056\textwidth,valign=c]{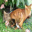}
 & 
\includegraphics[width=0.056\textwidth,valign=c]{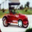}
 & 
\includegraphics[width=0.056\textwidth,valign=c]{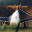}
 & 
\includegraphics[width=0.056\textwidth,valign=c]{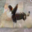}
\vspace{2mm}\\
\end{tabu}
\end{adjustbox}
\caption{\label{fig:cifar_overview}\textbf{More Visual Counterfactuals:} Random selection of  48 CIFAR10 test images misclassified by all models (top left shows original images together with ground truth labels) and the associated visual counterfactuals that are generated by maximizing the confidence in the ground truth class in a $l_2$ ball of radius 3. 
Note that some images like the frog on the left in the bottom row are clearly mislabeled.}\label{Fig:CIFAR10_OVERVIEW}
\end{figure}

%% file: res/appendix_od_cifar_main_paper_extended.tex
\begin{figure}
\begin{tabular}{p{1cm}x{\breite}x{\breite}x{\breite}x{\breite}x{\breite}x{\breite}x{\breite}x{\breite}}
Model  & Orig. & $\epsilon=0.5$ & $\epsilon=1.0$ & $\epsilon=1.5$ & $\epsilon=2.0$ & $\epsilon=2.5$ & $\epsilon=3.0$\\
\begin{turn}{90} \hspace{-.4cm} Plain \end{turn} & \multicolumn{7}{c}{
\includegraphics[width=0.91\textwidth,valign=c]{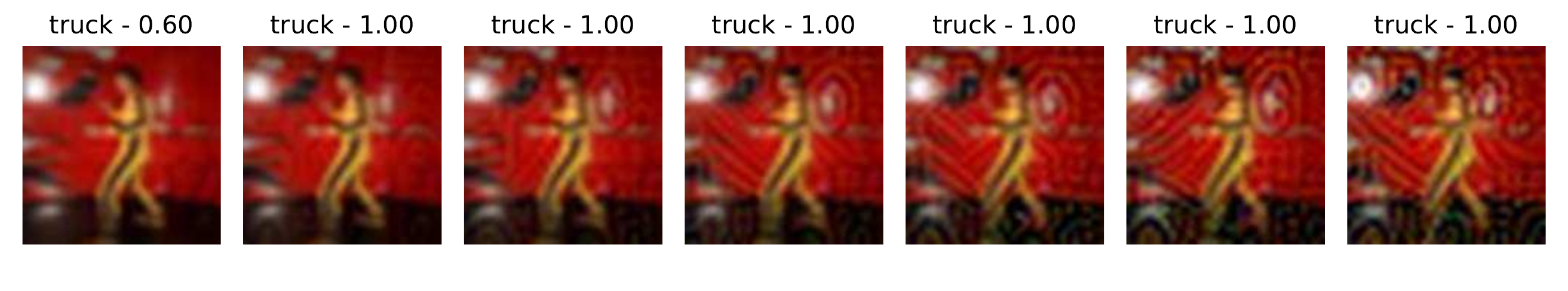}
} \\
\hline
\begin{turn}{90} \hspace{-.4cm} OE \end{turn} & \multicolumn{7}{c}{
\includegraphics[width=0.91\textwidth,valign=c]{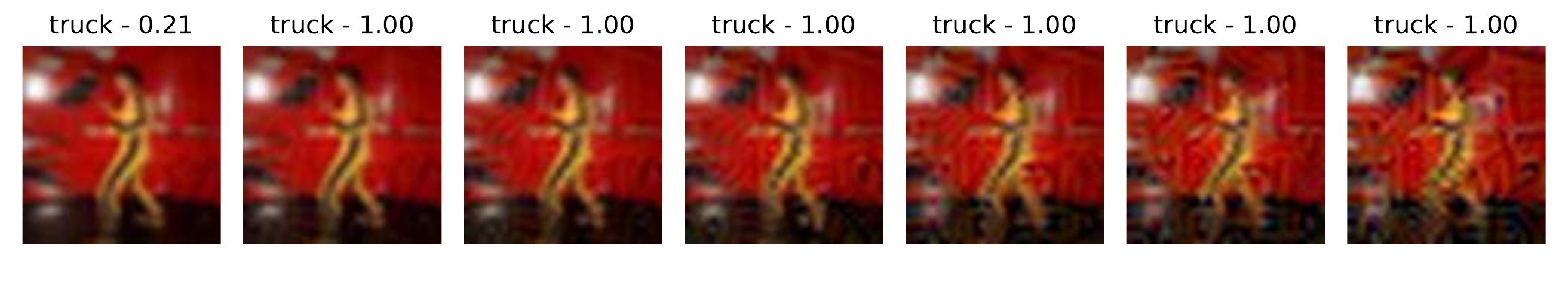}
} \\
\hline
\begin{turn}{90} \hspace{-.4cm} ACET \end{turn} & \multicolumn{7}{c}{
\includegraphics[width=0.91\textwidth,valign=c]{pics/CIFAR10/ACET/OD/img_97.pdf}
} \\
\hline
\begin{turn}{90} \hspace{-.4cm} M-0.50 \end{turn} & \multicolumn{7}{c}{
\includegraphics[width=0.91\textwidth,valign=c]{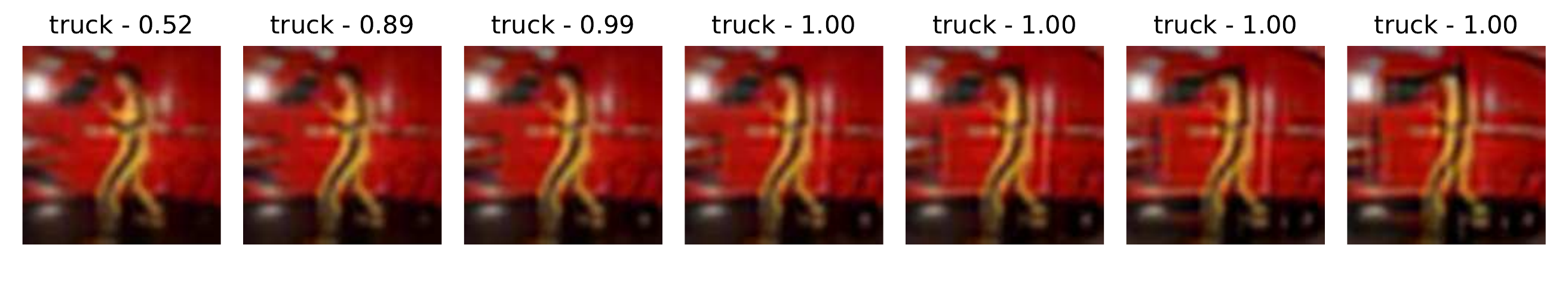}
} \\
\hline
\begin{turn}{90} \hspace{-.4cm} AT-0.50 \end{turn} & \multicolumn{7}{c}{
\includegraphics[width=0.91\textwidth,valign=c]{pics/CIFAR10/AT05/OD/img_97.pdf}
} \\
\hline
\begin{turn}{90} \hspace{-.4cm} AT-0.25 \end{turn} & \multicolumn{7}{c}{
\includegraphics[width=0.91\textwidth,valign=c]{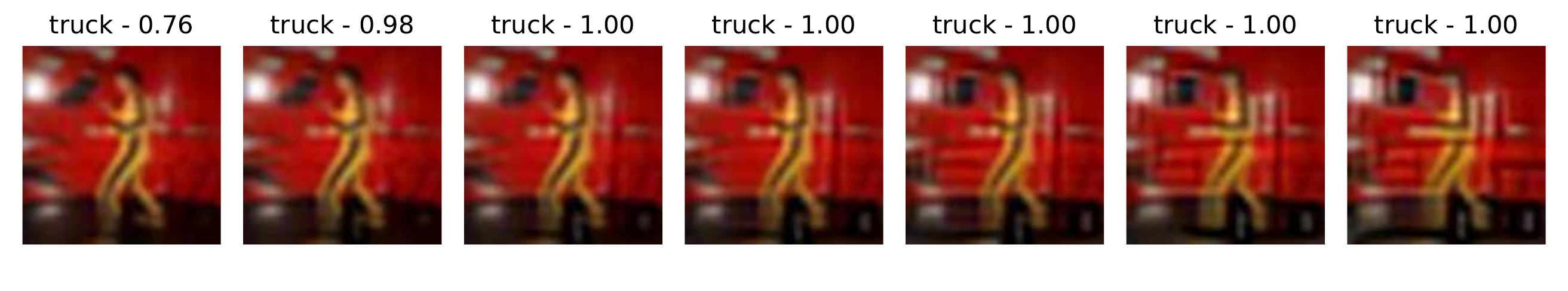}
} \\
\hline
\begin{turn}{90} \hspace{-.4cm} JEM-0 \end{turn} & \multicolumn{7}{c}{
\includegraphics[width=0.91\textwidth,valign=c]{pics/CIFAR10/EBM/OD/img_97.pdf}
} \\
\hline
\begin{turn}{90} \hspace{-.4cm} R-0.5 \end{turn} & \multicolumn{7}{c}{
\includegraphics[width=0.91\textwidth,valign=c]{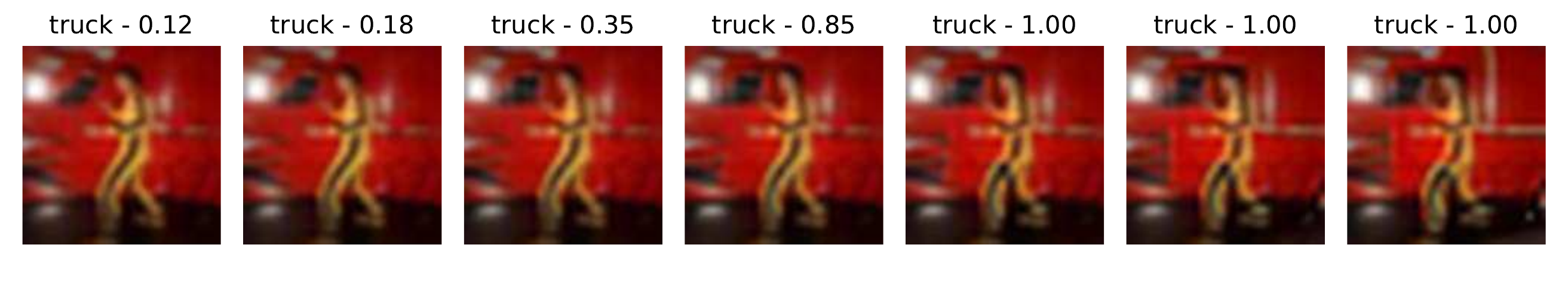}
} \\
\hline
\begin{turn}{90} \hspace{-.4cm} R-0.25 \end{turn} & \multicolumn{7}{c}{
\includegraphics[width=0.91\textwidth,valign=c]{pics/CIFAR10/Ratio025/OD/img_97.pdf}
} \\
\end{tabular}	
\vspace{-.5cm}
\caption{\label{fig:cifar_od_ext}\textbf{Feature Generation for out-distribution images:} Extension of Figure~\ref{Fig:CIFAR_OD} in the main paper to all models. }
\end{figure}

%% file: res/appendix_od_cifar_new.tex
\begin{figure}
\begin{tabular}{p{1cm}x{\breite}x{\breite}x{\breite}x{\breite}x{\breite}x{\breite}x{\breite}x{\breite}}
Model  & Orig. & $\epsilon=0.5$ & $\epsilon=1.0$ & $\epsilon=1.5$ & $\epsilon=2.0$ & $\epsilon=2.5$ & $\epsilon=3.0$\\
\begin{turn}{90} \hspace{-.4cm} ACET \end{turn} & \multicolumn{7}{c}{
\includegraphics[width=0.91\textwidth,valign=c]{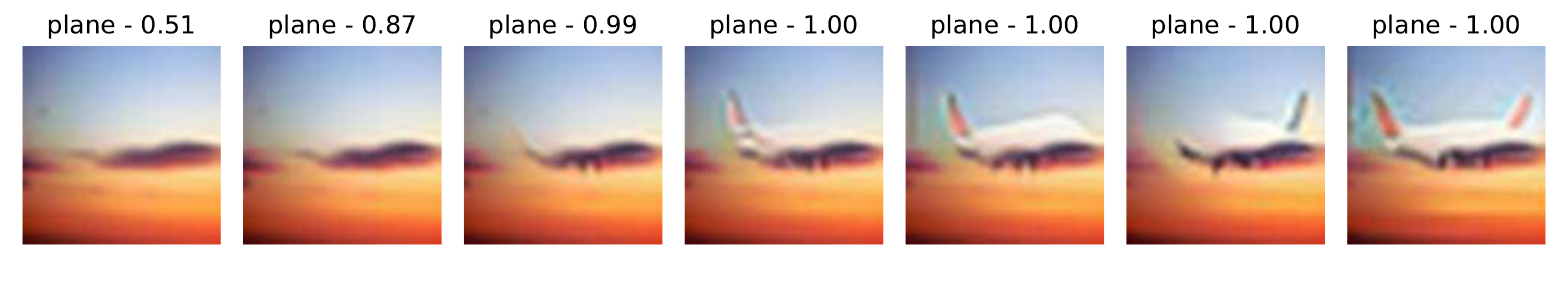}
} \\
\hline
\begin{turn}{90} \hspace{-.4cm} JEM-0 \end{turn} & \multicolumn{7}{c}{
\includegraphics[width=0.91\textwidth,valign=c]{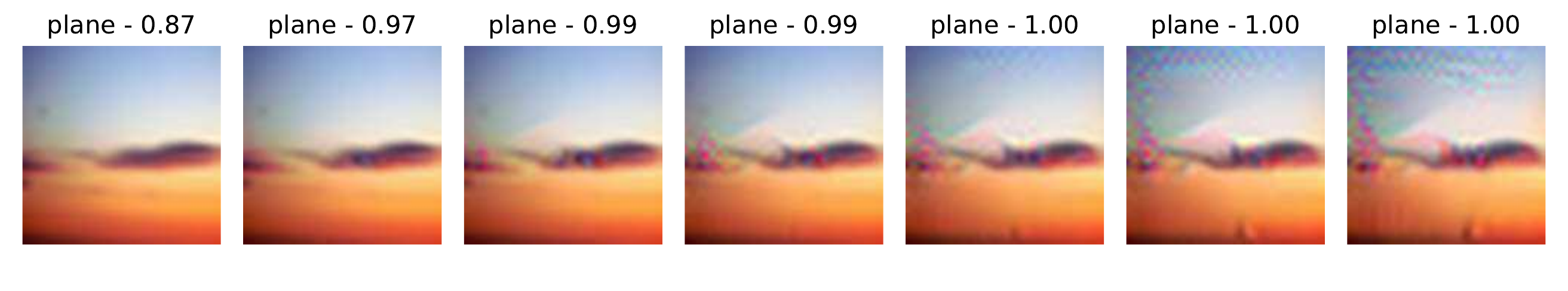}
} \\
\hline
\begin{turn}{90} \hspace{-.4cm} AT-0.50 \end{turn} & \multicolumn{7}{c}{
\includegraphics[width=0.91\textwidth,valign=c]{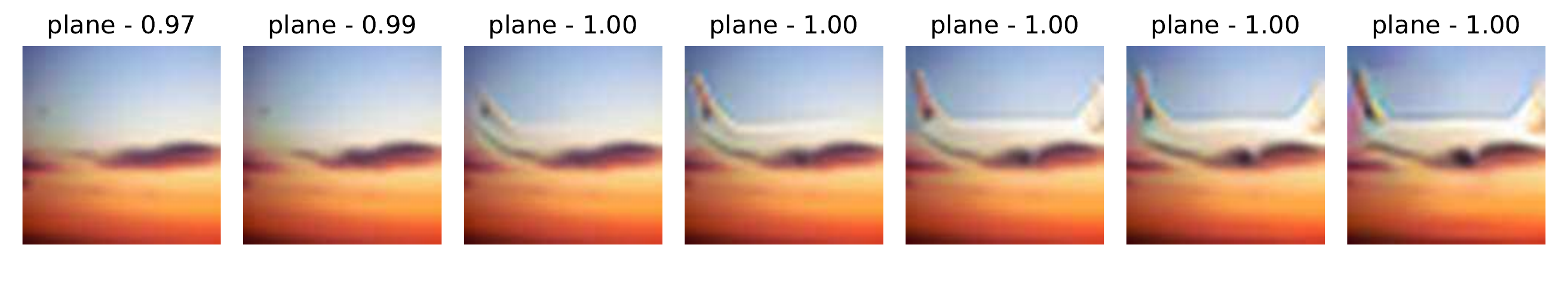}
} \\
\hline
\begin{turn}{90} \hspace{-.4cm} R-0.25 \end{turn} & \multicolumn{7}{c}{
\includegraphics[width=0.91\textwidth,valign=c]{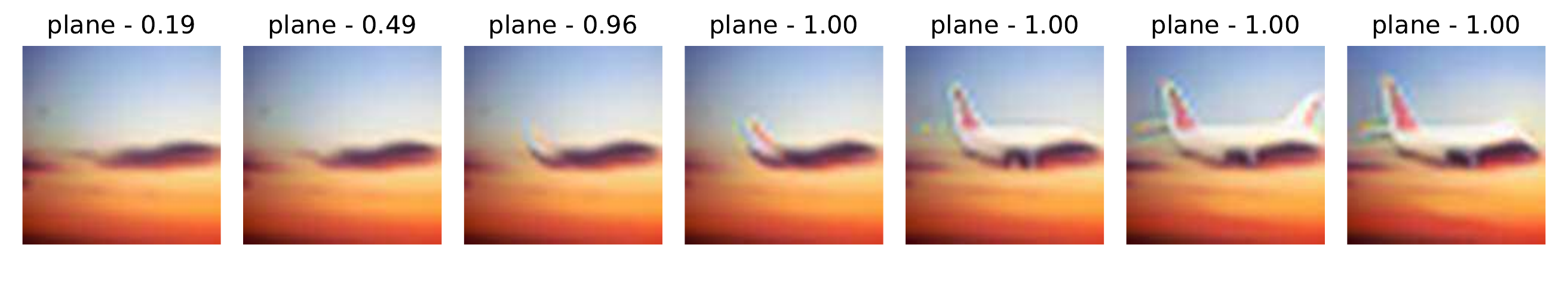}
} \\
\vspace{2mm}\\
\begin{turn}{90} \hspace{-.4cm} ACET \end{turn} & \multicolumn{7}{c}{
\includegraphics[width=0.91\textwidth,valign=c]{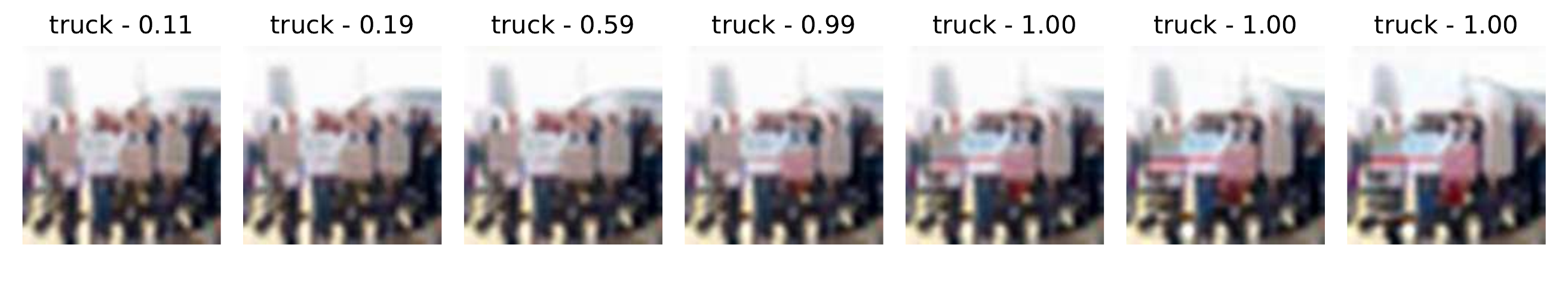}
} \\
\hline
\begin{turn}{90} \hspace{-.4cm} JEM-0 \end{turn} & \multicolumn{7}{c}{
\includegraphics[width=0.91\textwidth,valign=c]{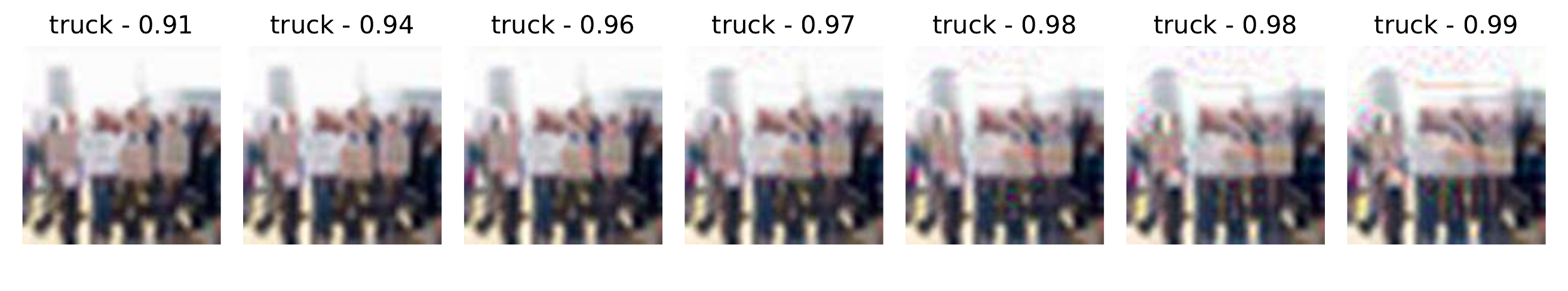}
} \\
\hline
\begin{turn}{90} \hspace{-.4cm} AT-0.50 \end{turn} & \multicolumn{7}{c}{
\includegraphics[width=0.91\textwidth,valign=c]{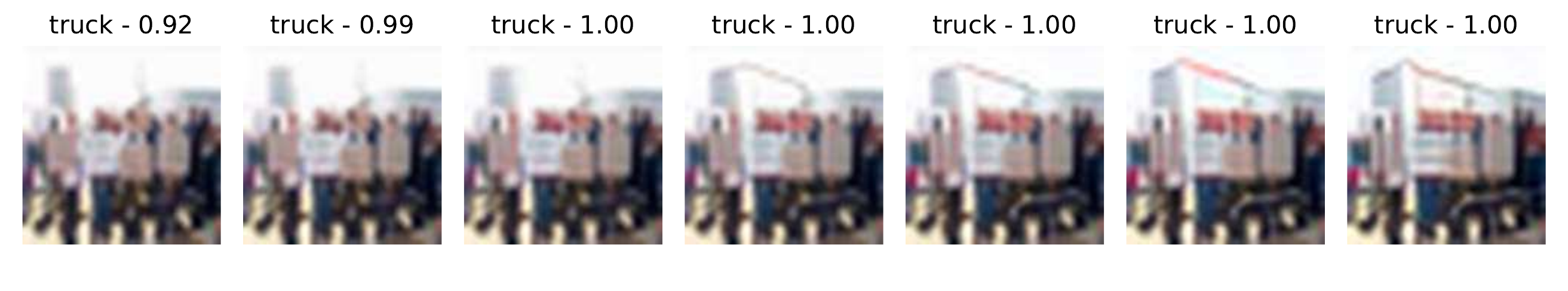}
} \\
\hline
\begin{turn}{90} \hspace{-.4cm} R-0.25 \end{turn} & \multicolumn{7}{c}{
\includegraphics[width=0.91\textwidth,valign=c]{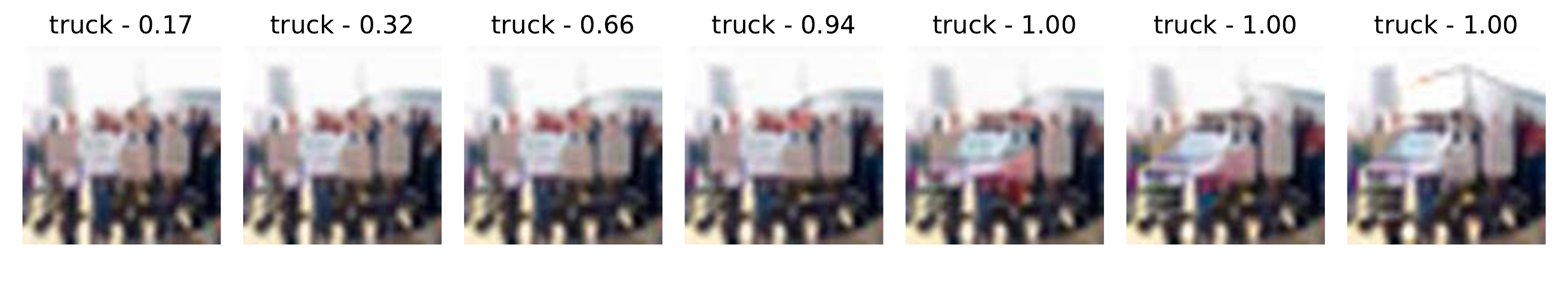}
} \\
\end{tabular}	
\vspace{-.5cm}
\caption{\label{fig:od_cifar_new1}\textbf{Feature Generation for OOD images:} for CIFAR10 models for unseen examples from 80 million tiny images. AT$_{0.5}$ is overconfident on the out-distribution samples, but produces high quality samples for sufficiently large radii. JEM-0 is also overly confident and generates its characteristic noise patterns. RATIO$_{0.25}$ produces high quality samples and the confidence increases as more class-specific features appear. }
\end{figure}

\begin{figure}
\begin{tabular}{p{1cm}x{\breite}x{\breite}x{\breite}x{\breite}x{\breite}x{\breite}x{\breite}x{\breite}}
Model  & Orig. & $\epsilon=0.5$ & $\epsilon=1.0$ & $\epsilon=1.5$ & $\epsilon=2.0$ & $\epsilon=2.5$ & $\epsilon=3.0$\\
\begin{turn}{90} \hspace{-.4cm} ACET \end{turn} & \multicolumn{7}{c}{
\includegraphics[width=0.91\textwidth,valign=c]{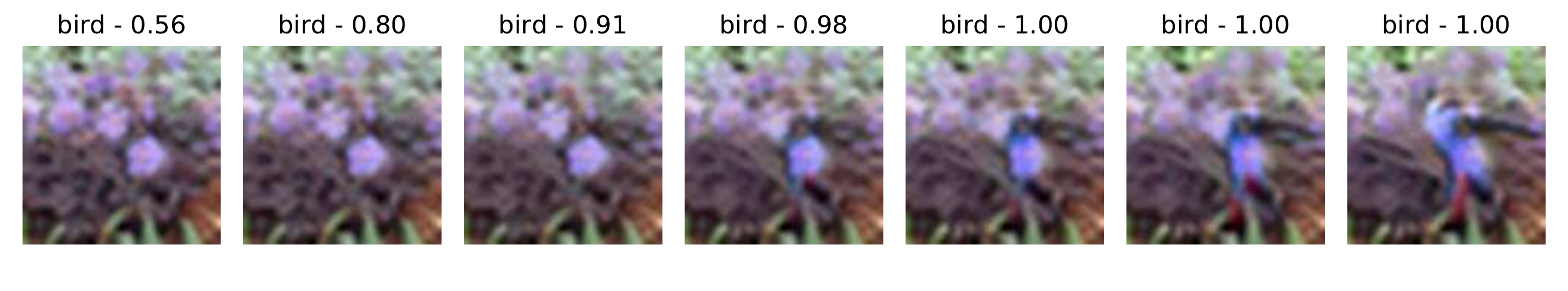}
} \\
\hline
\begin{turn}{90} \hspace{-.4cm} JEM-0 \end{turn} & \multicolumn{7}{c}{
\includegraphics[width=0.91\textwidth,valign=c]{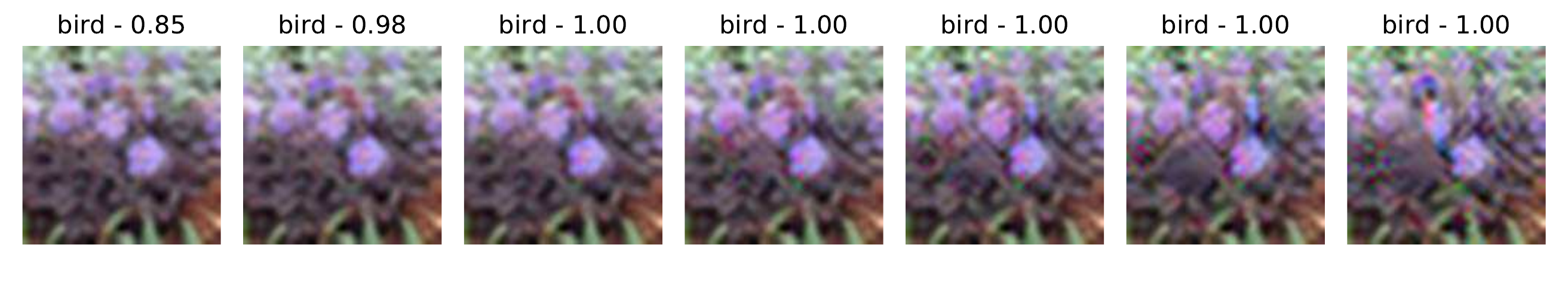}
} \\
\hline
\begin{turn}{90} \hspace{-.4cm} AT-0.50 \end{turn} & \multicolumn{7}{c}{
\includegraphics[width=0.91\textwidth,valign=c]{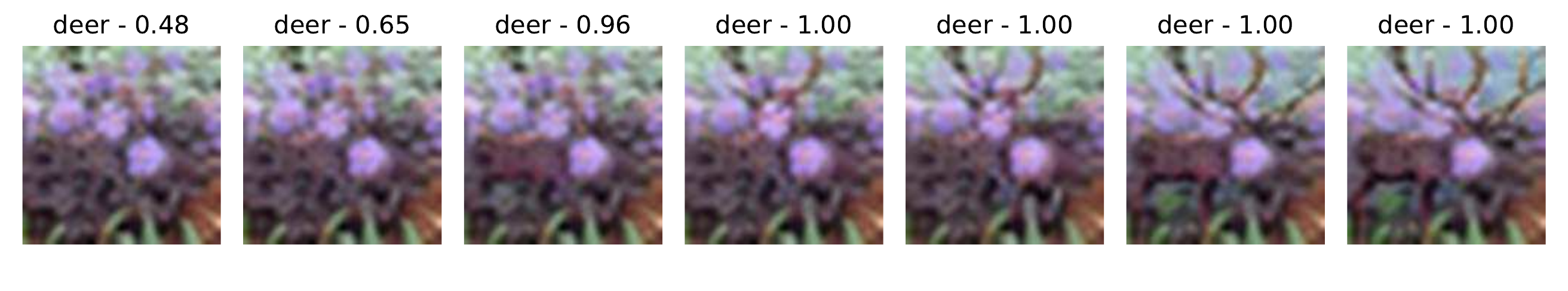}
} \\
\hline
\begin{turn}{90} \hspace{-.4cm} R-0.25 \end{turn} & \multicolumn{7}{c}{
\includegraphics[width=0.91\textwidth,valign=c]{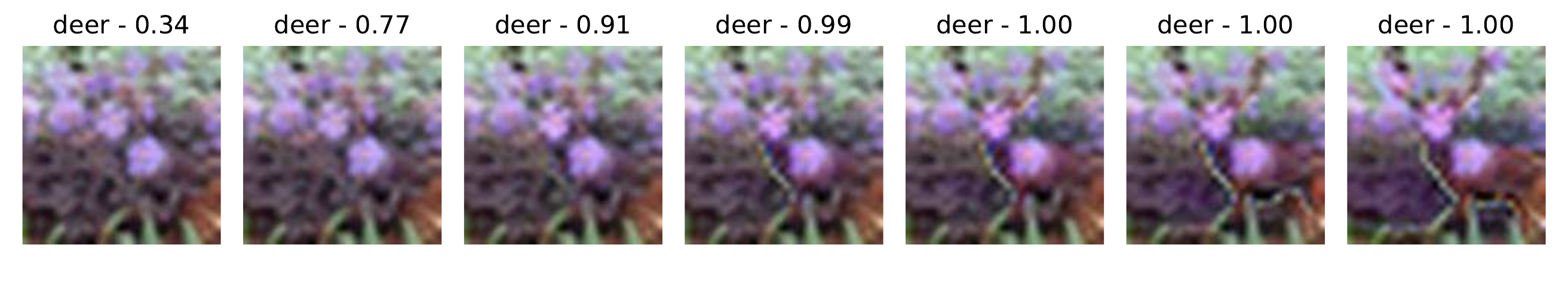}
} \\
\vspace{2mm}\\
\begin{turn}{90} \hspace{-.4cm} ACET \end{turn} & \multicolumn{7}{c}{
\includegraphics[width=0.91\textwidth,valign=c]{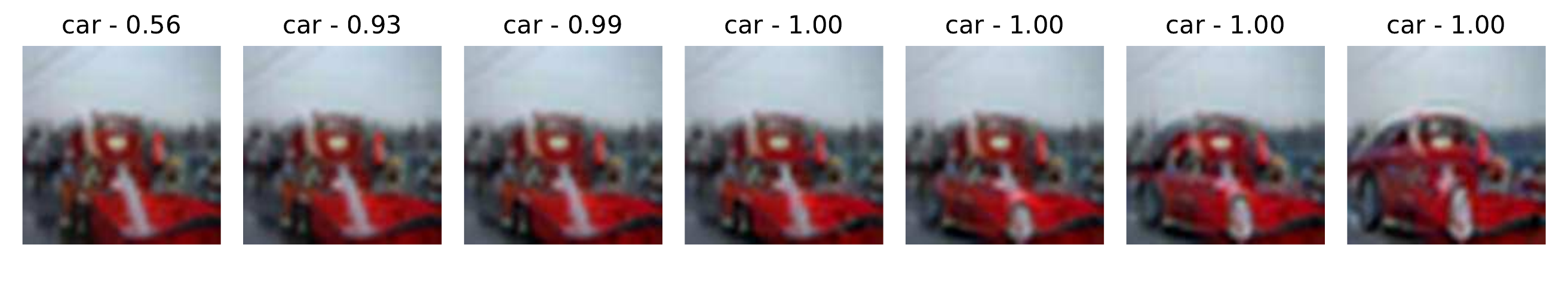}
} \\
\hline
\begin{turn}{90} \hspace{-.4cm} JEM-0 \end{turn} & \multicolumn{7}{c}{
\includegraphics[width=0.91\textwidth,valign=c]{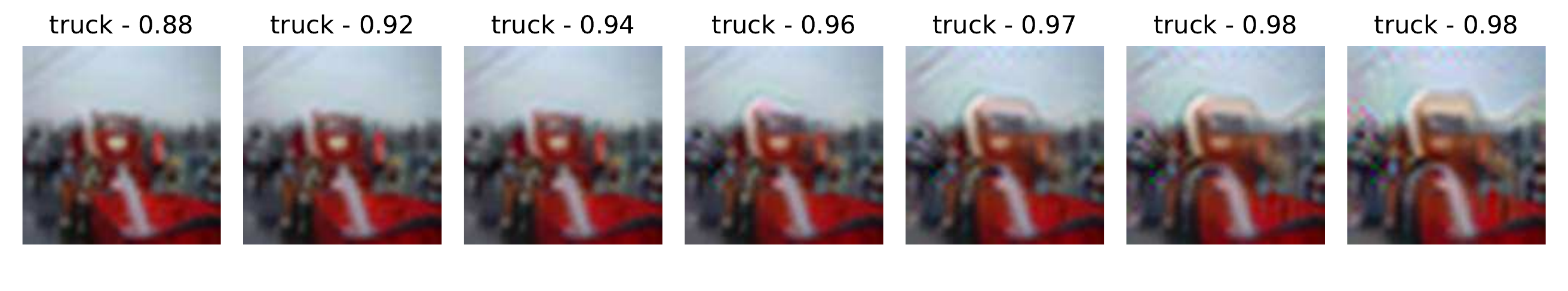}
} \\
\hline
\begin{turn}{90} \hspace{-.4cm} AT-0.50 \end{turn} & \multicolumn{7}{c}{
\includegraphics[width=0.91\textwidth,valign=c]{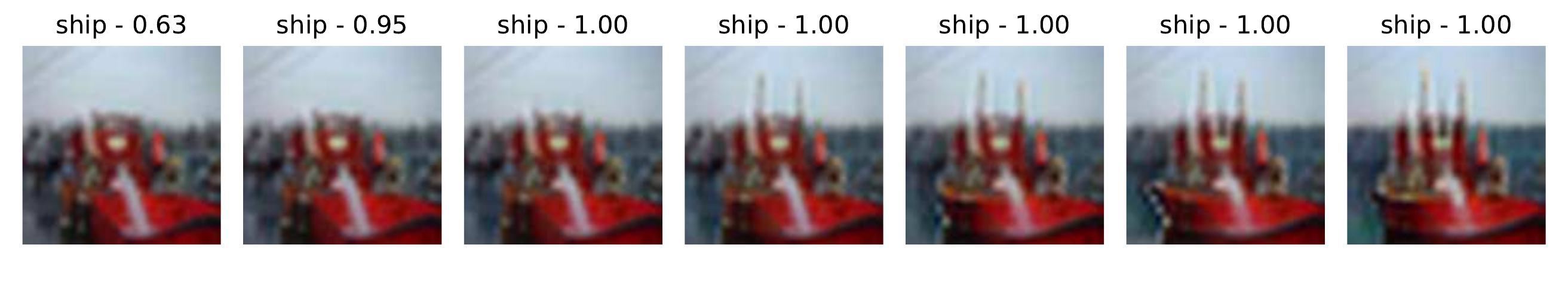}
} \\
\hline
\begin{turn}{90} \hspace{-.4cm} R-0.25 \end{turn} & \multicolumn{7}{c}{
\includegraphics[width=0.91\textwidth,valign=c]{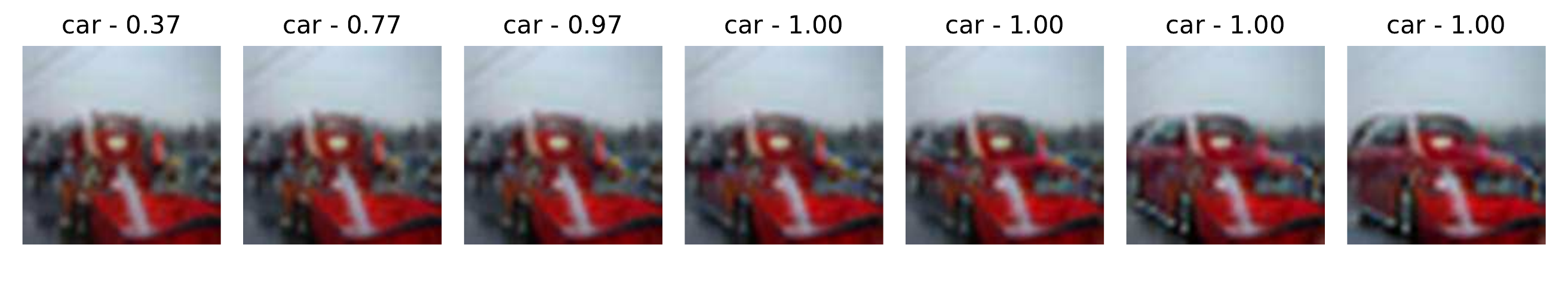}
} \\
\end{tabular}	
\vspace{-.5cm}
\caption{\label{fig:od_cifar_new2}\textbf{Feature Generation for OOD images:} for CIFAR10 models for unseen examples from 80 million tiny images. Despite the low robustness, ACET, AT$_{0.5}$ and RATIO$_{0.25}$ are able to generate meaningful class-specific features.}
\end{figure}

%% file: res/appendix_vc_svhn_main_paper_extended.tex
\begin{figure}[ht!]
\begin{tabular}{p{1cm}x{\breite}x{\breite}x{\breite}x{\breite}x{\breite}x{\breite}x{\breite}x{\breite}}
Model  & Orig. & $\epsilon=0.5$ & $\epsilon=1.0$ & $\epsilon=1.5$ & $\epsilon=2.0$ & $\epsilon=2.5$ & $\epsilon=3.0$\\
\begin{turn}{90} \hspace{-.4cm} Plain \end{turn} & \multicolumn{7}{c}{\includegraphics[width=0.91\textwidth,valign=c]{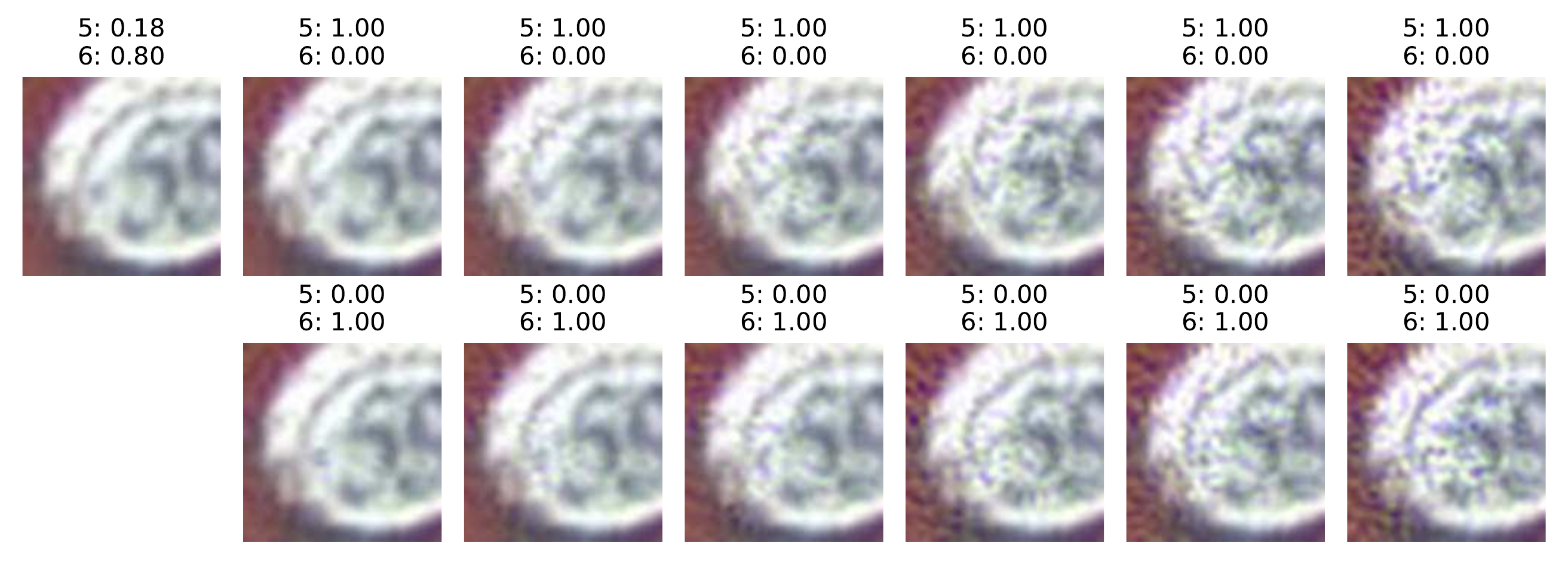}} \\
\hline
\begin{turn}{90} \hspace{-.4cm} OE \end{turn}  &  \multicolumn{7}{c}{\includegraphics[width=0.91\textwidth,valign=c]{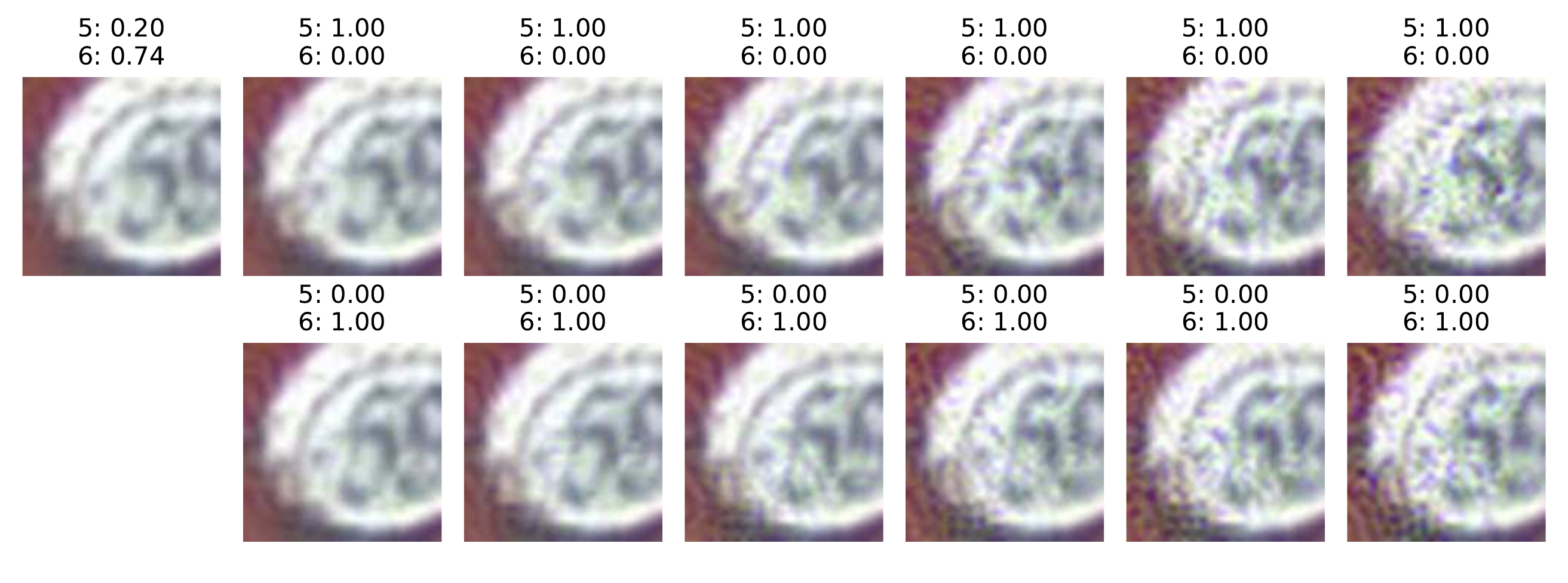}} \\
\hline
\begin{turn}{90} \hspace{-.4cm} ACET \end{turn} & \multicolumn{7}{c}{\includegraphics[width=0.91\textwidth,valign=c]{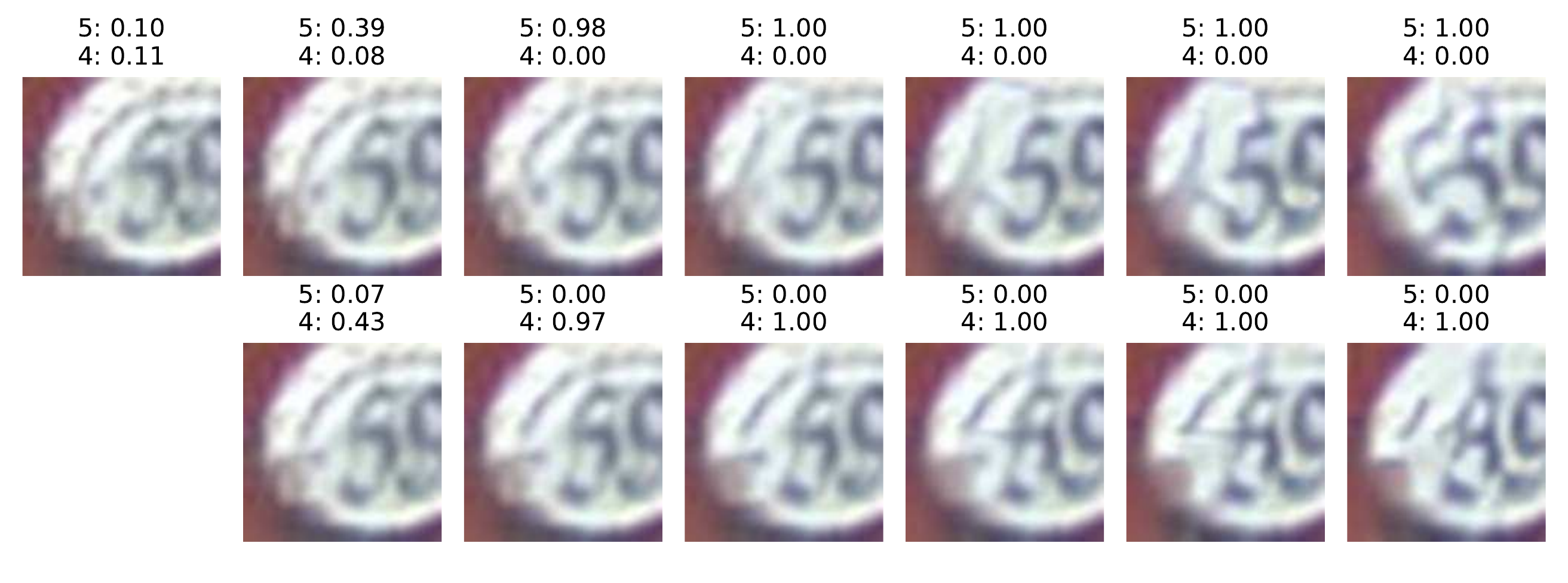}} \\
\end{tabular}	
\caption{\label{fig:vc_svhn_ext1} \textbf{Visual Counterfactuals} for all 7 SVHN models (extension of Figure~\ref{fig:cf_svhn} in the main paper). The test image (top left) has been misclassified by all models. Per model, we generate images which increase the confidence in the ground truth class (upper row) and falsely predicted class (bottom row) for varying $l_2$ budgets.  While non-robust models only generate noise patterns, the less robust ACET models produce higher quality images than AT$_{0.25}$. The samples from RATIO-0.25 show that adversarial out-distribution training clearly improves generative capabilities over AT$_{0.25}$ which has comparable in-distribution robustness. Continues on next page.
}
\vspace{-.3cm}
\end{figure}

\begin{figure}[ht!]
\begin{tabular}{p{1cm}x{\breite}x{\breite}x{\breite}x{\breite}x{\breite}x{\breite}x{\breite}x{\breite}}
Model  & Orig. & $\epsilon=0.5$ & $\epsilon=1.0$ & $\epsilon=1.5$ & $\epsilon=2.0$ & $\epsilon=2.5$ & $\epsilon=3.0$\\
\begin{turn}{90} \hspace{-.4cm} AT-0.50 \end{turn}  &  \multicolumn{7}{c}{\includegraphics[width=0.91\textwidth,valign=c]{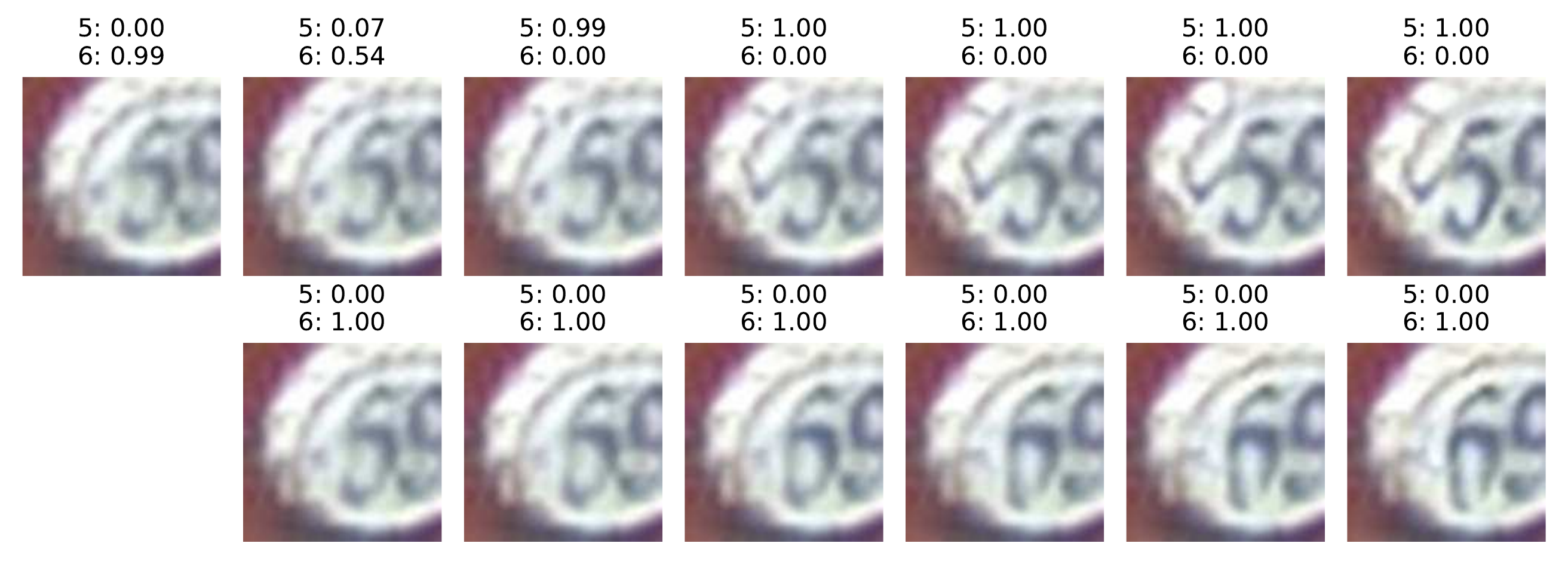}} \\
\hline
\begin{turn}{90} \hspace{-.4cm} AT-0.25 \end{turn}  &  \multicolumn{7}{c}{\includegraphics[width=0.91\textwidth,valign=c]{pics/SVHN/AT025/TestFailures/28.pdf}} \\
\hline
\begin{turn}{90} \hspace{-.4cm} RATIO-0.50 \end{turn} & \multicolumn{7}{c}{\includegraphics[width=0.91\textwidth,valign=c]{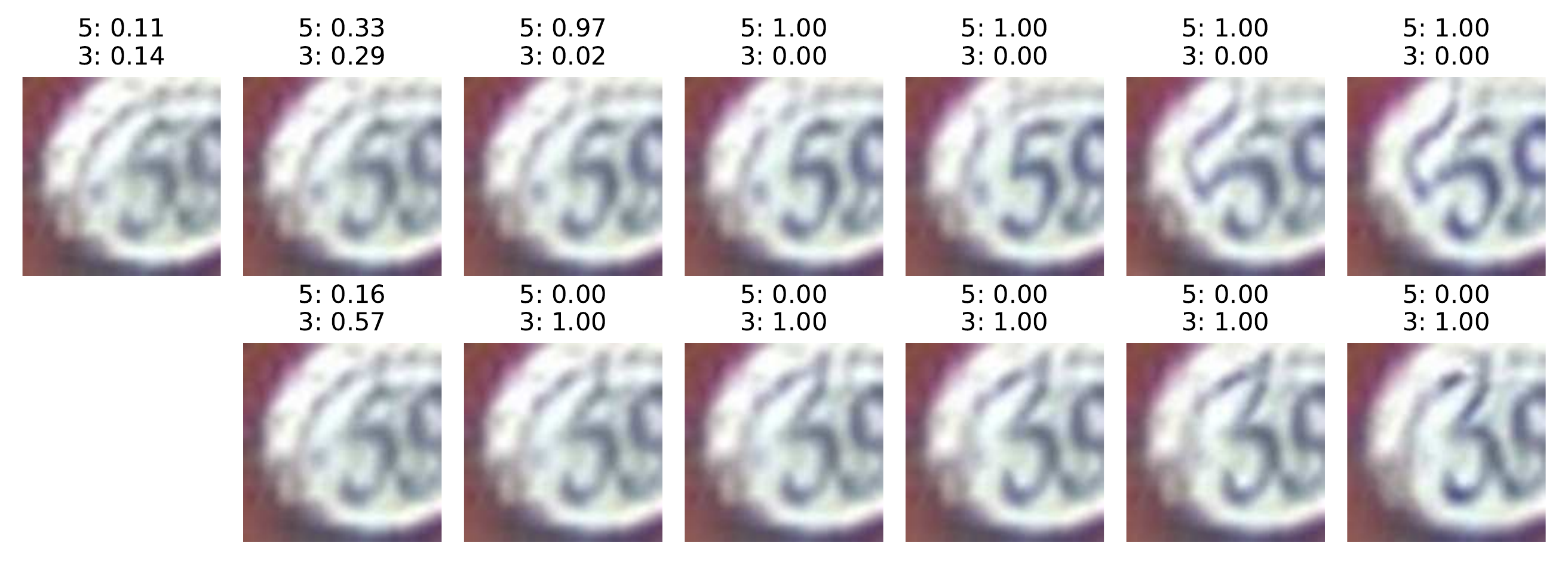}} \\
\hline
\begin{turn}{90} \hspace{-.4cm} RATIO-0.25 \end{turn} & \multicolumn{7}{c}{\includegraphics[width=0.91\textwidth,valign=c]{pics/SVHN/Ratio025/TestFailures/28.pdf}} \\
\end{tabular}	
\caption{\label{fig:vc_svhn_ext2}\textbf{Visual Counterfactuals} Figure \ref{fig:vc_svhn_ext1} continued, see page before for a description (extension of Figure \ref{fig:cf_svhn} in the main paper).
}
\vspace{-.3cm}
\end{figure}

%% file: res/appendix_vc_svhn_new.tex
\begin{figure}[ht!]
\begin{tabular}{p{1cm}x{\breite}x{\breite}x{\breite}x{\breite}x{\breite}x{\breite}x{\breite}x{\breite}}
Model  & Orig. & $\epsilon=0.5$ & $\epsilon=1.0$ & $\epsilon=1.5$ & $\epsilon=2.0$ & $\epsilon=2.5$ & $\epsilon=3.0$\\
\begin{turn}{90} \hspace{-.4cm} ACET \end{turn} & \multicolumn{7}{c}{\includegraphics[width=0.91\textwidth,valign=c]{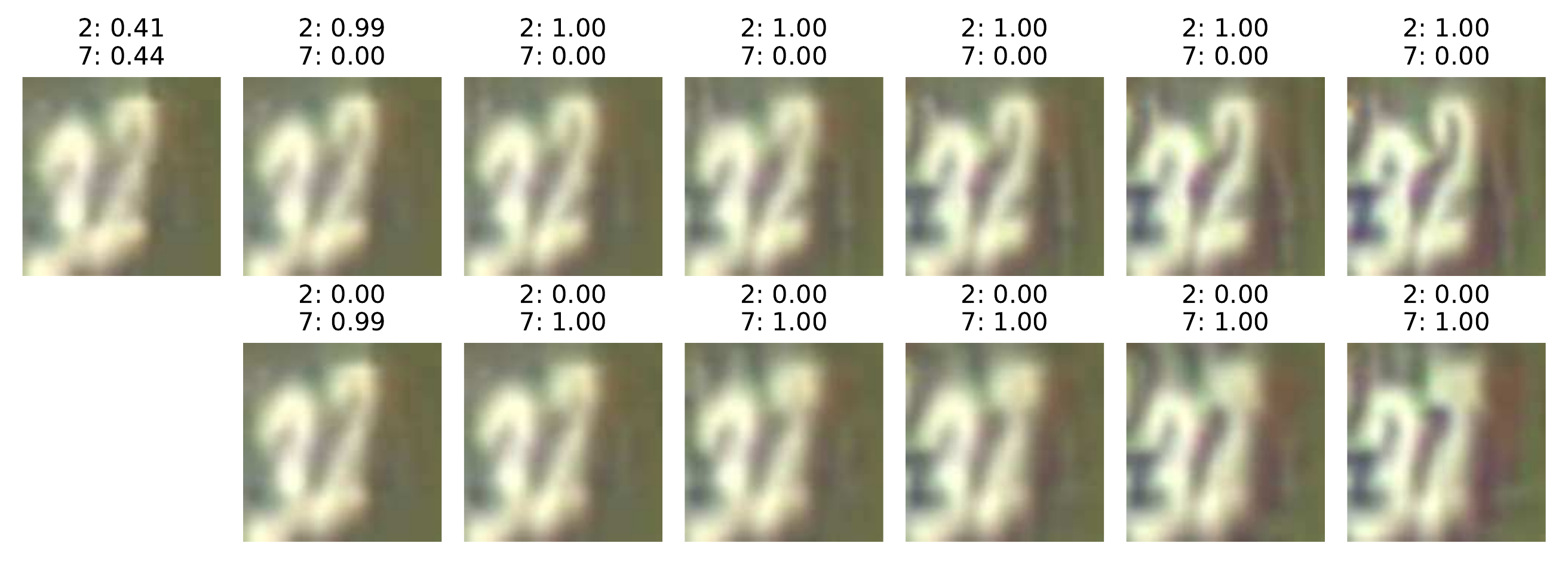}} \\
\hline
\begin{turn}{90} \hspace{-.4cm} AT-0.50 \end{turn}  &  \multicolumn{7}{c}{\includegraphics[width=0.91\textwidth,valign=c]{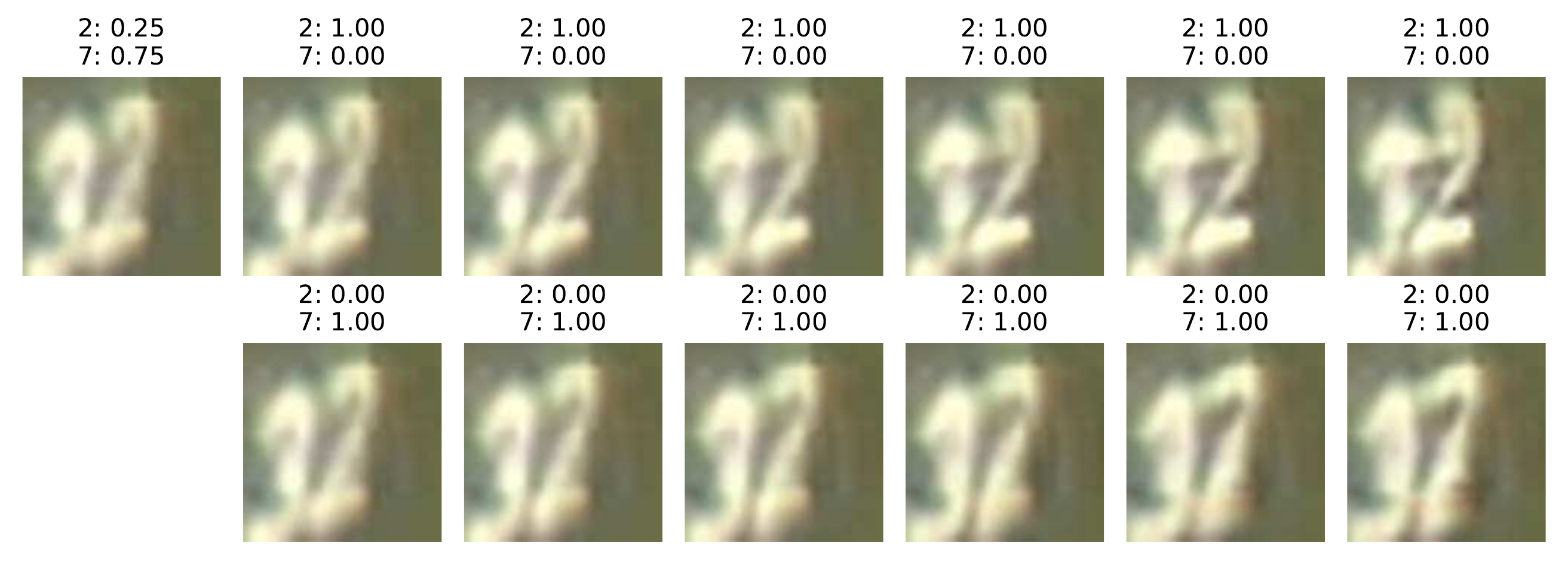}} \\
\hline
\begin{turn}{90} \hspace{-.4cm} AT-0.25 \end{turn}  &  \multicolumn{7}{c}{\includegraphics[width=0.91\textwidth,valign=c]{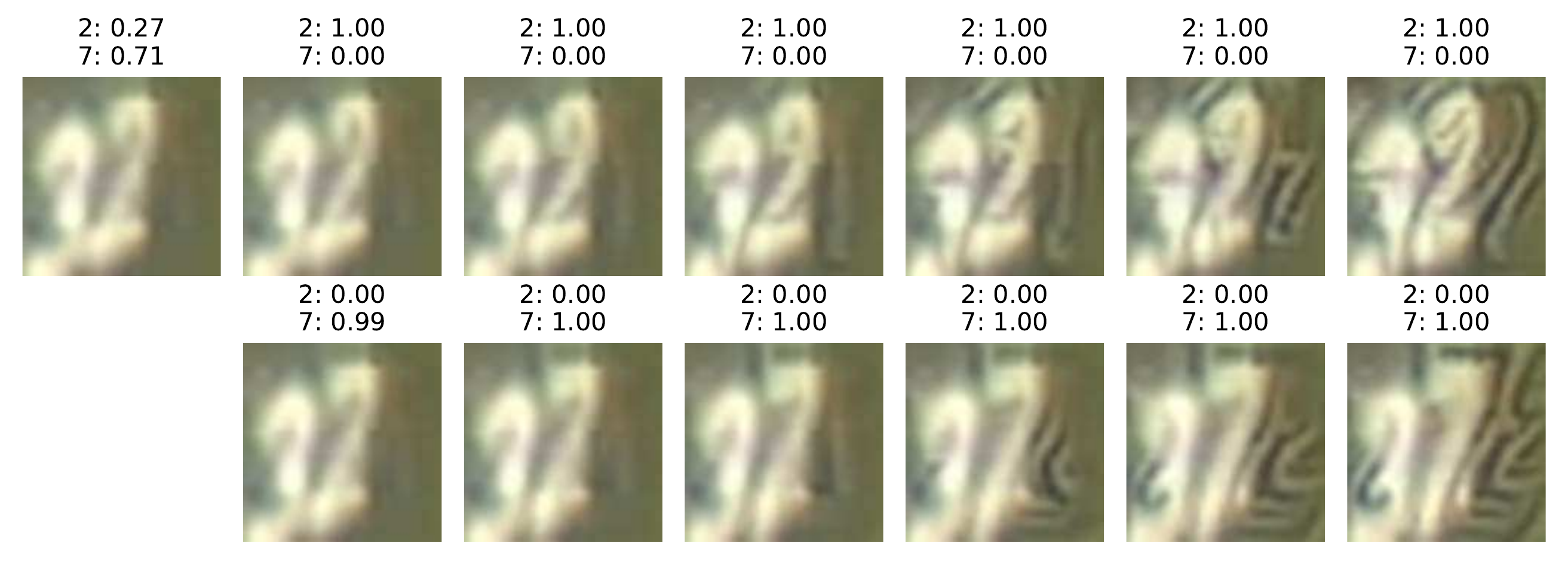}} \\
\hline
\begin{turn}{90} \hspace{-.9cm} RATIO-0.25 \end{turn} & \multicolumn{7}{c}{\includegraphics[width=0.91\textwidth,valign=c]{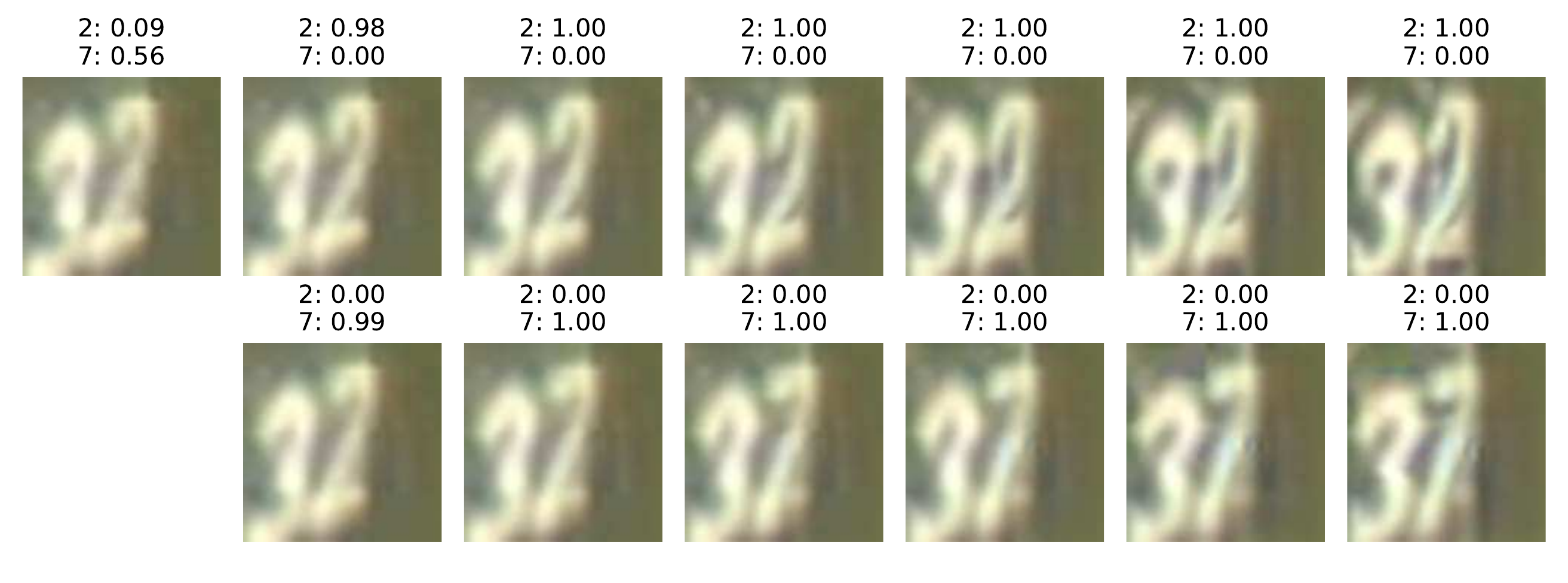}} \\
\end{tabular}	
\caption{\label{fig:vc_svhn_new1}\textbf{Visual Counterfactuals} on SVHN test samples misclassified by all methods. Especially AT$_{0.25}$ generates strong distortion patterns instead of class specific features, whereas RATIO$_{0.25}$ produces samples with a quality which at least matches that of AT$_{0.5}$.
}
\vspace{-.3cm}
\end{figure}

\begin{figure}[ht!]
\begin{tabular}{p{1cm}x{\breite}x{\breite}x{\breite}x{\breite}x{\breite}x{\breite}x{\breite}x{\breite}}
Model  & Orig. & $\epsilon=0.5$ & $\epsilon=1.0$ & $\epsilon=1.5$ & $\epsilon=2.0$ & $\epsilon=2.5$ & $\epsilon=3.0$\\
\begin{turn}{90} \hspace{-.4cm} ACET \end{turn} & \multicolumn{7}{c}{\includegraphics[width=0.91\textwidth,valign=c]{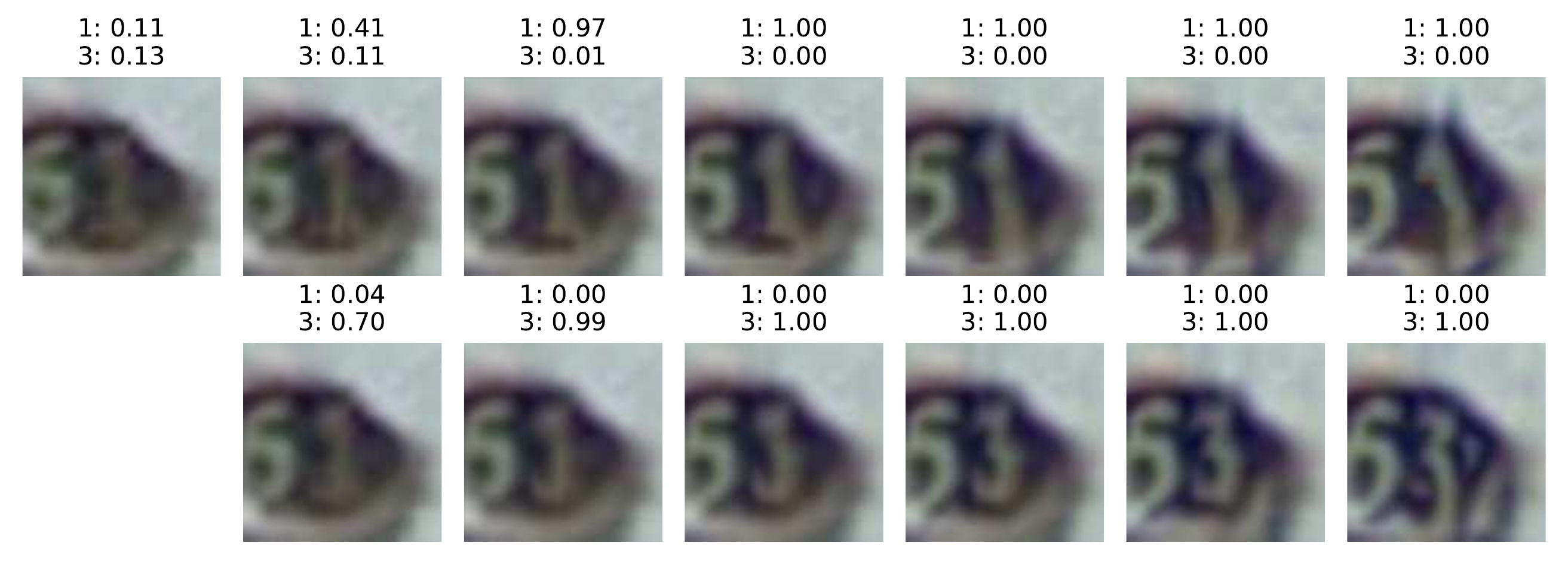}} \\
\hline
\begin{turn}{90} \hspace{-.4cm} AT-0.50 \end{turn}  &  \multicolumn{7}{c}{\includegraphics[width=0.91\textwidth,valign=c]{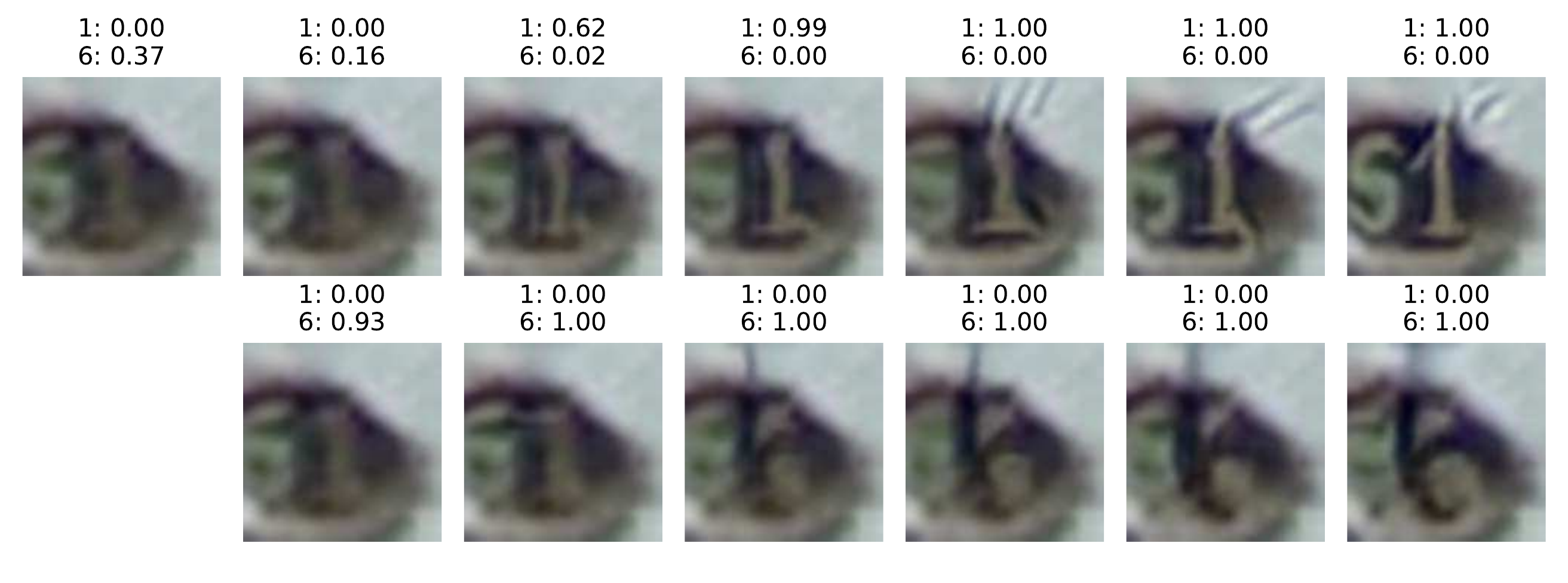}} \\
\hline
\begin{turn}{90} \hspace{-.4cm} AT-0.25 \end{turn}  &  \multicolumn{7}{c}{\includegraphics[width=0.91\textwidth,valign=c]{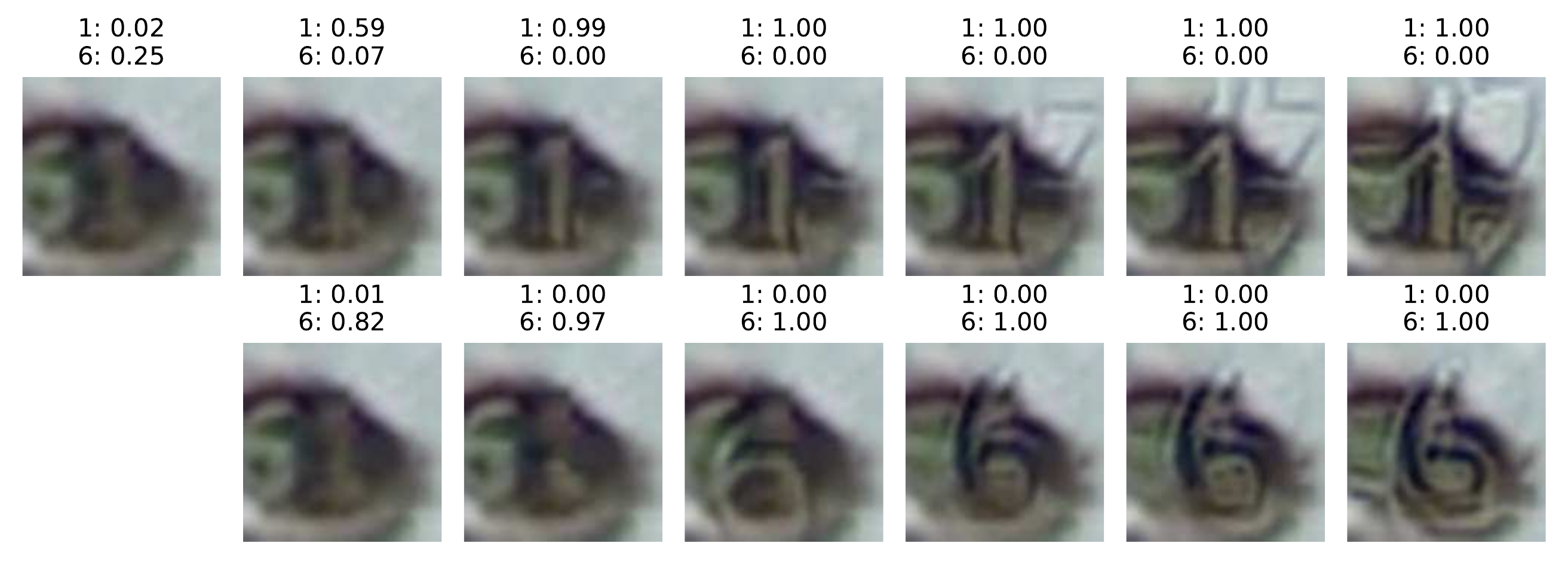}} \\
\hline
\begin{turn}{90} \hspace{-.9cm} RATIO-0.25 \end{turn} & \multicolumn{7}{c}{\includegraphics[width=0.91\textwidth,valign=c]{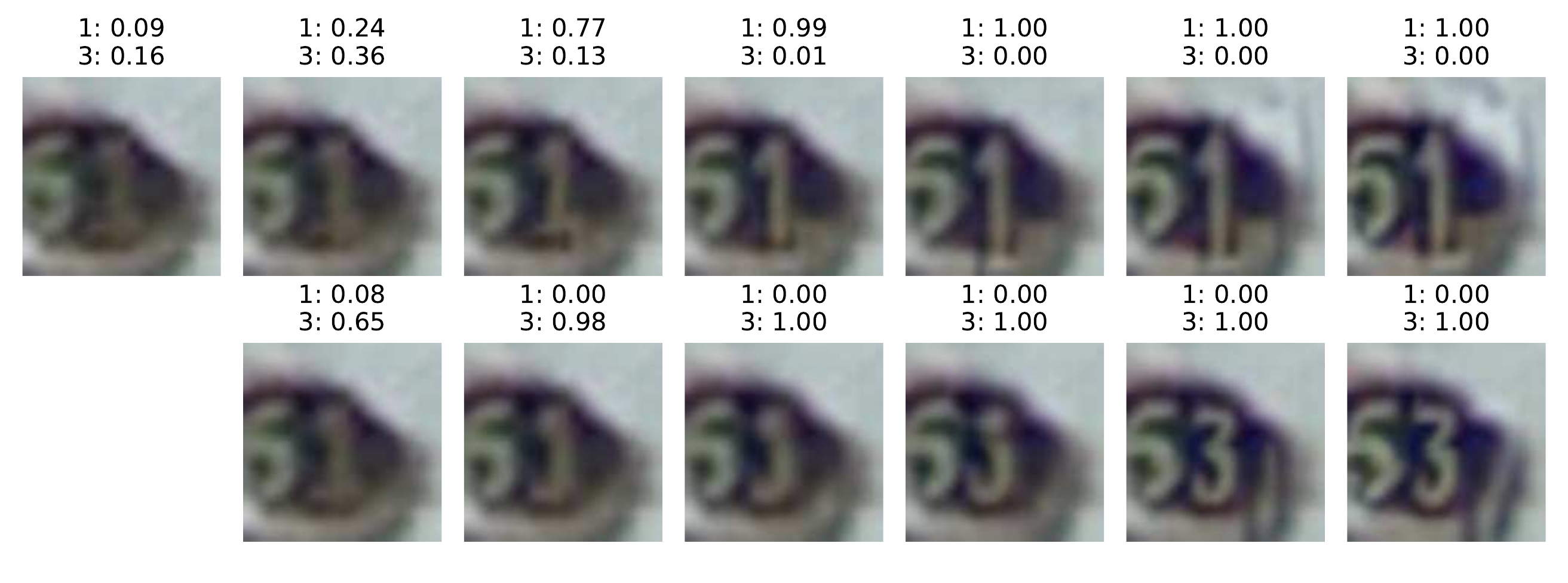}} \\
\end{tabular}	
\caption{\label{fig:vc_svhn_new2}\textbf{Visual Counterfactuals} on SVHN test samples misclassified by all methods. ACET, AT$_{0.5}$ and RATIO$_{0.25}$ produce for the true class super-resolution like images of higher quality than the original sample.
}
\vspace{-.3cm}
\end{figure}
\begin{figure}[ht!]
\begin{tabular}{p{1cm}x{\breite}x{\breite}x{\breite}x{\breite}x{\breite}x{\breite}x{\breite}x{\breite}}
Model  & Orig. & $\epsilon=0.5$ & $\epsilon=1.0$ & $\epsilon=1.5$ & $\epsilon=2.0$ & $\epsilon=2.5$ & $\epsilon=3.0$\\
\begin{turn}{90} \hspace{-.4cm} ACET \end{turn} & \multicolumn{7}{c}{\includegraphics[width=0.91\textwidth,valign=c]{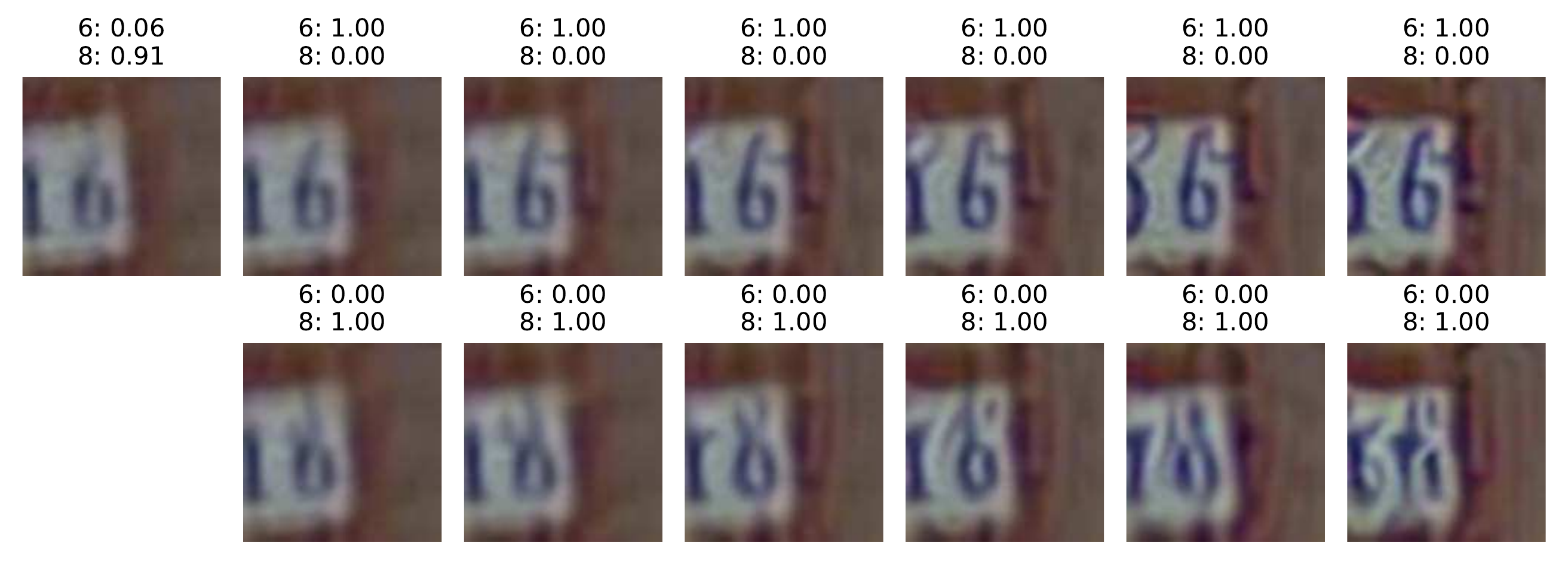}} \\
\hline
\begin{turn}{90} \hspace{-.4cm} AT-0.50 \end{turn}  &  \multicolumn{7}{c}{\includegraphics[width=0.91\textwidth,valign=c]{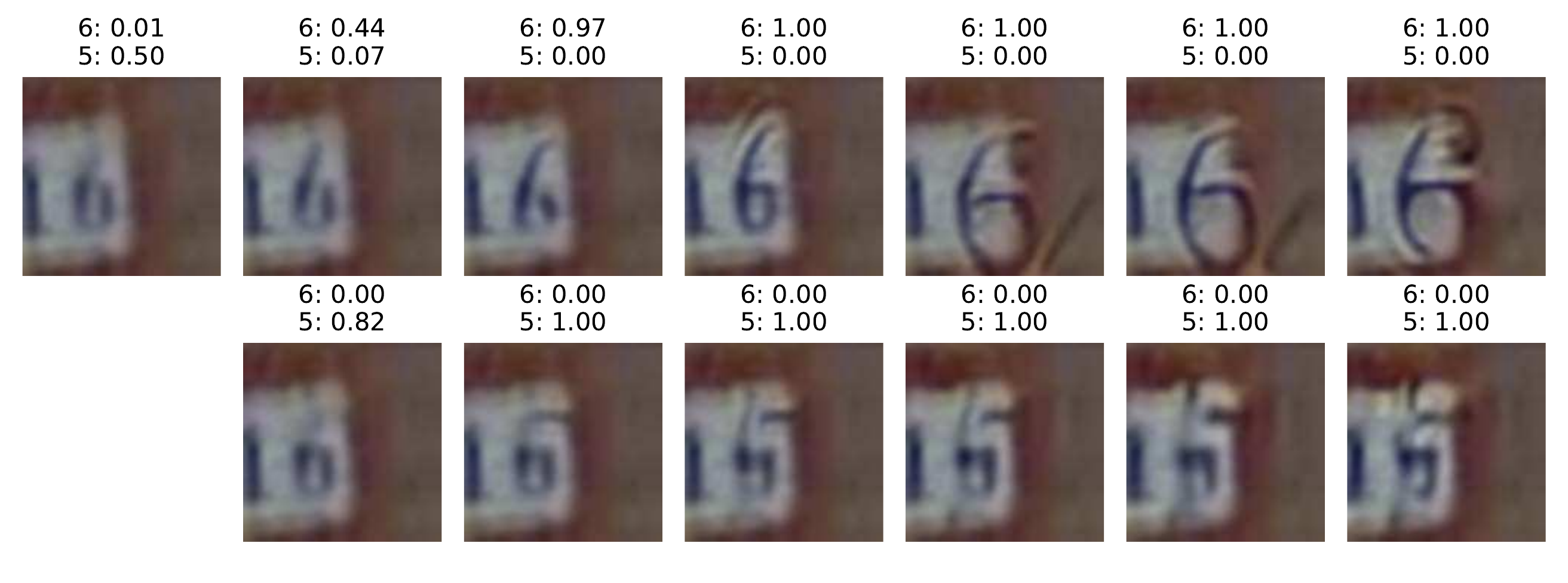}} \\
\hline
\begin{turn}{90} \hspace{-.4cm} AT-0.25 \end{turn}  &  \multicolumn{7}{c}{\includegraphics[width=0.91\textwidth,valign=c]{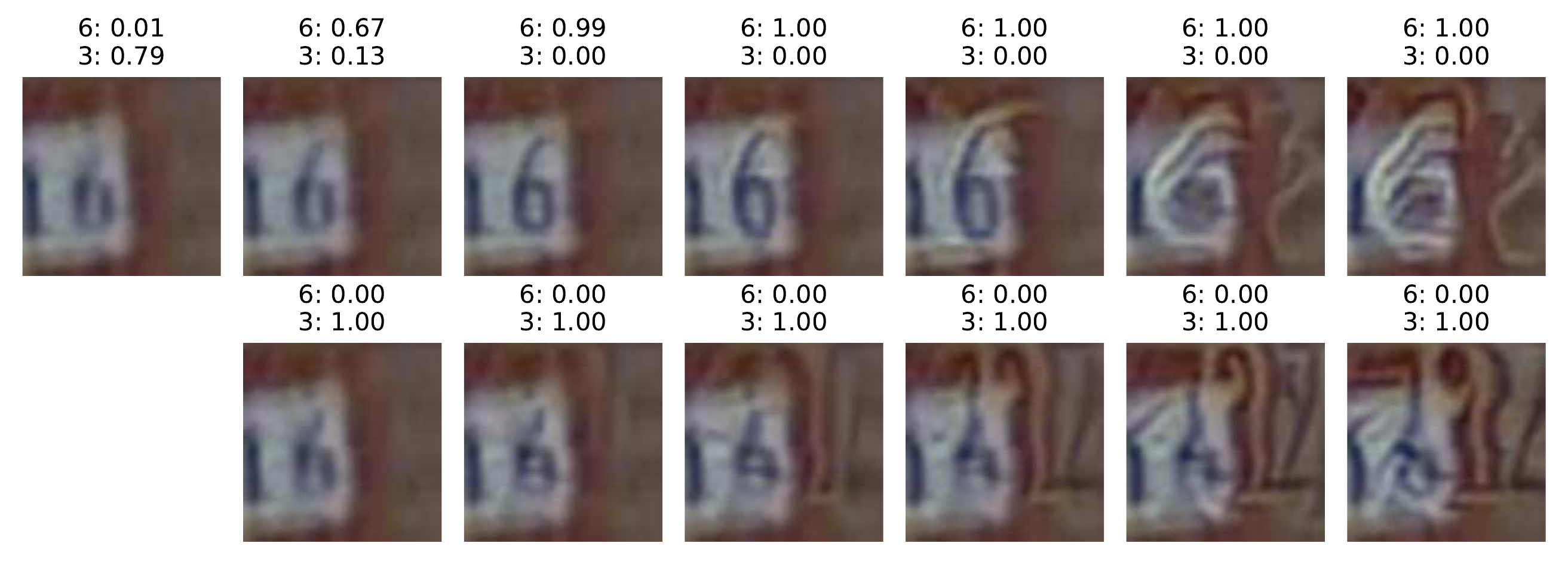}} \\
\hline
\begin{turn}{90} \hspace{-.9cm} RATIO-0.25 \end{turn} & \multicolumn{7}{c}{\includegraphics[width=0.91\textwidth,valign=c]{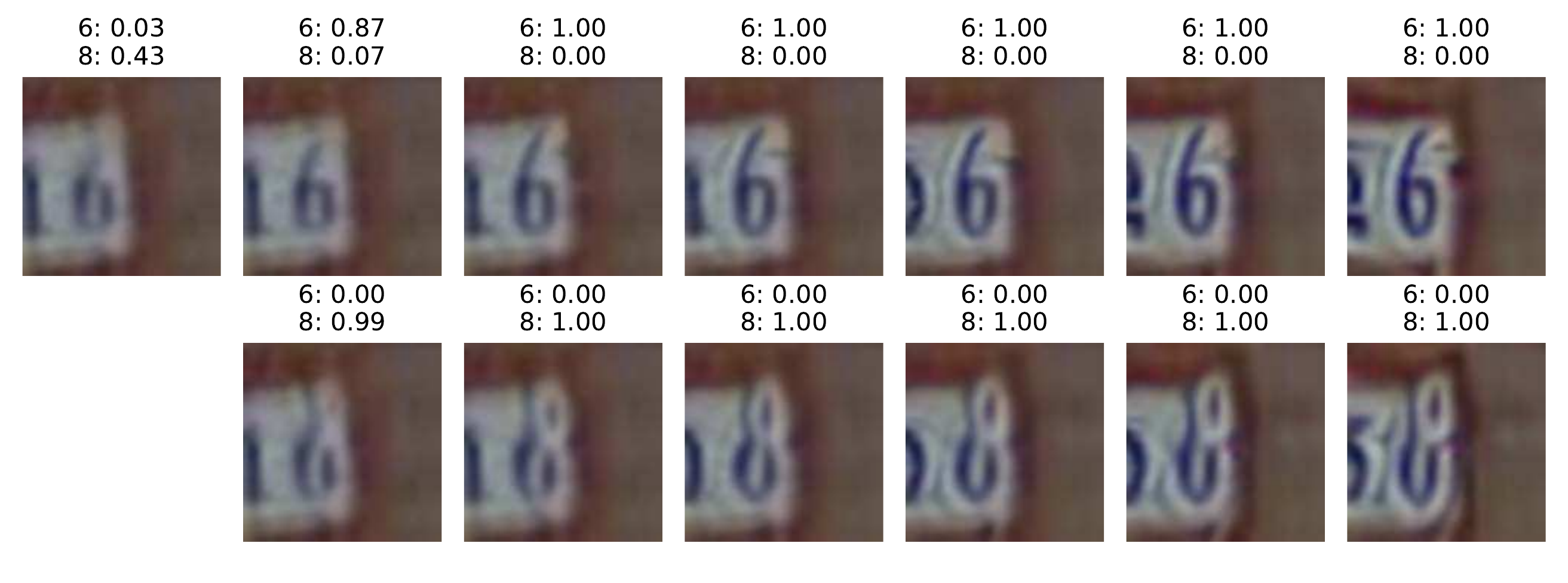}} \\
\end{tabular}	
\caption{\label{fig:vc_svhn_new3} \textbf{Visual Counterfactuals} on SVHN test samples misclassified by all methods.
}
\vspace{-.3cm}
\end{figure}

%% file: res/appendix_svhn_overview.tex
\begin{figure}
\begin{adjustbox}{max width=\textwidth}
\begin{tabu}{ccccccccccccccccc}
\multicolumn{8}{c}{Original} & & \multicolumn{8}{c}{ACET}\\
\rowfont{\tiny}
0
 & 
3
 & 
0
 & 
1
 & 
5
 & 
1
 & 
1
 & 
8
 & & 
0
 & 
3
 & 
0
 & 
1
 & 
5
 & 
1
 & 
1
 & 
8
\\
\includegraphics[width=0.056\textwidth,valign=c]{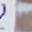}
 & 
\includegraphics[width=0.056\textwidth,valign=c]{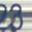}
 & 
\includegraphics[width=0.056\textwidth,valign=c]{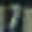}
 & 
\includegraphics[width=0.056\textwidth,valign=c]{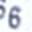}
 & 
\includegraphics[width=0.056\textwidth,valign=c]{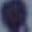}
 & 
\includegraphics[width=0.056\textwidth,valign=c]{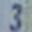}
 & 
\includegraphics[width=0.056\textwidth,valign=c]{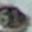}
 & 
\includegraphics[width=0.056\textwidth,valign=c]{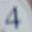}
 & 
\hspace{ 0.028 \textwidth} & 
\includegraphics[width=0.056\textwidth,valign=c]{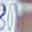}
 & 
\includegraphics[width=0.056\textwidth,valign=c]{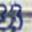}
 & 
\includegraphics[width=0.056\textwidth,valign=c]{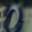}
 & 
\includegraphics[width=0.056\textwidth,valign=c]{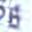}
 & 
\includegraphics[width=0.056\textwidth,valign=c]{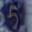}
 & 
\includegraphics[width=0.056\textwidth,valign=c]{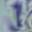}
 & 
\includegraphics[width=0.056\textwidth,valign=c]{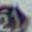}
 & 
\includegraphics[width=0.056\textwidth,valign=c]{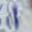}
\\
\rowfont{\tiny}
6
 & 
2
 & 
6
 & 
7
 & 
1
 & 
5
 & 
1
 & 
6
 & & 
6
 & 
2
 & 
6
 & 
7
 & 
1
 & 
5
 & 
1
 & 
6
\\
\includegraphics[width=0.056\textwidth,valign=c]{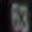}
 & 
\includegraphics[width=0.056\textwidth,valign=c]{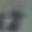}
 & 
\includegraphics[width=0.056\textwidth,valign=c]{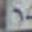}
 & 
\includegraphics[width=0.056\textwidth,valign=c]{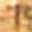}
 & 
\includegraphics[width=0.056\textwidth,valign=c]{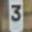}
 & 
\includegraphics[width=0.056\textwidth,valign=c]{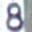}
 & 
\includegraphics[width=0.056\textwidth,valign=c]{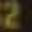}
 & 
\includegraphics[width=0.056\textwidth,valign=c]{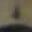}
 & 
\hspace{ 0.028 \textwidth} & 
\includegraphics[width=0.056\textwidth,valign=c]{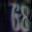}
 & 
\includegraphics[width=0.056\textwidth,valign=c]{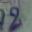}
 & 
\includegraphics[width=0.056\textwidth,valign=c]{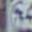}
 & 
\includegraphics[width=0.056\textwidth,valign=c]{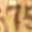}
 & 
\includegraphics[width=0.056\textwidth,valign=c]{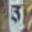}
 & 
\includegraphics[width=0.056\textwidth,valign=c]{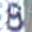}
 & 
\includegraphics[width=0.056\textwidth,valign=c]{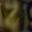}
 & 
\includegraphics[width=0.056\textwidth,valign=c]{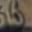}
\\
\rowfont{\tiny}
0
 & 
8
 & 
1
 & 
9
 & 
7
 & 
7
 & 
5
 & 
6
 & & 
0
 & 
8
 & 
1
 & 
9
 & 
7
 & 
7
 & 
5
 & 
6
\\
\includegraphics[width=0.056\textwidth,valign=c]{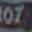}
 & 
\includegraphics[width=0.056\textwidth,valign=c]{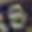}
 & 
\includegraphics[width=0.056\textwidth,valign=c]{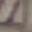}
 & 
\includegraphics[width=0.056\textwidth,valign=c]{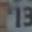}
 & 
\includegraphics[width=0.056\textwidth,valign=c]{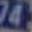}
 & 
\includegraphics[width=0.056\textwidth,valign=c]{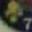}
 & 
\includegraphics[width=0.056\textwidth,valign=c]{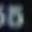}
 & 
\includegraphics[width=0.056\textwidth,valign=c]{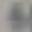}
 & 
\hspace{ 0.028 \textwidth} & 
\includegraphics[width=0.056\textwidth,valign=c]{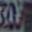}
 & 
\includegraphics[width=0.056\textwidth,valign=c]{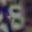}
 & 
\includegraphics[width=0.056\textwidth,valign=c]{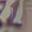}
 & 
\includegraphics[width=0.056\textwidth,valign=c]{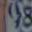}
 & 
\includegraphics[width=0.056\textwidth,valign=c]{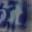}
 & 
\includegraphics[width=0.056\textwidth,valign=c]{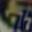}
 & 
\includegraphics[width=0.056\textwidth,valign=c]{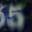}
 & 
\includegraphics[width=0.056\textwidth,valign=c]{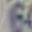}
\\
\rowfont{\tiny}
9
 & 
4
 & 
3
 & 
1
 & 
5
 & 
7
 & 
8
 & 
6
 & & 
9
 & 
4
 & 
3
 & 
1
 & 
5
 & 
7
 & 
8
 & 
6
\\
\includegraphics[width=0.056\textwidth,valign=c]{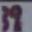}
 & 
\includegraphics[width=0.056\textwidth,valign=c]{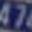}
 & 
\includegraphics[width=0.056\textwidth,valign=c]{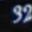}
 & 
\includegraphics[width=0.056\textwidth,valign=c]{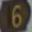}
 & 
\includegraphics[width=0.056\textwidth,valign=c]{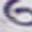}
 & 
\includegraphics[width=0.056\textwidth,valign=c]{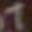}
 & 
\includegraphics[width=0.056\textwidth,valign=c]{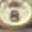}
 & 
\includegraphics[width=0.056\textwidth,valign=c]{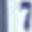}
 & 
\hspace{ 0.028 \textwidth} & 
\includegraphics[width=0.056\textwidth,valign=c]{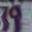}
 & 
\includegraphics[width=0.056\textwidth,valign=c]{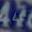}
 & 
\includegraphics[width=0.056\textwidth,valign=c]{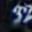}
 & 
\includegraphics[width=0.056\textwidth,valign=c]{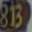}
 & 
\includegraphics[width=0.056\textwidth,valign=c]{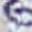}
 & 
\includegraphics[width=0.056\textwidth,valign=c]{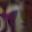}
 & 
\includegraphics[width=0.056\textwidth,valign=c]{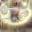}
 & 
\includegraphics[width=0.056\textwidth,valign=c]{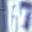}
\\
\rowfont{\tiny}
5
 & 
1
 & 
6
 & 
3
 & 
3
 & 
8
 & 
8
 & 
7
 & & 
5
 & 
1
 & 
6
 & 
3
 & 
3
 & 
8
 & 
8
 & 
7
\\
\includegraphics[width=0.056\textwidth,valign=c]{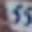}
 & 
\includegraphics[width=0.056\textwidth,valign=c]{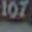}
 & 
\includegraphics[width=0.056\textwidth,valign=c]{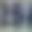}
 & 
\includegraphics[width=0.056\textwidth,valign=c]{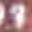}
 & 
\includegraphics[width=0.056\textwidth,valign=c]{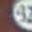}
 & 
\includegraphics[width=0.056\textwidth,valign=c]{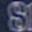}
 & 
\includegraphics[width=0.056\textwidth,valign=c]{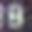}
 & 
\includegraphics[width=0.056\textwidth,valign=c]{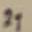}
 & 
\hspace{ 0.028 \textwidth} & 
\includegraphics[width=0.056\textwidth,valign=c]{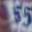}
 & 
\includegraphics[width=0.056\textwidth,valign=c]{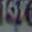}
 & 
\includegraphics[width=0.056\textwidth,valign=c]{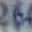}
 & 
\includegraphics[width=0.056\textwidth,valign=c]{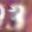}
 & 
\includegraphics[width=0.056\textwidth,valign=c]{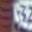}
 & 
\includegraphics[width=0.056\textwidth,valign=c]{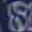}
 & 
\includegraphics[width=0.056\textwidth,valign=c]{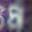}
 & 
\includegraphics[width=0.056\textwidth,valign=c]{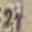}
\\
\rowfont{\tiny}
3
 & 
6
 & 
8
 & 
1
 & 
8
 & 
3
 & 
8
 & 
9
 & & 
3
 & 
6
 & 
8
 & 
1
 & 
8
 & 
3
 & 
8
 & 
9
\\
\includegraphics[width=0.056\textwidth,valign=c]{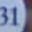}
 & 
\includegraphics[width=0.056\textwidth,valign=c]{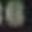}
 & 
\includegraphics[width=0.056\textwidth,valign=c]{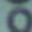}
 & 
\includegraphics[width=0.056\textwidth,valign=c]{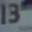}
 & 
\includegraphics[width=0.056\textwidth,valign=c]{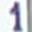}
 & 
\includegraphics[width=0.056\textwidth,valign=c]{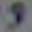}
 & 
\includegraphics[width=0.056\textwidth,valign=c]{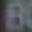}
 & 
\includegraphics[width=0.056\textwidth,valign=c]{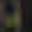}
 & 
\hspace{ 0.028 \textwidth} & 
\includegraphics[width=0.056\textwidth,valign=c]{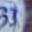}
 & 
\includegraphics[width=0.056\textwidth,valign=c]{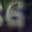}
 & 
\includegraphics[width=0.056\textwidth,valign=c]{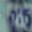}
 & 
\includegraphics[width=0.056\textwidth,valign=c]{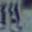}
 & 
\includegraphics[width=0.056\textwidth,valign=c]{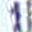}
 & 
\includegraphics[width=0.056\textwidth,valign=c]{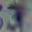}
 & 
\includegraphics[width=0.056\textwidth,valign=c]{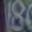}
 & 
\includegraphics[width=0.056\textwidth,valign=c]{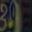}
\vspace{2mm}\\
\multicolumn{8}{c}{AT-0.5} & & \multicolumn{8}{c}{AT-0.25}\\
\includegraphics[width=0.056\textwidth,valign=c]{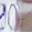}
 & 
\includegraphics[width=0.056\textwidth,valign=c]{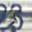}
 & 
\includegraphics[width=0.056\textwidth,valign=c]{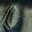}
 & 
\includegraphics[width=0.056\textwidth,valign=c]{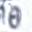}
 & 
\includegraphics[width=0.056\textwidth,valign=c]{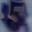}
 & 
\includegraphics[width=0.056\textwidth,valign=c]{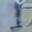}
 & 
\includegraphics[width=0.056\textwidth,valign=c]{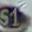}
 & 
\includegraphics[width=0.056\textwidth,valign=c]{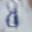}
 & 
\hspace{ 0.028 \textwidth} & 
\includegraphics[width=0.056\textwidth,valign=c]{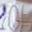}
 & 
\includegraphics[width=0.056\textwidth,valign=c]{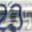}
 & 
\includegraphics[width=0.056\textwidth,valign=c]{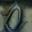}
 & 
\includegraphics[width=0.056\textwidth,valign=c]{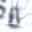}
 & 
\includegraphics[width=0.056\textwidth,valign=c]{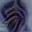}
 & 
\includegraphics[width=0.056\textwidth,valign=c]{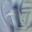}
 & 
\includegraphics[width=0.056\textwidth,valign=c]{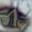}
 & 
\includegraphics[width=0.056\textwidth,valign=c]{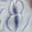}
\vspace{1mm}\\
\includegraphics[width=0.056\textwidth,valign=c]{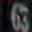}
 & 
\includegraphics[width=0.056\textwidth,valign=c]{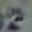}
 & 
\includegraphics[width=0.056\textwidth,valign=c]{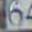}
 & 
\includegraphics[width=0.056\textwidth,valign=c]{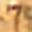}
 & 
\includegraphics[width=0.056\textwidth,valign=c]{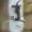}
 & 
\includegraphics[width=0.056\textwidth,valign=c]{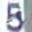}
 & 
\includegraphics[width=0.056\textwidth,valign=c]{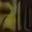}
 & 
\includegraphics[width=0.056\textwidth,valign=c]{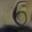}
 & 
\hspace{ 0.028 \textwidth} & 
\includegraphics[width=0.056\textwidth,valign=c]{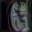}
 & 
\includegraphics[width=0.056\textwidth,valign=c]{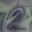}
 & 
\includegraphics[width=0.056\textwidth,valign=c]{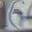}
 & 
\includegraphics[width=0.056\textwidth,valign=c]{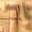}
 & 
\includegraphics[width=0.056\textwidth,valign=c]{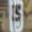}
 & 
\includegraphics[width=0.056\textwidth,valign=c]{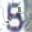}
 & 
\includegraphics[width=0.056\textwidth,valign=c]{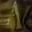}
 & 
\includegraphics[width=0.056\textwidth,valign=c]{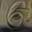}
\vspace{1mm}\\
\includegraphics[width=0.056\textwidth,valign=c]{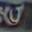}
 & 
\includegraphics[width=0.056\textwidth,valign=c]{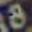}
 & 
\includegraphics[width=0.056\textwidth,valign=c]{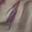}
 & 
\includegraphics[width=0.056\textwidth,valign=c]{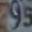}
 & 
\includegraphics[width=0.056\textwidth,valign=c]{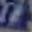}
 & 
\includegraphics[width=0.056\textwidth,valign=c]{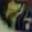}
 & 
\includegraphics[width=0.056\textwidth,valign=c]{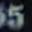}
 & 
\includegraphics[width=0.056\textwidth,valign=c]{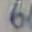}
 & 
\hspace{ 0.028 \textwidth} & 
\includegraphics[width=0.056\textwidth,valign=c]{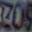}
 & 
\includegraphics[width=0.056\textwidth,valign=c]{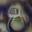}
 & 
\includegraphics[width=0.056\textwidth,valign=c]{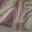}
 & 
\includegraphics[width=0.056\textwidth,valign=c]{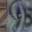}
 & 
\includegraphics[width=0.056\textwidth,valign=c]{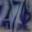}
 & 
\includegraphics[width=0.056\textwidth,valign=c]{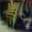}
 & 
\includegraphics[width=0.056\textwidth,valign=c]{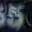}
 & 
\includegraphics[width=0.056\textwidth,valign=c]{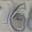}
\vspace{1mm}\\
\includegraphics[width=0.056\textwidth,valign=c]{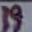}
 & 
\includegraphics[width=0.056\textwidth,valign=c]{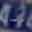}
 & 
\includegraphics[width=0.056\textwidth,valign=c]{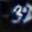}
 & 
\includegraphics[width=0.056\textwidth,valign=c]{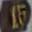}
 & 
\includegraphics[width=0.056\textwidth,valign=c]{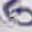}
 & 
\includegraphics[width=0.056\textwidth,valign=c]{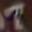}
 & 
\includegraphics[width=0.056\textwidth,valign=c]{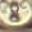}
 & 
\includegraphics[width=0.056\textwidth,valign=c]{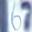}
 & 
\hspace{ 0.028 \textwidth} & 
\includegraphics[width=0.056\textwidth,valign=c]{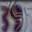}
 & 
\includegraphics[width=0.056\textwidth,valign=c]{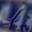}
 & 
\includegraphics[width=0.056\textwidth,valign=c]{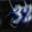}
 & 
\includegraphics[width=0.056\textwidth,valign=c]{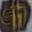}
 & 
\includegraphics[width=0.056\textwidth,valign=c]{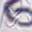}
 & 
\includegraphics[width=0.056\textwidth,valign=c]{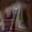}
 & 
\includegraphics[width=0.056\textwidth,valign=c]{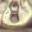}
 & 
\includegraphics[width=0.056\textwidth,valign=c]{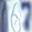}
\vspace{1mm}\\
\includegraphics[width=0.056\textwidth,valign=c]{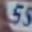}
 & 
\includegraphics[width=0.056\textwidth,valign=c]{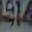}
 & 
\includegraphics[width=0.056\textwidth,valign=c]{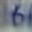}
 & 
\includegraphics[width=0.056\textwidth,valign=c]{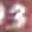}
 & 
\includegraphics[width=0.056\textwidth,valign=c]{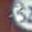}
 & 
\includegraphics[width=0.056\textwidth,valign=c]{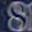}
 & 
\includegraphics[width=0.056\textwidth,valign=c]{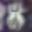}
 & 
\includegraphics[width=0.056\textwidth,valign=c]{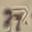}
 & 
\hspace{ 0.028 \textwidth} & 
\includegraphics[width=0.056\textwidth,valign=c]{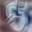}
 & 
\includegraphics[width=0.056\textwidth,valign=c]{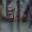}
 & 
\includegraphics[width=0.056\textwidth,valign=c]{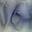}
 & 
\includegraphics[width=0.056\textwidth,valign=c]{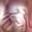}
 & 
\includegraphics[width=0.056\textwidth,valign=c]{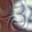}
 & 
\includegraphics[width=0.056\textwidth,valign=c]{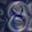}
 & 
\includegraphics[width=0.056\textwidth,valign=c]{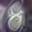}
 & 
\includegraphics[width=0.056\textwidth,valign=c]{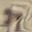}
\vspace{1mm}\\
\includegraphics[width=0.056\textwidth,valign=c]{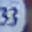}
 & 
\includegraphics[width=0.056\textwidth,valign=c]{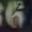}
 & 
\includegraphics[width=0.056\textwidth,valign=c]{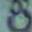}
 & 
\includegraphics[width=0.056\textwidth,valign=c]{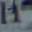}
 & 
\includegraphics[width=0.056\textwidth,valign=c]{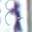}
 & 
\includegraphics[width=0.056\textwidth,valign=c]{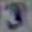}
 & 
\includegraphics[width=0.056\textwidth,valign=c]{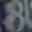}
 & 
\includegraphics[width=0.056\textwidth,valign=c]{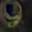}
 & 
\hspace{ 0.028 \textwidth} & 
\includegraphics[width=0.056\textwidth,valign=c]{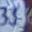}
 & 
\includegraphics[width=0.056\textwidth,valign=c]{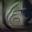}
 & 
\includegraphics[width=0.056\textwidth,valign=c]{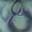}
 & 
\includegraphics[width=0.056\textwidth,valign=c]{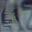}
 & 
\includegraphics[width=0.056\textwidth,valign=c]{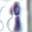}
 & 
\includegraphics[width=0.056\textwidth,valign=c]{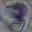}
 & 
\includegraphics[width=0.056\textwidth,valign=c]{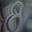}
 & 
\includegraphics[width=0.056\textwidth,valign=c]{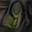}
\vspace{2mm}\\
\multicolumn{8}{c}{RATIO-0.5} & & \multicolumn{8}{c}{RATIO-0.25}\\
\includegraphics[width=0.056\textwidth,valign=c]{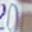}
 & 
\includegraphics[width=0.056\textwidth,valign=c]{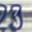}
 & 
\includegraphics[width=0.056\textwidth,valign=c]{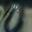}
 & 
\includegraphics[width=0.056\textwidth,valign=c]{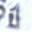}
 & 
\includegraphics[width=0.056\textwidth,valign=c]{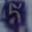}
 & 
\includegraphics[width=0.056\textwidth,valign=c]{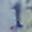}
 & 
\includegraphics[width=0.056\textwidth,valign=c]{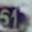}
 & 
\includegraphics[width=0.056\textwidth,valign=c]{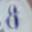}
 & 
\hspace{ 0.028 \textwidth} & 
\includegraphics[width=0.056\textwidth,valign=c]{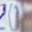}
 & 
\includegraphics[width=0.056\textwidth,valign=c]{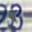}
 & 
\includegraphics[width=0.056\textwidth,valign=c]{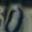}
 & 
\includegraphics[width=0.056\textwidth,valign=c]{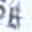}
 & 
\includegraphics[width=0.056\textwidth,valign=c]{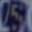}
 & 
\includegraphics[width=0.056\textwidth,valign=c]{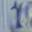}
 & 
\includegraphics[width=0.056\textwidth,valign=c]{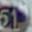}
 & 
\includegraphics[width=0.056\textwidth,valign=c]{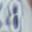}
\vspace{1mm}\\
\includegraphics[width=0.056\textwidth,valign=c]{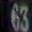}
 & 
\includegraphics[width=0.056\textwidth,valign=c]{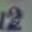}
 & 
\includegraphics[width=0.056\textwidth,valign=c]{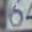}
 & 
\includegraphics[width=0.056\textwidth,valign=c]{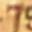}
 & 
\includegraphics[width=0.056\textwidth,valign=c]{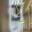}
 & 
\includegraphics[width=0.056\textwidth,valign=c]{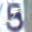}
 & 
\includegraphics[width=0.056\textwidth,valign=c]{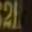}
 & 
\includegraphics[width=0.056\textwidth,valign=c]{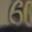}
 & 
\hspace{ 0.028 \textwidth} & 
\includegraphics[width=0.056\textwidth,valign=c]{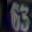}
 & 
\includegraphics[width=0.056\textwidth,valign=c]{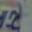}
 & 
\includegraphics[width=0.056\textwidth,valign=c]{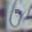}
 & 
\includegraphics[width=0.056\textwidth,valign=c]{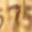}
 & 
\includegraphics[width=0.056\textwidth,valign=c]{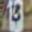}
 & 
\includegraphics[width=0.056\textwidth,valign=c]{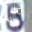}
 & 
\includegraphics[width=0.056\textwidth,valign=c]{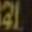}
 & 
\includegraphics[width=0.056\textwidth,valign=c]{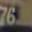}
\vspace{1mm}\\
\includegraphics[width=0.056\textwidth,valign=c]{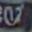}
 & 
\includegraphics[width=0.056\textwidth,valign=c]{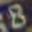}
 & 
\includegraphics[width=0.056\textwidth,valign=c]{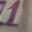}
 & 
\includegraphics[width=0.056\textwidth,valign=c]{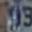}
 & 
\includegraphics[width=0.056\textwidth,valign=c]{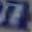}
 & 
\includegraphics[width=0.056\textwidth,valign=c]{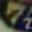}
 & 
\includegraphics[width=0.056\textwidth,valign=c]{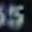}
 & 
\includegraphics[width=0.056\textwidth,valign=c]{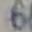}
 & 
\hspace{ 0.028 \textwidth} & 
\includegraphics[width=0.056\textwidth,valign=c]{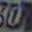}
 & 
\includegraphics[width=0.056\textwidth,valign=c]{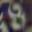}
 & 
\includegraphics[width=0.056\textwidth,valign=c]{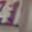}
 & 
\includegraphics[width=0.056\textwidth,valign=c]{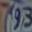}
 & 
\includegraphics[width=0.056\textwidth,valign=c]{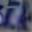}
 & 
\includegraphics[width=0.056\textwidth,valign=c]{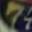}
 & 
\includegraphics[width=0.056\textwidth,valign=c]{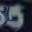}
 & 
\includegraphics[width=0.056\textwidth,valign=c]{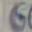}
\vspace{1mm}\\
\includegraphics[width=0.056\textwidth,valign=c]{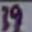}
 & 
\includegraphics[width=0.056\textwidth,valign=c]{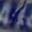}
 & 
\includegraphics[width=0.056\textwidth,valign=c]{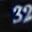}
 & 
\includegraphics[width=0.056\textwidth,valign=c]{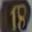}
 & 
\includegraphics[width=0.056\textwidth,valign=c]{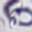}
 & 
\includegraphics[width=0.056\textwidth,valign=c]{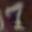}
 & 
\includegraphics[width=0.056\textwidth,valign=c]{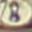}
 & 
\includegraphics[width=0.056\textwidth,valign=c]{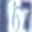}
 & 
\hspace{ 0.028 \textwidth} & 
\includegraphics[width=0.056\textwidth,valign=c]{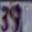}
 & 
\includegraphics[width=0.056\textwidth,valign=c]{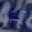}
 & 
\includegraphics[width=0.056\textwidth,valign=c]{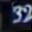}
 & 
\includegraphics[width=0.056\textwidth,valign=c]{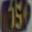}
 & 
\includegraphics[width=0.056\textwidth,valign=c]{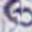}
 & 
\includegraphics[width=0.056\textwidth,valign=c]{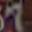}
 & 
\includegraphics[width=0.056\textwidth,valign=c]{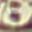}
 & 
\includegraphics[width=0.056\textwidth,valign=c]{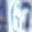}
\vspace{1mm}\\
\includegraphics[width=0.056\textwidth,valign=c]{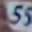}
 & 
\includegraphics[width=0.056\textwidth,valign=c]{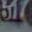}
 & 
\includegraphics[width=0.056\textwidth,valign=c]{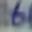}
 & 
\includegraphics[width=0.056\textwidth,valign=c]{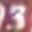}
 & 
\includegraphics[width=0.056\textwidth,valign=c]{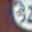}
 & 
\includegraphics[width=0.056\textwidth,valign=c]{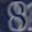}
 & 
\includegraphics[width=0.056\textwidth,valign=c]{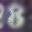}
 & 
\includegraphics[width=0.056\textwidth,valign=c]{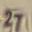}
 & 
\hspace{ 0.028 \textwidth} & 
\includegraphics[width=0.056\textwidth,valign=c]{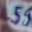}
 & 
\includegraphics[width=0.056\textwidth,valign=c]{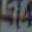}
 & 
\includegraphics[width=0.056\textwidth,valign=c]{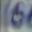}
 & 
\includegraphics[width=0.056\textwidth,valign=c]{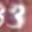}
 & 
\includegraphics[width=0.056\textwidth,valign=c]{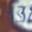}
 & 
\includegraphics[width=0.056\textwidth,valign=c]{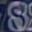}
 & 
\includegraphics[width=0.056\textwidth,valign=c]{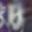}
 & 
\includegraphics[width=0.056\textwidth,valign=c]{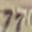}
\vspace{1mm}\\
\includegraphics[width=0.056\textwidth,valign=c]{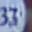}
 & 
\includegraphics[width=0.056\textwidth,valign=c]{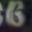}
 & 
\includegraphics[width=0.056\textwidth,valign=c]{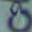}
 & 
\includegraphics[width=0.056\textwidth,valign=c]{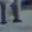}
 & 
\includegraphics[width=0.056\textwidth,valign=c]{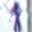}
 & 
\includegraphics[width=0.056\textwidth,valign=c]{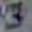}
 & 
\includegraphics[width=0.056\textwidth,valign=c]{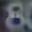}
 & 
\includegraphics[width=0.056\textwidth,valign=c]{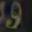}
 & 
\hspace{ 0.028 \textwidth} & 
\includegraphics[width=0.056\textwidth,valign=c]{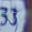}
 & 
\includegraphics[width=0.056\textwidth,valign=c]{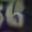}
 & 
\includegraphics[width=0.056\textwidth,valign=c]{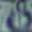}
 & 
\includegraphics[width=0.056\textwidth,valign=c]{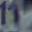}
 & 
\includegraphics[width=0.056\textwidth,valign=c]{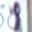}
 & 
\includegraphics[width=0.056\textwidth,valign=c]{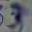}
 & 
\includegraphics[width=0.056\textwidth,valign=c]{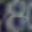}
 & 
\includegraphics[width=0.056\textwidth,valign=c]{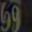}
\vspace{2mm}\\
\end{tabu}
\end{adjustbox}
\caption{\label{fig:svhn_overview}\textbf{More Visual Counterfactuals:} Random selection of  48 SVHN test images misclassified by all models (top left shows original images together with ground truth labels) and the associated visual counterfactuals that are generated by maximizing the confidence in the ground truth class in a $l_2$ ball of radius 3.  
Note that a lot of test images are falsely labeled, \ie the label either refers to the digit which is not located in the center of the image or to a completely different number. }\label{Fig:SVHN_OVERVIEW}
\end{figure}

%% file: res/appendix_od_svhn_main_paper_extended.tex
\begin{figure}[ht!]
\begin{tabular}{p{1cm}x{\breite}x{\breite}x{\breite}x{\breite}x{\breite}x{\breite}x{\breite}x{\breite}}
Model  & Orig. & $\epsilon=0.5$ & $\epsilon=1.0$ & $\epsilon=1.5$ & $\epsilon=2.0$ & $\epsilon=2.5$ & $\epsilon=3.0$\\
\begin{turn}{90} \hspace{-.4cm} Plain \end{turn} & \multicolumn{7}{c}{\includegraphics[width=0.91\textwidth,valign=c]{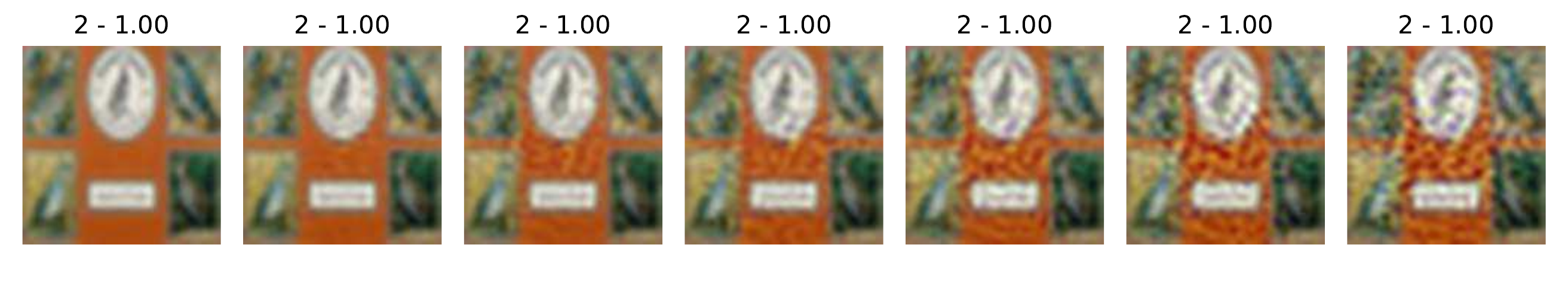}} \\
\hline
\begin{turn}{90} \hspace{-.4cm} OE \end{turn}  &  \multicolumn{7}{c}{\includegraphics[width=0.91\textwidth,valign=c]{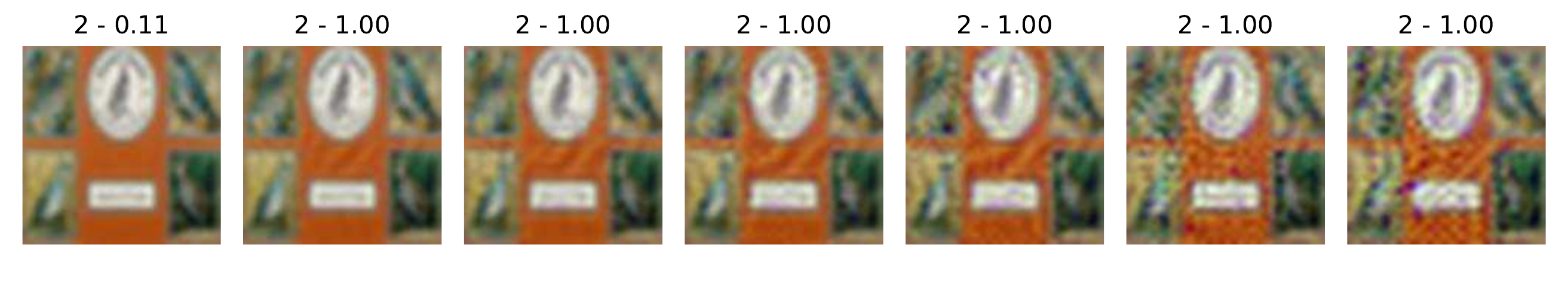}} \\
\hline
\begin{turn}{90} \hspace{-.4cm} ACET \end{turn} & \multicolumn{7}{c}{\includegraphics[width=0.91\textwidth,valign=c]{pics/SVHN/ACET/OD/img_209.pdf}} \\
\hline
\begin{turn}{90} \hspace{-.4cm} AT-0.50 \end{turn}  &  \multicolumn{7}{c}{\includegraphics[width=0.91\textwidth,valign=c]{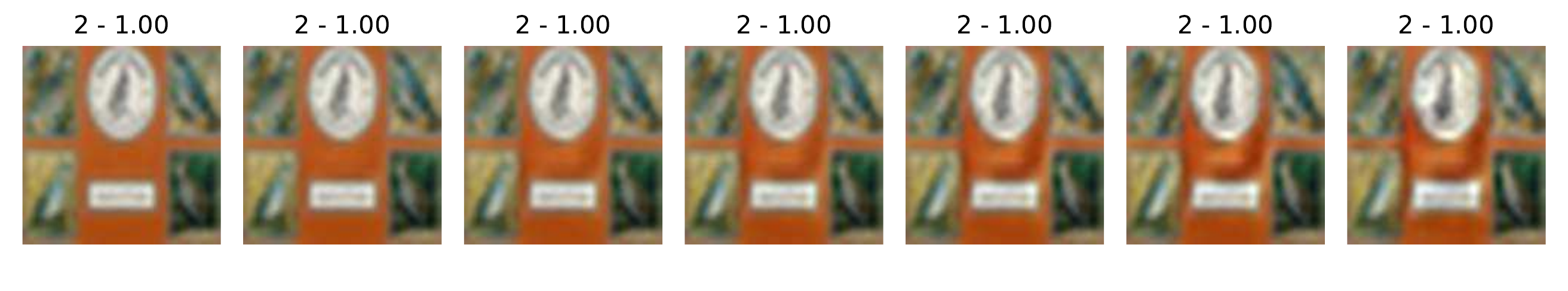}} \\
\hline
\begin{turn}{90} \hspace{-.4cm} AT-0.25 \end{turn}  &  \multicolumn{7}{c}{\includegraphics[width=0.91\textwidth,valign=c]{pics/SVHN/AT025/OD/img_209.pdf}} \\
\hline
\begin{turn}{90} \hspace{-.4cm} R-0.50 \end{turn} & \multicolumn{7}{c}{\includegraphics[width=0.91\textwidth,valign=c]{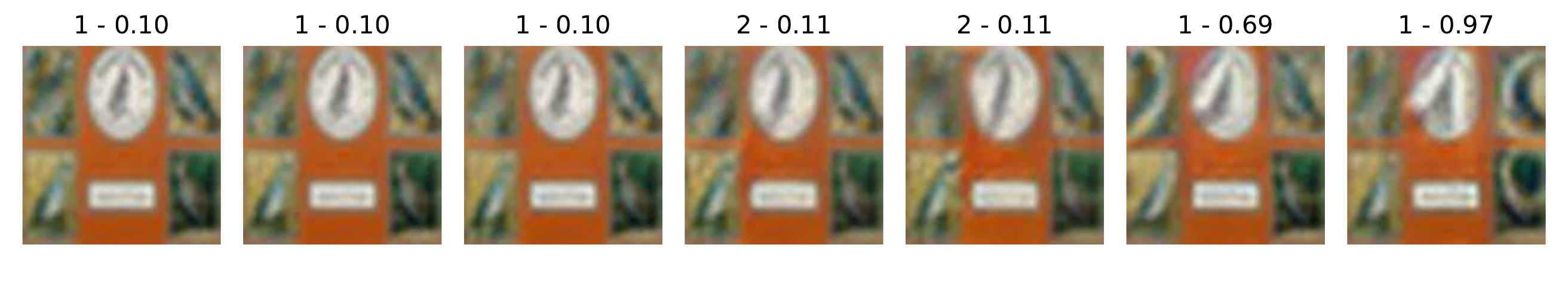}} \\
\hline
\begin{turn}{90} \hspace{-.4cm} R-0.25 \end{turn} & \multicolumn{7}{c}{\includegraphics[width=0.91\textwidth,valign=c]{pics/SVHN/Ratio025/OD/img_209.pdf}} \\
\end{tabular}	
\vspace{-.5cm}
\caption{\label{fig:svhn_od_ext}\textbf{Feature Generation for out-distribution images:} Extension of Figure~\ref{Fig:CIFAR_OD} in the main paper to all models.
Only ACET and both RATIO models are able to generate clearly visible digits, whereas the remaining models make high confidence predictions in the absence of class specific features.} 
\end{figure}

%% file: res/appendix_od_svhn_new.tex
\begin{figure}[ht!]
\begin{tabular}{p{1cm}x{\breite}x{\breite}x{\breite}x{\breite}x{\breite}x{\breite}x{\breite}x{\breite}}
Model  & Orig. & $\epsilon=0.5$ & $\epsilon=1.0$ & $\epsilon=1.5$ & $\epsilon=2.0$ & $\epsilon=2.5$ & $\epsilon=3.0$\\
\begin{turn}{90} \hspace{-.4cm} ACET \end{turn} & \multicolumn{7}{c}{\includegraphics[width=0.91\textwidth,valign=c]{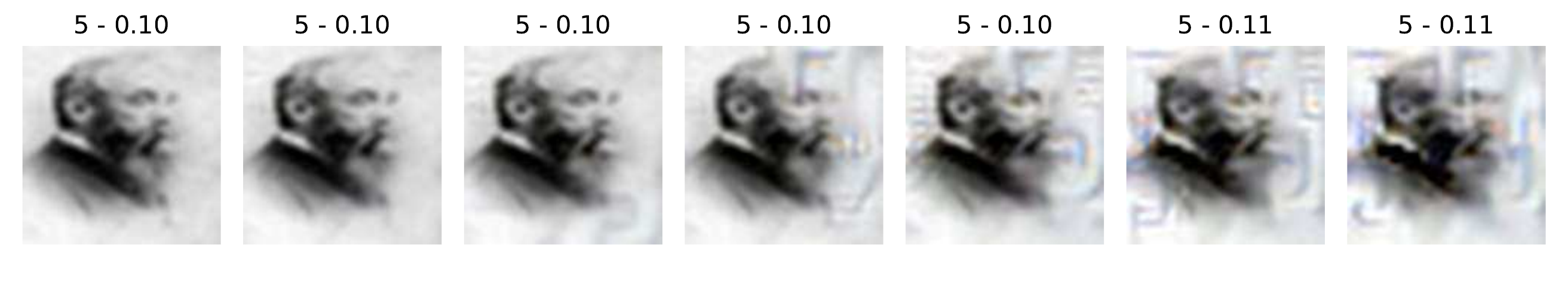}} \\
\hline
\begin{turn}{90} \hspace{-.4cm} AT-0.50 \end{turn}  &  \multicolumn{7}{c}{\includegraphics[width=0.91\textwidth,valign=c]{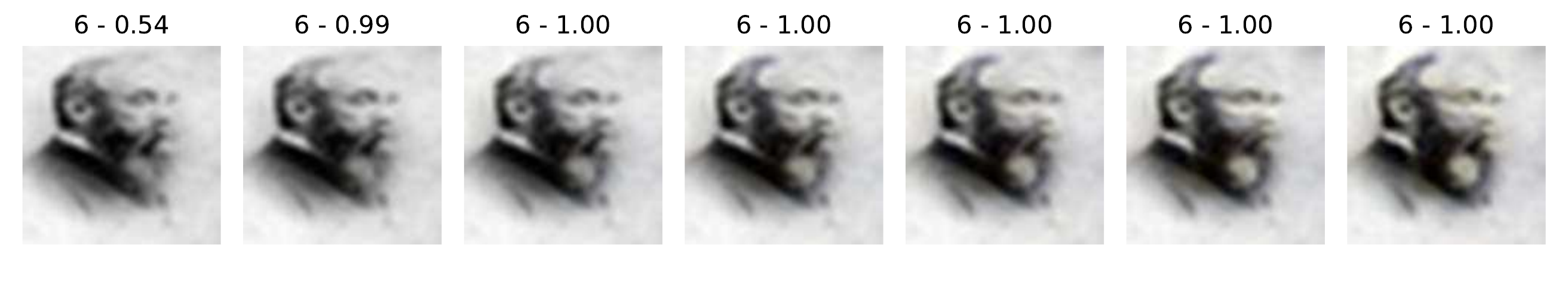}} \\
\hline
\begin{turn}{90} \hspace{-.4cm} AT-0.25 \end{turn} & \multicolumn{7}{c}{\includegraphics[width=0.91\textwidth,valign=c]{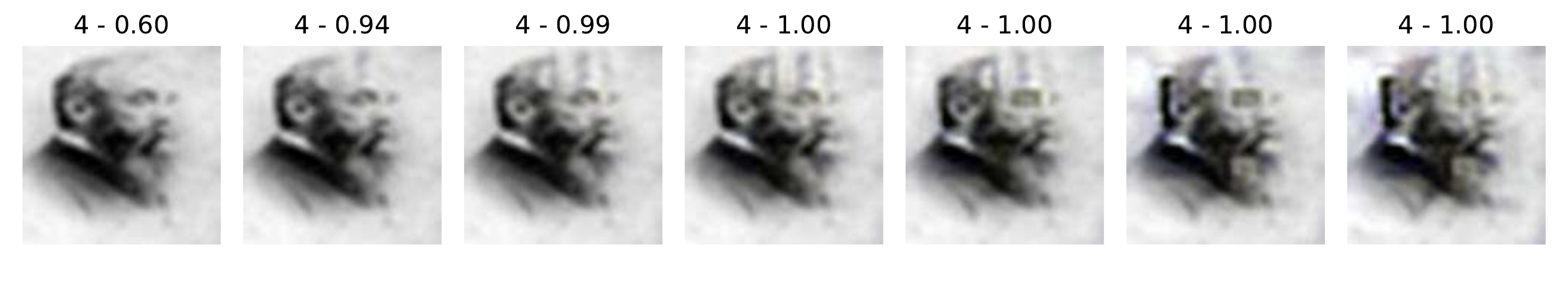}} \\
\hline
\begin{turn}{90} \hspace{-.4cm} R-0.25 \end{turn}  &  \multicolumn{7}{c}{\includegraphics[width=0.91\textwidth,valign=c]{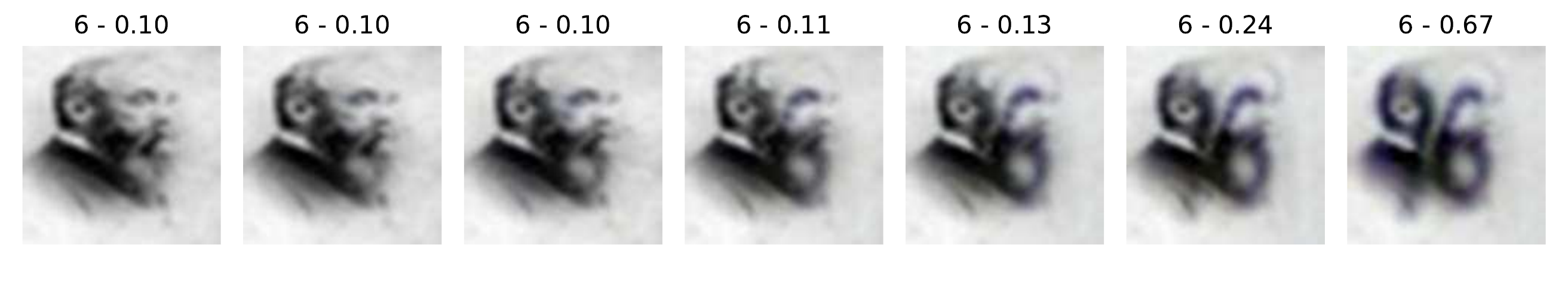}} \\
\vspace{2mm}\\
\begin{turn}{90} \hspace{-.4cm} ACET \end{turn} & \multicolumn{7}{c}{\includegraphics[width=0.91\textwidth,valign=c]{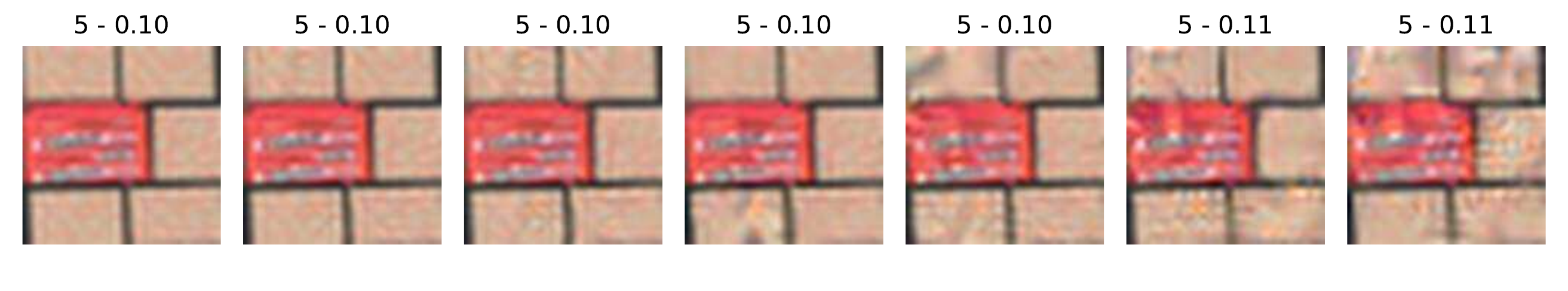}} \\
\hline
\begin{turn}{90} \hspace{-.4cm} AT-0.50 \end{turn}  &  \multicolumn{7}{c}{\includegraphics[width=0.91\textwidth,valign=c]{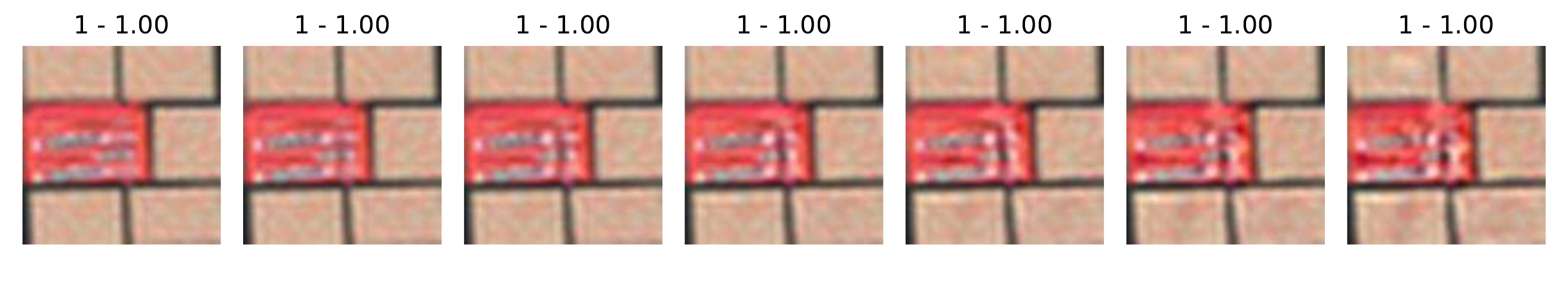}} \\
\hline
\begin{turn}{90} \hspace{-.4cm} AT-0.25 \end{turn} & \multicolumn{7}{c}{\includegraphics[width=0.91\textwidth,valign=c]{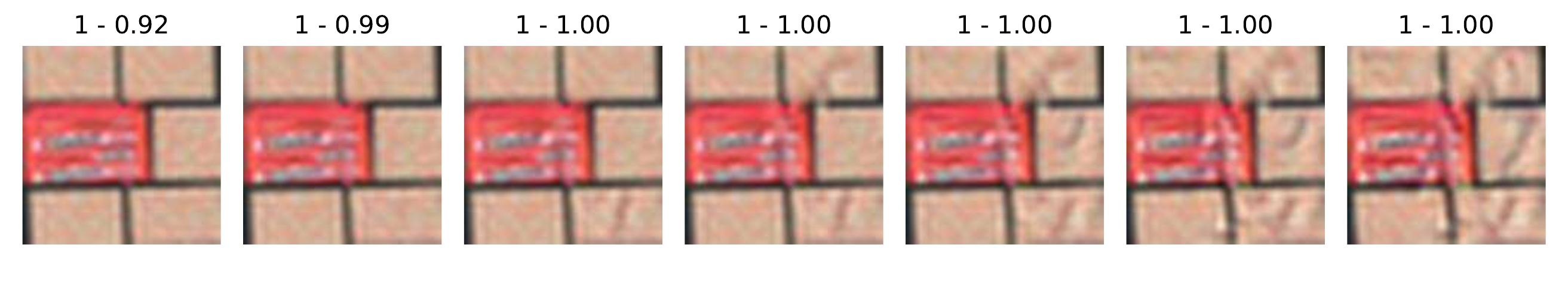}} \\
\hline
\begin{turn}{90} \hspace{-.4cm} R-0.25 \end{turn}  &  \multicolumn{7}{c}{\includegraphics[width=0.91\textwidth,valign=c]{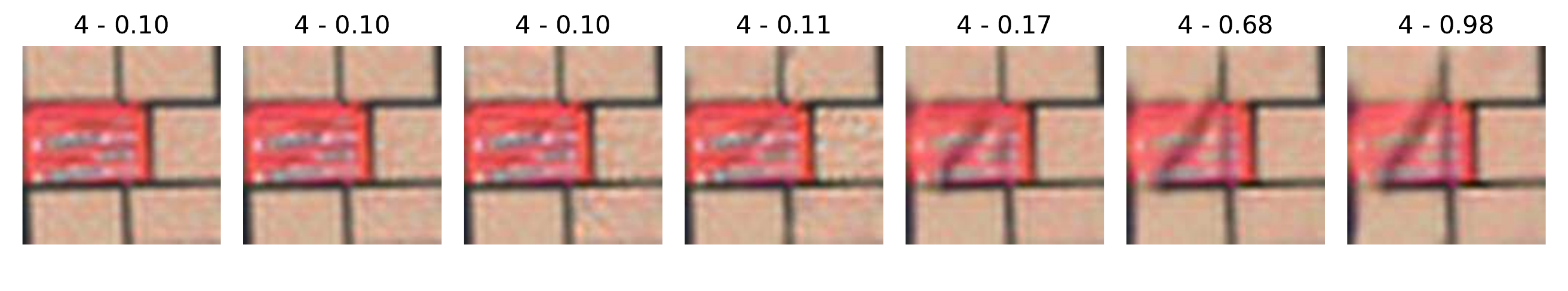}} \\
\end{tabular}	
\vspace{-.5cm}
\caption{\label{fig:od_svhn_new1}\textbf{Feature Generation on OOD images} for SVHN models for unseen images from 80 million tiny images. RATIO$_{0.25}$ is the only model capable of generating realistic digits, whereas both AT models make high confidence predictions in the absence of class-specific features. ACET consistently has low confidences for all budgets as the targeted attack cannot generate with this budget class-specific features of the predicted target class.}

\vspace{-.3cm}
\end{figure}

\begin{figure}[ht!]
\begin{tabular}{p{1cm}x{\breite}x{\breite}x{\breite}x{\breite}x{\breite}x{\breite}x{\breite}x{\breite}}
Model  & Orig. & $\epsilon=0.5$ & $\epsilon=1.0$ & $\epsilon=1.5$ & $\epsilon=2.0$ & $\epsilon=2.5$ & $\epsilon=3.0$\\
\begin{turn}{90} \hspace{-.4cm} ACET \end{turn} & \multicolumn{7}{c}{\includegraphics[width=0.91\textwidth,valign=c]{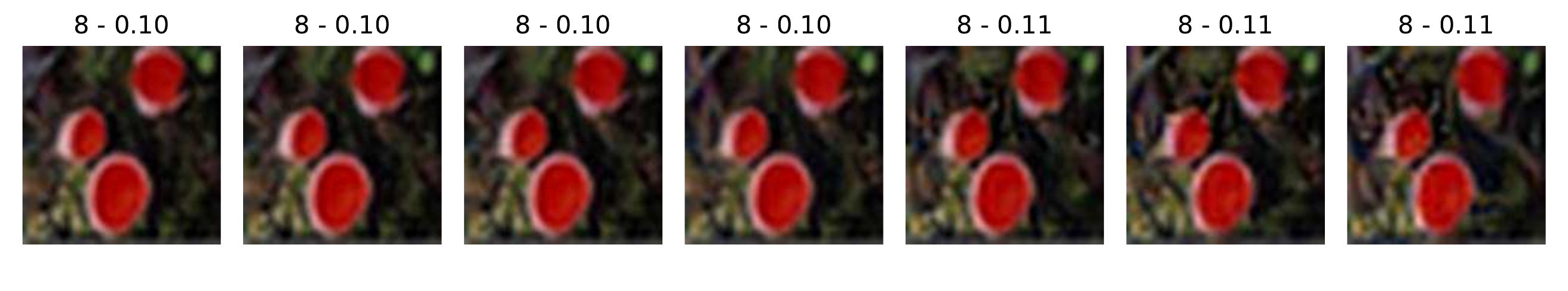}} \\
\hline
\begin{turn}{90} \hspace{-.4cm} AT-0.50 \end{turn}  &  \multicolumn{7}{c}{\includegraphics[width=0.91\textwidth,valign=c]{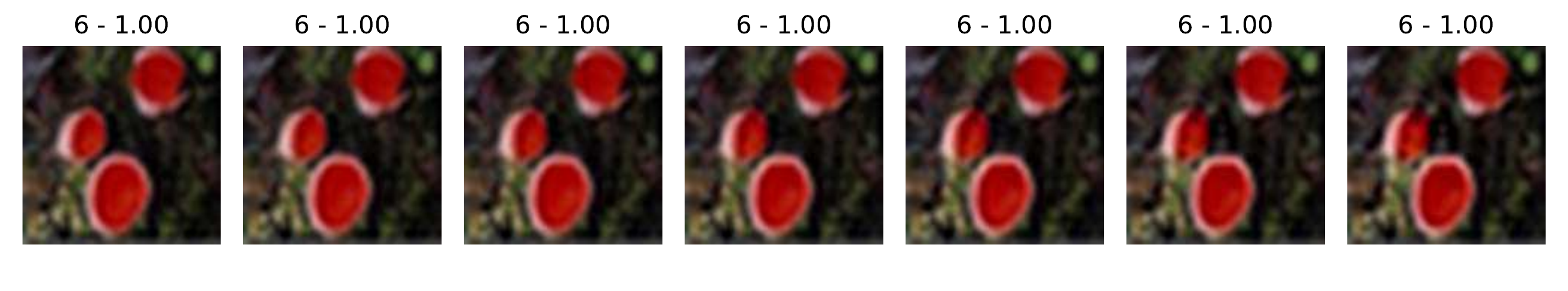}} \\
\hline
\begin{turn}{90} \hspace{-.4cm} AT-0.25 \end{turn} & \multicolumn{7}{c}{\includegraphics[width=0.91\textwidth,valign=c]{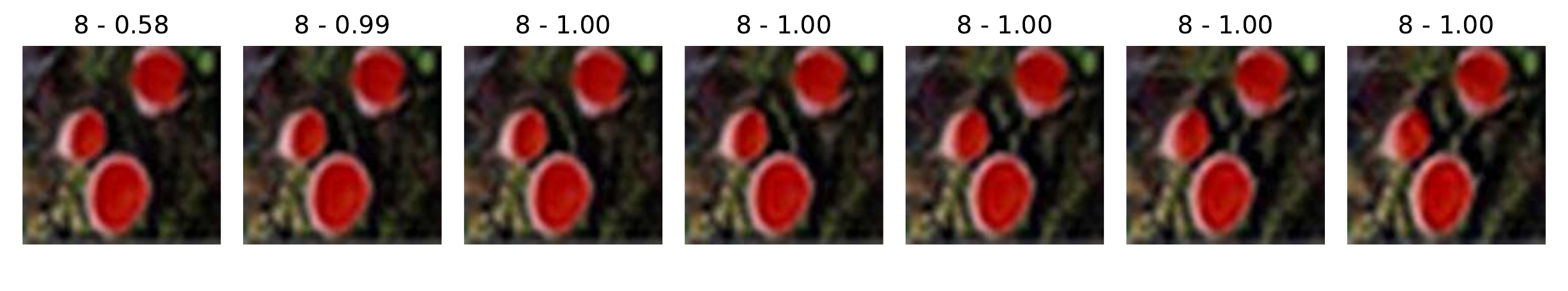}} \\
\hline
\begin{turn}{90} \hspace{-.4cm} R-0.25 \end{turn}  &  \multicolumn{7}{c}{\includegraphics[width=0.91\textwidth,valign=c]{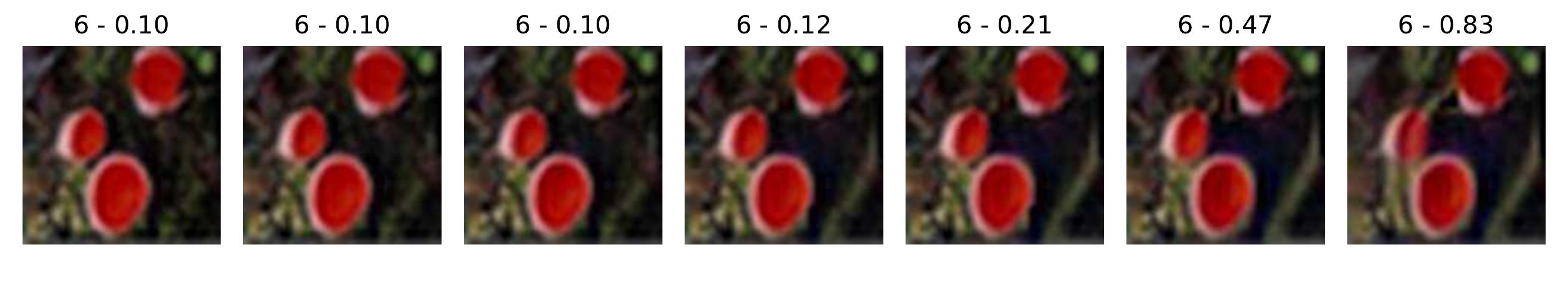}} \\
\vspace{2mm}\\
\begin{turn}{90} \hspace{-.4cm} ACET \end{turn} & \multicolumn{7}{c}{\includegraphics[width=0.91\textwidth,valign=c]{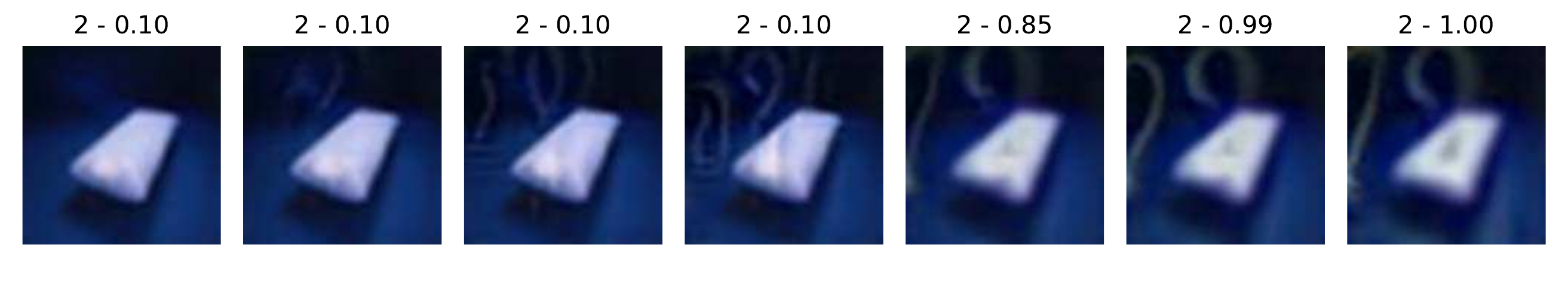}} \\
\hline
\begin{turn}{90} \hspace{-.4cm} AT-0.50 \end{turn}  &  \multicolumn{7}{c}{\includegraphics[width=0.91\textwidth,valign=c]{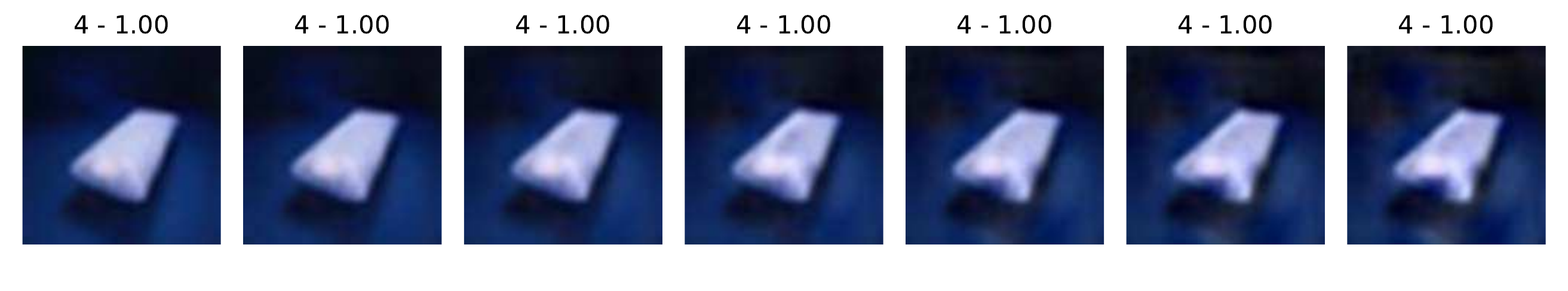}} \\
\hline
\begin{turn}{90} \hspace{-.4cm} AT-0.25 \end{turn} & \multicolumn{7}{c}{\includegraphics[width=0.91\textwidth,valign=c]{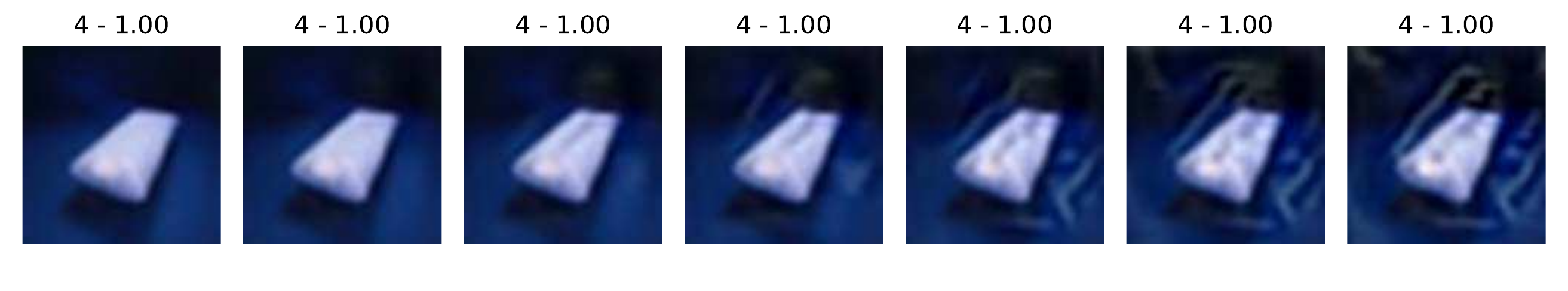}} \\
\hline
\begin{turn}{90} \hspace{-.4cm} R-0.25 \end{turn}  &  \multicolumn{7}{c}{\includegraphics[width=0.91\textwidth,valign=c]{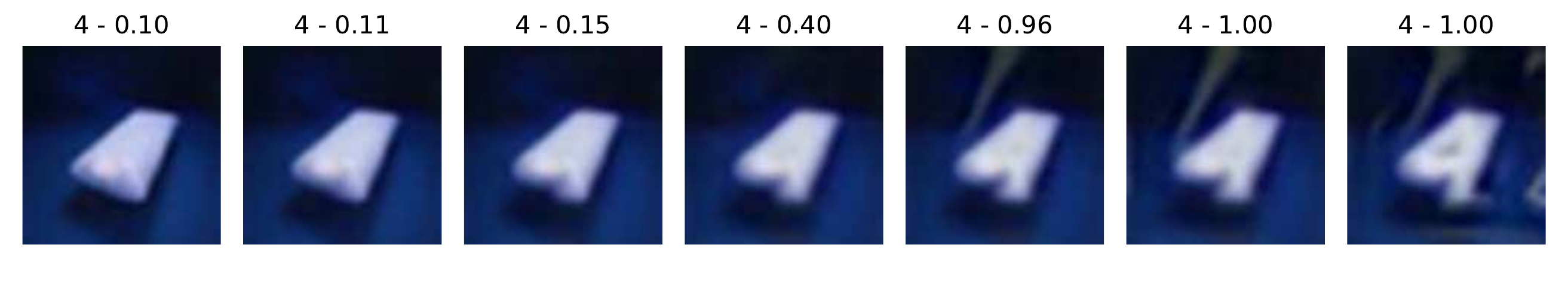}} \\

\end{tabular}	
\vspace{-.5cm}
\caption{\label{fig:od_svhn_new2}\textbf{Feature Generation on OOD images} for SVHN models for unseen images from 80 million tiny images. RATIO$_{0.25}$ generates the best digits and has low confidence on the initial images and the less images with lower radii which do not show yet class-specific features.}

\vspace{-.3cm}
\end{figure}

%% file: res/appendix_vc_cifar100_new.tex
\begin{figure}[ht!]
\begin{tabular}{p{1cm}x{\breite}x{\breite}x{\breite}x{\breite}x{\breite}x{\breite}x{\breite}x{\breite}}
Model  & Orig. & $\epsilon=0.5$ & $\epsilon=1.0$ & $\epsilon=1.5$ & $\epsilon=2.0$ & $\epsilon=2.5$ & $\epsilon=3.0$\\
\begin{turn}{90} \hspace{-.4cm} AT-0.50 \end{turn}  &  \multicolumn{7}{c}{\includegraphics[width=0.91\textwidth,valign=c]{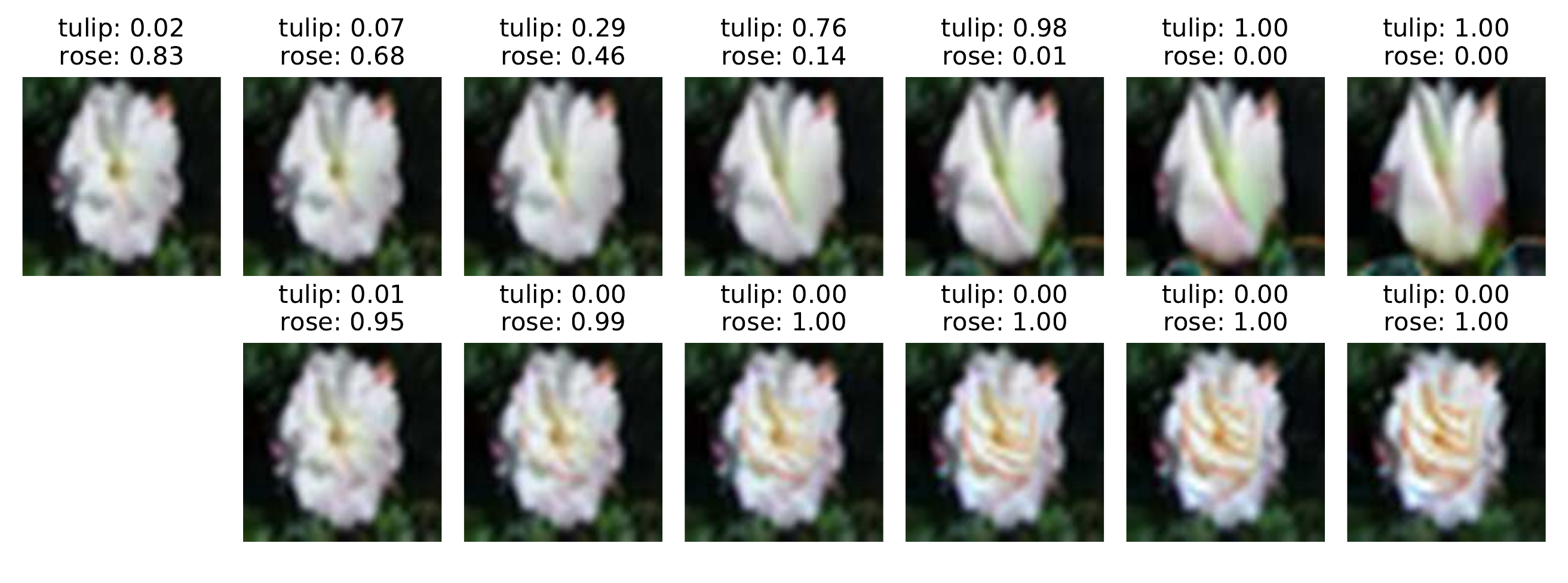}} \\
\hline
\begin{turn}{90} \hspace{-.4cm} AT-0.25 \end{turn}  &  \multicolumn{7}{c}{\includegraphics[width=0.91\textwidth,valign=c]{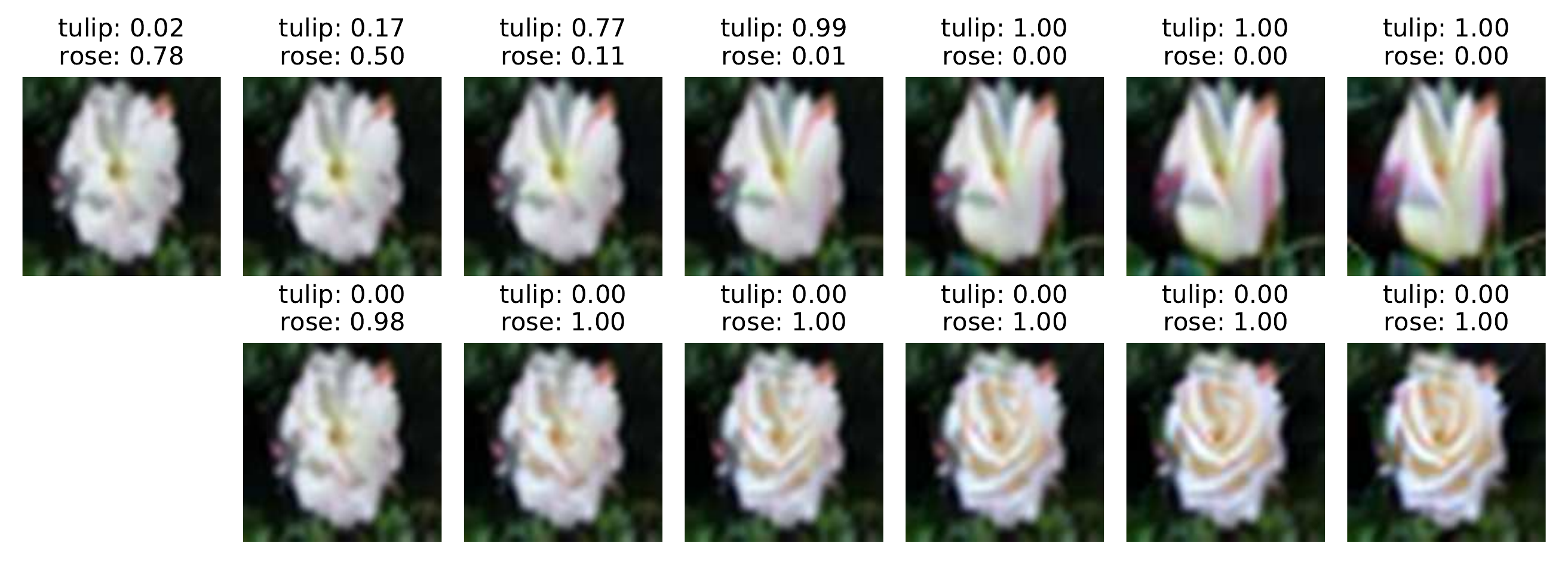}} \\
\hline
\begin{turn}{90} \hspace{-.9cm} RATIO-0.50 \end{turn} & \multicolumn{7}{c}{\includegraphics[width=0.91\textwidth,valign=c]{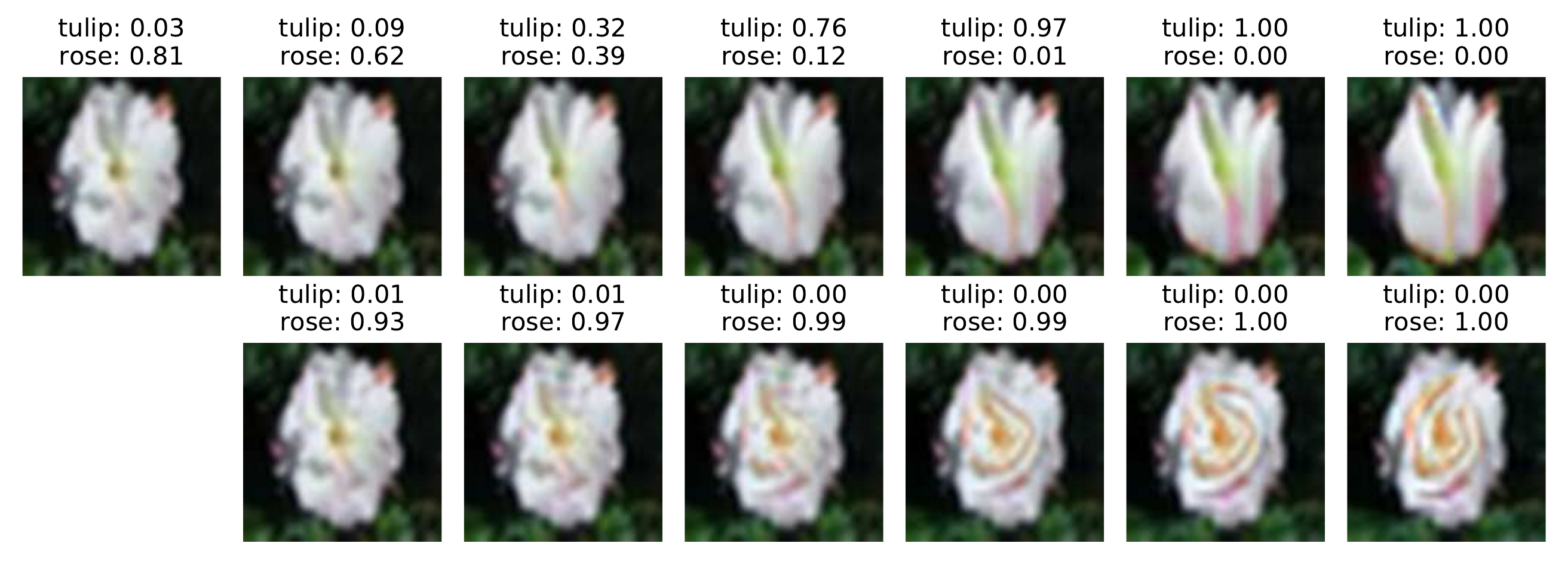}} \\
\hline
\begin{turn}{90} \hspace{-.9cm} RATIO-0.25 \end{turn} & \multicolumn{7}{c}{\includegraphics[width=0.91\textwidth,valign=c]{pics/CIFAR100/Ratio025/VC/157.pdf}} \\
\end{tabular}	
\caption{\label{fig:vc_cifar100_new1}\textbf{Visual Counterfactuals} on CIFAR100 test samples misclassified by all methods. While all models are able to generate class specific features, even for similar classes, the RATIO generated images show less distortions and overall higher image quality.
}
\end{figure}

\begin{figure}[ht!]
\begin{tabular}{p{1cm}x{\breite}x{\breite}x{\breite}x{\breite}x{\breite}x{\breite}x{\breite}x{\breite}}
Model  & Orig. & $\epsilon=0.5$ & $\epsilon=1.0$ & $\epsilon=1.5$ & $\epsilon=2.0$ & $\epsilon=2.5$ & $\epsilon=3.0$\\
\begin{turn}{90} \hspace{-.4cm} AT-0.50 \end{turn}  &  \multicolumn{7}{c}{\includegraphics[width=0.91\textwidth,valign=c]{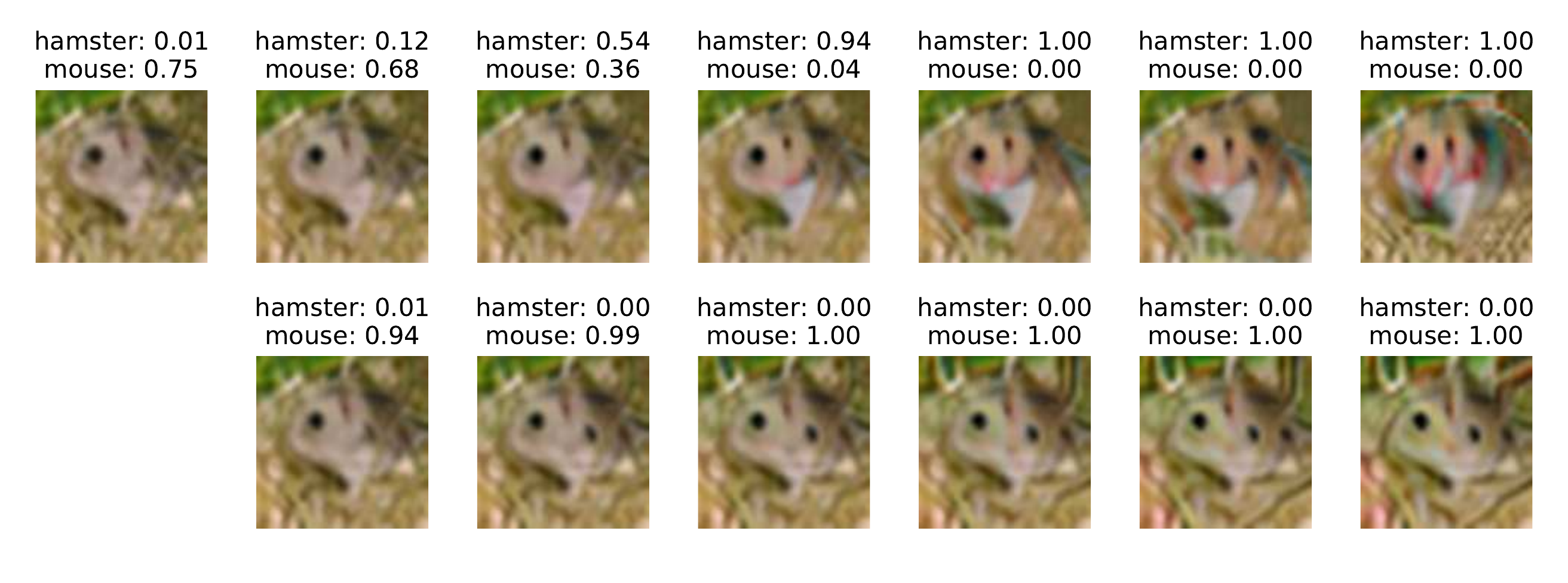}} \\
\hline
\begin{turn}{90} \hspace{-.4cm} AT-0.25 \end{turn}  &  \multicolumn{7}{c}{\includegraphics[width=0.91\textwidth,valign=c]{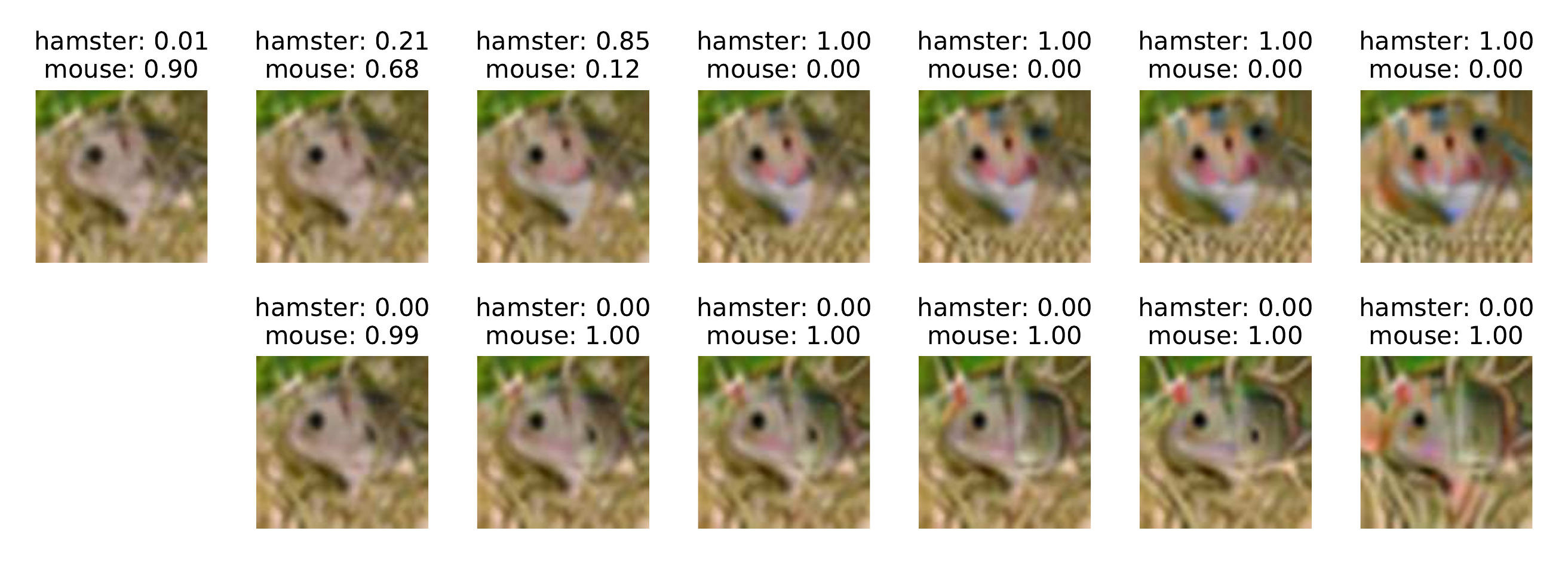}} \\
\hline
\begin{turn}{90} \hspace{-.9cm} RATIO-0.50 \end{turn} & \multicolumn{7}{c}{\includegraphics[width=0.91\textwidth,valign=c]{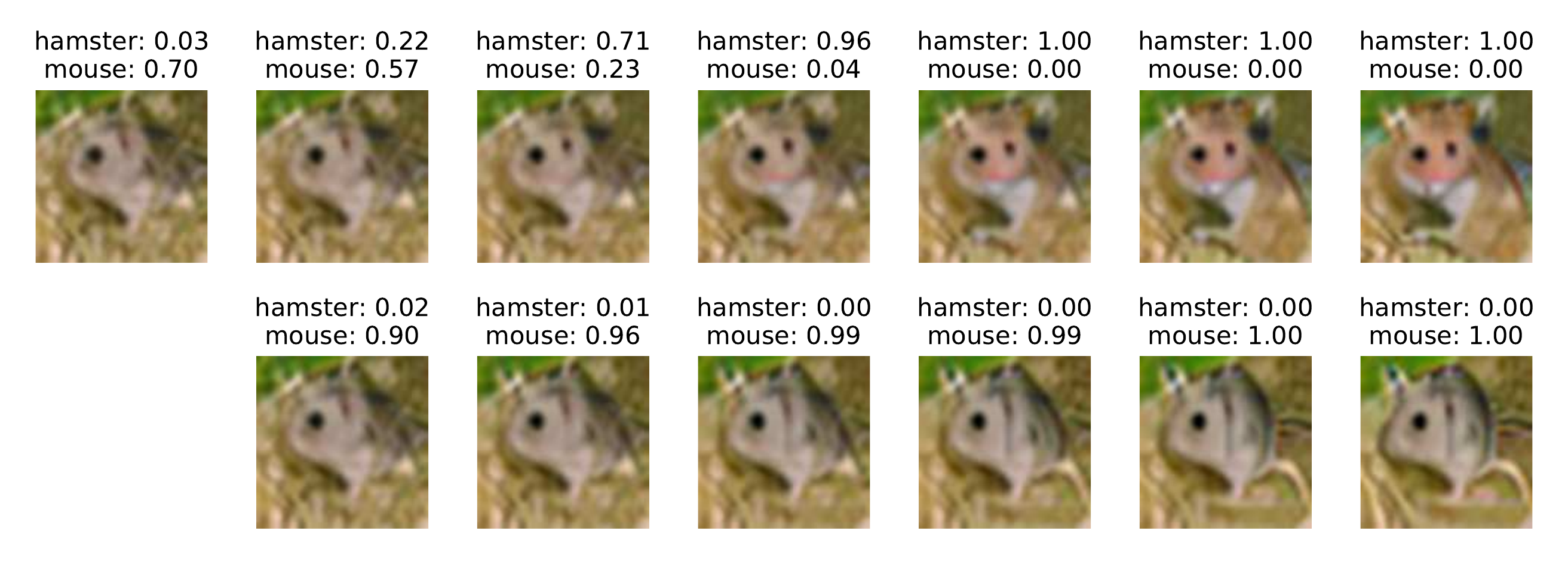}} \\
\hline
\begin{turn}{90} \hspace{-.9cm} RATIO-0.25 \end{turn} & \multicolumn{7}{c}{\includegraphics[width=0.91\textwidth,valign=c]{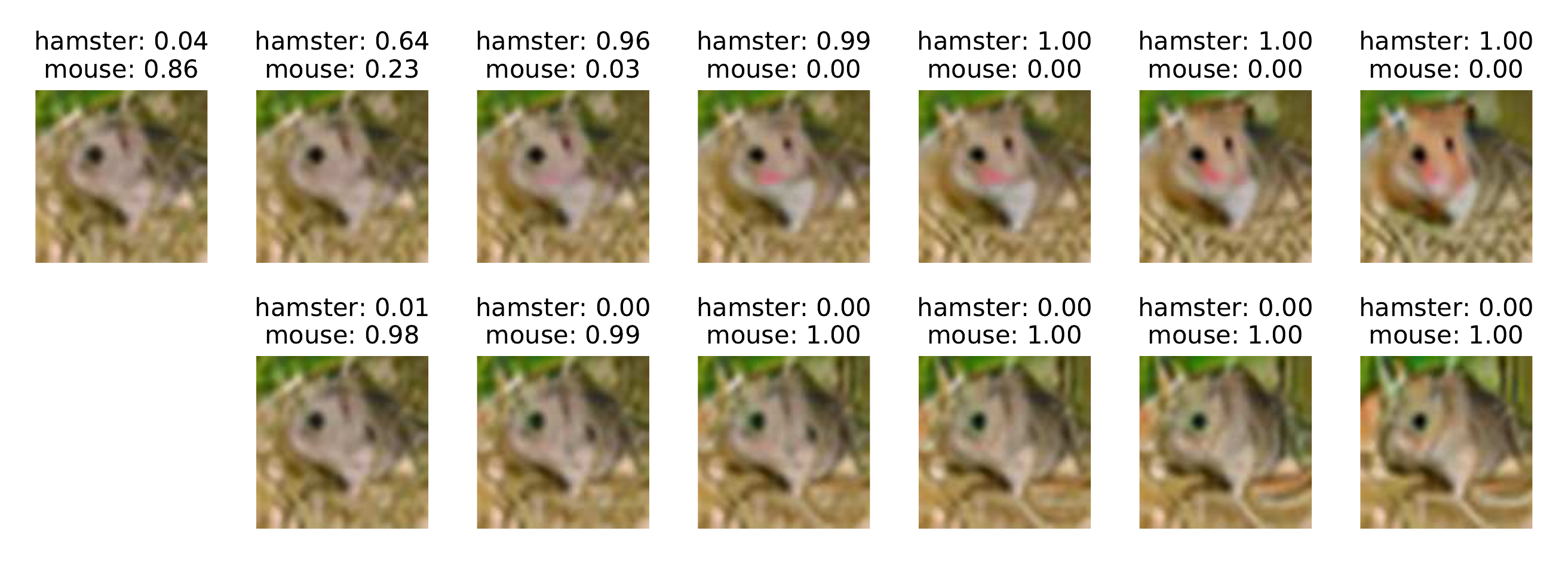}} \\
\end{tabular}	
\caption{\label{fig:vc_cifar100_new2}\textbf{Visual Counterfactuals} on CIFAR100 test samples misclassified by all methods. While all models are able to generate class specific features, even for similar classes, the RATIO generated images show less distortions and overall higher image quality.
}
\end{figure}

\begin{figure}[ht!]
\begin{tabular}{p{1cm}x{\breite}x{\breite}x{\breite}x{\breite}x{\breite}x{\breite}x{\breite}x{\breite}}
Model  & Orig. & $\epsilon=0.5$ & $\epsilon=1.0$ & $\epsilon=1.5$ & $\epsilon=2.0$ & $\epsilon=2.5$ & $\epsilon=3.0$\\
\begin{turn}{90} \hspace{-.4cm} AT-0.50 \end{turn}  &  \multicolumn{7}{c}{\includegraphics[width=0.91\textwidth,valign=c]{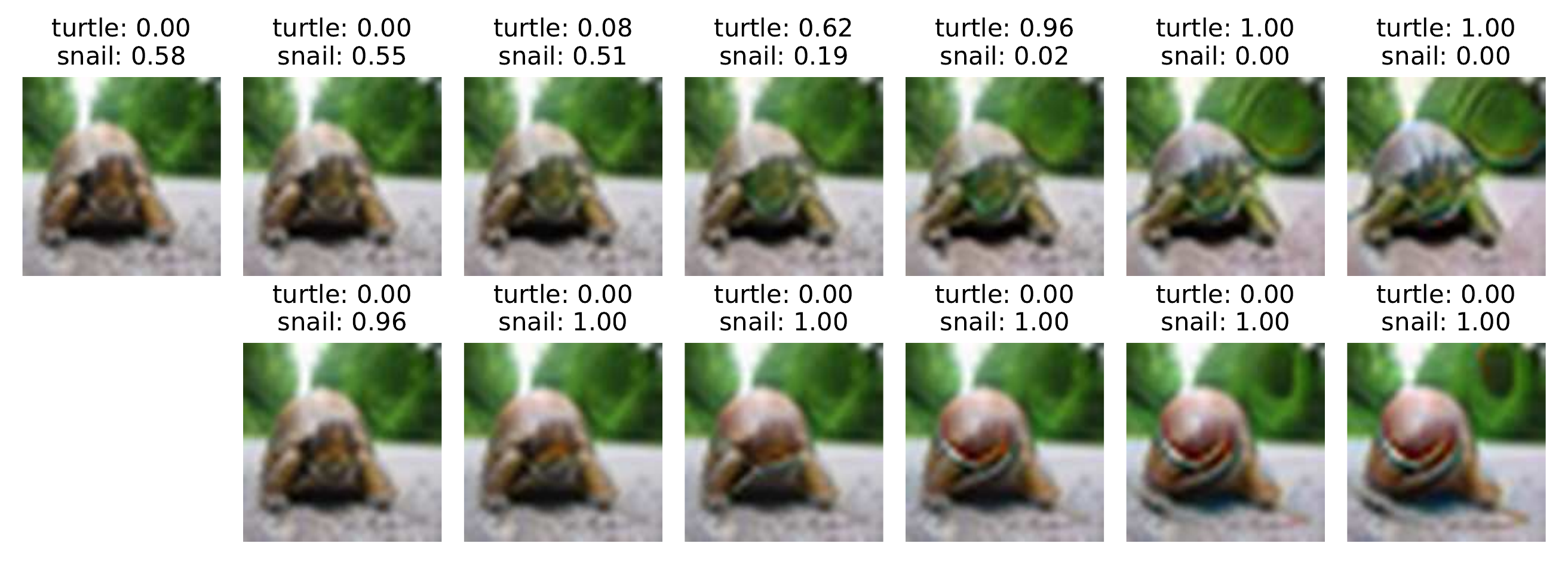}} \\
\hline
\begin{turn}{90} \hspace{-.4cm} AT-0.25 \end{turn}  &  \multicolumn{7}{c}{\includegraphics[width=0.91\textwidth,valign=c]{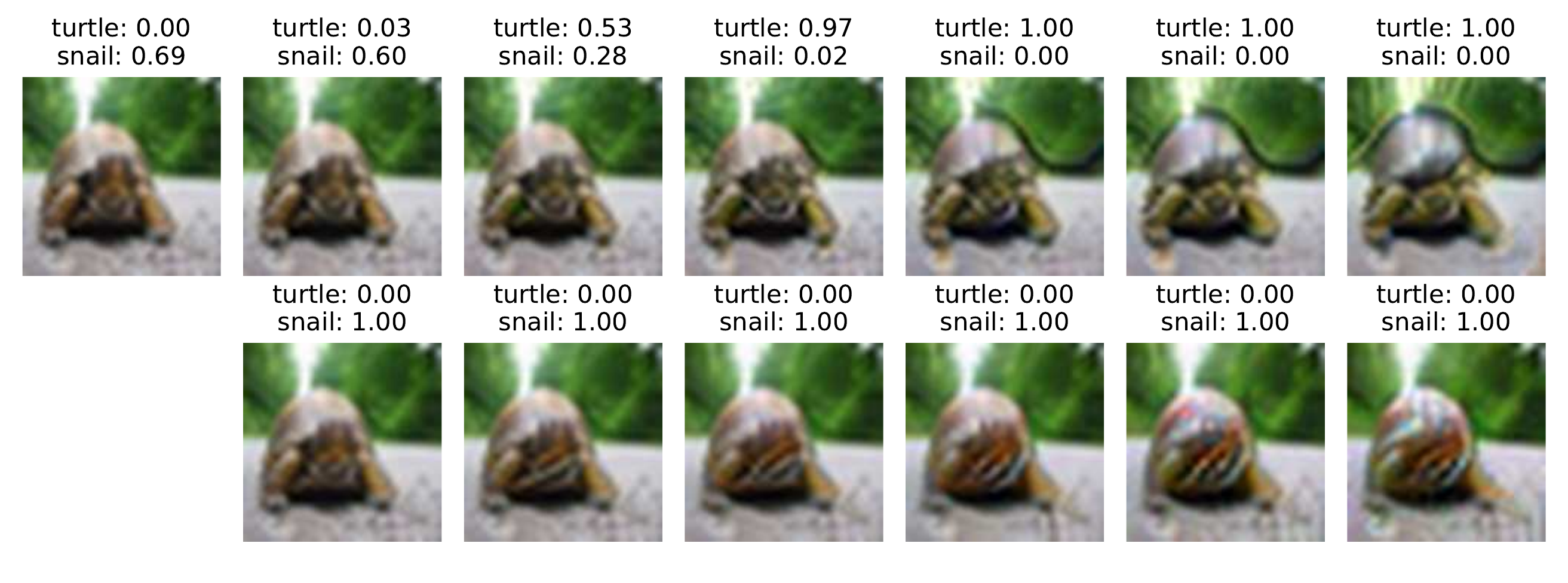}} \\
\hline
\begin{turn}{90} \hspace{-.9cm} RATIO-0.50 \end{turn} & \multicolumn{7}{c}{\includegraphics[width=0.91\textwidth,valign=c]{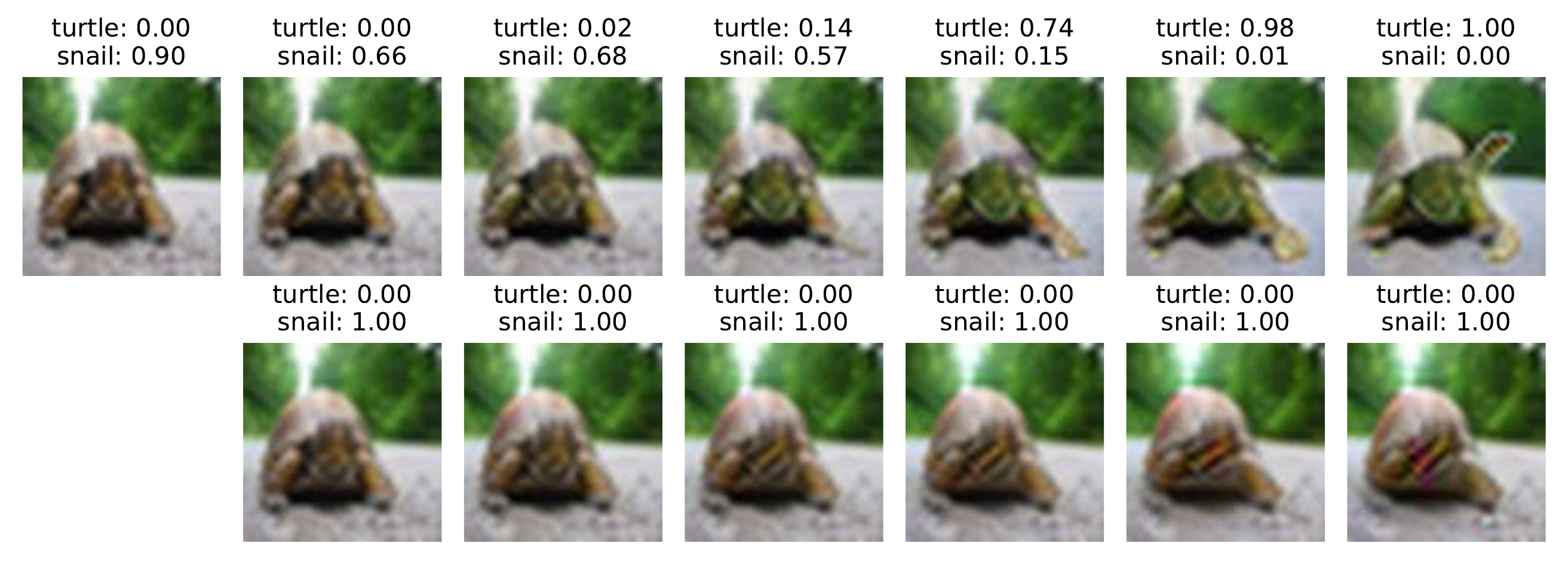}} \\
\hline
\begin{turn}{90} \hspace{-.9cm} RATIO-0.25 \end{turn} & \multicolumn{7}{c}{\includegraphics[width=0.91\textwidth,valign=c]{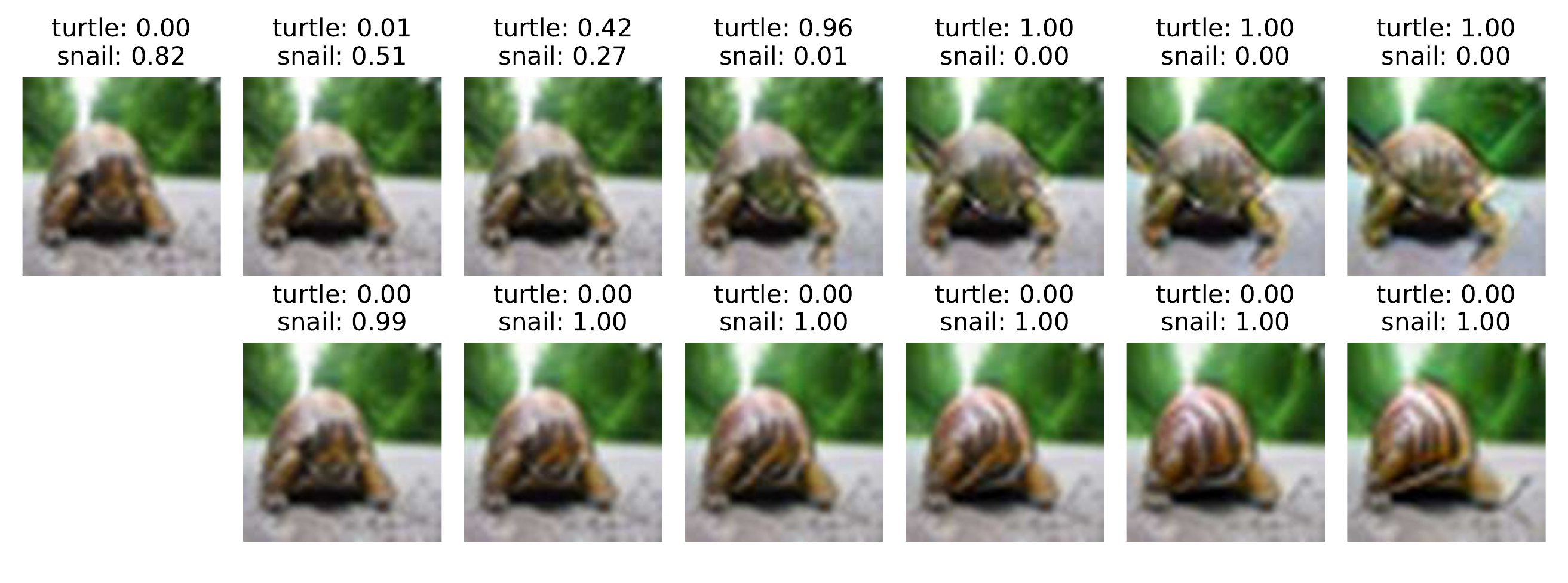}} \\
\end{tabular}	
\caption{\label{fig:vc_cifar100_new3}\textbf{Visual Counterfactuals} on CIFAR100 test samples misclassified by all methods. The two RATIO models are able to generate higher quality images for both the ground truth and falsely predicted class.
}
\end{figure}

%% file: res/appendix_cifar100_overview.tex
\begin{figure}
\begin{adjustbox}{max width=\textwidth}
\begin{tabu}{ccccccccccccccccc}
\multicolumn{8}{c}{Original} & & \multicolumn{8}{c}{Plain}\\
\rowfont{\tiny}
telephone & bus & forest & oak & girl & caterpillar & man & palm & & telephone & bus & forest & oak & girl & caterpillar & man & palm\\
\includegraphics[width=0.056\textwidth,valign=c]{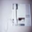}
 & 
\includegraphics[width=0.056\textwidth,valign=c]{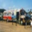}
 & 
\includegraphics[width=0.056\textwidth,valign=c]{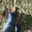}
 & 
\includegraphics[width=0.056\textwidth,valign=c]{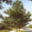}
 & 
\includegraphics[width=0.056\textwidth,valign=c]{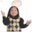}
 & 
\includegraphics[width=0.056\textwidth,valign=c]{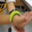}
 & 
\includegraphics[width=0.056\textwidth,valign=c]{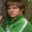}
 & 
\includegraphics[width=0.056\textwidth,valign=c]{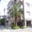}
 & 
\hspace{ 0.028 \textwidth} & 
\includegraphics[width=0.056\textwidth,valign=c]{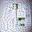}
 & 
\includegraphics[width=0.056\textwidth,valign=c]{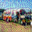}
 & 
\includegraphics[width=0.056\textwidth,valign=c]{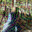}
 & 
\includegraphics[width=0.056\textwidth,valign=c]{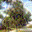}
 & 
\includegraphics[width=0.056\textwidth,valign=c]{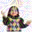}
 & 
\includegraphics[width=0.056\textwidth,valign=c]{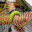}
 & 
\includegraphics[width=0.056\textwidth,valign=c]{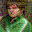}
 & 
\includegraphics[width=0.056\textwidth,valign=c]{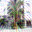}
\\
\rowfont{\tiny}
camel & beaver & beaver & house & oak & beaver & couch & beaver & & camel & beaver & beaver & house & oak & beaver & couch & beaver\\
\includegraphics[width=0.056\textwidth,valign=c]{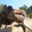}
 & 
\includegraphics[width=0.056\textwidth,valign=c]{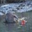}
 & 
\includegraphics[width=0.056\textwidth,valign=c]{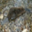}
 & 
\includegraphics[width=0.056\textwidth,valign=c]{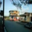}
 & 
\includegraphics[width=0.056\textwidth,valign=c]{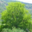}
 & 
\includegraphics[width=0.056\textwidth,valign=c]{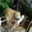}
 & 
\includegraphics[width=0.056\textwidth,valign=c]{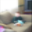}
 & 
\includegraphics[width=0.056\textwidth,valign=c]{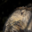}
 & 
\hspace{ 0.028 \textwidth} & 
\includegraphics[width=0.056\textwidth,valign=c]{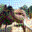}
 & 
\includegraphics[width=0.056\textwidth,valign=c]{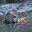}
 & 
\includegraphics[width=0.056\textwidth,valign=c]{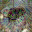}
 & 
\includegraphics[width=0.056\textwidth,valign=c]{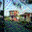}
 & 
\includegraphics[width=0.056\textwidth,valign=c]{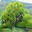}
 & 
\includegraphics[width=0.056\textwidth,valign=c]{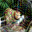}
 & 
\includegraphics[width=0.056\textwidth,valign=c]{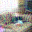}
 & 
\includegraphics[width=0.056\textwidth,valign=c]{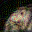}
\\
\rowfont{\tiny}
house & raccoon & bowl & cloud & table & girl & table & bed & & house & raccoon & bowl & cloud & table & girl & table & bed\\
\includegraphics[width=0.056\textwidth,valign=c]{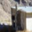}
 & 
\includegraphics[width=0.056\textwidth,valign=c]{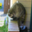}
 & 
\includegraphics[width=0.056\textwidth,valign=c]{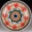}
 & 
\includegraphics[width=0.056\textwidth,valign=c]{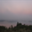}
 & 
\includegraphics[width=0.056\textwidth,valign=c]{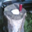}
 & 
\includegraphics[width=0.056\textwidth,valign=c]{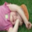}
 & 
\includegraphics[width=0.056\textwidth,valign=c]{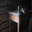}
 & 
\includegraphics[width=0.056\textwidth,valign=c]{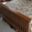}
 & 
\hspace{ 0.028 \textwidth} & 
\includegraphics[width=0.056\textwidth,valign=c]{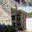}
 & 
\includegraphics[width=0.056\textwidth,valign=c]{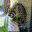}
 & 
\includegraphics[width=0.056\textwidth,valign=c]{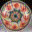}
 & 
\includegraphics[width=0.056\textwidth,valign=c]{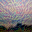}
 & 
\includegraphics[width=0.056\textwidth,valign=c]{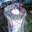}
 & 
\includegraphics[width=0.056\textwidth,valign=c]{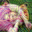}
 & 
\includegraphics[width=0.056\textwidth,valign=c]{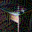}
 & 
\includegraphics[width=0.056\textwidth,valign=c]{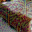}
\\
\rowfont{\tiny}
kangaroo & possum & girl & lamp & bowl & snake & turtle & snail & & kangaroo & possum & girl & lamp & bowl & snake & turtle & snail\\
\includegraphics[width=0.056\textwidth,valign=c]{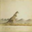}
 & 
\includegraphics[width=0.056\textwidth,valign=c]{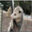}
 & 
\includegraphics[width=0.056\textwidth,valign=c]{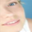}
 & 
\includegraphics[width=0.056\textwidth,valign=c]{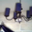}
 & 
\includegraphics[width=0.056\textwidth,valign=c]{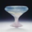}
 & 
\includegraphics[width=0.056\textwidth,valign=c]{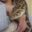}
 & 
\includegraphics[width=0.056\textwidth,valign=c]{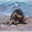}
 & 
\includegraphics[width=0.056\textwidth,valign=c]{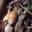}
 & 
\hspace{ 0.028 \textwidth} & 
\includegraphics[width=0.056\textwidth,valign=c]{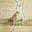}
 & 
\includegraphics[width=0.056\textwidth,valign=c]{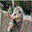}
 & 
\includegraphics[width=0.056\textwidth,valign=c]{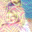}
 & 
\includegraphics[width=0.056\textwidth,valign=c]{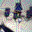}
 & 
\includegraphics[width=0.056\textwidth,valign=c]{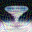}
 & 
\includegraphics[width=0.056\textwidth,valign=c]{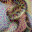}
 & 
\includegraphics[width=0.056\textwidth,valign=c]{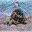}
 & 
\includegraphics[width=0.056\textwidth,valign=c]{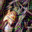}
\\
\rowfont{\tiny}
man & willow & woman & cattle & lobster & seal & seal & bus & & man & willow & woman & cattle & lobster & seal & seal & bus\\
\includegraphics[width=0.056\textwidth,valign=c]{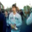}
 & 
\includegraphics[width=0.056\textwidth,valign=c]{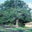}
 & 
\includegraphics[width=0.056\textwidth,valign=c]{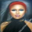}
 & 
\includegraphics[width=0.056\textwidth,valign=c]{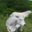}
 & 
\includegraphics[width=0.056\textwidth,valign=c]{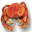}
 & 
\includegraphics[width=0.056\textwidth,valign=c]{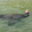}
 & 
\includegraphics[width=0.056\textwidth,valign=c]{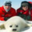}
 & 
\includegraphics[width=0.056\textwidth,valign=c]{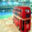}
 & 
\hspace{ 0.028 \textwidth} & 
\includegraphics[width=0.056\textwidth,valign=c]{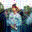}
 & 
\includegraphics[width=0.056\textwidth,valign=c]{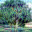}
 & 
\includegraphics[width=0.056\textwidth,valign=c]{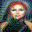}
 & 
\includegraphics[width=0.056\textwidth,valign=c]{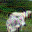}
 & 
\includegraphics[width=0.056\textwidth,valign=c]{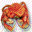}
 & 
\includegraphics[width=0.056\textwidth,valign=c]{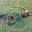}
 & 
\includegraphics[width=0.056\textwidth,valign=c]{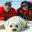}
 & 
\includegraphics[width=0.056\textwidth,valign=c]{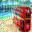}
\\
\rowfont{\tiny}
boy & tulip & bee & mushroom & hamster & seal & otter & kangaroo & & boy & tulip & bee & mushroom & hamster & seal & otter & kangaroo\\
\includegraphics[width=0.056\textwidth,valign=c]{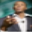}
 & 
\includegraphics[width=0.056\textwidth,valign=c]{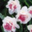}
 & 
\includegraphics[width=0.056\textwidth,valign=c]{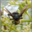}
 & 
\includegraphics[width=0.056\textwidth,valign=c]{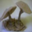}
 & 
\includegraphics[width=0.056\textwidth,valign=c]{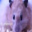}
 & 
\includegraphics[width=0.056\textwidth,valign=c]{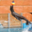}
 & 
\includegraphics[width=0.056\textwidth,valign=c]{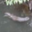}
 & 
\includegraphics[width=0.056\textwidth,valign=c]{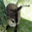}
 & 
\hspace{ 0.028 \textwidth} & 
\includegraphics[width=0.056\textwidth,valign=c]{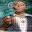}
 & 
\includegraphics[width=0.056\textwidth,valign=c]{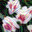}
 & 
\includegraphics[width=0.056\textwidth,valign=c]{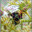}
 & 
\includegraphics[width=0.056\textwidth,valign=c]{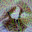}
 & 
\includegraphics[width=0.056\textwidth,valign=c]{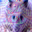}
 & 
\includegraphics[width=0.056\textwidth,valign=c]{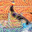}
 & 
\includegraphics[width=0.056\textwidth,valign=c]{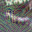}
 & 
\includegraphics[width=0.056\textwidth,valign=c]{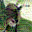}
\vspace{2mm}\\
\multicolumn{8}{c}{AT-0.5} & & \multicolumn{8}{c}{AT-0.25}\\
\includegraphics[width=0.056\textwidth,valign=c]{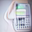}
 & 
\includegraphics[width=0.056\textwidth,valign=c]{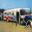}
 & 
\includegraphics[width=0.056\textwidth,valign=c]{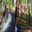}
 & 
\includegraphics[width=0.056\textwidth,valign=c]{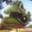}
 & 
\includegraphics[width=0.056\textwidth,valign=c]{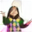}
 & 
\includegraphics[width=0.056\textwidth,valign=c]{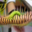}
 & 
\includegraphics[width=0.056\textwidth,valign=c]{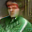}
 & 
\includegraphics[width=0.056\textwidth,valign=c]{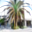}
 & 
\hspace{ 0.028 \textwidth} & 
\includegraphics[width=0.056\textwidth,valign=c]{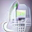}
 & 
\includegraphics[width=0.056\textwidth,valign=c]{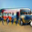}
 & 
\includegraphics[width=0.056\textwidth,valign=c]{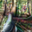}
 & 
\includegraphics[width=0.056\textwidth,valign=c]{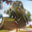}
 & 
\includegraphics[width=0.056\textwidth,valign=c]{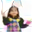}
 & 
\includegraphics[width=0.056\textwidth,valign=c]{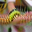}
 & 
\includegraphics[width=0.056\textwidth,valign=c]{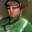}
 & 
\includegraphics[width=0.056\textwidth,valign=c]{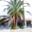}
\vspace{1mm}\\
\includegraphics[width=0.056\textwidth,valign=c]{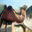}
 & 
\includegraphics[width=0.056\textwidth,valign=c]{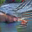}
 & 
\includegraphics[width=0.056\textwidth,valign=c]{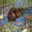}
 & 
\includegraphics[width=0.056\textwidth,valign=c]{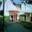}
 & 
\includegraphics[width=0.056\textwidth,valign=c]{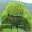}
 & 
\includegraphics[width=0.056\textwidth,valign=c]{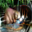}
 & 
\includegraphics[width=0.056\textwidth,valign=c]{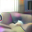}
 & 
\includegraphics[width=0.056\textwidth,valign=c]{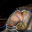}
 & 
\hspace{ 0.028 \textwidth} & 
\includegraphics[width=0.056\textwidth,valign=c]{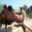}
 & 
\includegraphics[width=0.056\textwidth,valign=c]{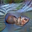}
 & 
\includegraphics[width=0.056\textwidth,valign=c]{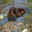}
 & 
\includegraphics[width=0.056\textwidth,valign=c]{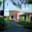}
 & 
\includegraphics[width=0.056\textwidth,valign=c]{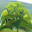}
 & 
\includegraphics[width=0.056\textwidth,valign=c]{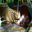}
 & 
\includegraphics[width=0.056\textwidth,valign=c]{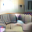}
 & 
\includegraphics[width=0.056\textwidth,valign=c]{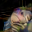}
\vspace{1mm}\\
\includegraphics[width=0.056\textwidth,valign=c]{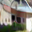}
 & 
\includegraphics[width=0.056\textwidth,valign=c]{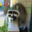}
 & 
\includegraphics[width=0.056\textwidth,valign=c]{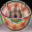}
 & 
\includegraphics[width=0.056\textwidth,valign=c]{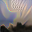}
 & 
\includegraphics[width=0.056\textwidth,valign=c]{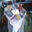}
 & 
\includegraphics[width=0.056\textwidth,valign=c]{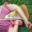}
 & 
\includegraphics[width=0.056\textwidth,valign=c]{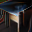}
 & 
\includegraphics[width=0.056\textwidth,valign=c]{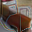}
 & 
\hspace{ 0.028 \textwidth} & 
\includegraphics[width=0.056\textwidth,valign=c]{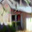}
 & 
\includegraphics[width=0.056\textwidth,valign=c]{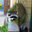}
 & 
\includegraphics[width=0.056\textwidth,valign=c]{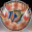}
 & 
\includegraphics[width=0.056\textwidth,valign=c]{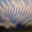}
 & 
\includegraphics[width=0.056\textwidth,valign=c]{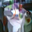}
 & 
\includegraphics[width=0.056\textwidth,valign=c]{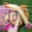}
 & 
\includegraphics[width=0.056\textwidth,valign=c]{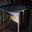}
 & 
\includegraphics[width=0.056\textwidth,valign=c]{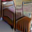}
\vspace{1mm}\\
\includegraphics[width=0.056\textwidth,valign=c]{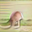}
 & 
\includegraphics[width=0.056\textwidth,valign=c]{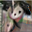}
 & 
\includegraphics[width=0.056\textwidth,valign=c]{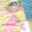}
 & 
\includegraphics[width=0.056\textwidth,valign=c]{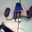}
 & 
\includegraphics[width=0.056\textwidth,valign=c]{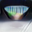}
 & 
\includegraphics[width=0.056\textwidth,valign=c]{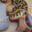}
 & 
\includegraphics[width=0.056\textwidth,valign=c]{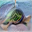}
 & 
\includegraphics[width=0.056\textwidth,valign=c]{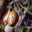}
 & 
\hspace{ 0.028 \textwidth} & 
\includegraphics[width=0.056\textwidth,valign=c]{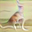}
 & 
\includegraphics[width=0.056\textwidth,valign=c]{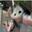}
 & 
\includegraphics[width=0.056\textwidth,valign=c]{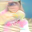}
 & 
\includegraphics[width=0.056\textwidth,valign=c]{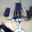}
 & 
\includegraphics[width=0.056\textwidth,valign=c]{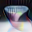}
 & 
\includegraphics[width=0.056\textwidth,valign=c]{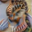}
 & 
\includegraphics[width=0.056\textwidth,valign=c]{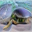}
 & 
\includegraphics[width=0.056\textwidth,valign=c]{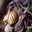}
\vspace{1mm}\\
\includegraphics[width=0.056\textwidth,valign=c]{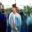}
 & 
\includegraphics[width=0.056\textwidth,valign=c]{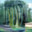}
 & 
\includegraphics[width=0.056\textwidth,valign=c]{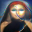}
 & 
\includegraphics[width=0.056\textwidth,valign=c]{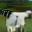}
 & 
\includegraphics[width=0.056\textwidth,valign=c]{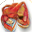}
 & 
\includegraphics[width=0.056\textwidth,valign=c]{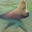}
 & 
\includegraphics[width=0.056\textwidth,valign=c]{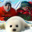}
 & 
\includegraphics[width=0.056\textwidth,valign=c]{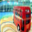}
 & 
\hspace{ 0.028 \textwidth} & 
\includegraphics[width=0.056\textwidth,valign=c]{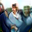}
 & 
\includegraphics[width=0.056\textwidth,valign=c]{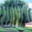}
 & 
\includegraphics[width=0.056\textwidth,valign=c]{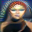}
 & 
\includegraphics[width=0.056\textwidth,valign=c]{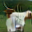}
 & 
\includegraphics[width=0.056\textwidth,valign=c]{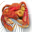}
 & 
\includegraphics[width=0.056\textwidth,valign=c]{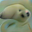}
 & 
\includegraphics[width=0.056\textwidth,valign=c]{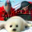}
 & 
\includegraphics[width=0.056\textwidth,valign=c]{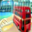}
\vspace{1mm}\\
\includegraphics[width=0.056\textwidth,valign=c]{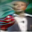}
 & 
\includegraphics[width=0.056\textwidth,valign=c]{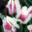}
 & 
\includegraphics[width=0.056\textwidth,valign=c]{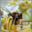}
 & 
\includegraphics[width=0.056\textwidth,valign=c]{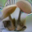}
 & 
\includegraphics[width=0.056\textwidth,valign=c]{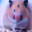}
 & 
\includegraphics[width=0.056\textwidth,valign=c]{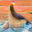}
 & 
\includegraphics[width=0.056\textwidth,valign=c]{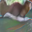}
 & 
\includegraphics[width=0.056\textwidth,valign=c]{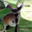}
 & 
\hspace{ 0.028 \textwidth} & 
\includegraphics[width=0.056\textwidth,valign=c]{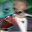}
 & 
\includegraphics[width=0.056\textwidth,valign=c]{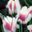}
 & 
\includegraphics[width=0.056\textwidth,valign=c]{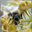}
 & 
\includegraphics[width=0.056\textwidth,valign=c]{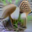}
 & 
\includegraphics[width=0.056\textwidth,valign=c]{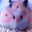}
 & 
\includegraphics[width=0.056\textwidth,valign=c]{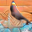}
 & 
\includegraphics[width=0.056\textwidth,valign=c]{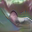}
 & 
\includegraphics[width=0.056\textwidth,valign=c]{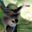}
\vspace{2mm}\\
\multicolumn{8}{c}{RATIO-0.5} & & \multicolumn{8}{c}{RATIO-0.25}\\
\includegraphics[width=0.056\textwidth,valign=c]{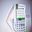}
 & 
\includegraphics[width=0.056\textwidth,valign=c]{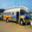}
 & 
\includegraphics[width=0.056\textwidth,valign=c]{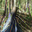}
 & 
\includegraphics[width=0.056\textwidth,valign=c]{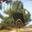}
 & 
\includegraphics[width=0.056\textwidth,valign=c]{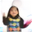}
 & 
\includegraphics[width=0.056\textwidth,valign=c]{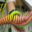}
 & 
\includegraphics[width=0.056\textwidth,valign=c]{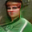}
 & 
\includegraphics[width=0.056\textwidth,valign=c]{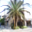}
 & 
\hspace{ 0.028 \textwidth} & 
\includegraphics[width=0.056\textwidth,valign=c]{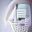}
 & 
\includegraphics[width=0.056\textwidth,valign=c]{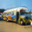}
 & 
\includegraphics[width=0.056\textwidth,valign=c]{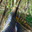}
 & 
\includegraphics[width=0.056\textwidth,valign=c]{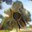}
 & 
\includegraphics[width=0.056\textwidth,valign=c]{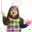}
 & 
\includegraphics[width=0.056\textwidth,valign=c]{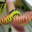}
 & 
\includegraphics[width=0.056\textwidth,valign=c]{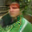}
 & 
\includegraphics[width=0.056\textwidth,valign=c]{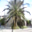}
\vspace{1mm}\\
\includegraphics[width=0.056\textwidth,valign=c]{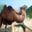}
 & 
\includegraphics[width=0.056\textwidth,valign=c]{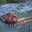}
 & 
\includegraphics[width=0.056\textwidth,valign=c]{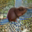}
 & 
\includegraphics[width=0.056\textwidth,valign=c]{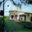}
 & 
\includegraphics[width=0.056\textwidth,valign=c]{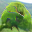}
 & 
\includegraphics[width=0.056\textwidth,valign=c]{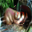}
 & 
\includegraphics[width=0.056\textwidth,valign=c]{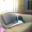}
 & 
\includegraphics[width=0.056\textwidth,valign=c]{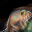}
 & 
\hspace{ 0.028 \textwidth} & 
\includegraphics[width=0.056\textwidth,valign=c]{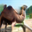}
 & 
\includegraphics[width=0.056\textwidth,valign=c]{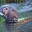}
 & 
\includegraphics[width=0.056\textwidth,valign=c]{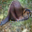}
 & 
\includegraphics[width=0.056\textwidth,valign=c]{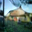}
 & 
\includegraphics[width=0.056\textwidth,valign=c]{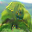}
 & 
\includegraphics[width=0.056\textwidth,valign=c]{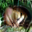}
 & 
\includegraphics[width=0.056\textwidth,valign=c]{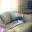}
 & 
\includegraphics[width=0.056\textwidth,valign=c]{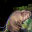}
\vspace{1mm}\\
\includegraphics[width=0.056\textwidth,valign=c]{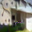}
 & 
\includegraphics[width=0.056\textwidth,valign=c]{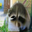}
 & 
\includegraphics[width=0.056\textwidth,valign=c]{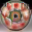}
 & 
\includegraphics[width=0.056\textwidth,valign=c]{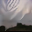}
 & 
\includegraphics[width=0.056\textwidth,valign=c]{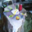}
 & 
\includegraphics[width=0.056\textwidth,valign=c]{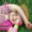}
 & 
\includegraphics[width=0.056\textwidth,valign=c]{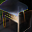}
 & 
\includegraphics[width=0.056\textwidth,valign=c]{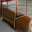}
 & 
\hspace{ 0.028 \textwidth} & 
\includegraphics[width=0.056\textwidth,valign=c]{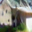}
 & 
\includegraphics[width=0.056\textwidth,valign=c]{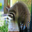}
 & 
\includegraphics[width=0.056\textwidth,valign=c]{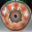}
 & 
\includegraphics[width=0.056\textwidth,valign=c]{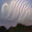}
 & 
\includegraphics[width=0.056\textwidth,valign=c]{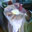}
 & 
\includegraphics[width=0.056\textwidth,valign=c]{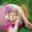}
 & 
\includegraphics[width=0.056\textwidth,valign=c]{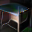}
 & 
\includegraphics[width=0.056\textwidth,valign=c]{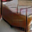}
\vspace{1mm}\\
\includegraphics[width=0.056\textwidth,valign=c]{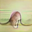}
 & 
\includegraphics[width=0.056\textwidth,valign=c]{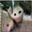}
 & 
\includegraphics[width=0.056\textwidth,valign=c]{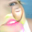}
 & 
\includegraphics[width=0.056\textwidth,valign=c]{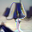}
 & 
\includegraphics[width=0.056\textwidth,valign=c]{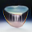}
 & 
\includegraphics[width=0.056\textwidth,valign=c]{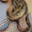}
 & 
\includegraphics[width=0.056\textwidth,valign=c]{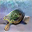}
 & 
\includegraphics[width=0.056\textwidth,valign=c]{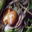}
 & 
\hspace{ 0.028 \textwidth} & 
\includegraphics[width=0.056\textwidth,valign=c]{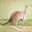}
 & 
\includegraphics[width=0.056\textwidth,valign=c]{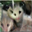}
 & 
\includegraphics[width=0.056\textwidth,valign=c]{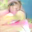}
 & 
\includegraphics[width=0.056\textwidth,valign=c]{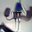}
 & 
\includegraphics[width=0.056\textwidth,valign=c]{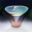}
 & 
\includegraphics[width=0.056\textwidth,valign=c]{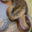}
 & 
\includegraphics[width=0.056\textwidth,valign=c]{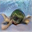}
 & 
\includegraphics[width=0.056\textwidth,valign=c]{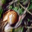}
\vspace{1mm}\\
\includegraphics[width=0.056\textwidth,valign=c]{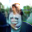}
 & 
\includegraphics[width=0.056\textwidth,valign=c]{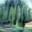}
 & 
\includegraphics[width=0.056\textwidth,valign=c]{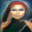}
 & 
\includegraphics[width=0.056\textwidth,valign=c]{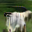}
 & 
\includegraphics[width=0.056\textwidth,valign=c]{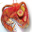}
 & 
\includegraphics[width=0.056\textwidth,valign=c]{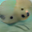}
 & 
\includegraphics[width=0.056\textwidth,valign=c]{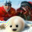}
 & 
\includegraphics[width=0.056\textwidth,valign=c]{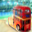}
 & 
\hspace{ 0.028 \textwidth} & 
\includegraphics[width=0.056\textwidth,valign=c]{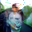}
 & 
\includegraphics[width=0.056\textwidth,valign=c]{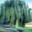}
 & 
\includegraphics[width=0.056\textwidth,valign=c]{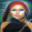}
 & 
\includegraphics[width=0.056\textwidth,valign=c]{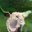}
 & 
\includegraphics[width=0.056\textwidth,valign=c]{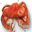}
 & 
\includegraphics[width=0.056\textwidth,valign=c]{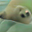}
 & 
\includegraphics[width=0.056\textwidth,valign=c]{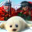}
 & 
\includegraphics[width=0.056\textwidth,valign=c]{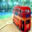}
\vspace{1mm}\\
\includegraphics[width=0.056\textwidth,valign=c]{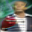}
 & 
\includegraphics[width=0.056\textwidth,valign=c]{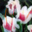}
 & 
\includegraphics[width=0.056\textwidth,valign=c]{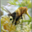}
 & 
\includegraphics[width=0.056\textwidth,valign=c]{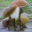}
 & 
\includegraphics[width=0.056\textwidth,valign=c]{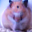}
 & 
\includegraphics[width=0.056\textwidth,valign=c]{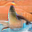}
 & 
\includegraphics[width=0.056\textwidth,valign=c]{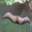}
 & 
\includegraphics[width=0.056\textwidth,valign=c]{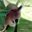}
 & 
\hspace{ 0.028 \textwidth} & 
\includegraphics[width=0.056\textwidth,valign=c]{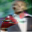}
 & 
\includegraphics[width=0.056\textwidth,valign=c]{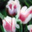}
 & 
\includegraphics[width=0.056\textwidth,valign=c]{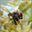}
 & 
\includegraphics[width=0.056\textwidth,valign=c]{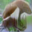}
 & 
\includegraphics[width=0.056\textwidth,valign=c]{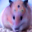}
 & 
\includegraphics[width=0.056\textwidth,valign=c]{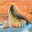}
 & 
\includegraphics[width=0.056\textwidth,valign=c]{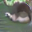}
 & 
\includegraphics[width=0.056\textwidth,valign=c]{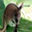}
\vspace{2mm}\\
\end{tabu}
\end{adjustbox}
\caption{Random selection of  48 images from the CIFAR100 test set which are misclassified by all our models and the associated visual counterfactuals that are generated by maximizing the confidence in the gt-class in a $l_2$ ball of radius 3. The numbers over the individual images indicate the target class and were omitted for the latter models. }\label{fig:cifar100_overview}
\end{figure}

%% file: res/appendix_od_cifar100_new.tex
\begin{figure}[ht!]
\begin{tabular}{p{1cm}x{\breite}x{\breite}x{\breite}x{\breite}x{\breite}x{\breite}x{\breite}x{\breite}}
Model  & Orig. & $\epsilon=0.5$ & $\epsilon=1.0$ & $\epsilon=1.5$ & $\epsilon=2.0$ & $\epsilon=2.5$ & $\epsilon=3.0$\\ 
\begin{turn}{90} \hspace{-.4cm} AT-0.50 \end{turn}  &  \multicolumn{7}{c}{\includegraphics[width=0.91\textwidth,valign=c]{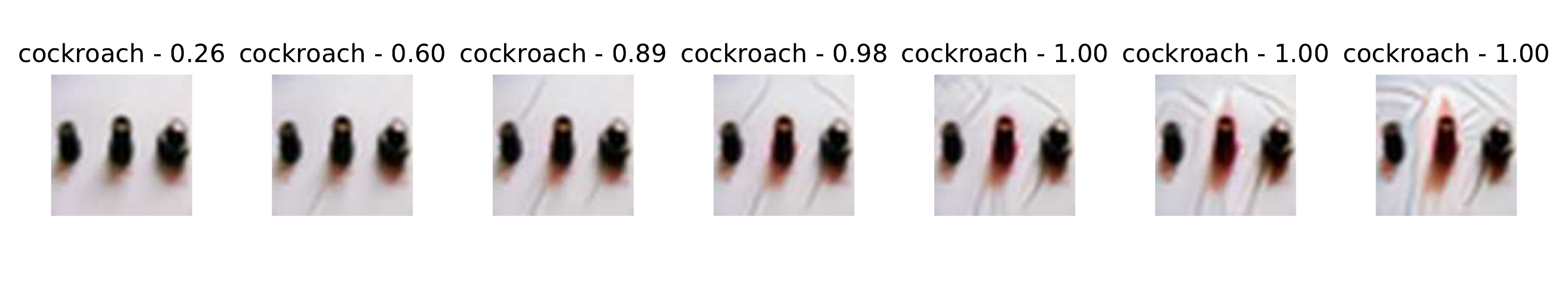}} \\
\hline
\begin{turn}{90} \hspace{-.4cm} AT-0.25 \end{turn} & \multicolumn{7}{c}{\includegraphics[width=0.91\textwidth,valign=c]{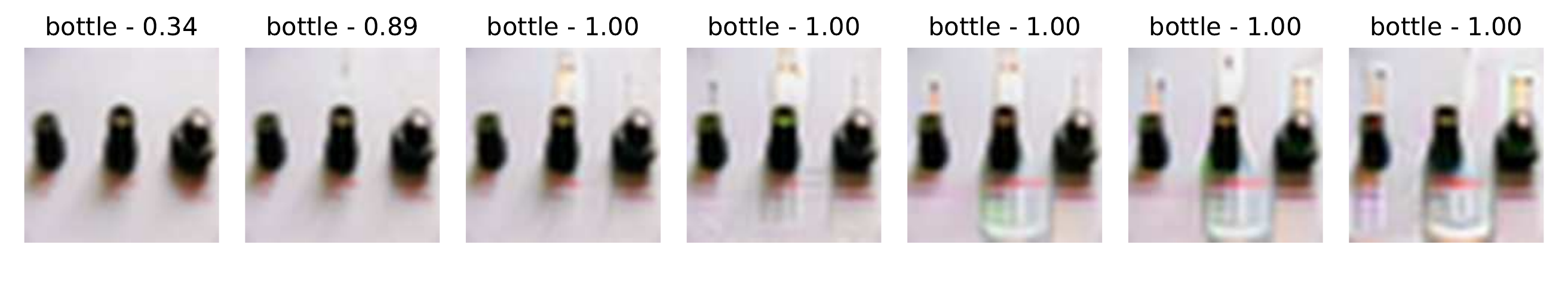}} \\
\hline
\begin{turn}{90} \hspace{-.4cm} R-0.50 \end{turn} & \multicolumn{7}{c}{\includegraphics[width=0.91\textwidth,valign=c]{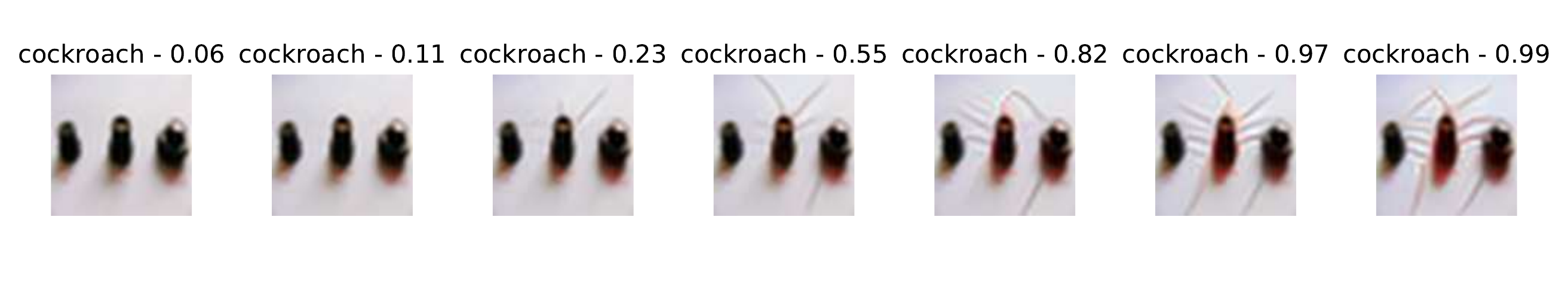}} \\
\hline
\begin{turn}{90} \hspace{-.4cm} R-0.25 \end{turn}  &  \multicolumn{7}{c}{\includegraphics[width=0.91\textwidth,valign=c]{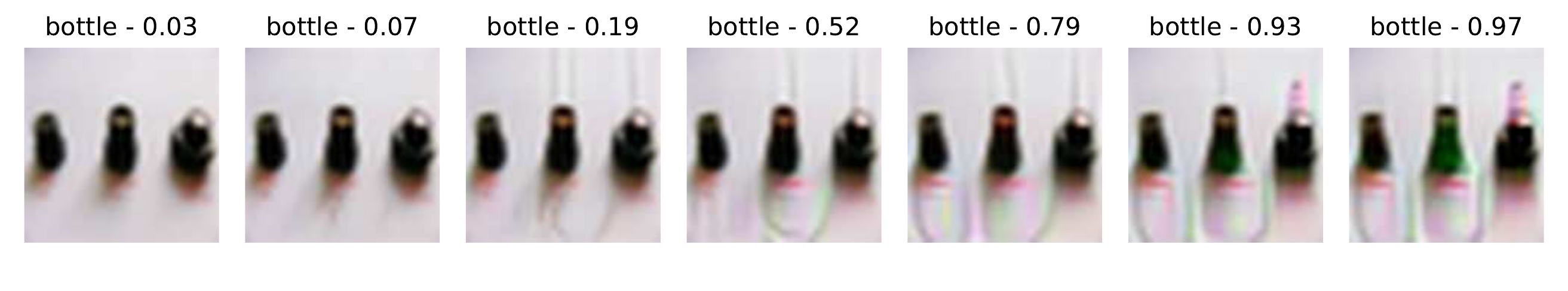}} \\
\vspace{2mm}\\
\begin{turn}{90} \hspace{-.4cm} AT-0.50 \end{turn}  &  \multicolumn{7}{c}{\includegraphics[width=0.91\textwidth,valign=c]{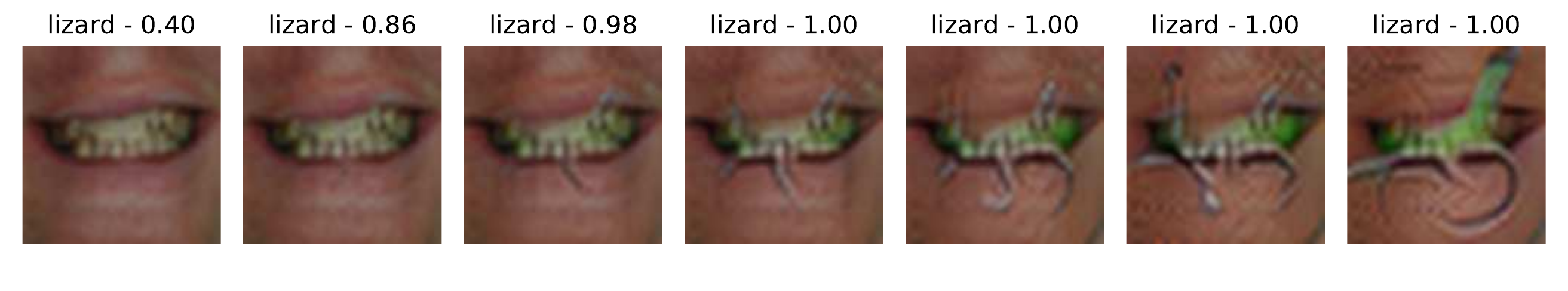}} \\
\hline
\begin{turn}{90} \hspace{-.4cm} AT-0.25 \end{turn} & \multicolumn{7}{c}{\includegraphics[width=0.91\textwidth,valign=c]{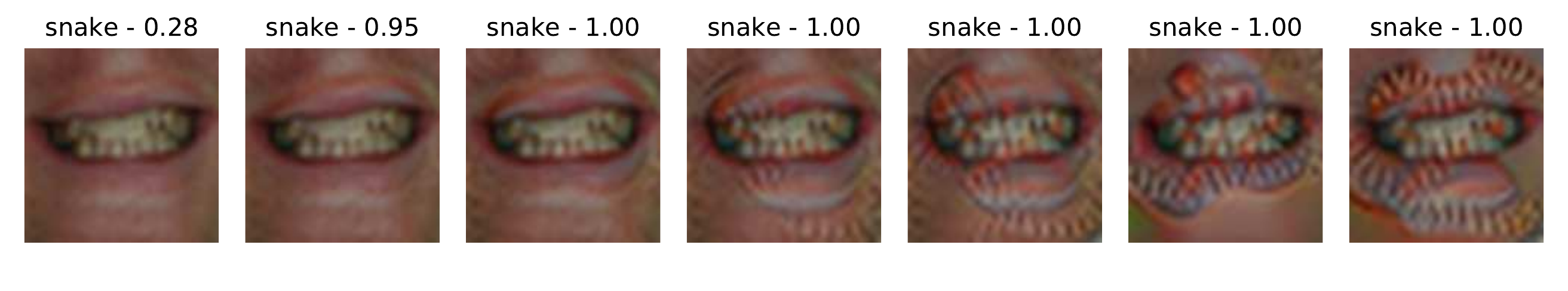}} \\
\hline
\begin{turn}{90} \hspace{-.4cm} R-0.50 \end{turn} & \multicolumn{7}{c}{\includegraphics[width=0.91\textwidth,valign=c]{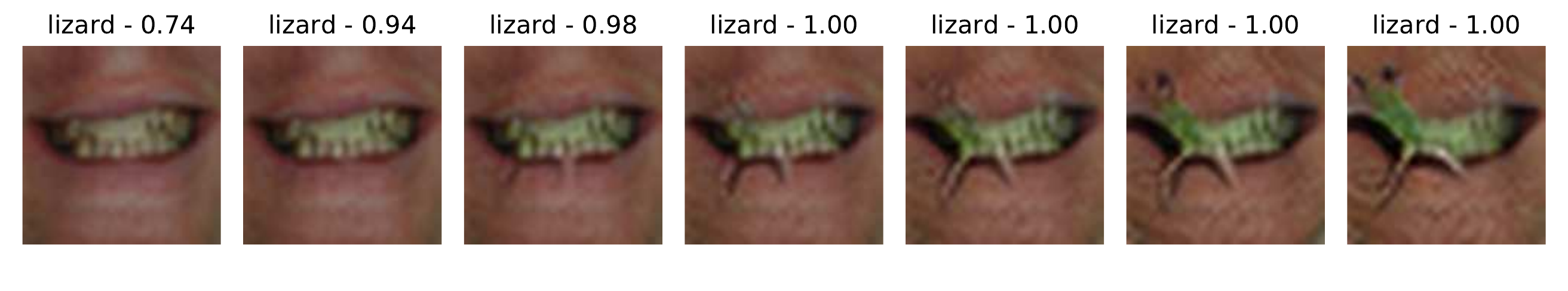}} \\
\hline
\begin{turn}{90} \hspace{-.4cm} R-0.25 \end{turn}  &  \multicolumn{7}{c}{\includegraphics[width=0.91\textwidth,valign=c]{pics/CIFAR100/Ratio025/OD/img_238.pdf}} \\
\end{tabular}	
\vspace{-.5cm}
\caption{\label{fig:od_cifar100_new1}\textbf{Feature Generation on OOD images (CIFAR100)} for  models for unseen images from 80 million tiny images. The two RATIO models are able to create less distorted images than the simple adversarially trained models.}
\end{figure}

\begin{figure}[ht!]
\begin{tabular}{p{1cm}x{\breite}x{\breite}x{\breite}x{\breite}x{\breite}x{\breite}x{\breite}x{\breite}}
Model  & Orig. & $\epsilon=0.5$ & $\epsilon=1.0$ & $\epsilon=1.5$ & $\epsilon=2.0$ & $\epsilon=2.5$ & $\epsilon=3.0$\\ 
\begin{turn}{90} \hspace{-.4cm} AT-0.50 \end{turn}  &  \multicolumn{7}{c}{\includegraphics[width=0.91\textwidth,valign=c]{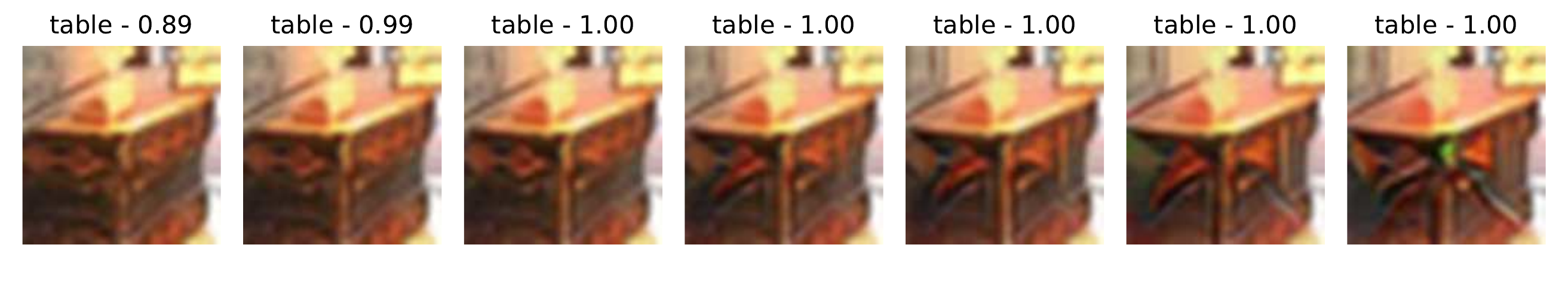}} \\
\hline
\begin{turn}{90} \hspace{-.4cm} AT-0.25 \end{turn} & \multicolumn{7}{c}{\includegraphics[width=0.91\textwidth,valign=c]{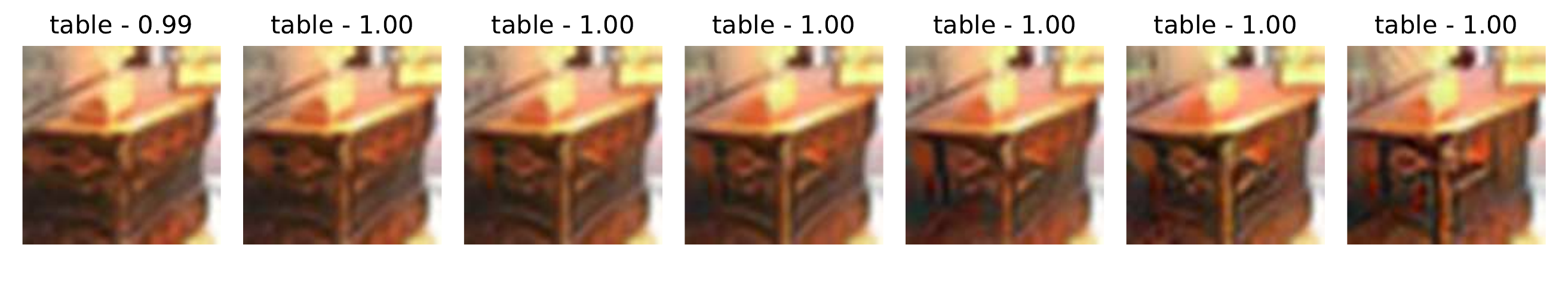}} \\
\hline
\begin{turn}{90} \hspace{-.4cm} R-0.50 \end{turn} & \multicolumn{7}{c}{\includegraphics[width=0.91\textwidth,valign=c]{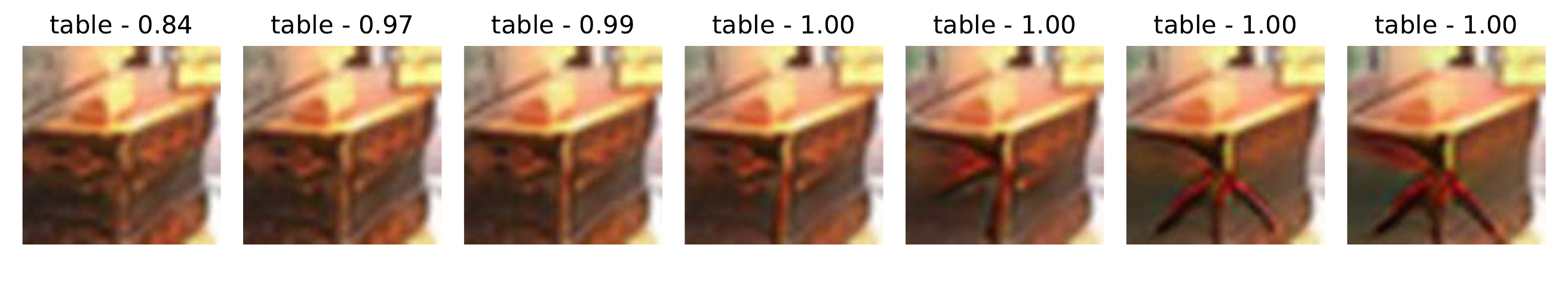}} \\
\hline
\begin{turn}{90} \hspace{-.4cm} R-0.25 \end{turn}  &  \multicolumn{7}{c}{\includegraphics[width=0.91\textwidth,valign=c]{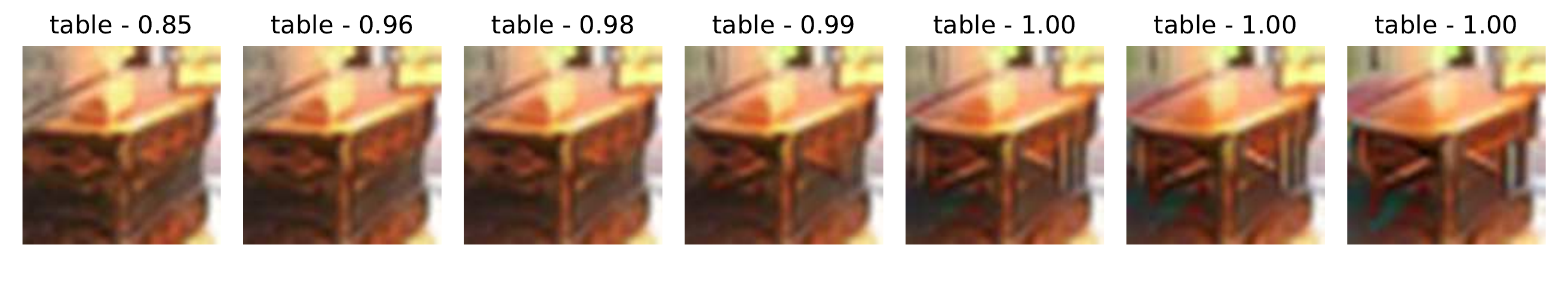}} \\
\vspace{2mm}\\
\begin{turn}{90} \hspace{-.4cm} AT-0.50 \end{turn}  &  \multicolumn{7}{c}{\includegraphics[width=0.91\textwidth,valign=c]{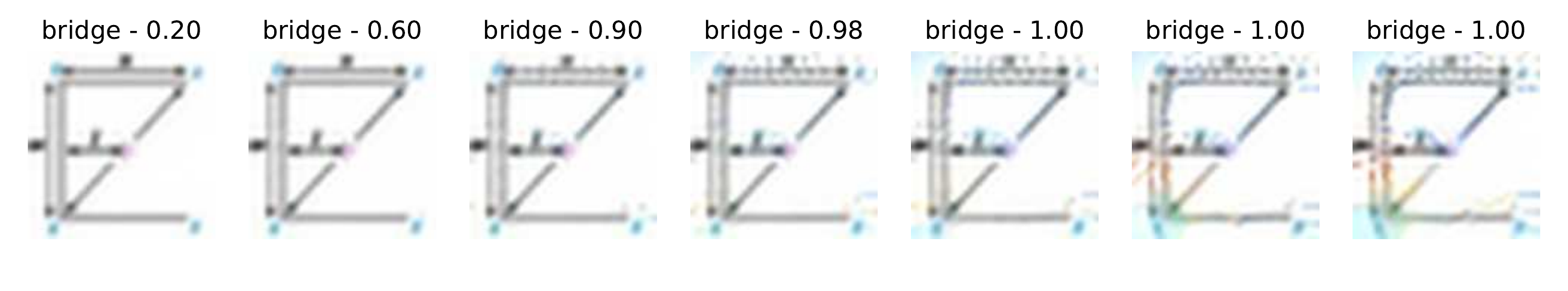}} \\
\hline
\begin{turn}{90} \hspace{-.4cm} AT-0.25 \end{turn} & \multicolumn{7}{c}{\includegraphics[width=0.91\textwidth,valign=c]{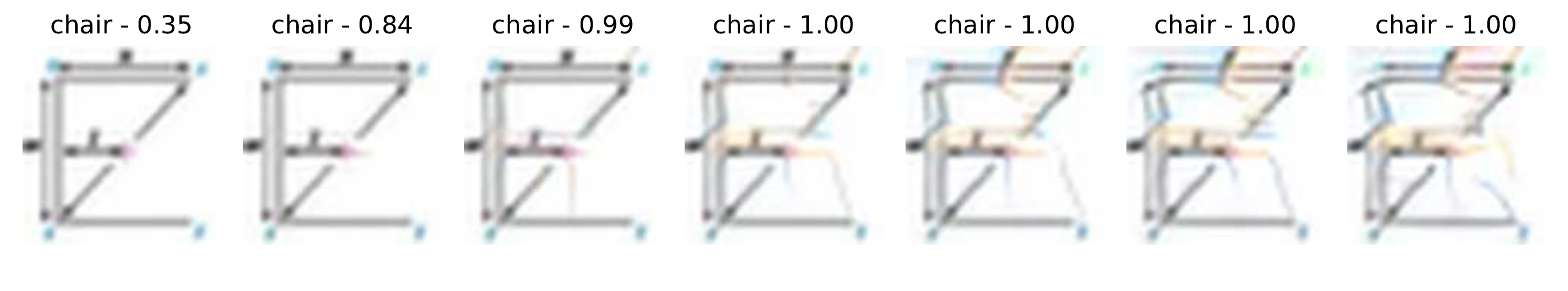}} \\
\hline
\begin{turn}{90} \hspace{-.4cm} R-0.50 \end{turn} & \multicolumn{7}{c}{\includegraphics[width=0.91\textwidth,valign=c]{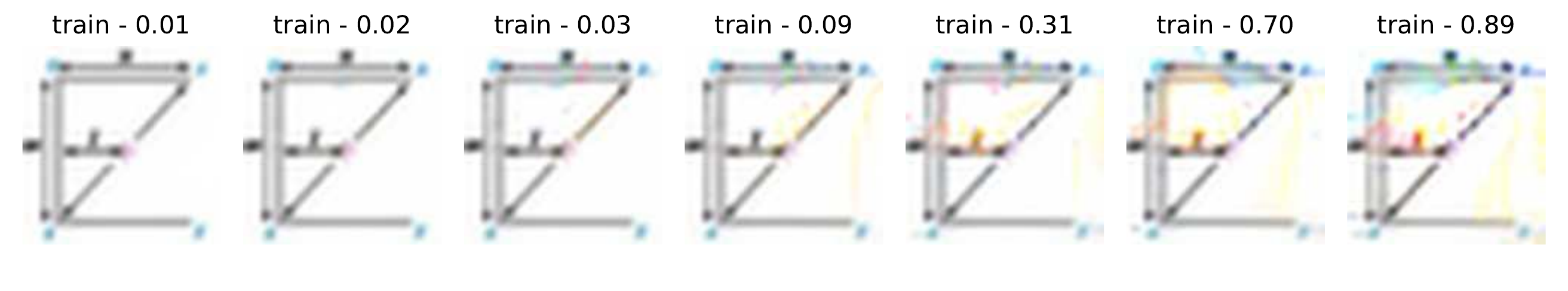}} \\
\hline
\begin{turn}{90} \hspace{-.4cm} R-0.25 \end{turn}  &  \multicolumn{7}{c}{\includegraphics[width=0.91\textwidth,valign=c]{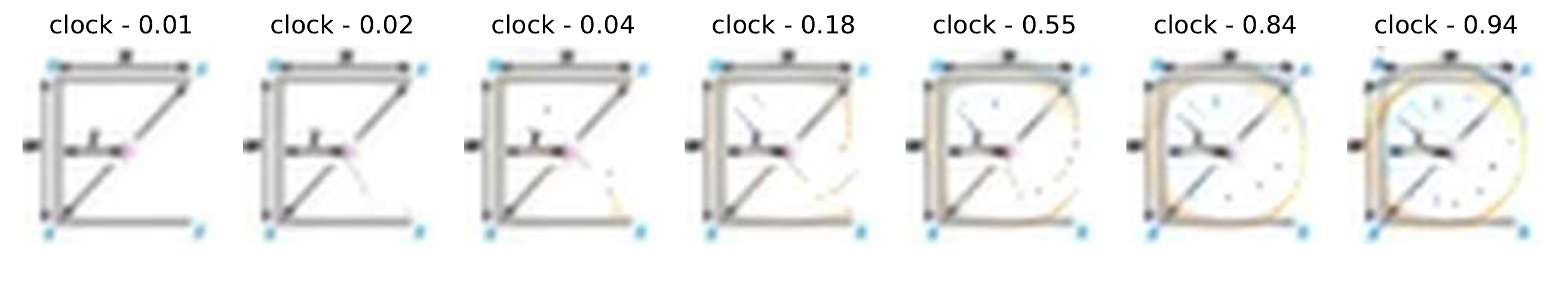}} \\
\end{tabular}	
\vspace{-.5cm}
\caption{\label{fig:od_cifar100_new2}\textbf{Feature Generation on OOD images (CIFAR100)} for  models for unseen images from 80 million tiny images. The two RATIO models are able to create less distorted images than the simple adversarially trained models.}
\end{figure}

%% file: res/appendix_vc_restricted_new.tex
\begin{figure}[ht!]
\begin{tabular}{p{1cm}x{\breite}x{\breite}x{\breite}x{\breite}x{\breite}x{\breite}x{\breite}x{\breite}}
Model  & Orig. & $\epsilon=3.5$ & $\epsilon=7.0$ & $\epsilon=10.5$ & $\epsilon=14.0$ & $\epsilon=17.5$ & $\epsilon=21.0$\\
\begin{turn}{90} \hspace{-1.2cm} Madry AT-3.50 \end{turn}  &  \multicolumn{7}{c}{\includegraphics[width=0.91\textwidth,valign=c]{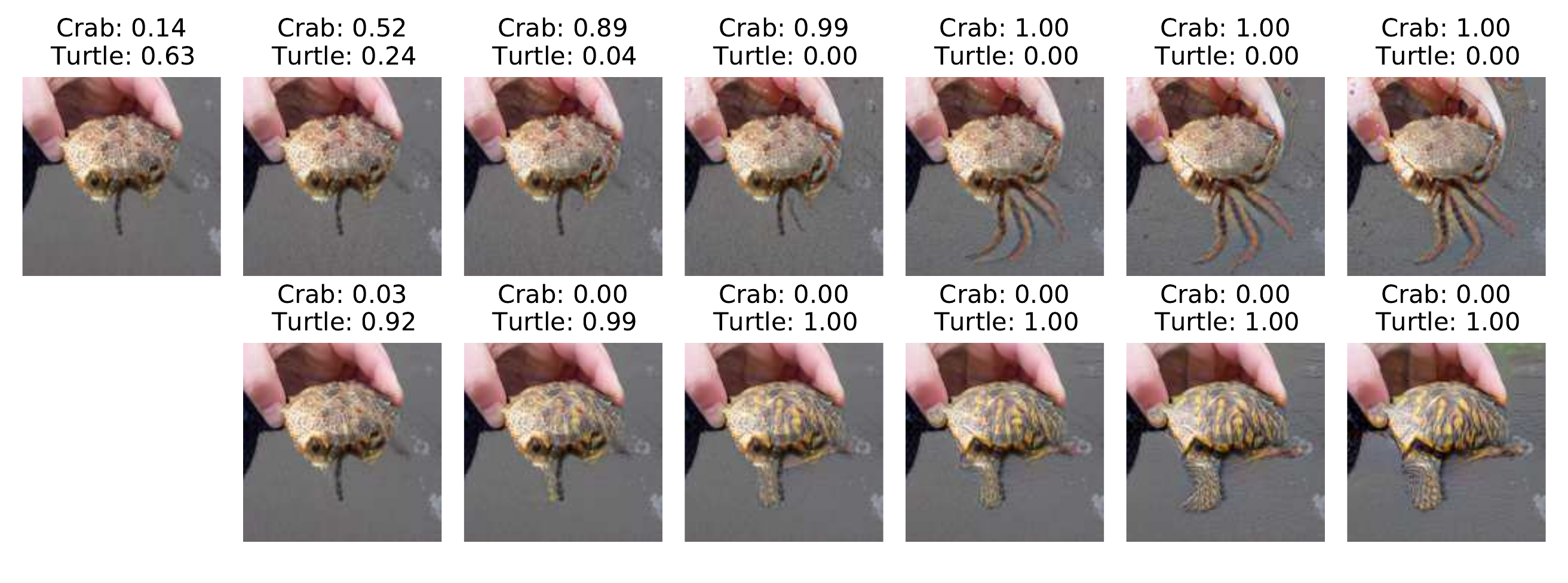}} \\
\hline
\begin{turn}{90} \hspace{-.4cm} AT-3.50 \end{turn}  &  \multicolumn{7}{c}{\includegraphics[width=0.91\textwidth,valign=c]{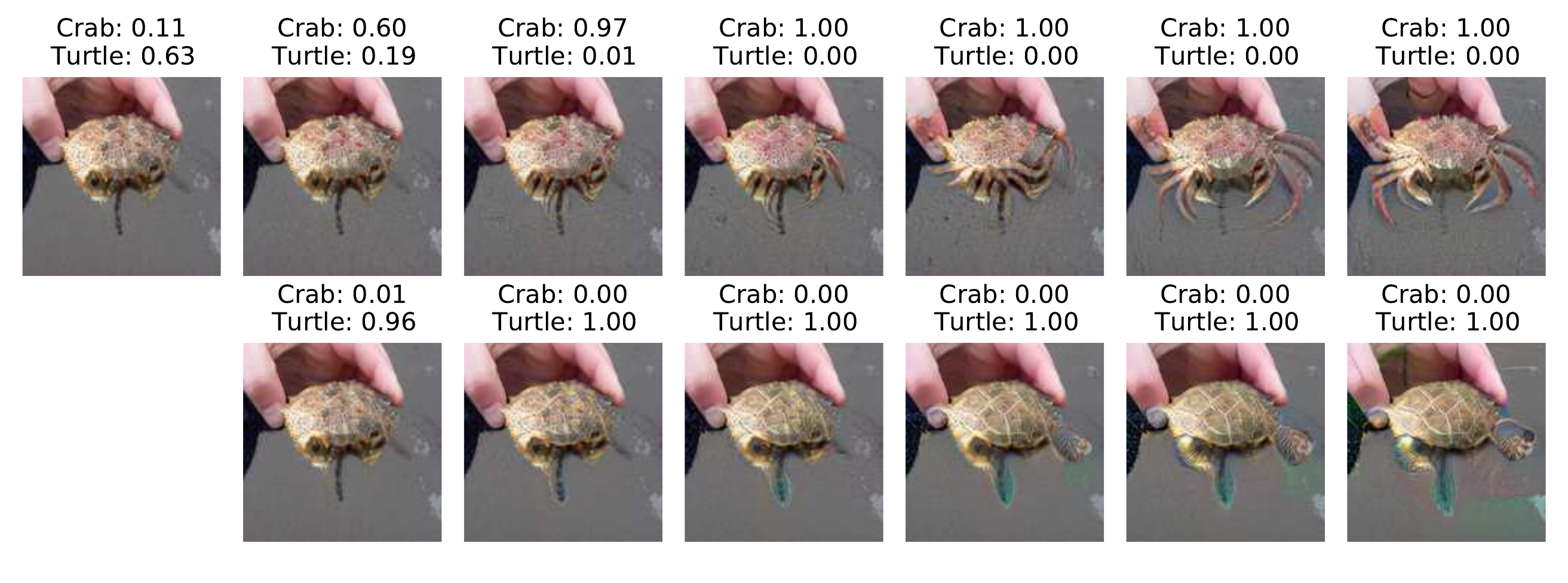}} \\
\hline
\begin{turn}{90} \hspace{-.9cm} RATIO-3.50 \end{turn} & \multicolumn{7}{c}{\includegraphics[width=0.91\textwidth,valign=c]{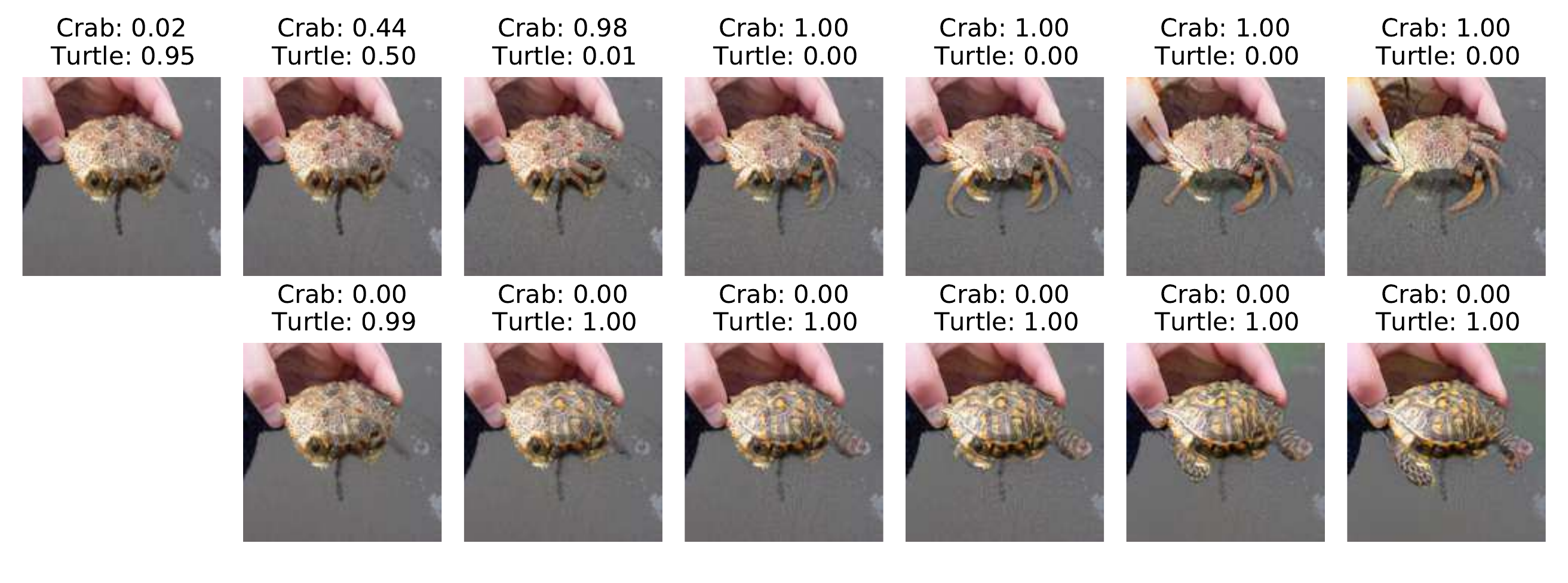}} \\
\hline
\begin{turn}{90} \hspace{-.9cm} RATIO-1.75 \end{turn} & \multicolumn{7}{c}{\includegraphics[width=0.91\textwidth,valign=c]{pics/imagenet/Ratio175Clean/VC/224.pdf}} \\
\end{tabular}	
\caption{\label{fig:vc_imagenet_new1}\textbf{Visual Counterfactuals} on restricted ImageNet test samples misclassified by all methods. The visual counterfactuals
of all AT and RATIO models are comparable in terms of quality.
}
\end{figure}

\begin{figure}[ht!]
\begin{tabular}{p{1cm}x{\breite}x{\breite}x{\breite}x{\breite}x{\breite}x{\breite}x{\breite}x{\breite}}
Model  & Orig. & $\epsilon=3.5$ & $\epsilon=7.0$ & $\epsilon=10.5$ & $\epsilon=14.0$ & $\epsilon=17.5$ & $\epsilon=21.0$\\
\begin{turn}{90} \hspace{-1.2cm} Madry AT-3.50 \end{turn}  &  \multicolumn{7}{c}{\includegraphics[width=0.91\textwidth,valign=c]{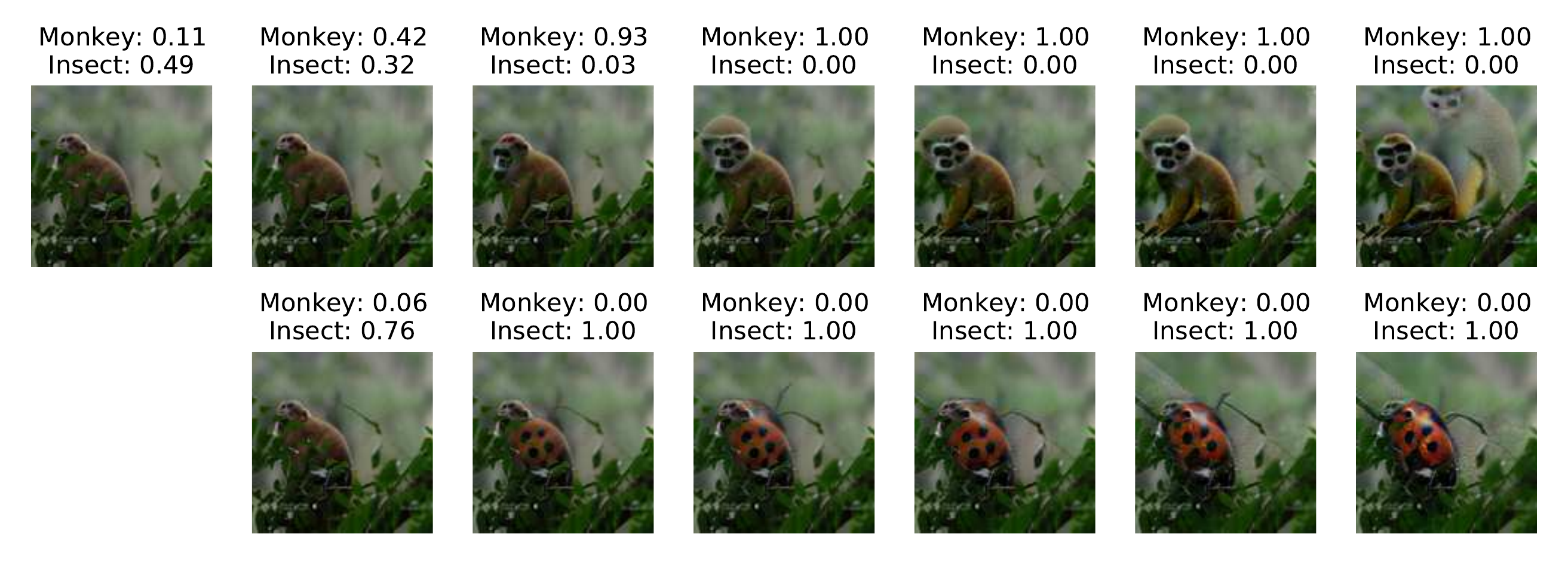}} \\
\hline
\begin{turn}{90} \hspace{-.4cm} AT-3.50 \end{turn}  &  \multicolumn{7}{c}{\includegraphics[width=0.91\textwidth,valign=c]{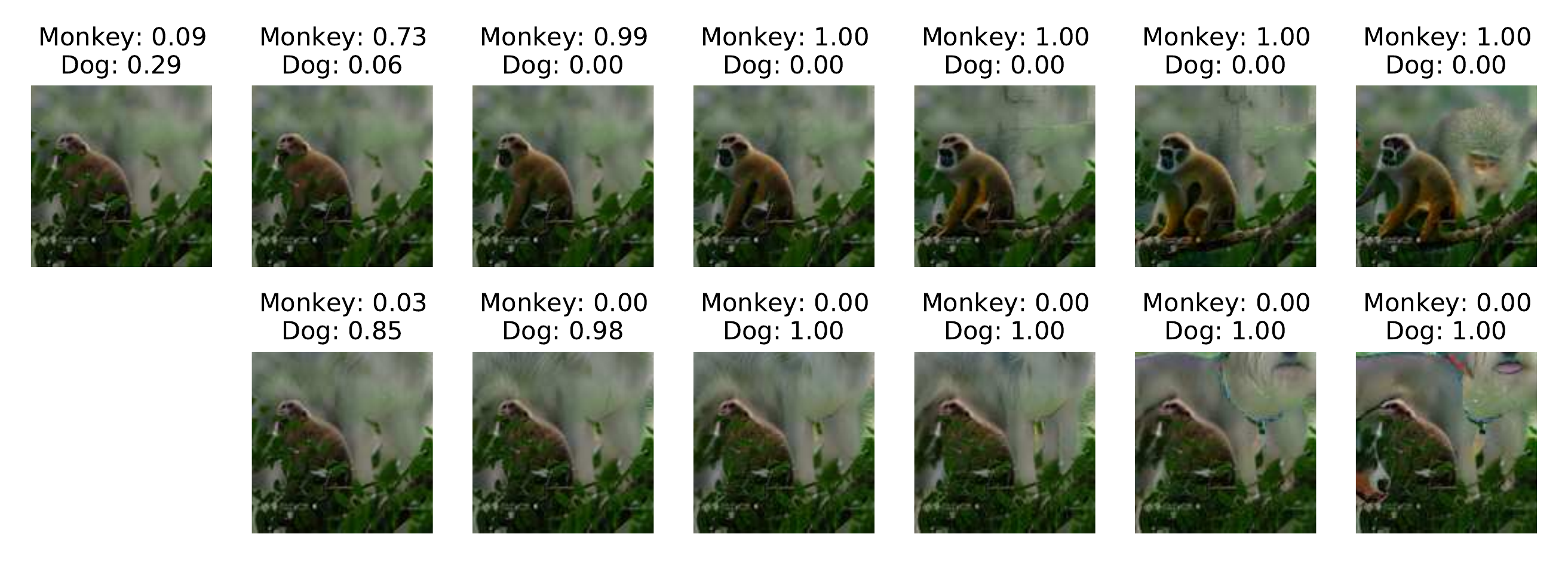}} \\
\hline
\begin{turn}{90} \hspace{-.9cm} RATIO-3.50 \end{turn} & \multicolumn{7}{c}{\includegraphics[width=0.91\textwidth,valign=c]{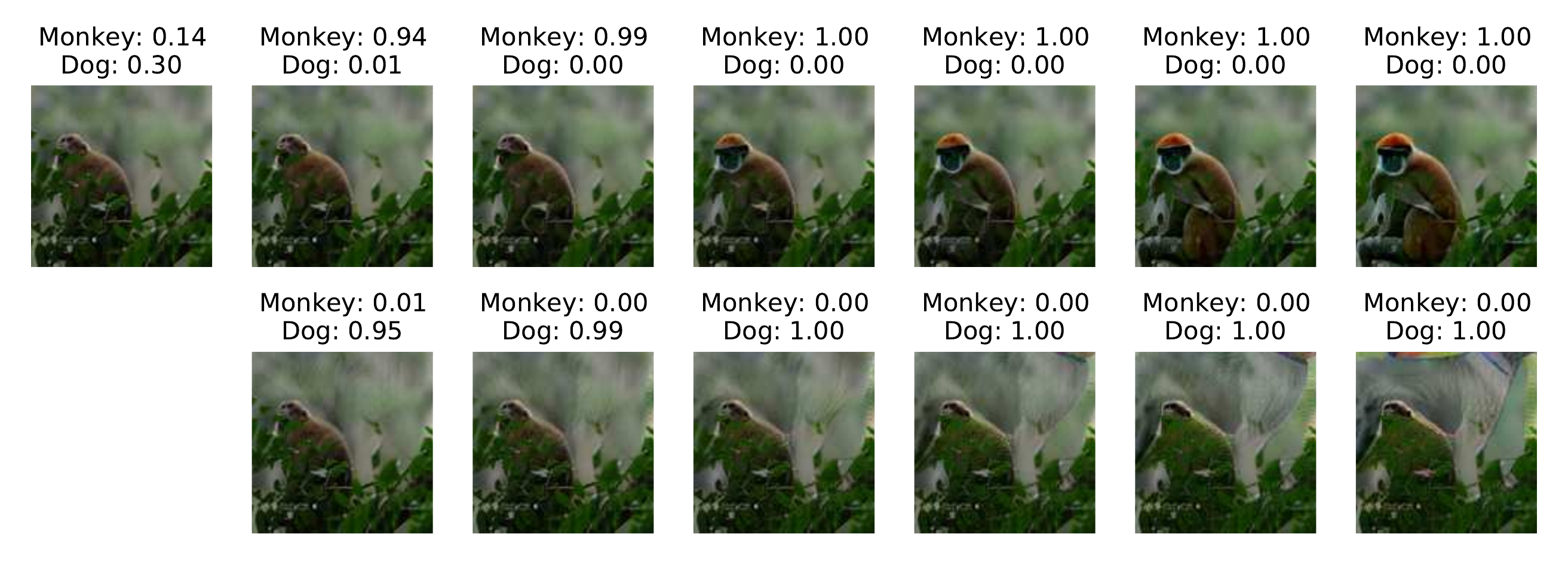}} \\
\hline
\begin{turn}{90} \hspace{-.9cm} RATIO-1.75 \end{turn} & \multicolumn{7}{c}{\includegraphics[width=0.91\textwidth,valign=c]{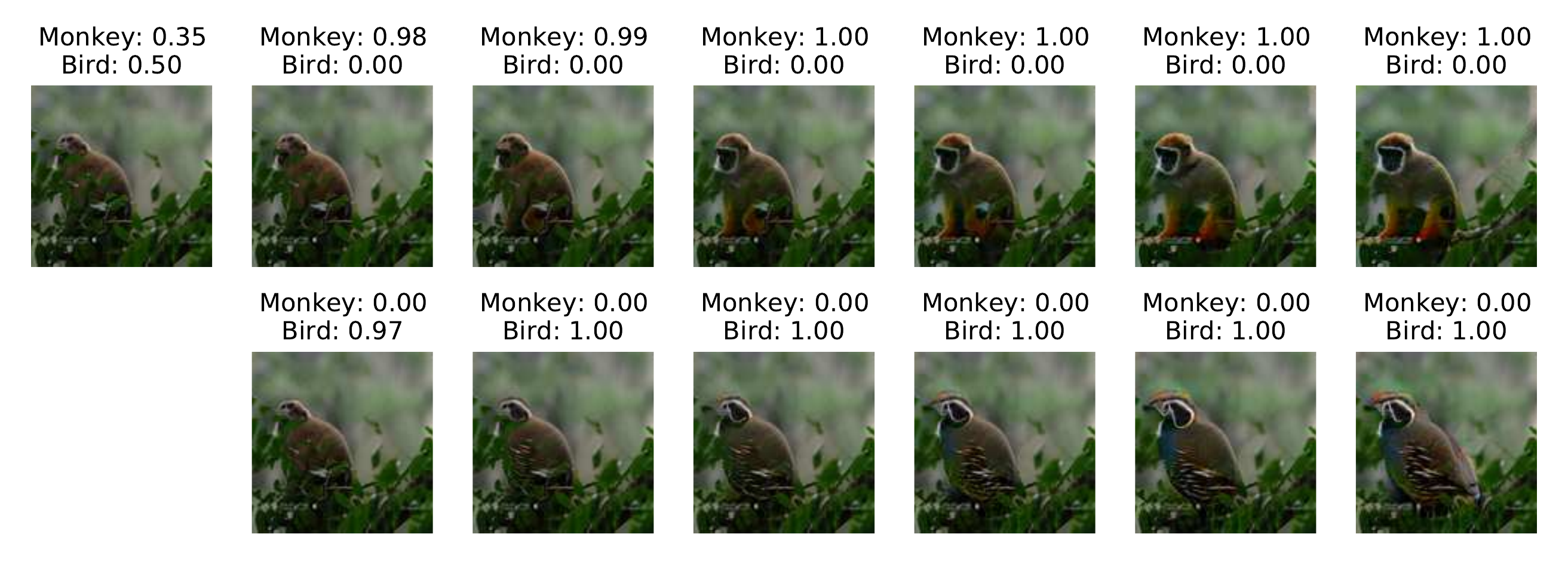}} \\
\end{tabular}	
\caption{\label{fig:vc_imagenet_new2}\textbf{Visual Counterfactuals} on restricted ImageNet test samples misclassified by all methods. The quality of adversarial
training and RATIO is quite similar. The change to dog is quite subtle but reasonable both for RATIO$_{3.5}$ and AT$_{3.5}$.
}
\end{figure}

\begin{figure}[ht!]
\begin{tabular}{p{1cm}x{\breite}x{\breite}x{\breite}x{\breite}x{\breite}x{\breite}x{\breite}x{\breite}}
Model  & Orig. & $\epsilon=3.5$ & $\epsilon=7.0$ & $\epsilon=10.5$ & $\epsilon=14.0$ & $\epsilon=17.5$ & $\epsilon=21.0$\\
\begin{turn}{90} \hspace{-1.2cm} Madry AT-3.50 \end{turn}  &  \multicolumn{7}{c}{\includegraphics[width=0.91\textwidth,valign=c]{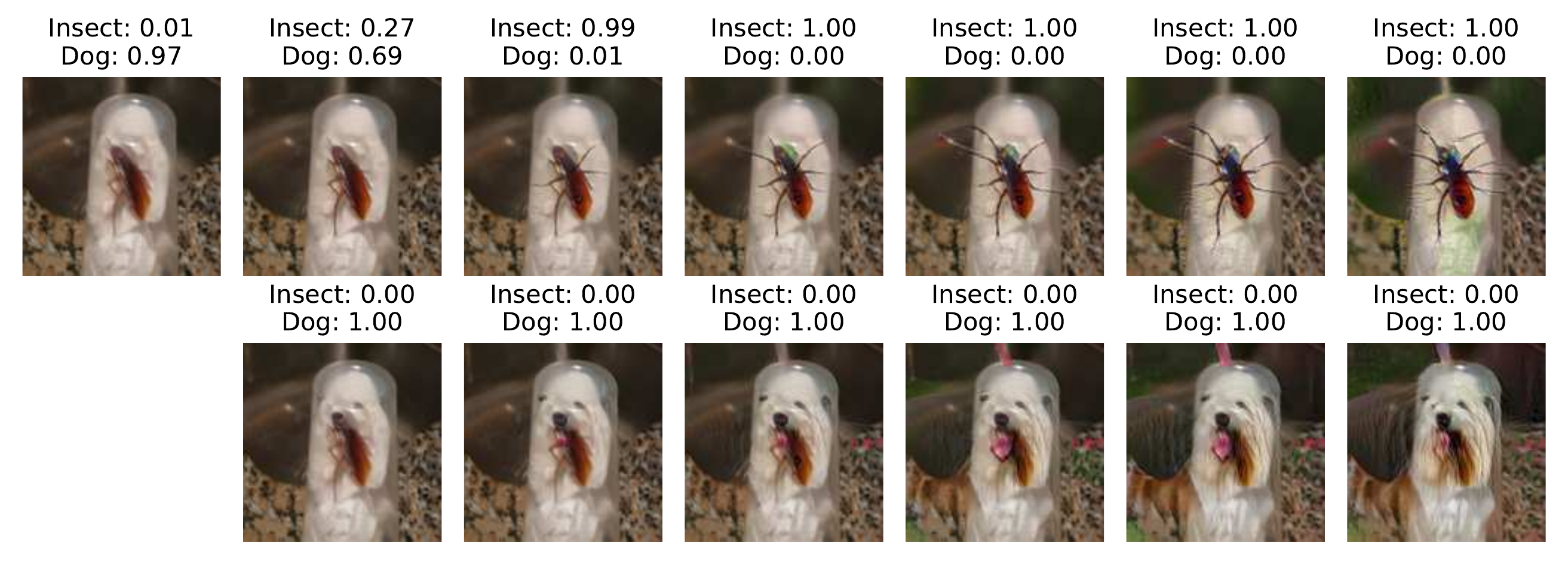}} \\
\hline
\begin{turn}{90} \hspace{-.4cm} AT-3.50 \end{turn}  &  \multicolumn{7}{c}{\includegraphics[width=0.91\textwidth,valign=c]{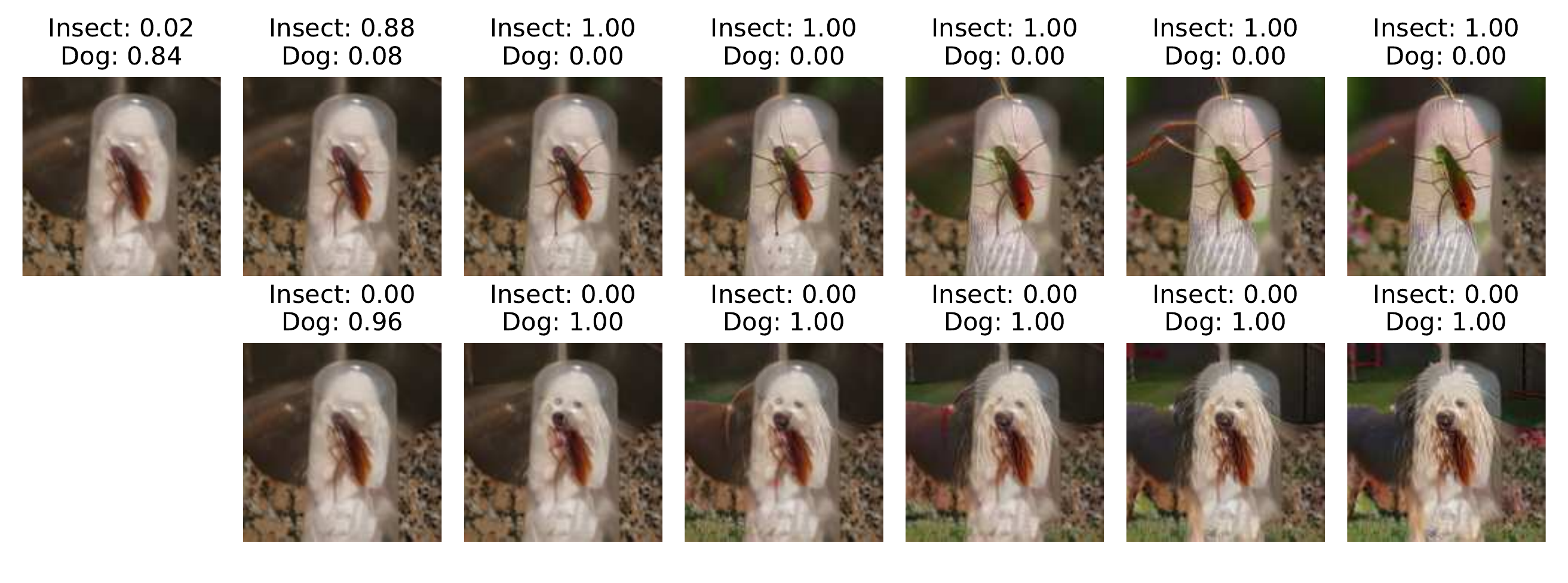}} \\
\hline
\begin{turn}{90} \hspace{-.9cm} RATIO-3.50 \end{turn} & \multicolumn{7}{c}{\includegraphics[width=0.91\textwidth,valign=c]{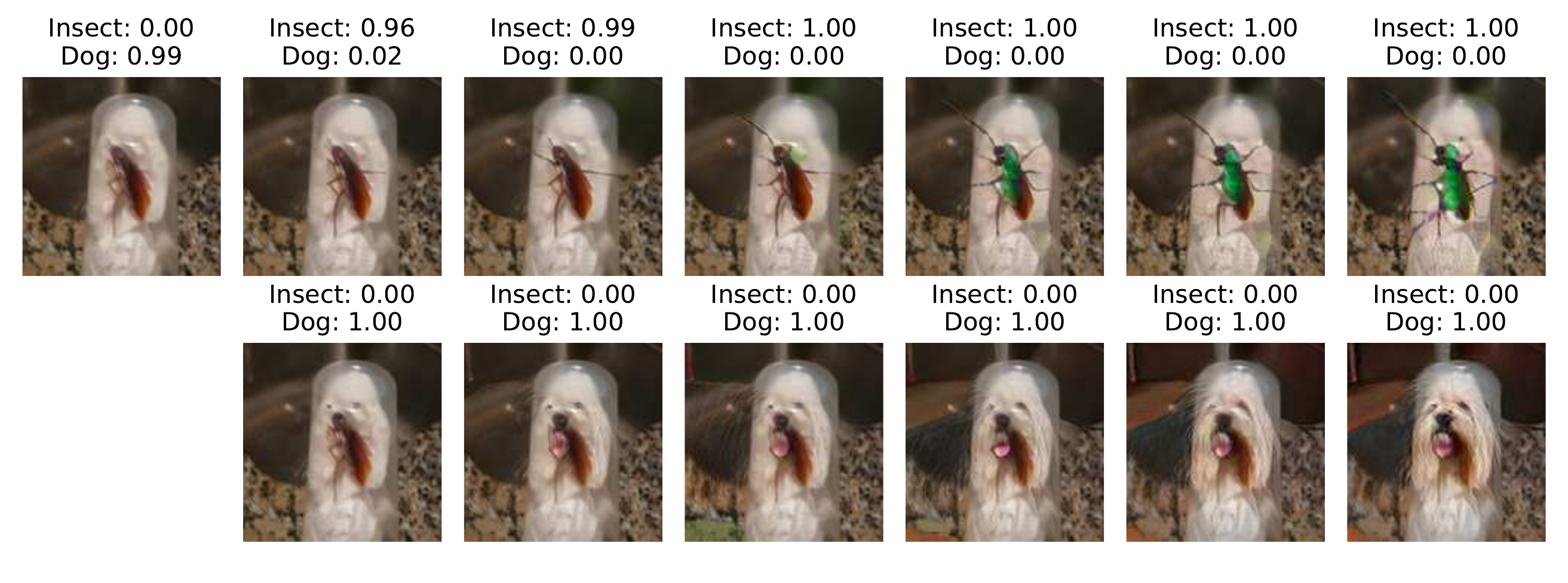}} \\
\hline
\begin{turn}{90} \hspace{-.9cm} RATIO-1.75 \end{turn} & \multicolumn{7}{c}{\includegraphics[width=0.91\textwidth,valign=c]{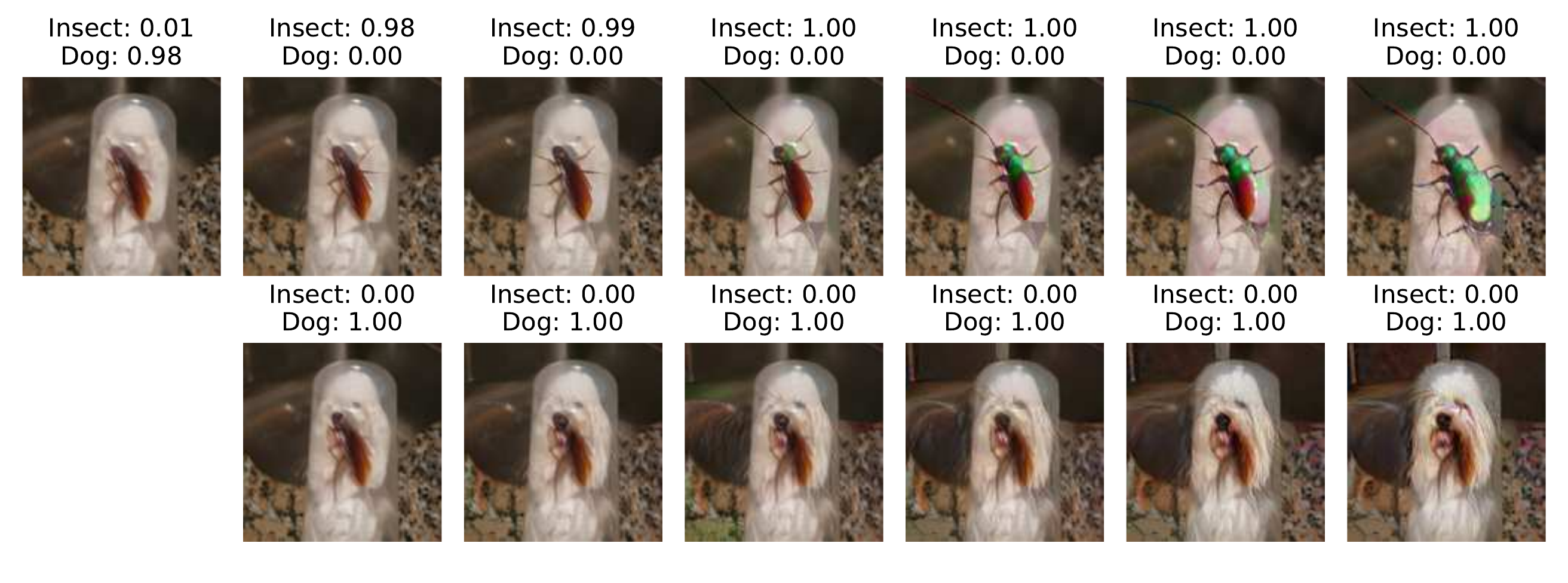}} \\
\end{tabular}	
\caption{\label{fig:vc_imagenet_new3}\textbf{Visual Counterfactuals} on restricted ImageNet test samples misclassified by all methods. The quality of AT and RATIO is quite similar.
}
\end{figure}

%% file: res/appendix_restricted_overview.tex
\begin{figure}
\begin{adjustbox}{max width=\textwidth}
\begin{tabu}{ccccccccccc}
\multicolumn{5}{c}{Original} & & \multicolumn{5}{c}{Madry}\\
\rowfont{\tiny}
Monkey & Fish & Crab & Turtle & Bird & & Monkey & Fish & Crab & Turtle & Bird\\
\includegraphics[width=0.086\textwidth,valign=c]{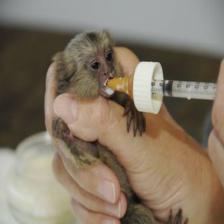}
 & 
\includegraphics[width=0.086\textwidth,valign=c]{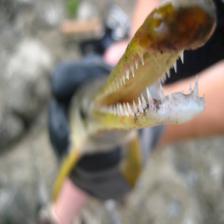}
 & 
\includegraphics[width=0.086\textwidth,valign=c]{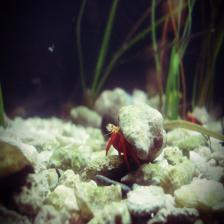}
 & 
\includegraphics[width=0.086\textwidth,valign=c]{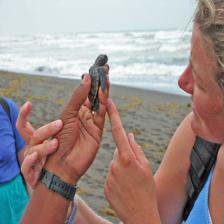}
 & 
\includegraphics[width=0.086\textwidth,valign=c]{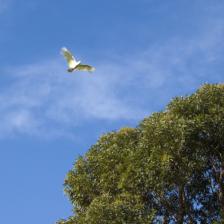}
 & 
\hspace{ 0.043 \textwidth} & 
\includegraphics[width=0.086\textwidth,valign=c]{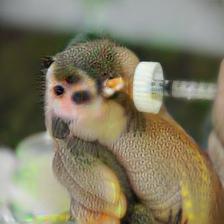}
 & 
\includegraphics[width=0.086\textwidth,valign=c]{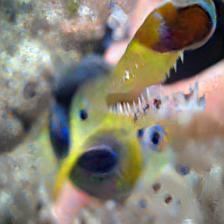}
 & 
\includegraphics[width=0.086\textwidth,valign=c]{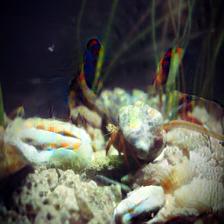}
 & 
\includegraphics[width=0.086\textwidth,valign=c]{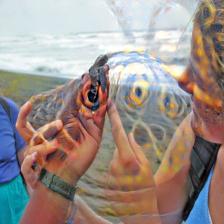}
 & 
\includegraphics[width=0.086\textwidth,valign=c]{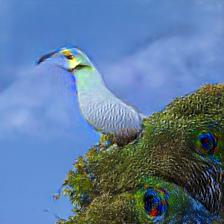}
\\
\rowfont{\tiny}
Fish & Monkey & Bird & Bird & Turtle & & Fish & Monkey & Bird & Bird & Turtle\\
\includegraphics[width=0.086\textwidth,valign=c]{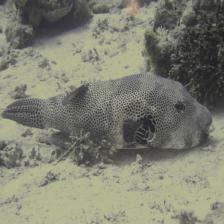}
 & 
\includegraphics[width=0.086\textwidth,valign=c]{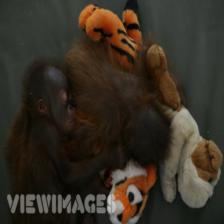}
 & 
\includegraphics[width=0.086\textwidth,valign=c]{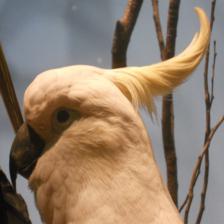}
 & 
\includegraphics[width=0.086\textwidth,valign=c]{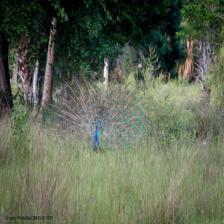}
 & 
\includegraphics[width=0.086\textwidth,valign=c]{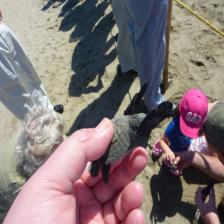}
 & 
\hspace{ 0.043 \textwidth} & 
\includegraphics[width=0.086\textwidth,valign=c]{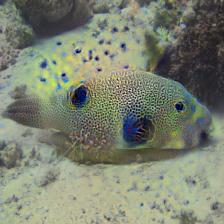}
 & 
\includegraphics[width=0.086\textwidth,valign=c]{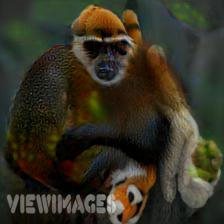}
 & 
\includegraphics[width=0.086\textwidth,valign=c]{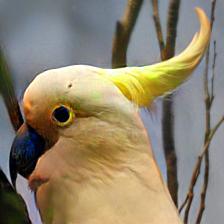}
 & 
\includegraphics[width=0.086\textwidth,valign=c]{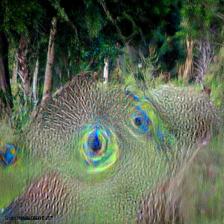}
 & 
\includegraphics[width=0.086\textwidth,valign=c]{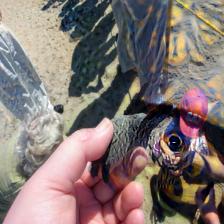}
\\
\rowfont{\tiny}
Insect & Frog & Fish & Fish & Cat & & Insect & Frog & Fish & Fish & Cat\\
\includegraphics[width=0.086\textwidth,valign=c]{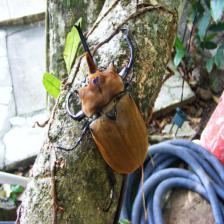}
 & 
\includegraphics[width=0.086\textwidth,valign=c]{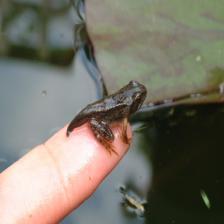}
 & 
\includegraphics[width=0.086\textwidth,valign=c]{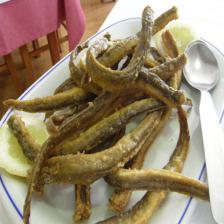}
 & 
\includegraphics[width=0.086\textwidth,valign=c]{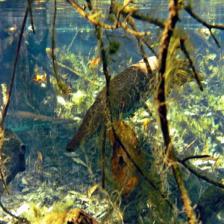}
 & 
\includegraphics[width=0.086\textwidth,valign=c]{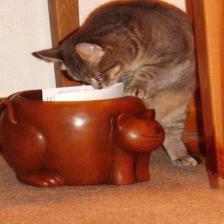}
 & 
\hspace{ 0.043 \textwidth} & 
\includegraphics[width=0.086\textwidth,valign=c]{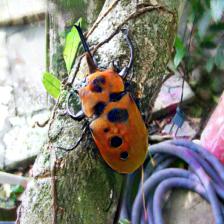}
 & 
\includegraphics[width=0.086\textwidth,valign=c]{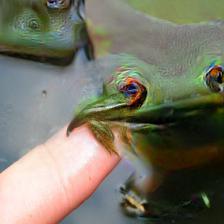}
 & 
\includegraphics[width=0.086\textwidth,valign=c]{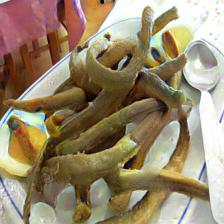}
 & 
\includegraphics[width=0.086\textwidth,valign=c]{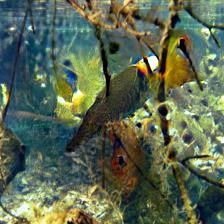}
 & 
\includegraphics[width=0.086\textwidth,valign=c]{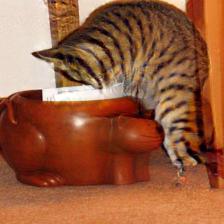}
\\
\rowfont{\tiny}
Crab & Fish & Cat & Monkey & Bird & & Crab & Fish & Cat & Monkey & Bird\\
\includegraphics[width=0.086\textwidth,valign=c]{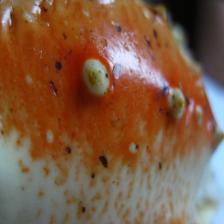}
 & 
\includegraphics[width=0.086\textwidth,valign=c]{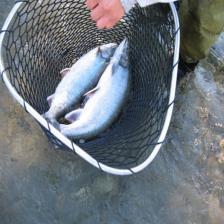}
 & 
\includegraphics[width=0.086\textwidth,valign=c]{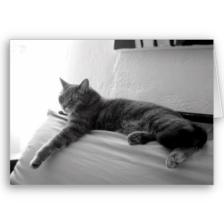}
 & 
\includegraphics[width=0.086\textwidth,valign=c]{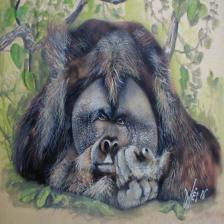}
 & 
\includegraphics[width=0.086\textwidth,valign=c]{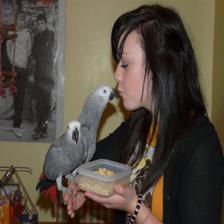}
 & 
\hspace{ 0.043 \textwidth} & 
\includegraphics[width=0.086\textwidth,valign=c]{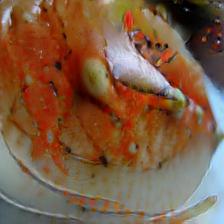}
 & 
\includegraphics[width=0.086\textwidth,valign=c]{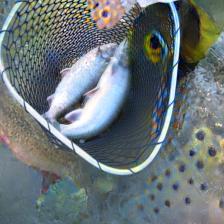}
 & 
\includegraphics[width=0.086\textwidth,valign=c]{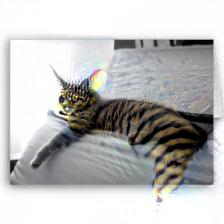}
 & 
\includegraphics[width=0.086\textwidth,valign=c]{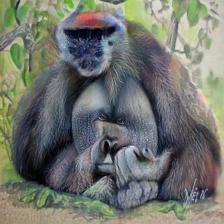}
 & 
\includegraphics[width=0.086\textwidth,valign=c]{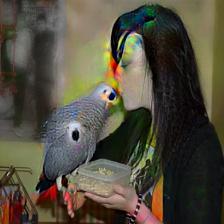}
\vspace{2mm}\\
\multicolumn{5}{c}{AT-3.5} & & \multicolumn{5}{c}{AT-1.75}\\
\includegraphics[width=0.086\textwidth,valign=c]{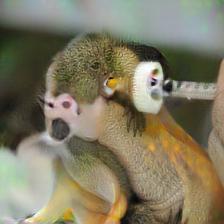}
 & 
\includegraphics[width=0.086\textwidth,valign=c]{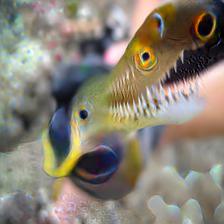}
 & 
\includegraphics[width=0.086\textwidth,valign=c]{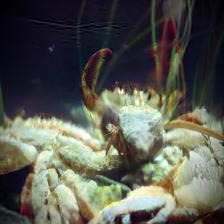}
 & 
\includegraphics[width=0.086\textwidth,valign=c]{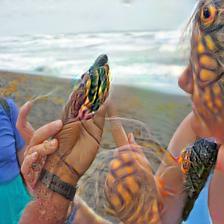}
 & 
\includegraphics[width=0.086\textwidth,valign=c]{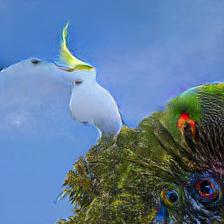}
 & 
\hspace{ 0.043 \textwidth} & 
\includegraphics[width=0.086\textwidth,valign=c]{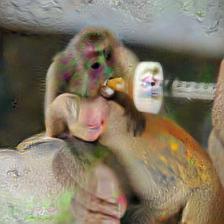}
 & 
\includegraphics[width=0.086\textwidth,valign=c]{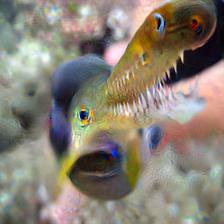}
 & 
\includegraphics[width=0.086\textwidth,valign=c]{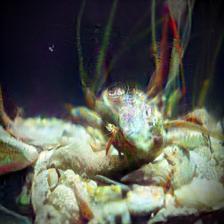}
 & 
\includegraphics[width=0.086\textwidth,valign=c]{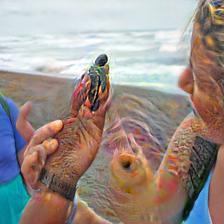}
 & 
\includegraphics[width=0.086\textwidth,valign=c]{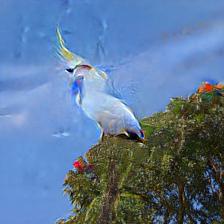}
\vspace{1mm}\\
\includegraphics[width=0.086\textwidth,valign=c]{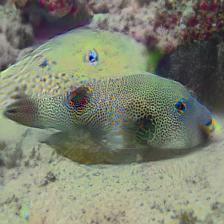}
 & 
\includegraphics[width=0.086\textwidth,valign=c]{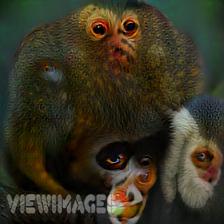}
 & 
\includegraphics[width=0.086\textwidth,valign=c]{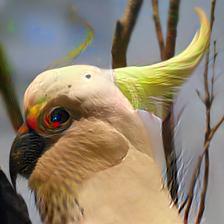}
 & 
\includegraphics[width=0.086\textwidth,valign=c]{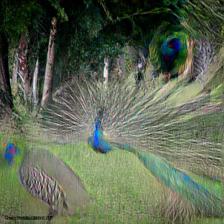}
 & 
\includegraphics[width=0.086\textwidth,valign=c]{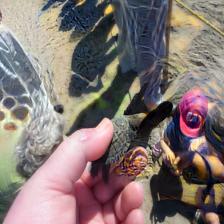}
 & 
\hspace{ 0.043 \textwidth} & 
\includegraphics[width=0.086\textwidth,valign=c]{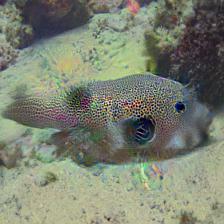}
 & 
\includegraphics[width=0.086\textwidth,valign=c]{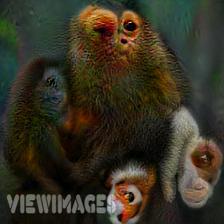}
 & 
\includegraphics[width=0.086\textwidth,valign=c]{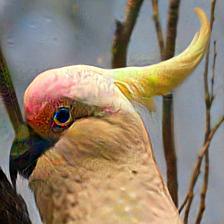}
 & 
\includegraphics[width=0.086\textwidth,valign=c]{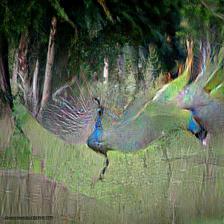}
 & 
\includegraphics[width=0.086\textwidth,valign=c]{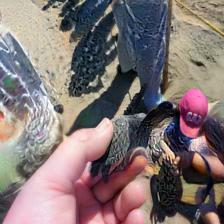}
\vspace{1mm}\\
\includegraphics[width=0.086\textwidth,valign=c]{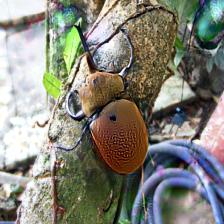}
 & 
\includegraphics[width=0.086\textwidth,valign=c]{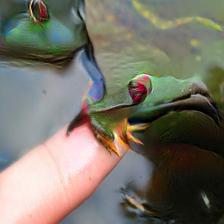}
 & 
\includegraphics[width=0.086\textwidth,valign=c]{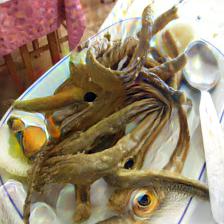}
 & 
\includegraphics[width=0.086\textwidth,valign=c]{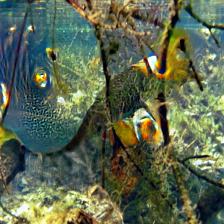}
 & 
\includegraphics[width=0.086\textwidth,valign=c]{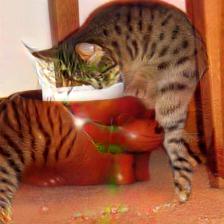}
 & 
\hspace{ 0.043 \textwidth} & 
\includegraphics[width=0.086\textwidth,valign=c]{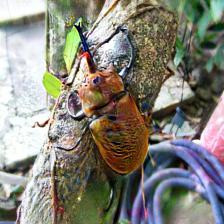}
 & 
\includegraphics[width=0.086\textwidth,valign=c]{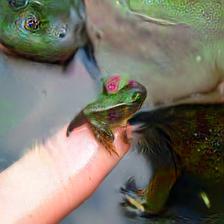}
 & 
\includegraphics[width=0.086\textwidth,valign=c]{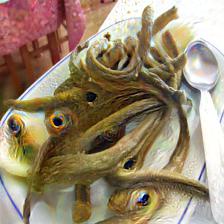}
 & 
\includegraphics[width=0.086\textwidth,valign=c]{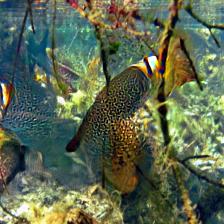}
 & 
\includegraphics[width=0.086\textwidth,valign=c]{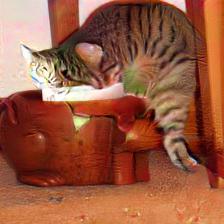}
\vspace{1mm}\\
\includegraphics[width=0.086\textwidth,valign=c]{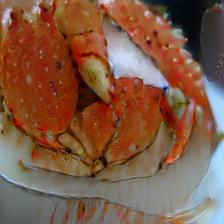}
 & 
\includegraphics[width=0.086\textwidth,valign=c]{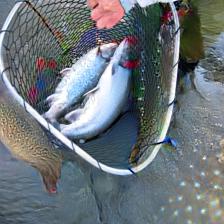}
 & 
\includegraphics[width=0.086\textwidth,valign=c]{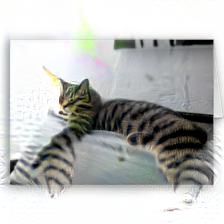}
 & 
\includegraphics[width=0.086\textwidth,valign=c]{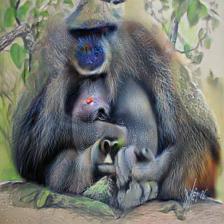}
 & 
\includegraphics[width=0.086\textwidth,valign=c]{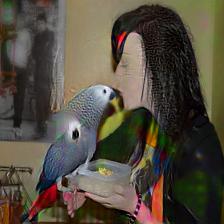}
 & 
\hspace{ 0.043 \textwidth} & 
\includegraphics[width=0.086\textwidth,valign=c]{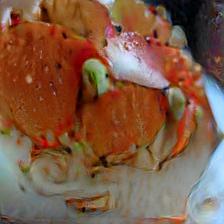}
 & 
\includegraphics[width=0.086\textwidth,valign=c]{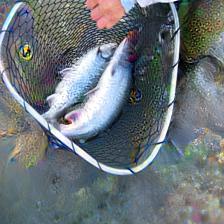}
 & 
\includegraphics[width=0.086\textwidth,valign=c]{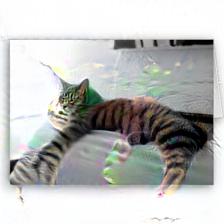}
 & 
\includegraphics[width=0.086\textwidth,valign=c]{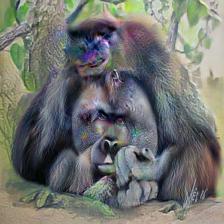}
 & 
\includegraphics[width=0.086\textwidth,valign=c]{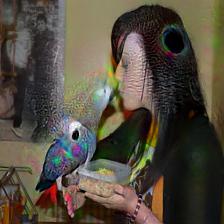}
\vspace{2mm}\\
\multicolumn{5}{c}{RATIO-3.5} & & \multicolumn{5}{c}{RATIO-1.75}\\
\includegraphics[width=0.086\textwidth,valign=c]{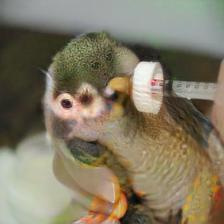}
 & 
\includegraphics[width=0.086\textwidth,valign=c]{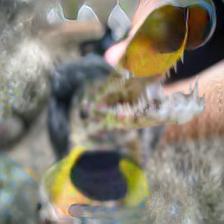}
 & 
\includegraphics[width=0.086\textwidth,valign=c]{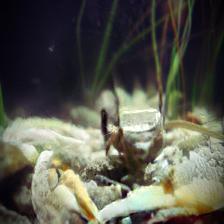}
 & 
\includegraphics[width=0.086\textwidth,valign=c]{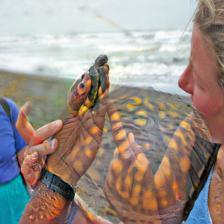}
 & 
\includegraphics[width=0.086\textwidth,valign=c]{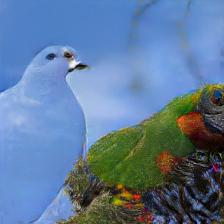}
 & 
\hspace{ 0.043 \textwidth} & 
\includegraphics[width=0.086\textwidth,valign=c]{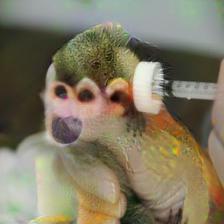}
 & 
\includegraphics[width=0.086\textwidth,valign=c]{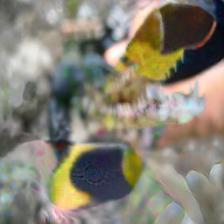}
 & 
\includegraphics[width=0.086\textwidth,valign=c]{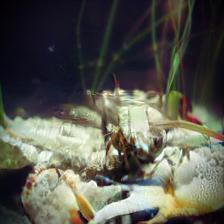}
 & 
\includegraphics[width=0.086\textwidth,valign=c]{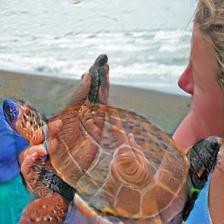}
 & 
\includegraphics[width=0.086\textwidth,valign=c]{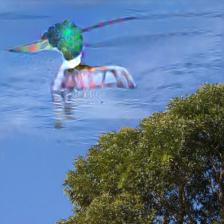}
\vspace{1mm}\\
\includegraphics[width=0.086\textwidth,valign=c]{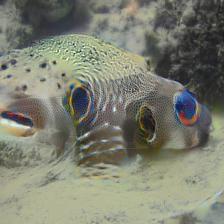}
 & 
\includegraphics[width=0.086\textwidth,valign=c]{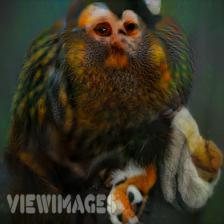}
 & 
\includegraphics[width=0.086\textwidth,valign=c]{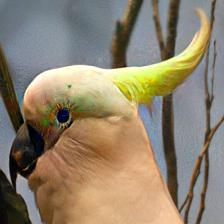}
 & 
\includegraphics[width=0.086\textwidth,valign=c]{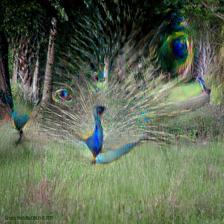}
 & 
\includegraphics[width=0.086\textwidth,valign=c]{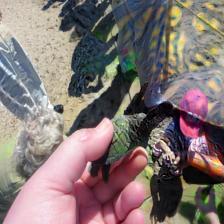}
 & 
\hspace{ 0.043 \textwidth} & 
\includegraphics[width=0.086\textwidth,valign=c]{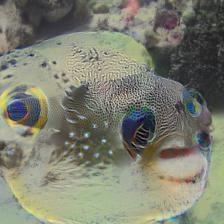}
 & 
\includegraphics[width=0.086\textwidth,valign=c]{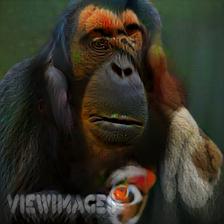}
 & 
\includegraphics[width=0.086\textwidth,valign=c]{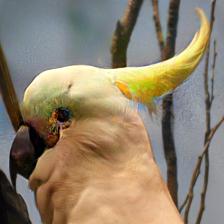}
 & 
\includegraphics[width=0.086\textwidth,valign=c]{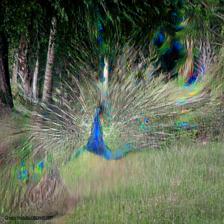}
 & 
\includegraphics[width=0.086\textwidth,valign=c]{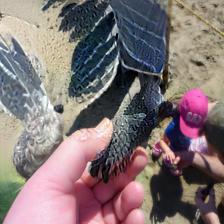}
\vspace{1mm}\\
\includegraphics[width=0.086\textwidth,valign=c]{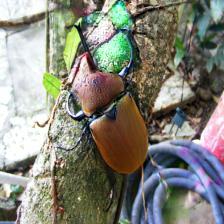}
 & 
\includegraphics[width=0.086\textwidth,valign=c]{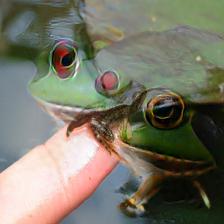}
 & 
\includegraphics[width=0.086\textwidth,valign=c]{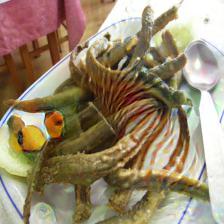}
 & 
\includegraphics[width=0.086\textwidth,valign=c]{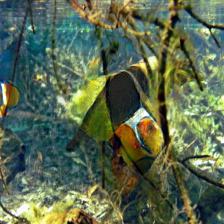}
 & 
\includegraphics[width=0.086\textwidth,valign=c]{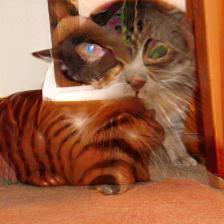}
 & 
\hspace{ 0.043 \textwidth} & 
\includegraphics[width=0.086\textwidth,valign=c]{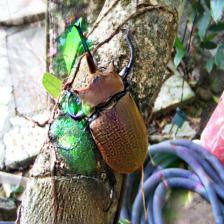}
 & 
\includegraphics[width=0.086\textwidth,valign=c]{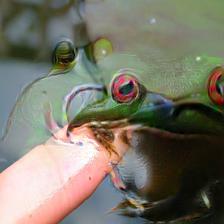}
 & 
\includegraphics[width=0.086\textwidth,valign=c]{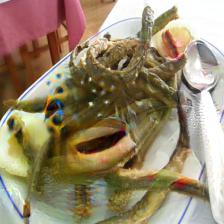}
 & 
\includegraphics[width=0.086\textwidth,valign=c]{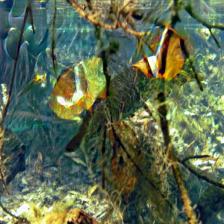}
 & 
\includegraphics[width=0.086\textwidth,valign=c]{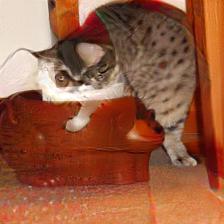}
\vspace{1mm}\\
\includegraphics[width=0.086\textwidth,valign=c]{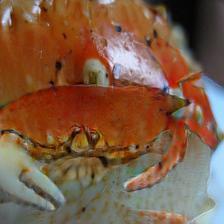}
 & 
\includegraphics[width=0.086\textwidth,valign=c]{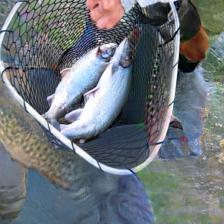}
 & 
\includegraphics[width=0.086\textwidth,valign=c]{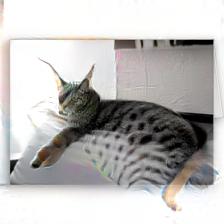}
 & 
\includegraphics[width=0.086\textwidth,valign=c]{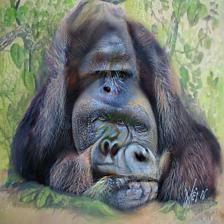}
 & 
\includegraphics[width=0.086\textwidth,valign=c]{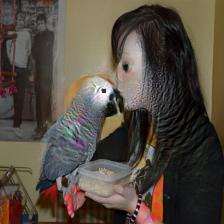}
 & 
\hspace{ 0.043 \textwidth} & 
\includegraphics[width=0.086\textwidth,valign=c]{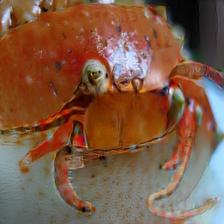}
 & 
\includegraphics[width=0.086\textwidth,valign=c]{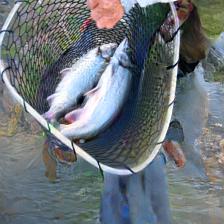}
 & 
\includegraphics[width=0.086\textwidth,valign=c]{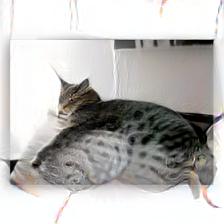}
 & 
\includegraphics[width=0.086\textwidth,valign=c]{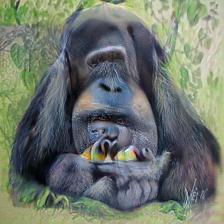}
 & 
\includegraphics[width=0.086\textwidth,valign=c]{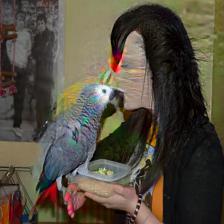}
\vspace{2mm}\\
\end{tabu}
\end{adjustbox}
\caption{Random selection of  20 images from the restricted ImageNet test set which are misclassified by all our models and the associated visual counterfactuals that are generated by maximizing the confidence in the gt-class in a $l_2$ ball of radius 21. The numbers over the individual images indicate the target class and were omitted for the latter models. }\label{fig:restricted_overview}
\end{figure}

%% file: res/appendix_od_restricted_new.tex
\begin{figure}[ht!]
\begin{tabular}{p{1cm}x{\breite}x{\breite}x{\breite}x{\breite}x{\breite}x{\breite}x{\breite}x{\breite}}
Model  & Orig. & $\epsilon=3.5$ & $\epsilon=7.0$ & $\epsilon=10.5$ & $\epsilon=14.0$ & $\epsilon=17.5$ & $\epsilon=21.0$\\  
\begin{turn}{90} \hspace{-.8cm} M. AT-3.50 \end{turn}  &  \multicolumn{7}{c}{\includegraphics[width=0.91\textwidth,valign=c]{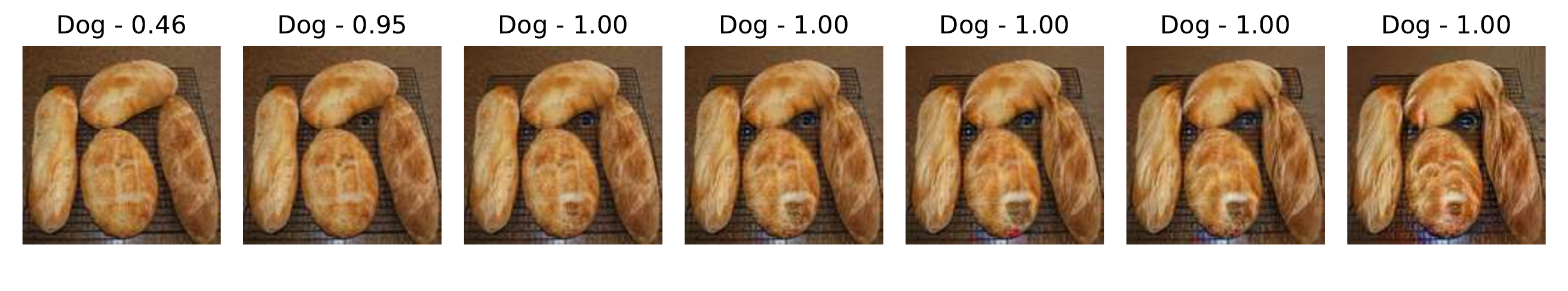}} \\
\hline
\begin{turn}{90} \hspace{-.5cm} AT-3.50 \end{turn}  &  \multicolumn{7}{c}{\includegraphics[width=0.91\textwidth,valign=c]{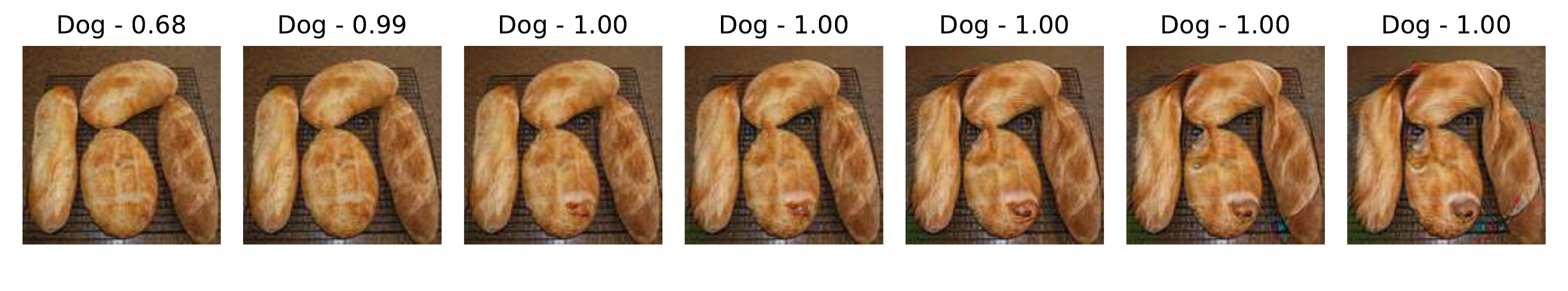}} \\
\hline
\begin{turn}{90} \hspace{-.4cm} R-3.50 \end{turn} & \multicolumn{7}{c}{\includegraphics[width=0.91\textwidth,valign=c]{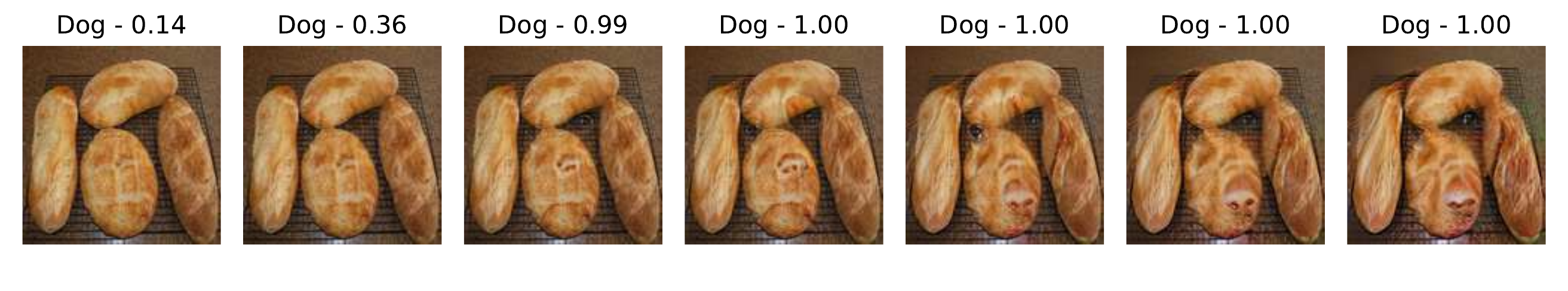}} \\
\hline
\begin{turn}{90} \hspace{-.4cm} R-1.75 \end{turn}  &  \multicolumn{7}{c}{\includegraphics[width=0.91\textwidth,valign=c]{pics/imagenet/Ratio175Clean/OD/img_466.pdf}} \\
%
\vspace{2mm}\\
\begin{turn}{90} \hspace{-.9cm} M. AT-3.50 \end{turn}  &  \multicolumn{7}{c}{\includegraphics[width=0.91\textwidth,valign=c]{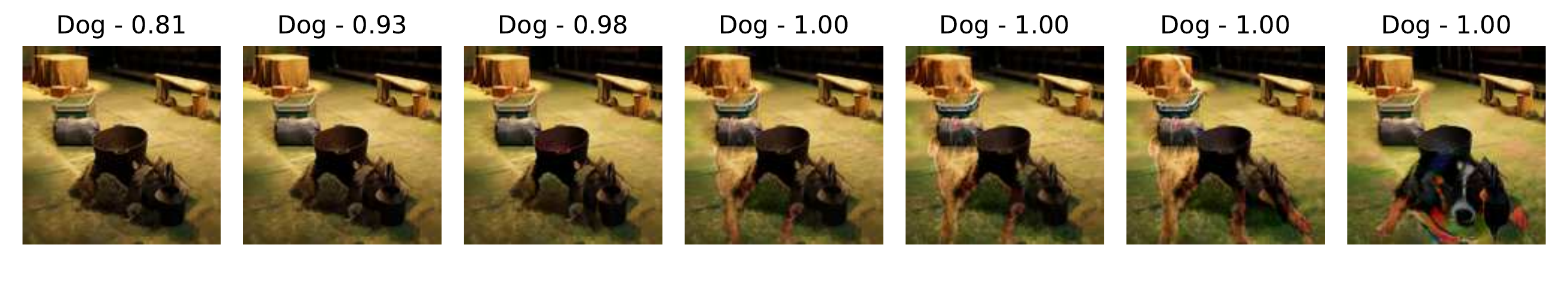}} \\
\hline
\begin{turn}{90} \hspace{-.5cm} AT-3.50 \end{turn}  &  \multicolumn{7}{c}{\includegraphics[width=0.91\textwidth,valign=c]{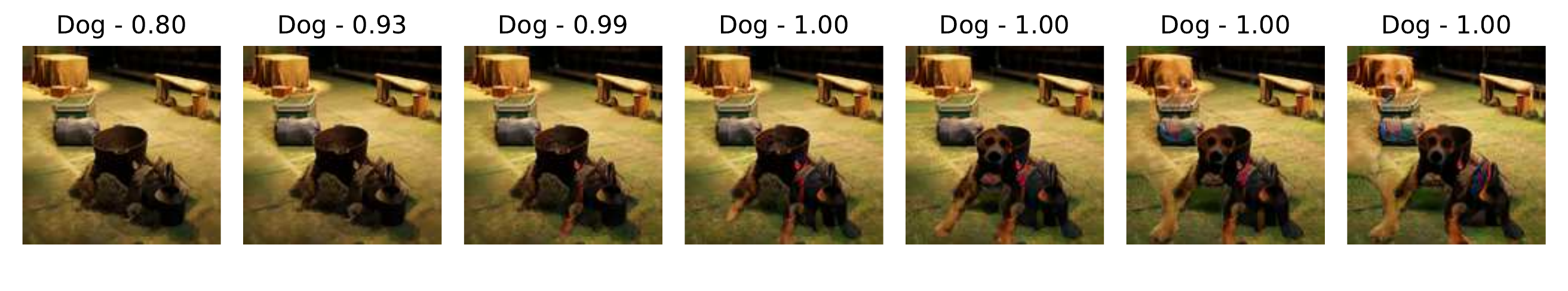}} \\
\hline
\begin{turn}{90} \hspace{-.4cm} R-3.50 \end{turn} & \multicolumn{7}{c}{\includegraphics[width=0.91\textwidth,valign=c]{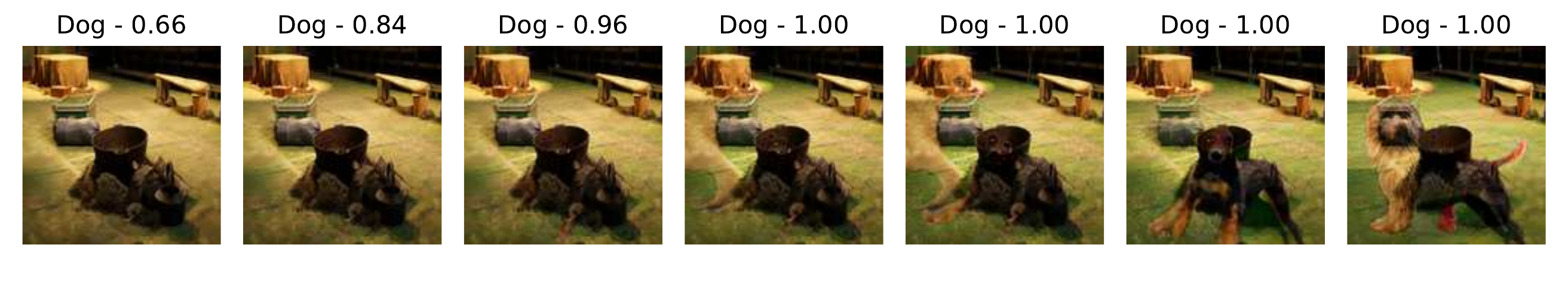}} \\
\hline
\begin{turn}{90} \hspace{-.4cm} R-1.75 \end{turn}  &  \multicolumn{7}{c}{\includegraphics[width=0.91\textwidth,valign=c]{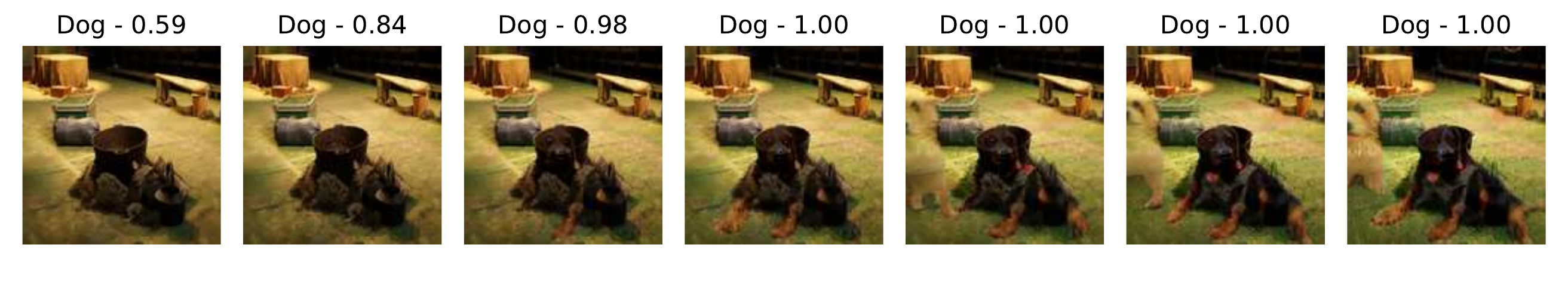}} \\
\end{tabular}	
\vspace{-.5cm}
\caption{\label{fig:od_imagenet_new1}\textbf{Feature Generation on OOD images (restricted ImageNet)} for images from the remaining classes of ImageNet. In both panels the AT models are overconfident on the OOD images and produce too fast high confidence predictions even though class specific features are not present yet. The RATIO models show high confidence only when class-specific features have appeared.}
\end{figure}

\begin{figure}[ht!]
\begin{tabular}{p{1cm}x{\breite}x{\breite}x{\breite}x{\breite}x{\breite}x{\breite}x{\breite}x{\breite}}
Model  & Orig. & $\epsilon=3.5$ & $\epsilon=7.0$ & $\epsilon=10.5$ & $\epsilon=14.0$ & $\epsilon=17.5$ & $\epsilon=21.0$\\  
\begin{turn}{90} \hspace{-.9cm} M. AT-3.50 \end{turn}  &  \multicolumn{7}{c}{\includegraphics[width=0.91\textwidth,valign=c]{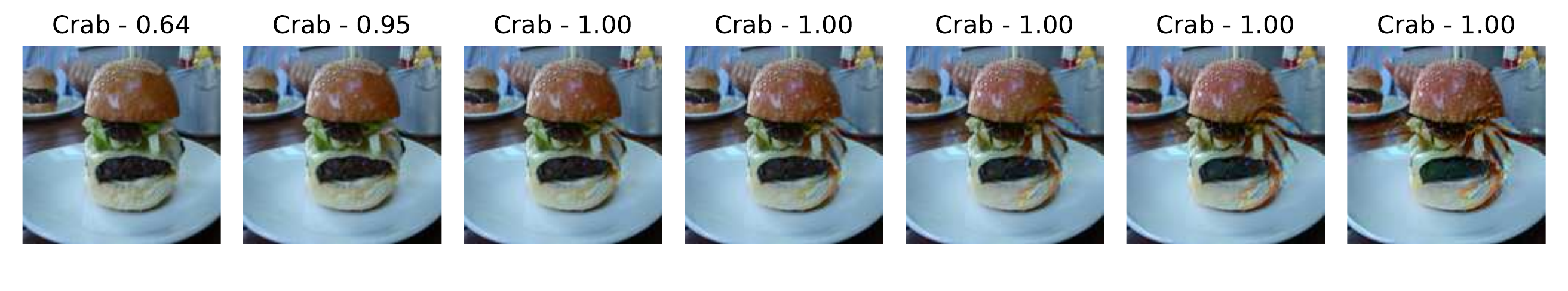}} \\
\hline
\begin{turn}{90} \hspace{-.5cm} AT-3.50 \end{turn}  &  \multicolumn{7}{c}{\includegraphics[width=0.91\textwidth,valign=c]{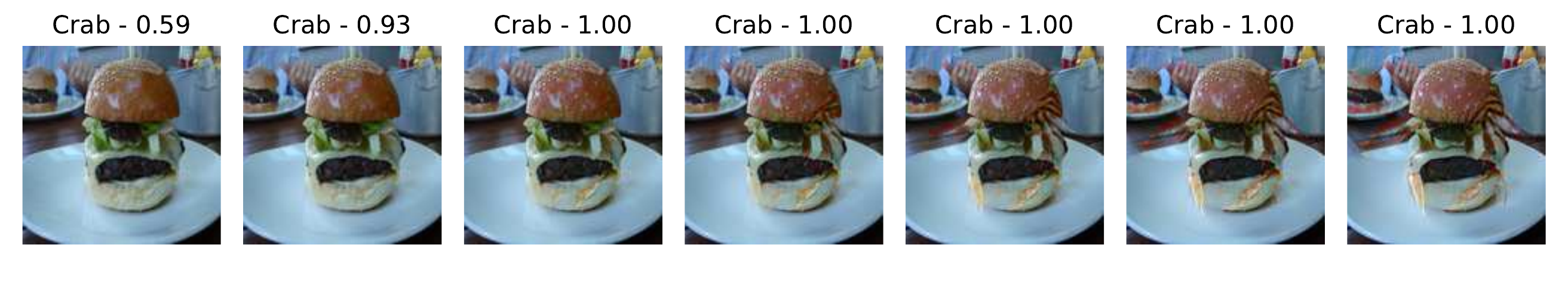}} \\
\hline
\begin{turn}{90} \hspace{-.4cm} R-3.50 \end{turn} & \multicolumn{7}{c}{\includegraphics[width=0.91\textwidth,valign=c]{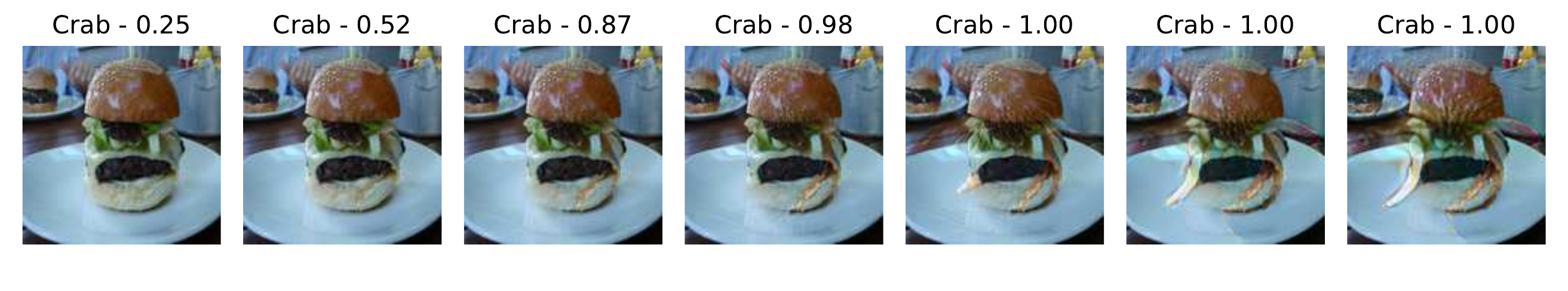}} \\
\hline
\begin{turn}{90} \hspace{-.4cm} R-1.75 \end{turn}  &  \multicolumn{7}{c}{\includegraphics[width=0.91\textwidth,valign=c]{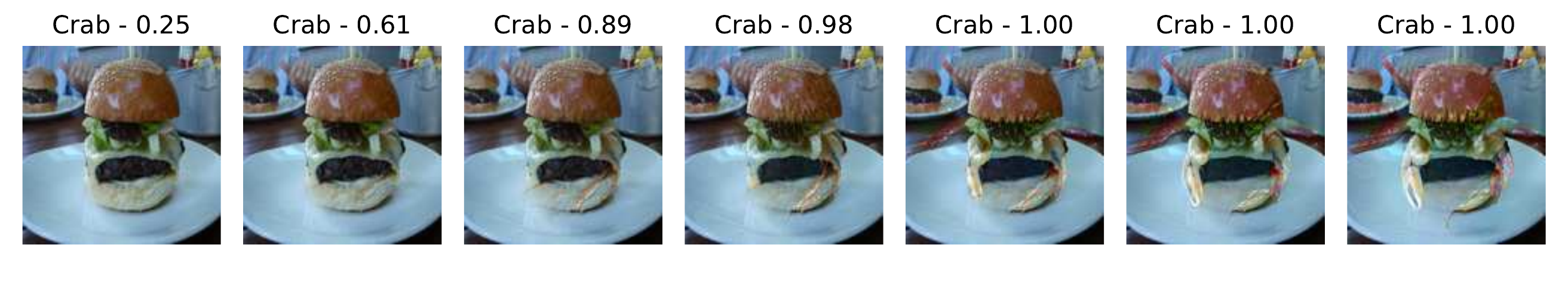}} \\
%
\vspace{2mm}\\
\begin{turn}{90} \hspace{-.9cm} M. AT-3.50 \end{turn}  &  \multicolumn{7}{c}{\includegraphics[width=0.91\textwidth,valign=c]{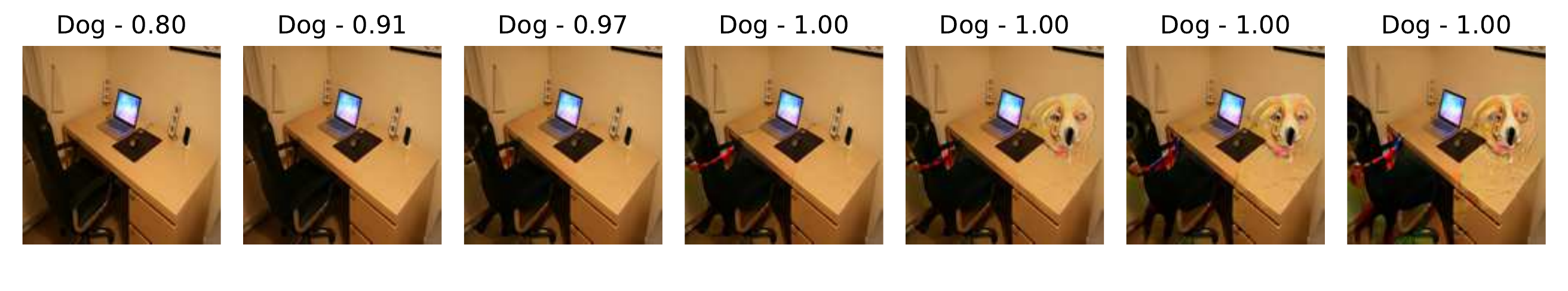}} \\
\hline
\begin{turn}{90} \hspace{-.5cm} AT-3.50 \end{turn}  &  \multicolumn{7}{c}{\includegraphics[width=0.91\textwidth,valign=c]{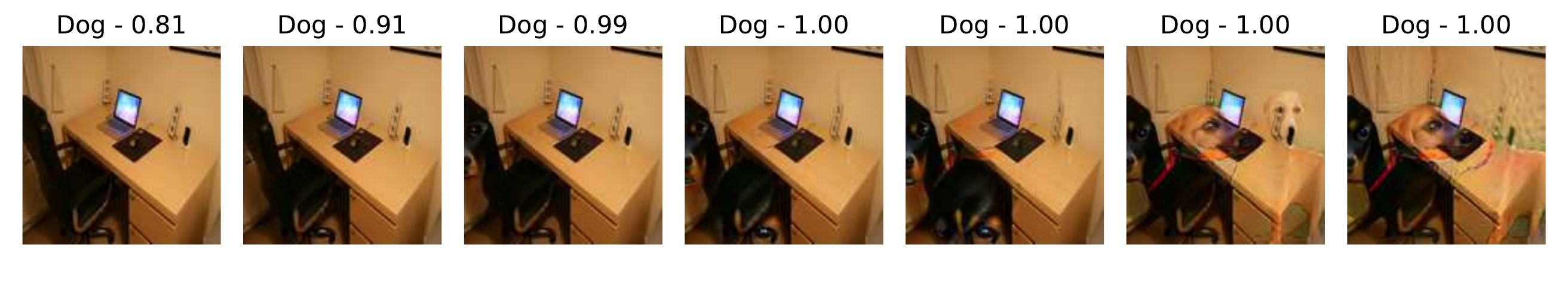}} \\
\hline
\begin{turn}{90} \hspace{-.4cm} R-3.50 \end{turn} & \multicolumn{7}{c}{\includegraphics[width=0.91\textwidth,valign=c]{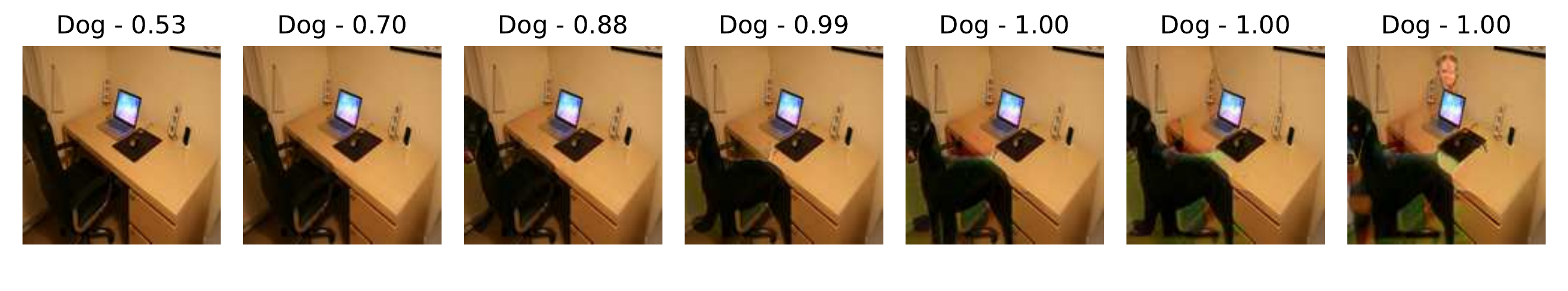}} \\
\hline
\begin{turn}{90} \hspace{-.4cm} R-1.75 \end{turn}  &  \multicolumn{7}{c}{\includegraphics[width=0.91\textwidth,valign=c]{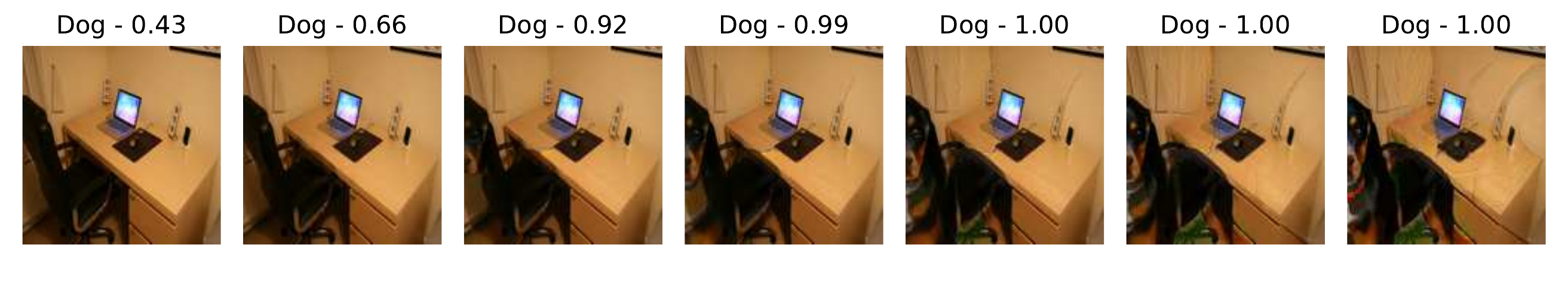}} \\
\end{tabular}	
\vspace{-.5cm}
\caption{\label{fig:od_imagenet_new2}\textbf{Feature Generation on OOD images (restricted ImageNet)} for images from the remaining classes of ImageNet. In both panels the AT models are overconfident on the OOD images and produce high confidence predictions too fast even though class specific features are not present yet. The RATIO models show high confidence only when class-specific features have appeared.}
\end{figure}

%% file: res/appendix_failure_vc_cifar10.tex
\begin{figure}[ht!]
\begin{tabular}{p{1cm}x{\breite}x{\breite}x{\breite}x{\breite}x{\breite}x{\breite}x{\breite}x{\breite}}
Model  & Orig. & $\epsilon=0.5$ & $\epsilon=1.0$ & $\epsilon=1.5$ & $\epsilon=2.0$ & $\epsilon=2.5$ & $\epsilon=3.0$\\
\begin{turn}{90} \hspace{-.4cm} ACET \end{turn} & \multicolumn{7}{c}{\includegraphics[width=0.91\textwidth,valign=c]{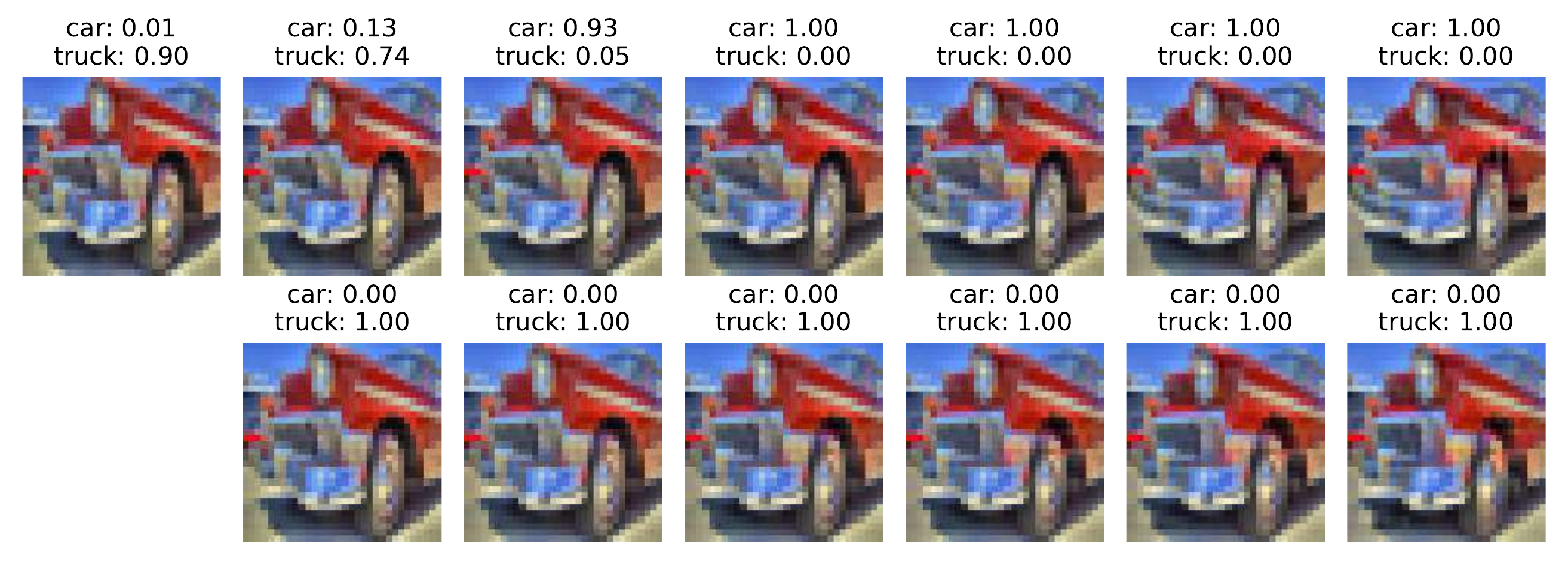}} \\
\hline
\begin{turn}{90} \hspace{-.33cm}  JEM-0 \end{turn}  &  \multicolumn{7}{c}{\includegraphics[width=0.91\textwidth,valign=c]{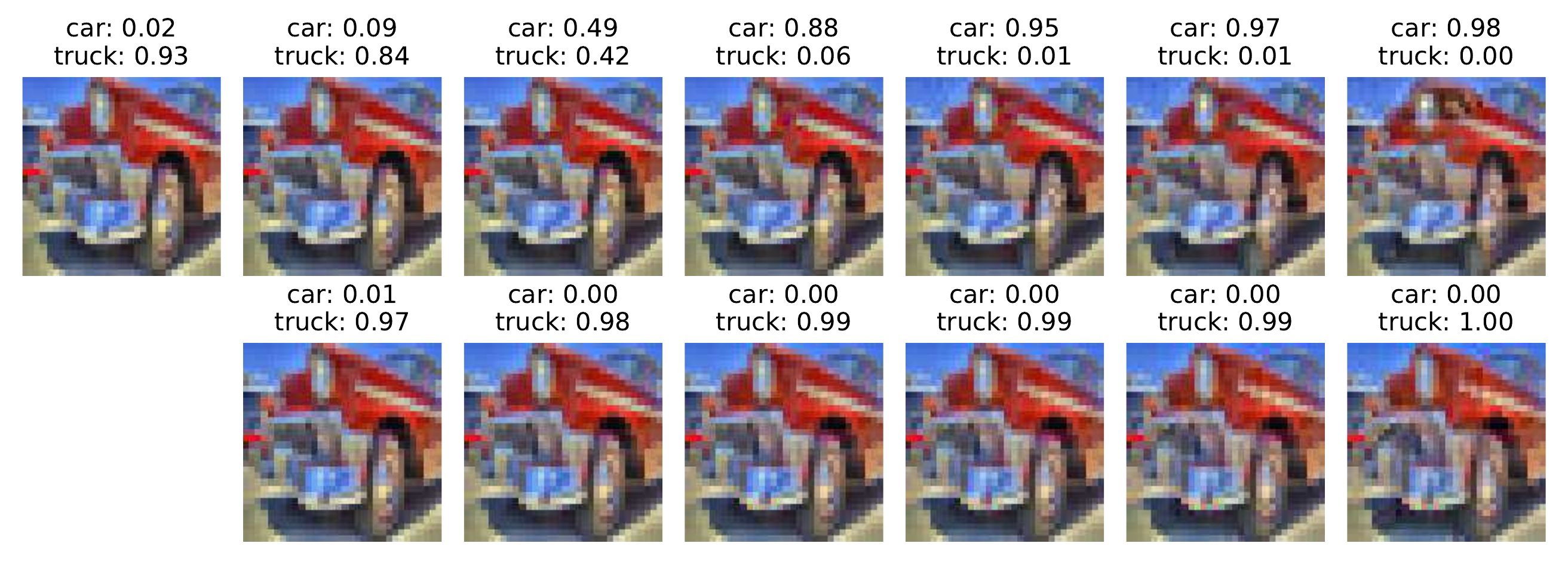}} \\
\hline
\begin{turn}{90} \hspace{-.4cm} AT-0.50 \end{turn}  &  \multicolumn{7}{c}{\includegraphics[width=0.91\textwidth,valign=c]{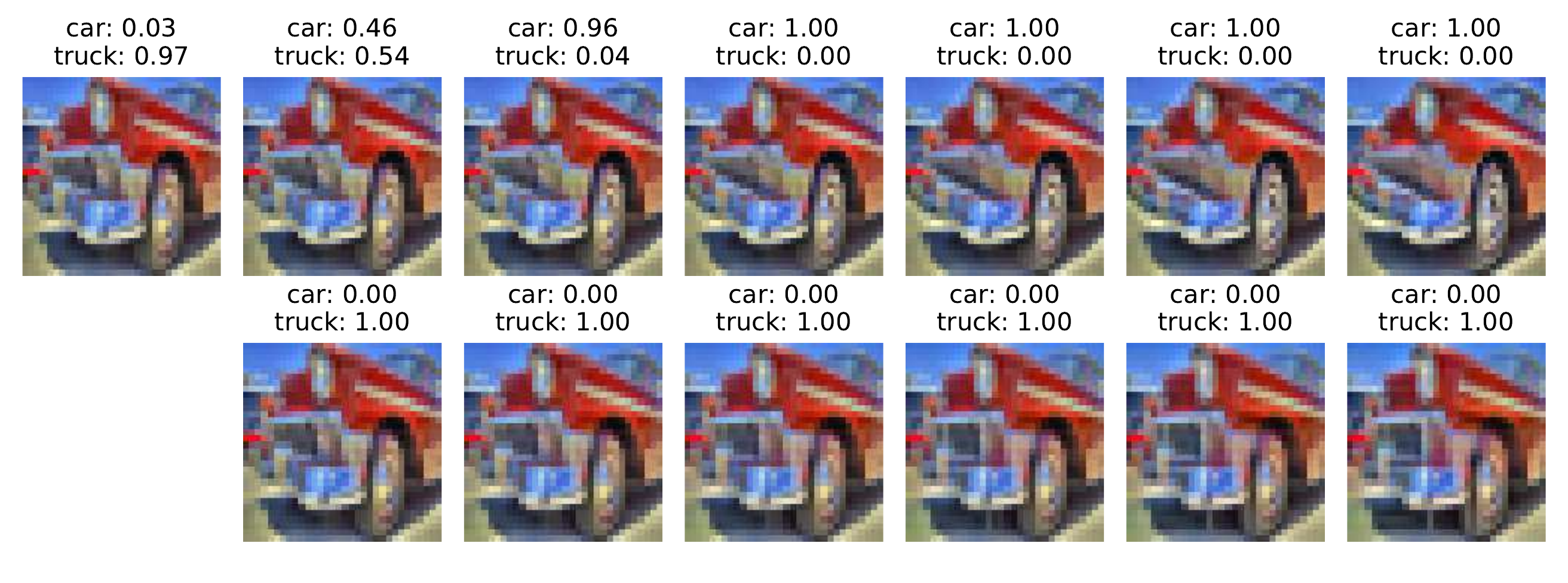}} \\
\hline
\begin{turn}{90} \hspace{-.9cm} RATIO-0.25 \end{turn}  &  \multicolumn{7}{c}{\includegraphics[width=0.91\textwidth,valign=c]{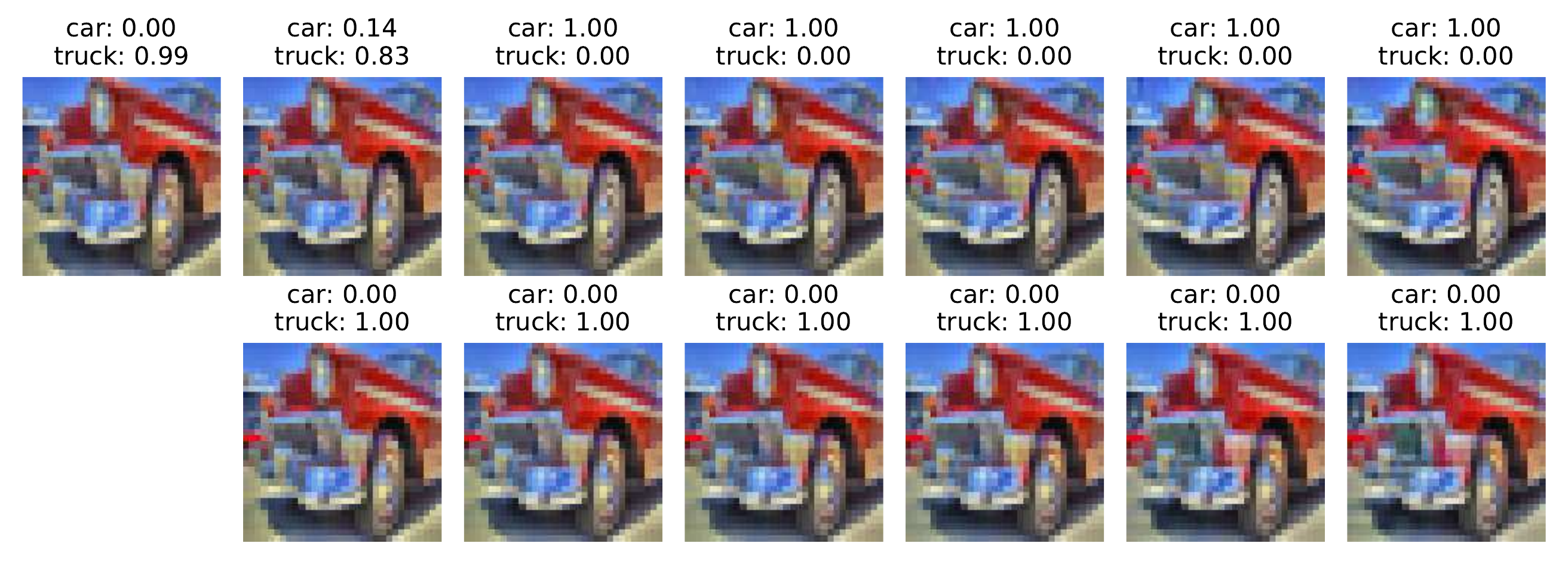}} \\
\end{tabular}	
\vspace{-.3cm}
\caption{\label{fig:vc_cifar_failure1}
\textbf{Visual Counterfactual - Failure Case for CIFAR10:}
For certain images visual counterfactuals yield little insight for the human observer even though the model confidences flip from 0 to 1. For CIFAR10, especially the ``car'' and ``truck'' classes are hard to distinguish for all models. }
\end{figure}

%% file: res/appendix_failure_od_cifar10.tex
\begin{figure}
\begin{tabular}{p{1cm}x{\breite}x{\breite}x{\breite}x{\breite}x{\breite}x{\breite}x{\breite}x{\breite}}
Model  & Orig. & $\epsilon=0.5$ & $\epsilon=1.0$ & $\epsilon=1.5$ & $\epsilon=2.0$ & $\epsilon=2.5$ & $\epsilon=3.0$\\
\begin{turn}{90} \hspace{-.4cm} ACET \end{turn} & \multicolumn{7}{c}{
\includegraphics[width=0.91\textwidth,valign=c]{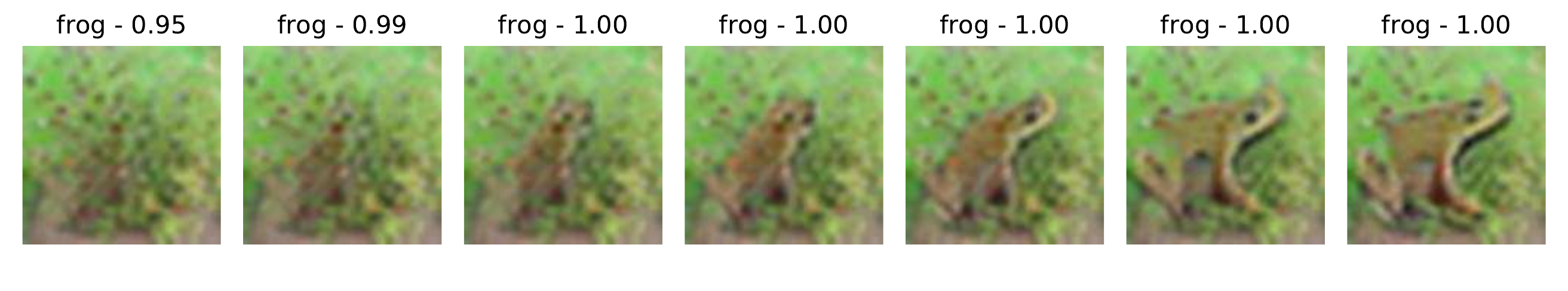}
} \\
\hline
\begin{turn}{90} \hspace{-.4cm} JEM-0 \end{turn} & \multicolumn{7}{c}{
\includegraphics[width=0.91\textwidth,valign=c]{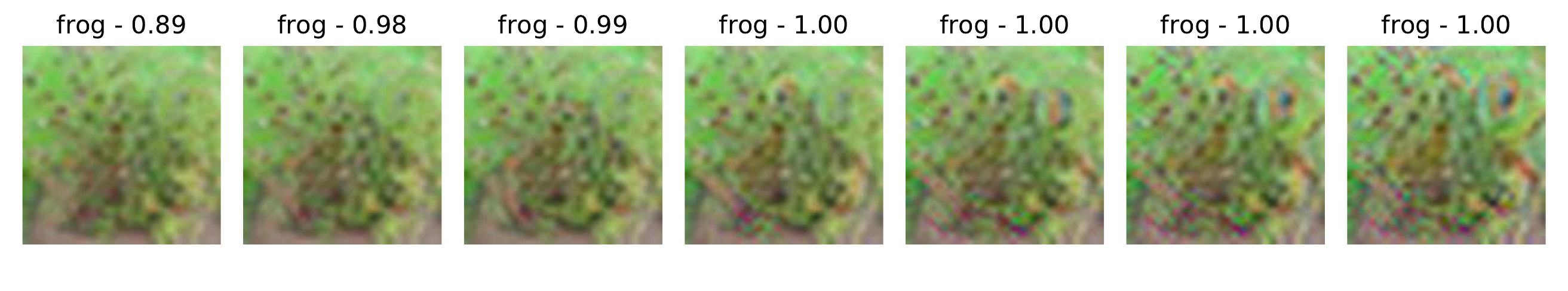}
} \\
\hline
\begin{turn}{90} \hspace{-.4cm} AT-0.50 \end{turn} & \multicolumn{7}{c}{
\includegraphics[width=0.91\textwidth,valign=c]{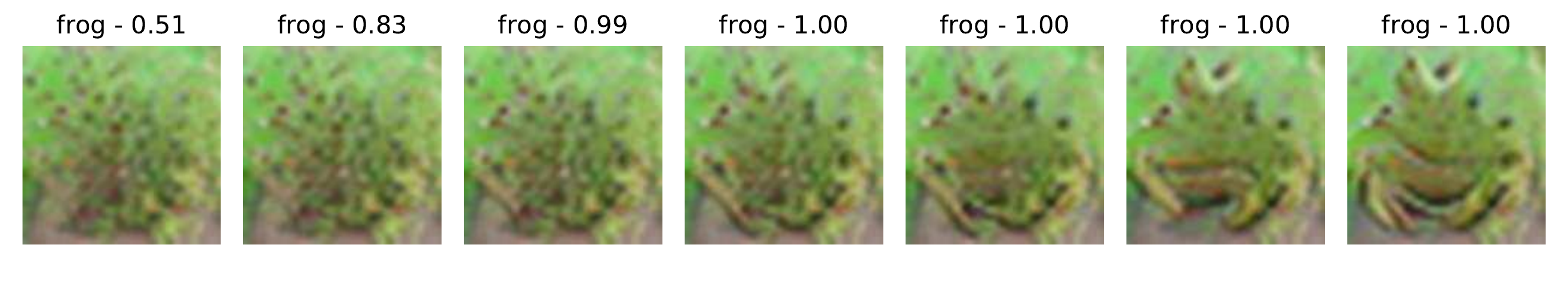}
} \\
\hline
\begin{turn}{90} \hspace{-.4cm} R-0.25 \end{turn} & \multicolumn{7}{c}{
\includegraphics[width=0.91\textwidth,valign=c]{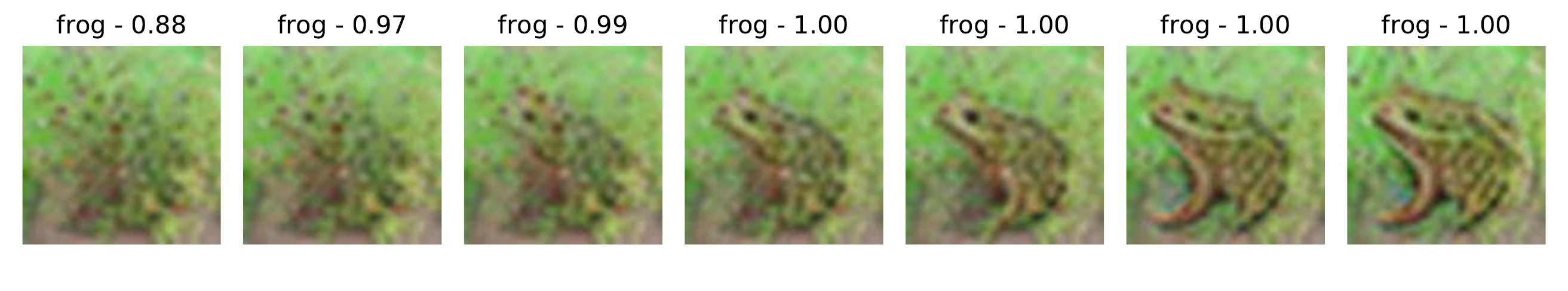}
} \\
\end{tabular}	
\vspace{-.5cm}
\caption{\textbf{Feature Generation on OOD images - Failure Case CIFAR10:} While RATIO is able to generate high-quality samples for comparatively small radii of 1.0 or larger, the initial confidence is too high as no class features are visible.}

\begin{tabular}{p{1cm}x{\breite}x{\breite}x{\breite}x{\breite}x{\breite}x{\breite}x{\breite}x{\breite}}
\begin{turn}{90} \hspace{-.4cm} ACET \end{turn} & \multicolumn{7}{c}{
\includegraphics[width=0.91\textwidth,valign=c]{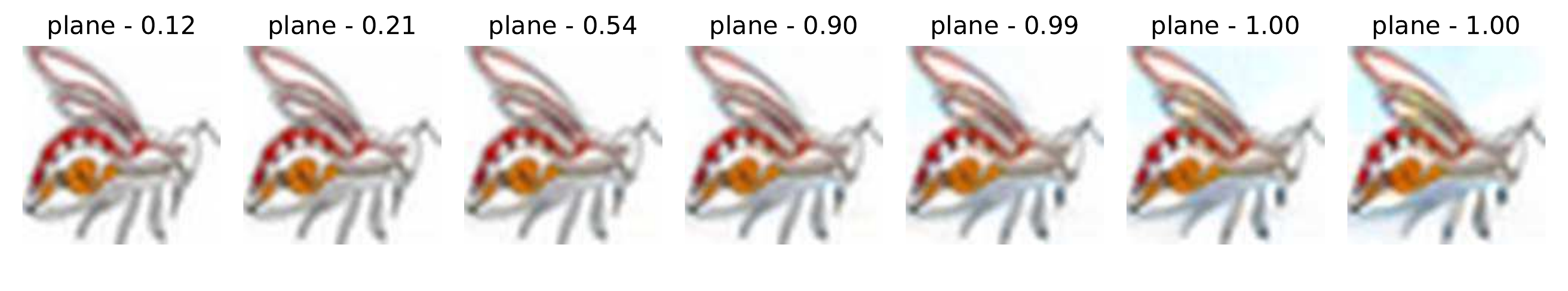}
} \\
\hline
\begin{turn}{90} \hspace{-.4cm} JEM-0 \end{turn} & \multicolumn{7}{c}{
\includegraphics[width=0.91\textwidth,valign=c]{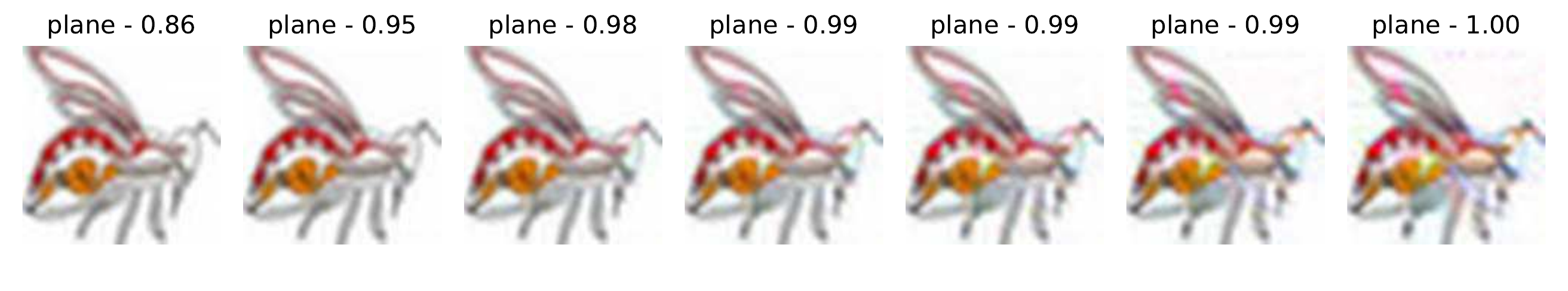}
} \\
\hline
\begin{turn}{90} \hspace{-.4cm} AT-0.50 \end{turn} & \multicolumn{7}{c}{
\includegraphics[width=0.91\textwidth,valign=c]{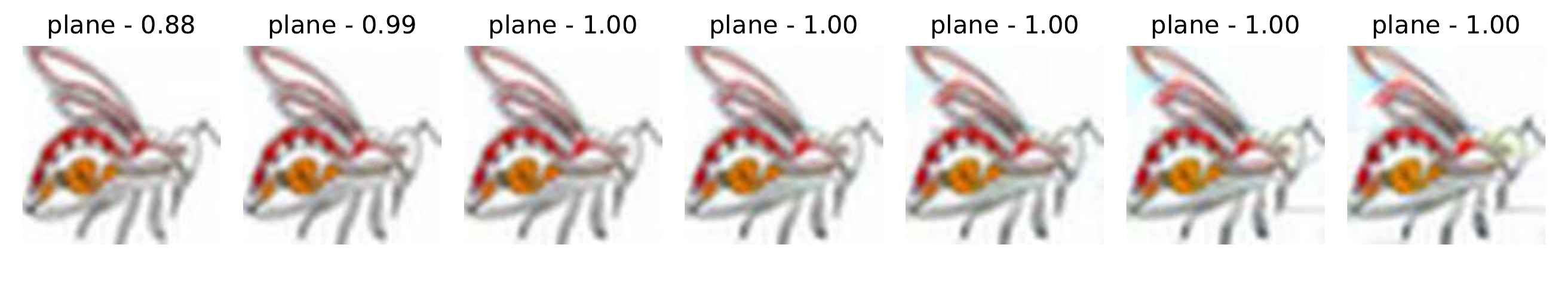}
} \\
\hline
\begin{turn}{90} \hspace{-.4cm} R-0.25 \end{turn} & \multicolumn{7}{c}{
\includegraphics[width=0.91\textwidth,valign=c]{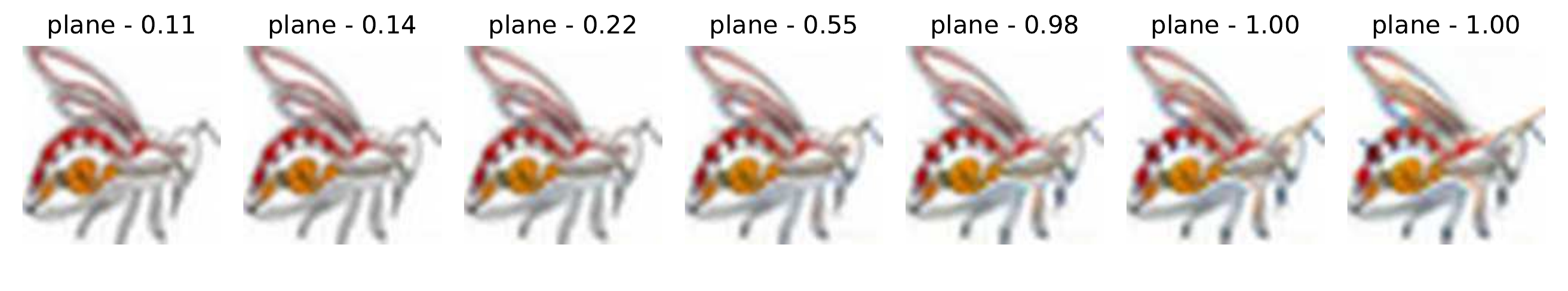}
} \\
\end{tabular}
\vspace{-.5cm}	
\caption{\label{fig:od_cifar_fail2}\textbf{Feature Generation on OOD images - Failure Case CIFAR10:} RATIO correctly assigns a low confidence to the given out-distribution samples (in contrast to JEM-0 and AT$_{0.5}$) but the confidence increases to 1.0 without the appearance of interpretable features.}
\end{figure}

%% file: res/appendix_failure_vc_svhn.tex
\begin{figure}[ht!]
\begin{tabular}{p{1cm}x{\breite}x{\breite}x{\breite}x{\breite}x{\breite}x{\breite}x{\breite}x{\breite}}
Model  & Orig. & $\epsilon=0.5$ & $\epsilon=1.0$ & $\epsilon=1.5$ & $\epsilon=2.0$ & $\epsilon=2.5$ & $\epsilon=3.0$\\
\begin{turn}{90} \hspace{-.4cm} ACET \end{turn} & \multicolumn{7}{c}{\includegraphics[width=0.91\textwidth,valign=c]{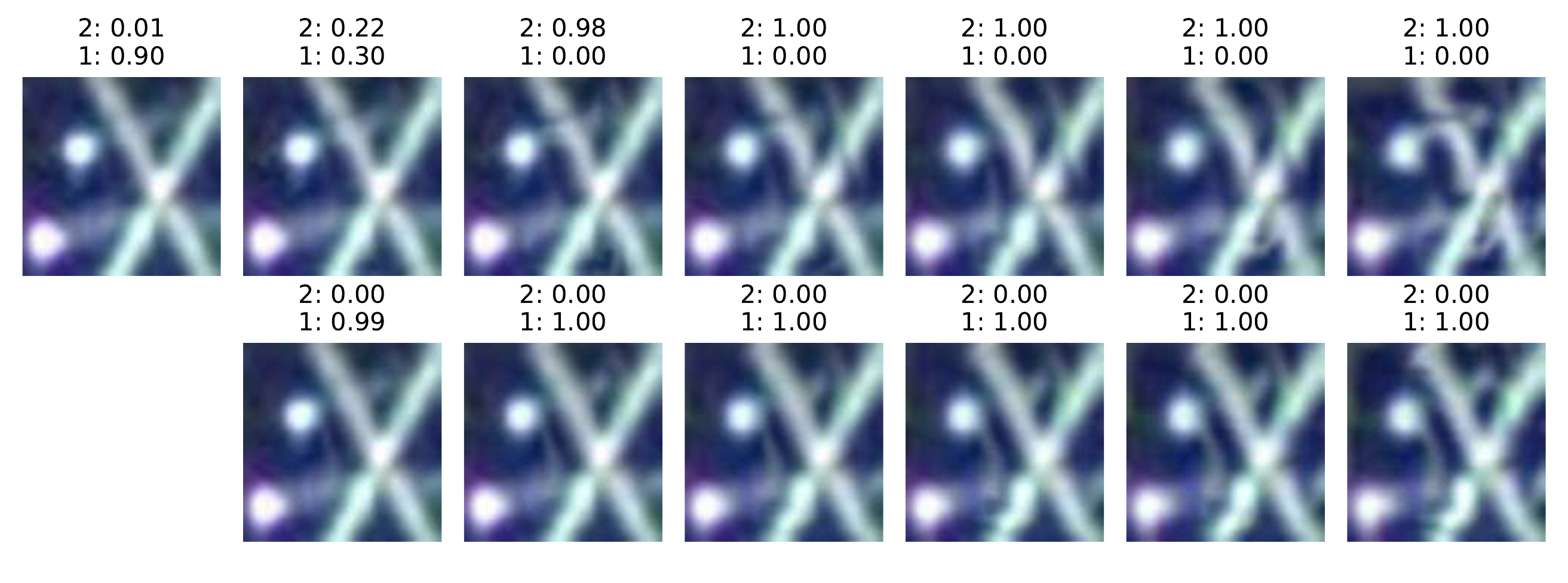}} \\
\hline
\begin{turn}{90} \hspace{-.4cm} AT-0.50 \end{turn}  &  \multicolumn{7}{c}{\includegraphics[width=0.91\textwidth,valign=c]{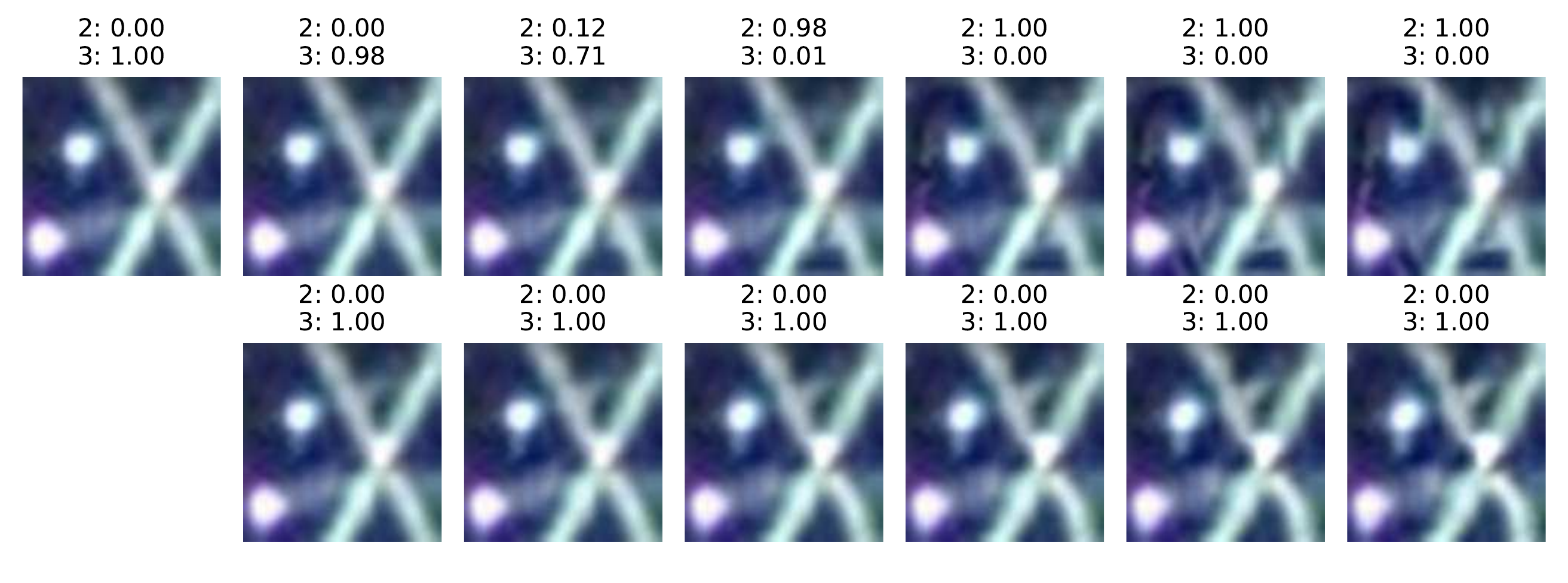}} \\
\hline
\begin{turn}{90} \hspace{-.4cm} AT-0.25 \end{turn}  &  \multicolumn{7}{c}{\includegraphics[width=0.91\textwidth,valign=c]{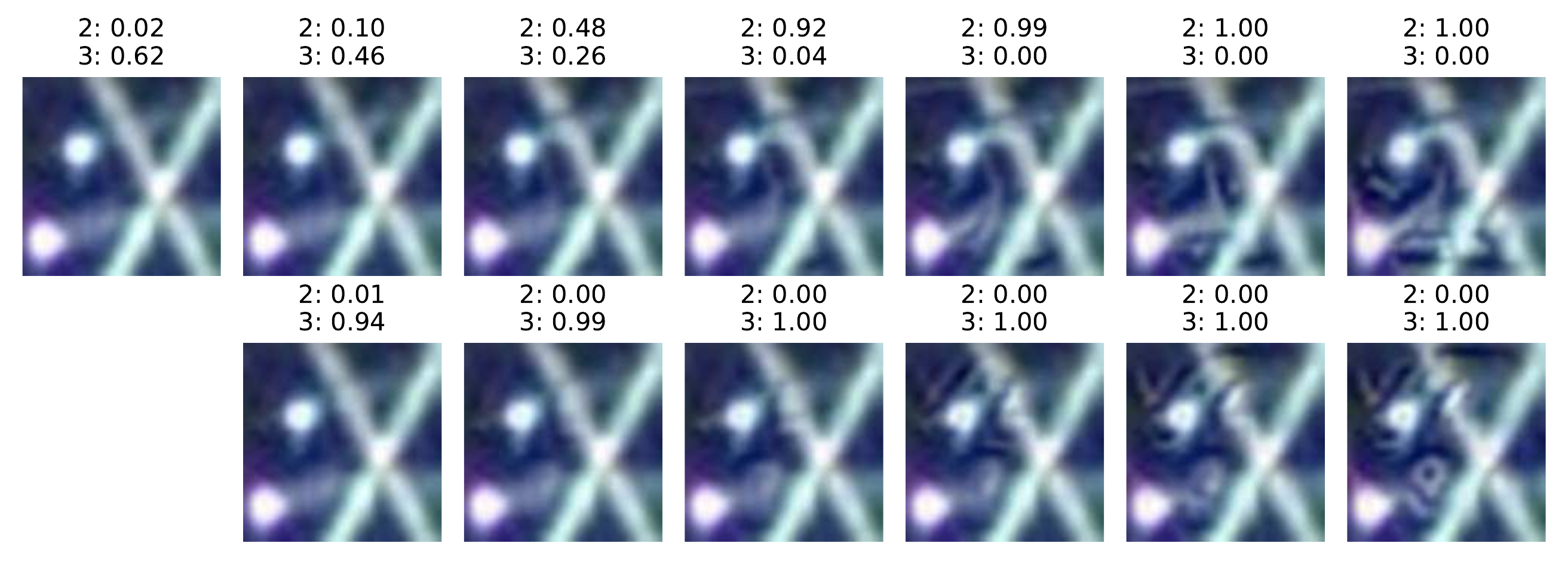}} \\
\hline
\begin{turn}{90} \hspace{-.9cm} RATIO-0.25 \end{turn} & \multicolumn{7}{c}{\includegraphics[width=0.91\textwidth,valign=c]{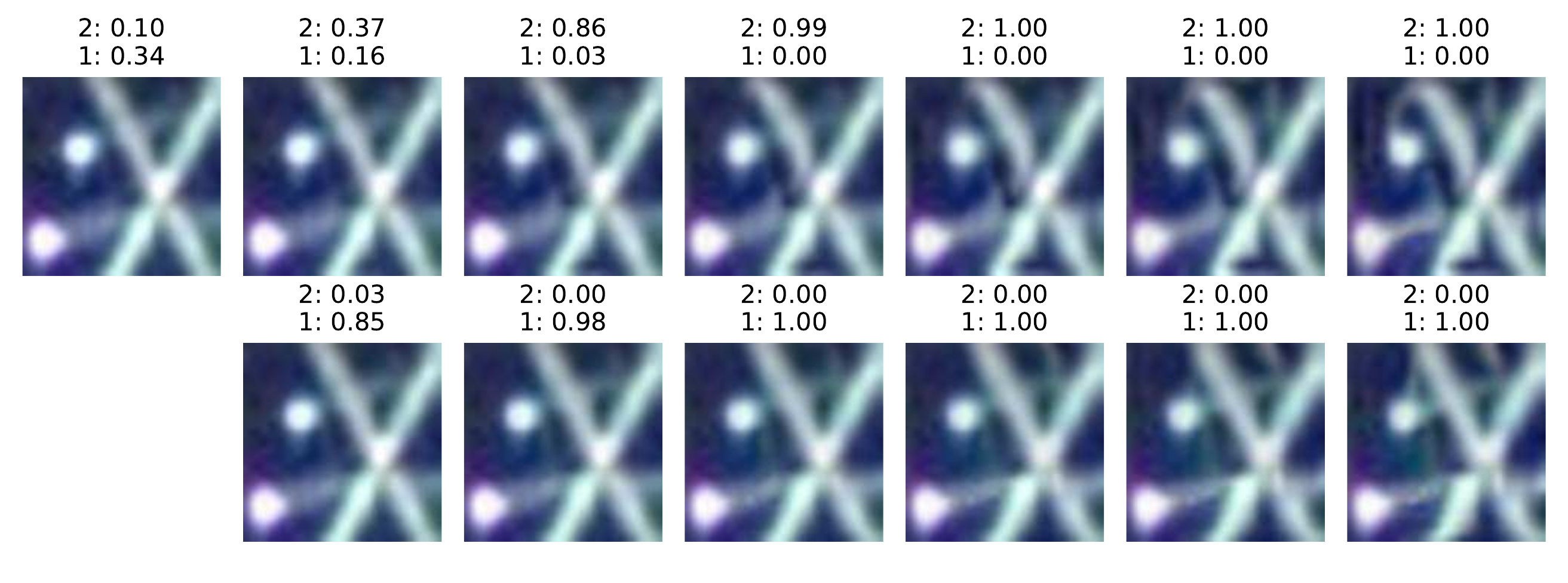}} \\
\end{tabular}	
\caption{\textbf{Visual Counterfactual - Failure Case for SVHN:}
On few SVHN test set examples, neither of the models is able to generate proper digits but the predicted confidence rises even for small radii without the appearance of class specific features. Note that RATIO is the only model able to generate meaningful features for larger radii, \eg the bow of the 2. 
}
\vspace{-.3cm}
\end{figure}

%

%% file: res/appendix_failure_od_svhn.tex
\begin{figure}[ht!]
\begin{tabular}{p{1cm}x{\breite}x{\breite}x{\breite}x{\breite}x{\breite}x{\breite}x{\breite}x{\breite}}
Model  & Orig. & $\epsilon=0.5$ & $\epsilon=1.0$ & $\epsilon=1.5$ & $\epsilon=2.0$ & $\epsilon=2.5$ & $\epsilon=3.0$\\
\begin{turn}{90} \hspace{-.4cm} ACET \end{turn} & \multicolumn{7}{c}{\includegraphics[width=0.91\textwidth,valign=c]{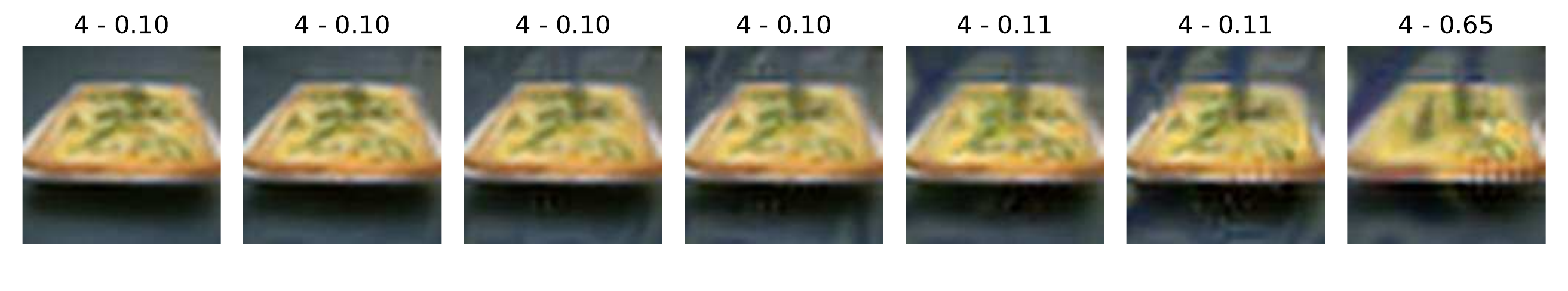}} \\
\hline
\begin{turn}{90} \hspace{-.4cm} AT-0.50 \end{turn}  &  \multicolumn{7}{c}{\includegraphics[width=0.91\textwidth,valign=c]{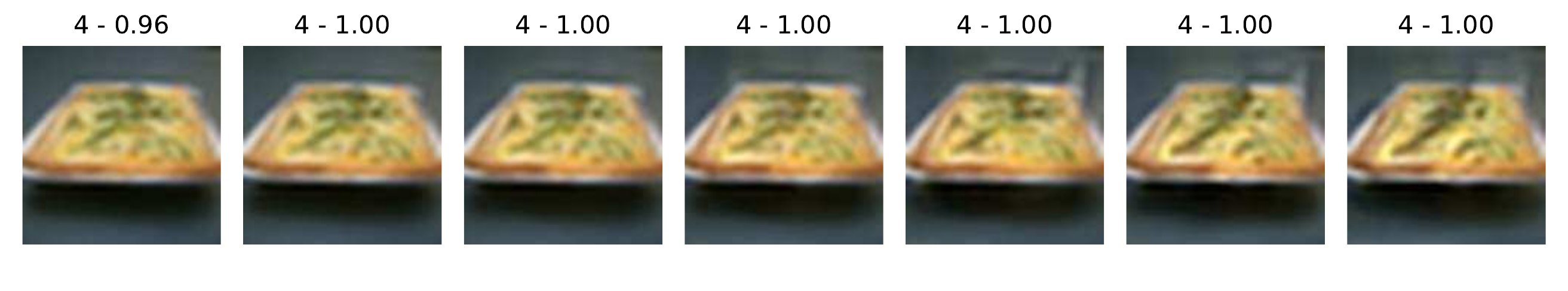}} \\
\hline
\begin{turn}{90} \hspace{-.4cm} AT-0.25 \end{turn} & \multicolumn{7}{c}{\includegraphics[width=0.91\textwidth,valign=c]{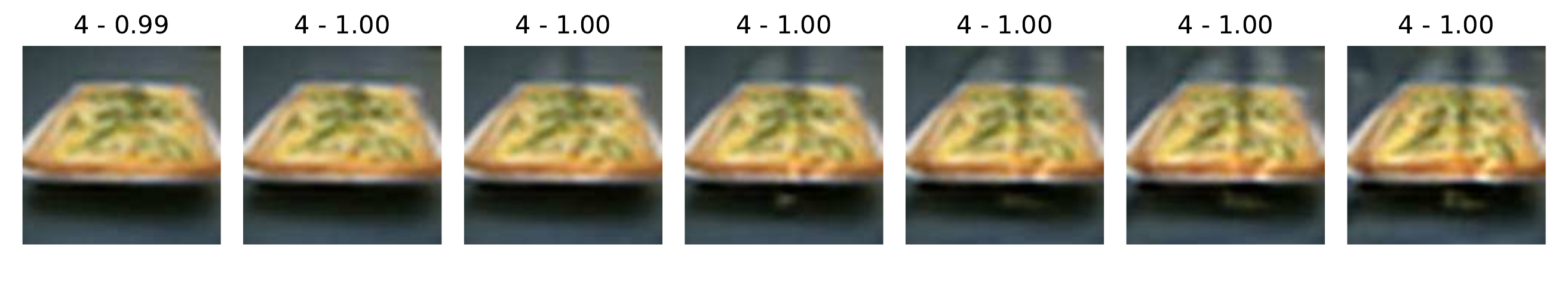}} \\
\hline
\begin{turn}{90} \hspace{-.4cm} R-0.25 \end{turn}  &  \multicolumn{7}{c}{\includegraphics[width=0.91\textwidth,valign=c]{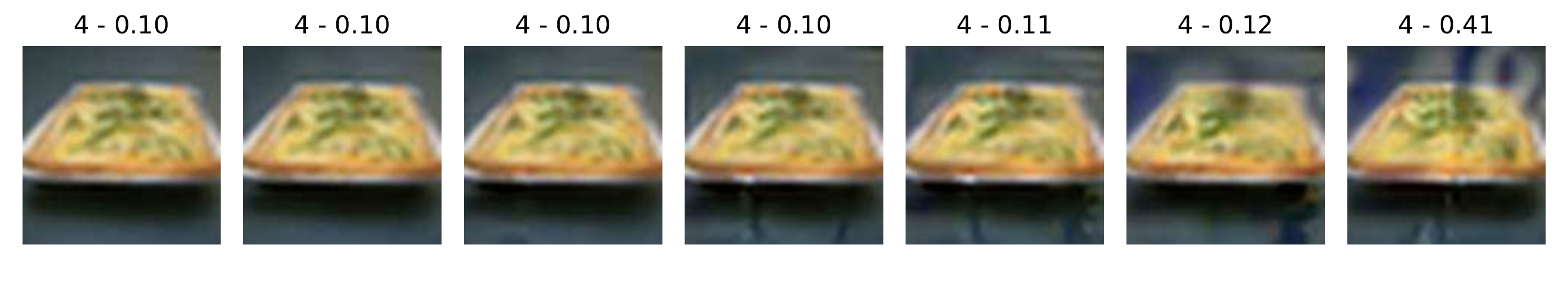}} \\
\end{tabular}	
\vspace{-.5cm}
\caption{\label{fig:od_svhn_fail1}\textbf{Feature Generation on OOD images - Failure Case SVHN:} On some OOD samples, the radius of the threat model is not large enough to produce proper digits. However ACET and RATIO are the only models assigning low confidences to the $\epsilon=3$ samples.}
\vspace{2mm}
\begin{tabular}{p{1cm}x{\breite}x{\breite}x{\breite}x{\breite}x{\breite}x{\breite}x{\breite}x{\breite}}
\begin{turn}{90} \hspace{-.4cm} ACET \end{turn} & \multicolumn{7}{c}{\includegraphics[width=0.91\textwidth,valign=c]{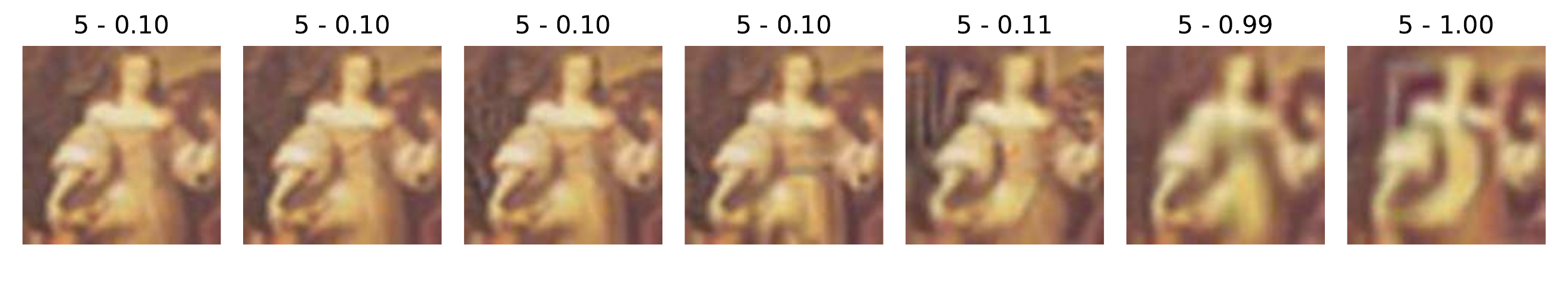}} \\
\hline
\begin{turn}{90} \hspace{-.4cm} AT-0.50 \end{turn}  &  \multicolumn{7}{c}{\includegraphics[width=0.91\textwidth,valign=c]{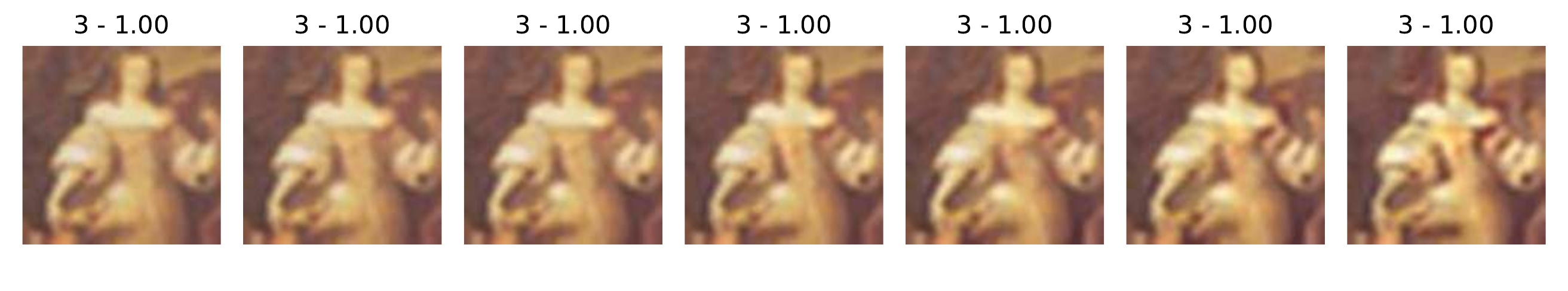}} \\
\hline
\begin{turn}{90} \hspace{-.4cm} AT-0.25 \end{turn} & \multicolumn{7}{c}{\includegraphics[width=0.91\textwidth,valign=c]{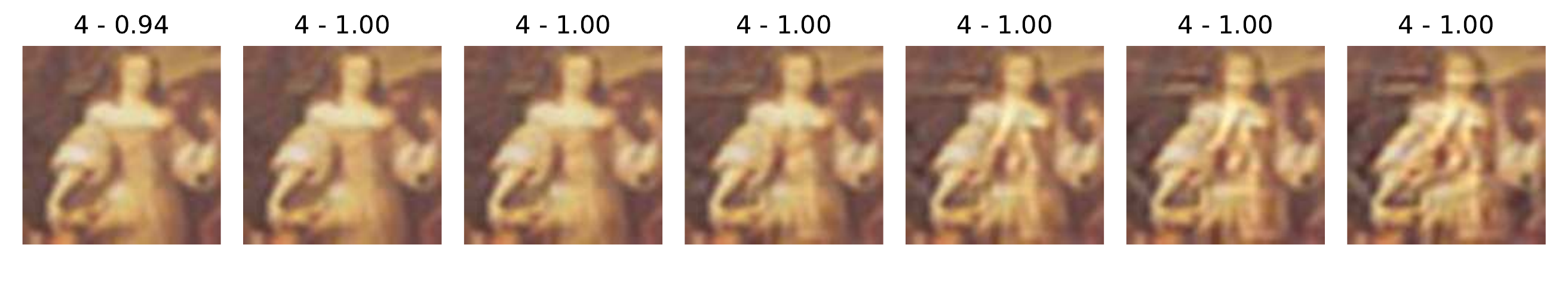}} \\
\hline
\begin{turn}{90} \hspace{-.4cm} R-0.25 \end{turn}  &  \multicolumn{7}{c}{\includegraphics[width=0.91\textwidth,valign=c]{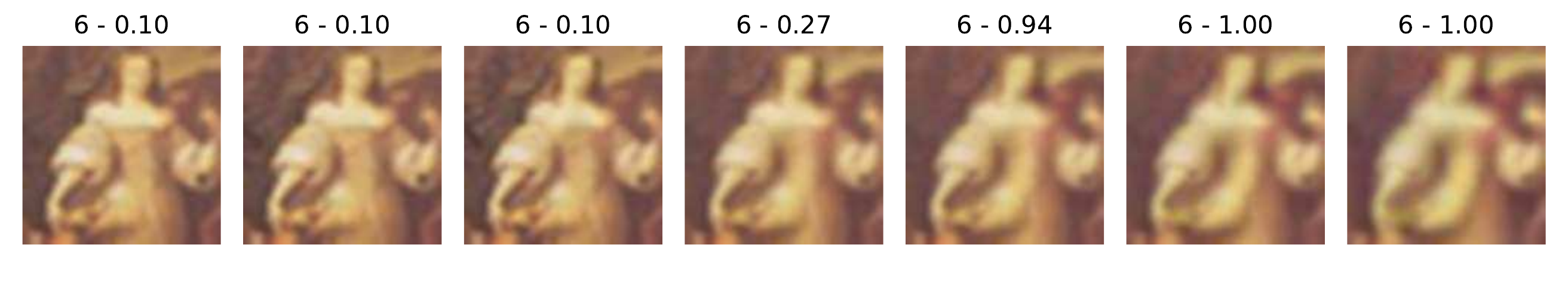}} \\
\end{tabular}	
\vspace{-.5cm}
\caption{\label{fig:od_svhn_fail2}\textbf{Feature Generation on OOD images - Failure Case SVHN:} On this OOD sample, all models assign very high confidence to the final image even though all but the ACET images clearly lack strong digit-like features.}
\end{figure}

%% file: res/appendix_failure_vc_cifar100.tex
\begin{figure}[ht!]
\begin{tabular}{p{1cm}x{\breite}x{\breite}x{\breite}x{\breite}x{\breite}x{\breite}x{\breite}x{\breite}}
Model  & Orig. & $\epsilon=0.5$ & $\epsilon=1.0$ & $\epsilon=1.5$ & $\epsilon=2.0$ & $\epsilon=2.5$ & $\epsilon=3.0$\\
\begin{turn}{90} \hspace{-.4cm} AT-0.50 \end{turn}  &  \multicolumn{7}{c}{\includegraphics[width=0.91\textwidth,valign=c]{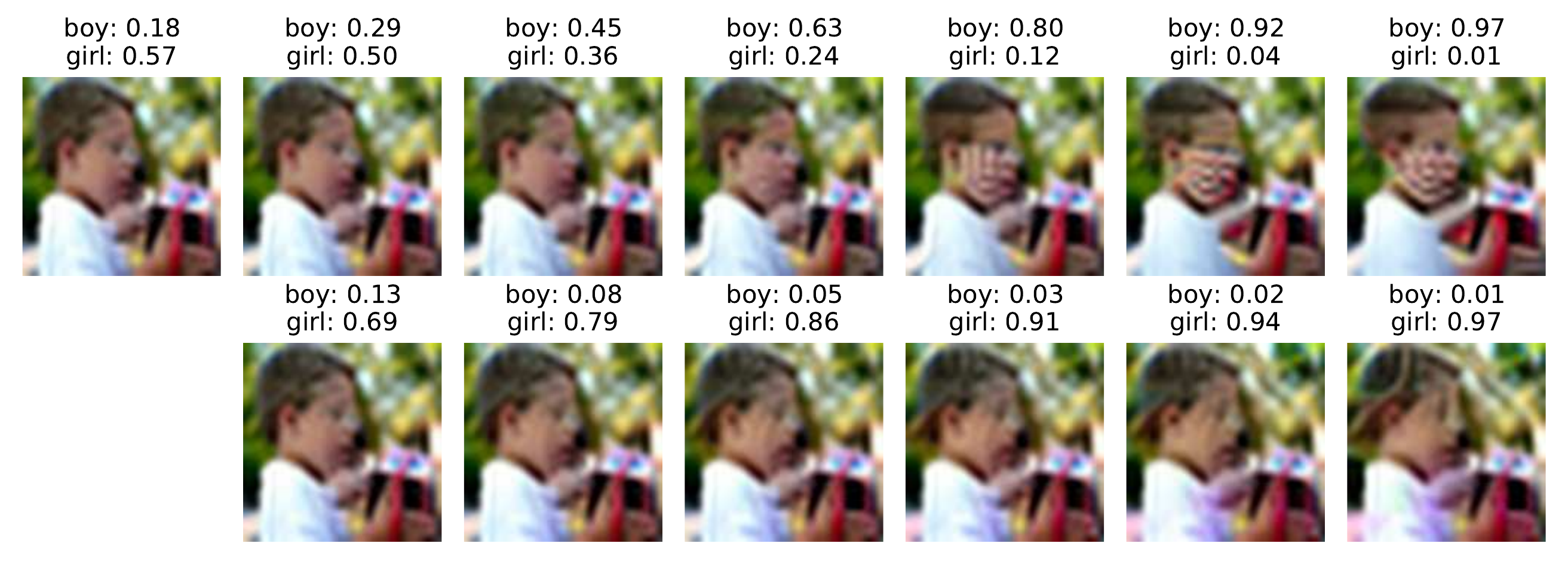}} \\
\hline
\begin{turn}{90} \hspace{-.4cm} AT-0.25 \end{turn}  &  \multicolumn{7}{c}{\includegraphics[width=0.91\textwidth,valign=c]{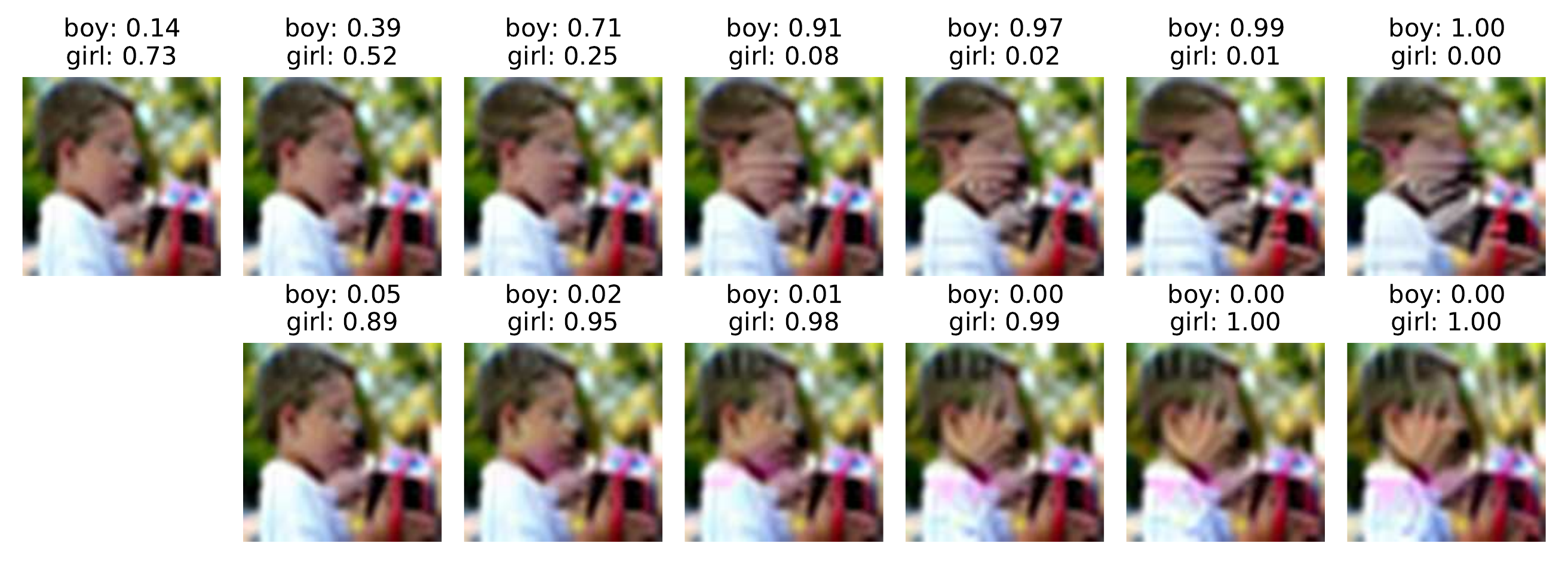}} \\
\hline
\begin{turn}{90} \hspace{-.9cm} RATIO-0.50 \end{turn} & \multicolumn{7}{c}{\includegraphics[width=0.91\textwidth,valign=c]{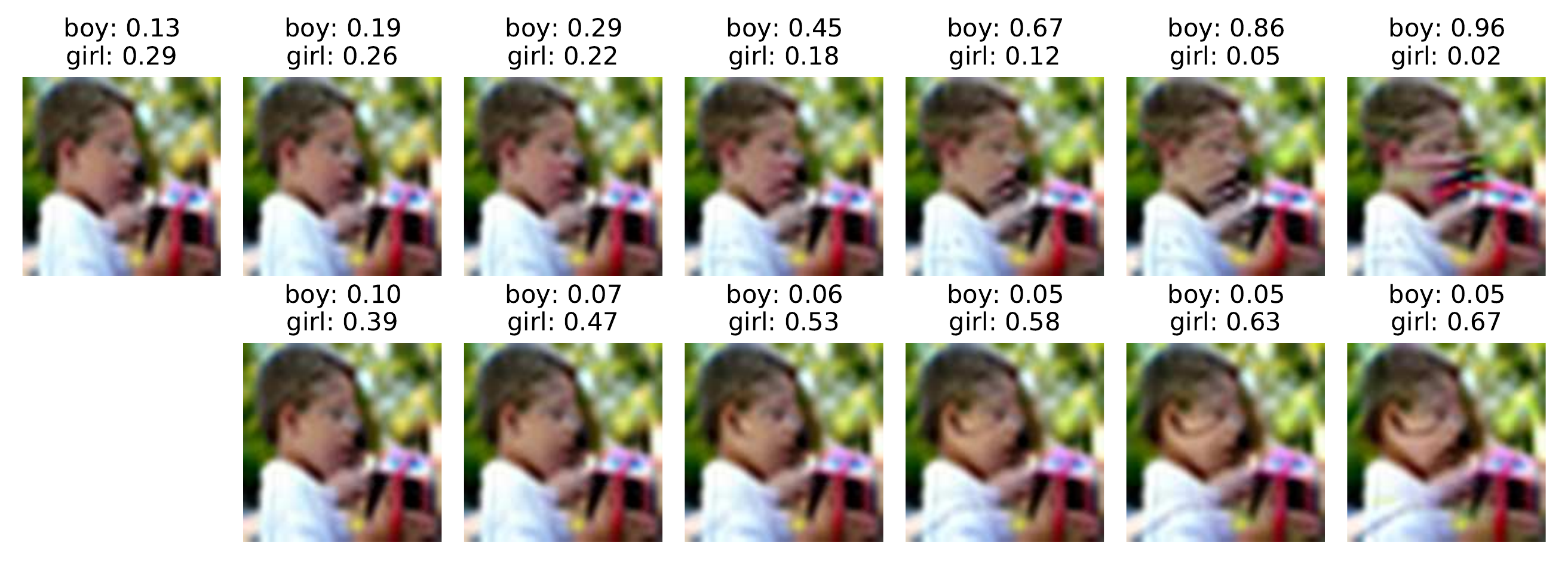}} \\
\hline
\begin{turn}{90} \hspace{-.9cm} RATIO-0.25 \end{turn} & \multicolumn{7}{c}{\includegraphics[width=0.91\textwidth,valign=c]{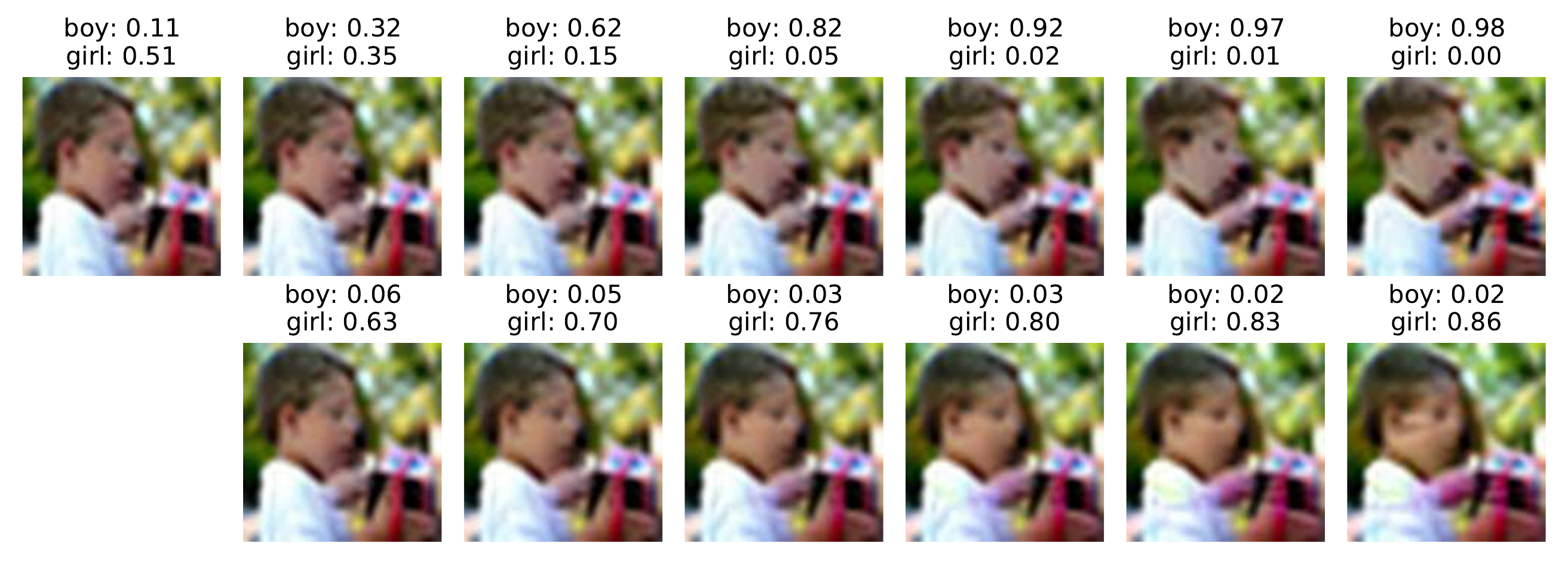}} \\
\end{tabular}	
\caption{\label{fig:vc_cifar100_failure1}
\textbf{Visual Counterfactual - Failure Case for CIFAR100:}
Especially for images from the classes "girl", "woman", "boy" and "man", visual counterfactuals often yield little insight into the model behaviour as they introduce meaningless distortions. }
\end{figure}

%% file: res/appendix_failure_od_cifar100.tex
\begin{figure}[ht!]
\begin{tabular}{p{1cm}x{\breite}x{\breite}x{\breite}x{\breite}x{\breite}x{\breite}x{\breite}x{\breite}}
Model  & Orig. & $\epsilon=0.5$ & $\epsilon=1.0$ & $\epsilon=1.5$ & $\epsilon=2.0$ & $\epsilon=2.5$ & $\epsilon=3.0$\\  
\begin{turn}{90} \hspace{-.4cm} AT-0.50 \end{turn}  &  \multicolumn{7}{c}{\includegraphics[width=0.91\textwidth,valign=c]{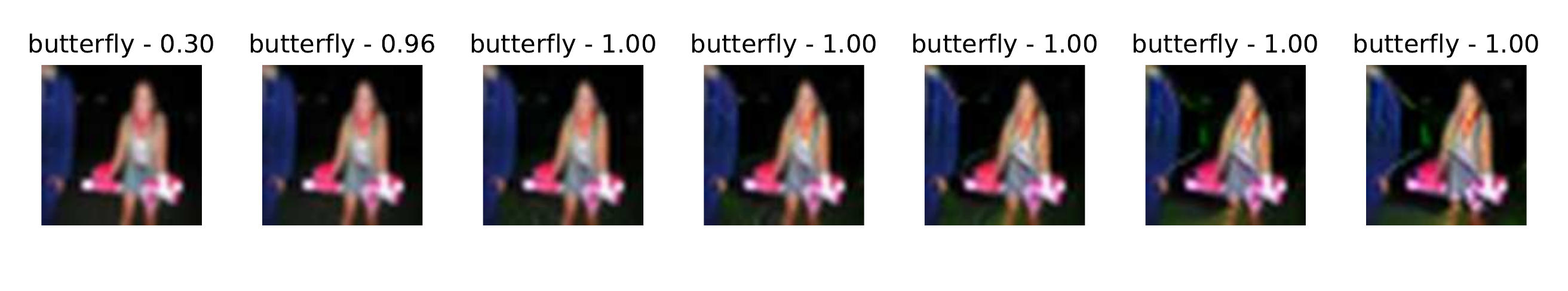}} \\
\hline
\begin{turn}{90} \hspace{-.4cm} AT-0.25 \end{turn} & \multicolumn{7}{c}{\includegraphics[width=0.91\textwidth,valign=c]{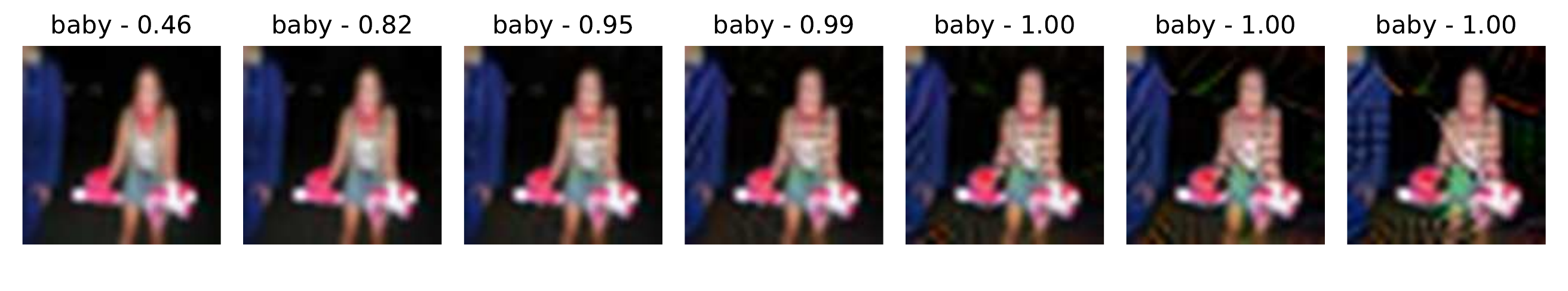}} \\
\hline
\begin{turn}{90} \hspace{-.4cm} R-0.50 \end{turn} & \multicolumn{7}{c}{\includegraphics[width=0.91\textwidth,valign=c]{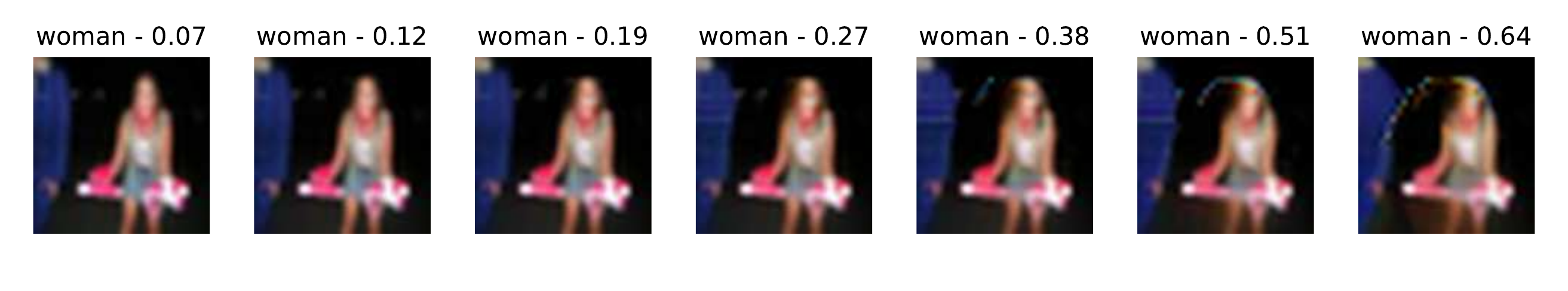}} \\
\hline
\begin{turn}{90} \hspace{-.4cm} R-0.25 \end{turn}  &  \multicolumn{7}{c}{\includegraphics[width=0.91\textwidth,valign=c]{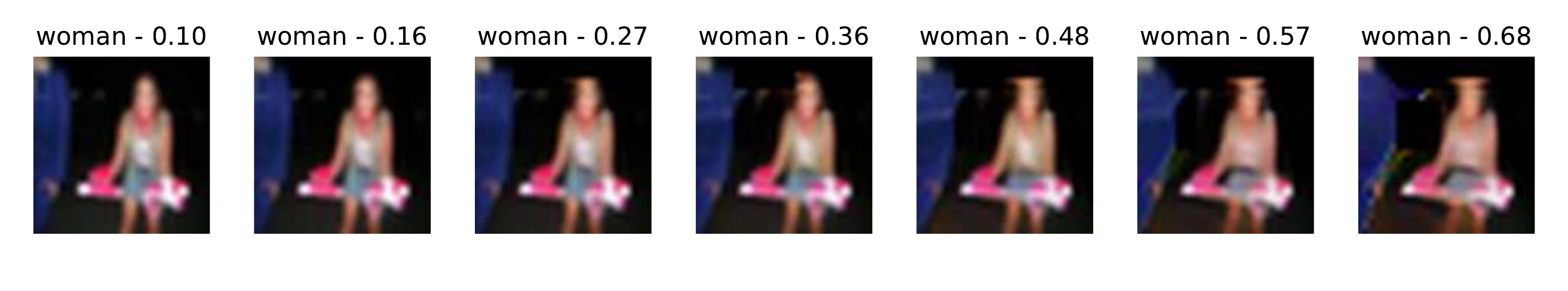}} \\
\end{tabular}	
\vspace{-.5cm}
\caption{\label{fig:od_cifar100_fail1}\textbf{Feature Generation on OOD images - Failure Case CIFAR100:} The 80M Tiny Images dataset contains a lot of images depicting humans. In a lot of cases, the model distorts the images and creates a strong uncanny valley-feeling. }

\begin{tabular}{p{1cm}x{\breite}x{\breite}x{\breite}x{\breite}x{\breite}x{\breite}x{\breite}x{\breite}}
\begin{turn}{90} \hspace{-.4cm} AT-0.50 \end{turn}  &  \multicolumn{7}{c}{\includegraphics[width=0.91\textwidth,valign=c]{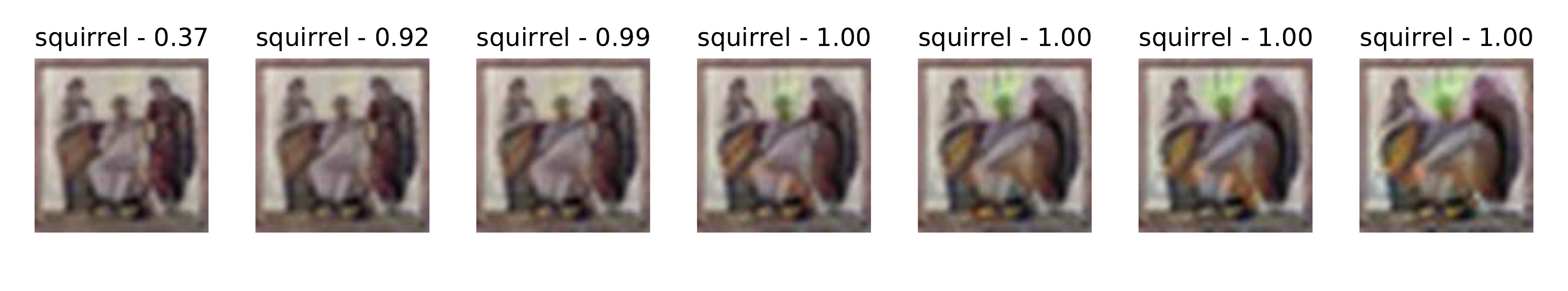}} \\
\hline
\begin{turn}{90} \hspace{-.4cm} AT-0.25 \end{turn} & \multicolumn{7}{c}{\includegraphics[width=0.91\textwidth,valign=c]{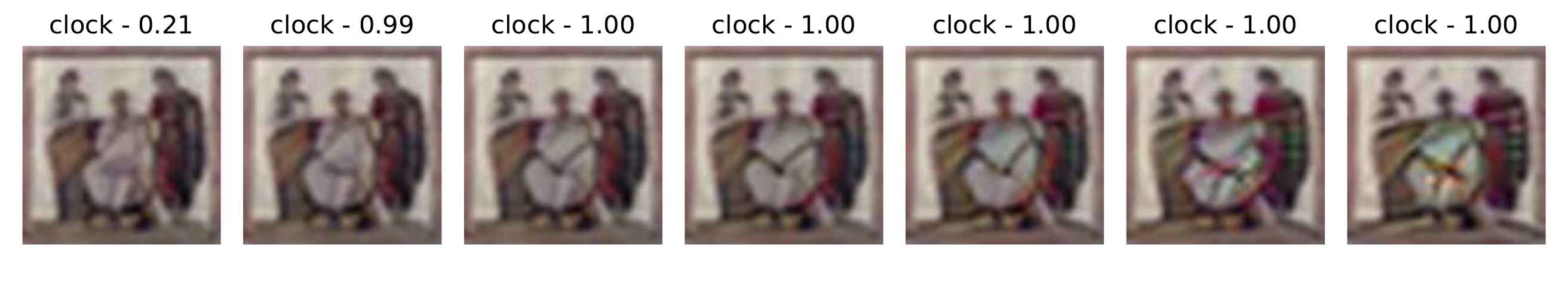}} \\
\hline
\begin{turn}{90} \hspace{-.4cm} R-0.50 \end{turn} & \multicolumn{7}{c}{\includegraphics[width=0.91\textwidth,valign=c]{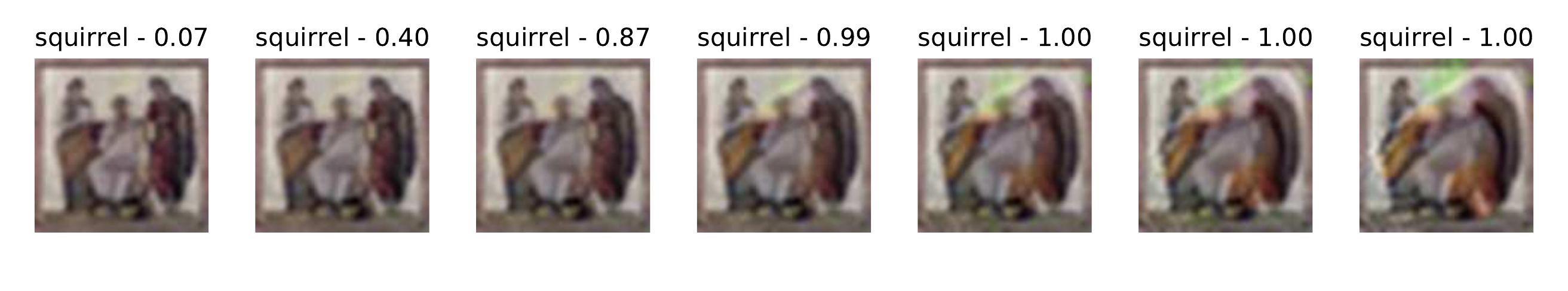}} \\
\hline
\begin{turn}{90} \hspace{-.4cm} R-0.25 \end{turn}  &  \multicolumn{7}{c}{\includegraphics[width=0.91\textwidth,valign=c]{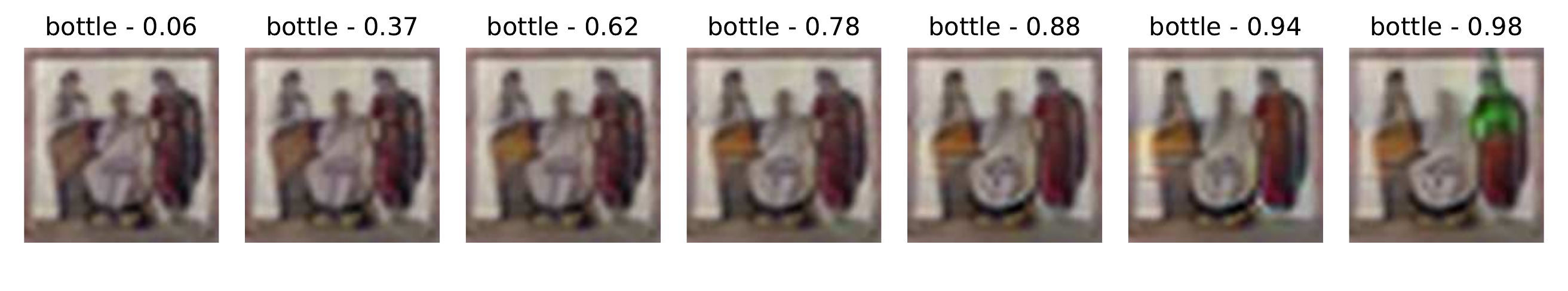}} \\
\end{tabular}	
\vspace{-.5cm}
\caption{\label{fig:od_cifar100_fail2}\textbf{Feature Generation on OOD images - Failure Case CIFAR100:} The image quality often is dependent on the highest scoring class in the original OD-image. While R-0.25 transforms the painting into three bottles, there are hardly any squirrel features visible for R-0.50 and AT-0.50, even though the confidence reaches 1.0. }
\end{figure}

%% file: res/appendix_failure_vc_restricted.tex
\begin{figure}[ht!]
\begin{tabular}{p{1cm}x{\breite}x{\breite}x{\breite}x{\breite}x{\breite}x{\breite}x{\breite}x{\breite}}
Model  & Orig. & $\epsilon=3.5$ & $\epsilon=7.0$ & $\epsilon=10.5$ & $\epsilon=14.0$ & $\epsilon=17.5$ & $\epsilon=21.0$\\
\begin{turn}{90} \hspace{-1.2cm} Madry AT-3.50 \end{turn} & \multicolumn{7}{c}{\includegraphics[width=0.91\textwidth,valign=c]{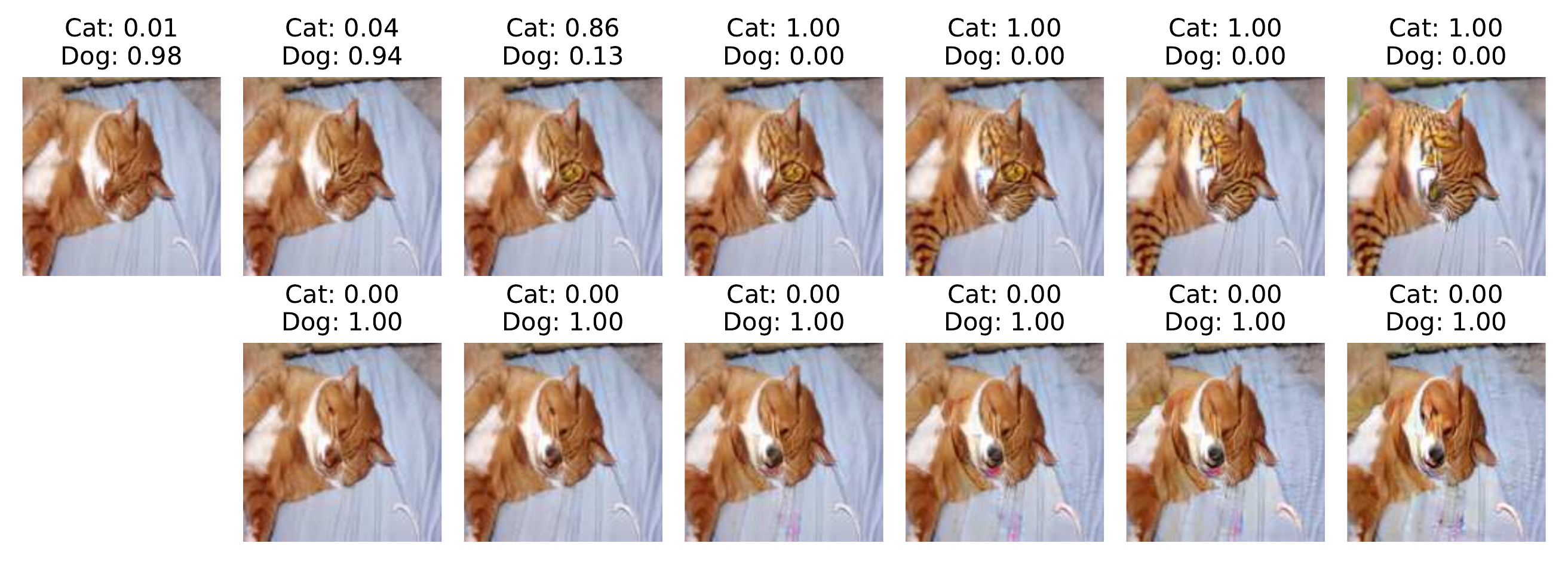}} \\
\hline
\begin{turn}{90} \hspace{-.4cm} AT-3.50 \end{turn}  &  \multicolumn{7}{c}{\includegraphics[width=0.91\textwidth,valign=c]{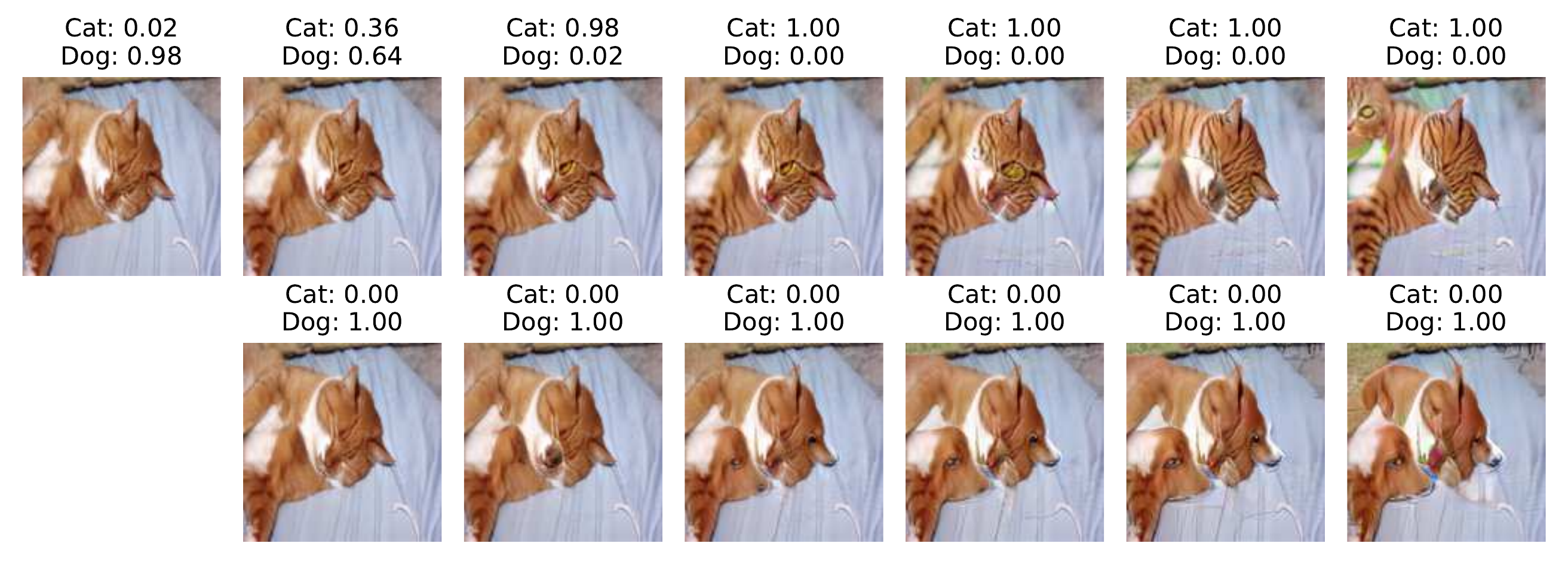}} \\
\hline
\begin{turn}{90} \hspace{-.9cm} RATIO-3.50 \end{turn} & \multicolumn{7}{c}{\includegraphics[width=0.91\textwidth,valign=c]{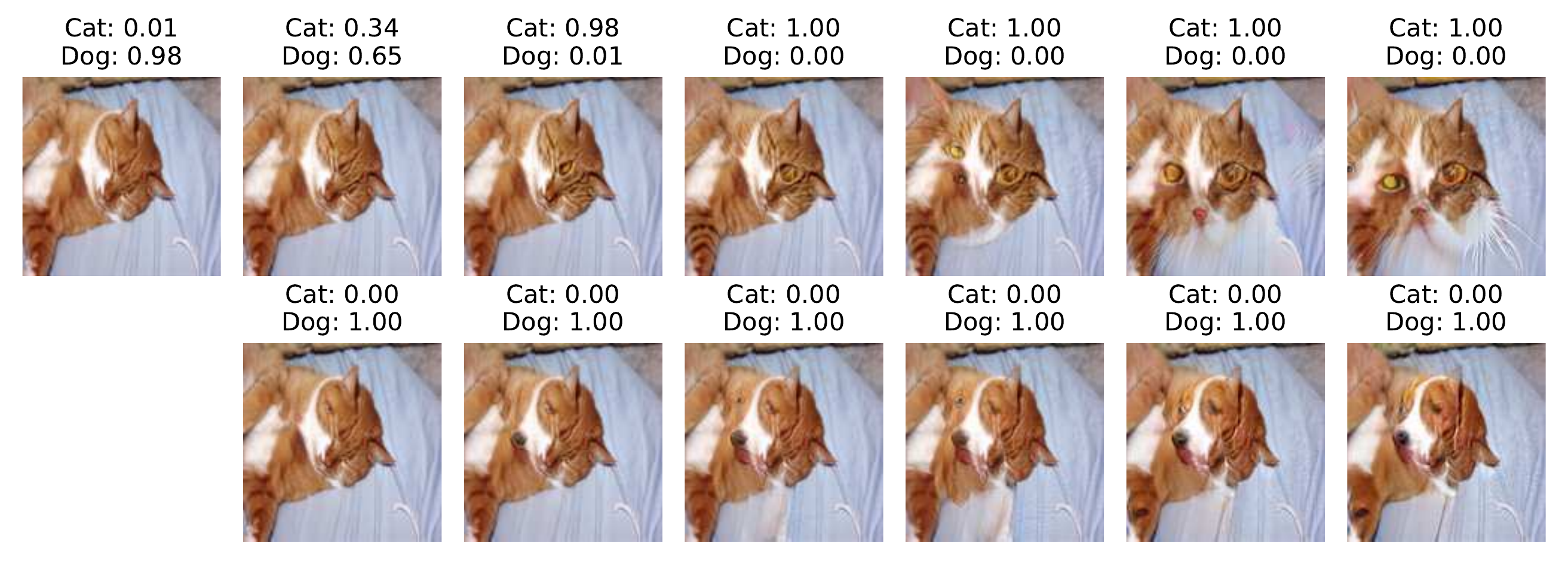}} \\
\hline
\begin{turn}{90} \hspace{-.9cm} RATIO-1.75 \end{turn} & \multicolumn{7}{c}{\includegraphics[width=0.91\textwidth,valign=c]{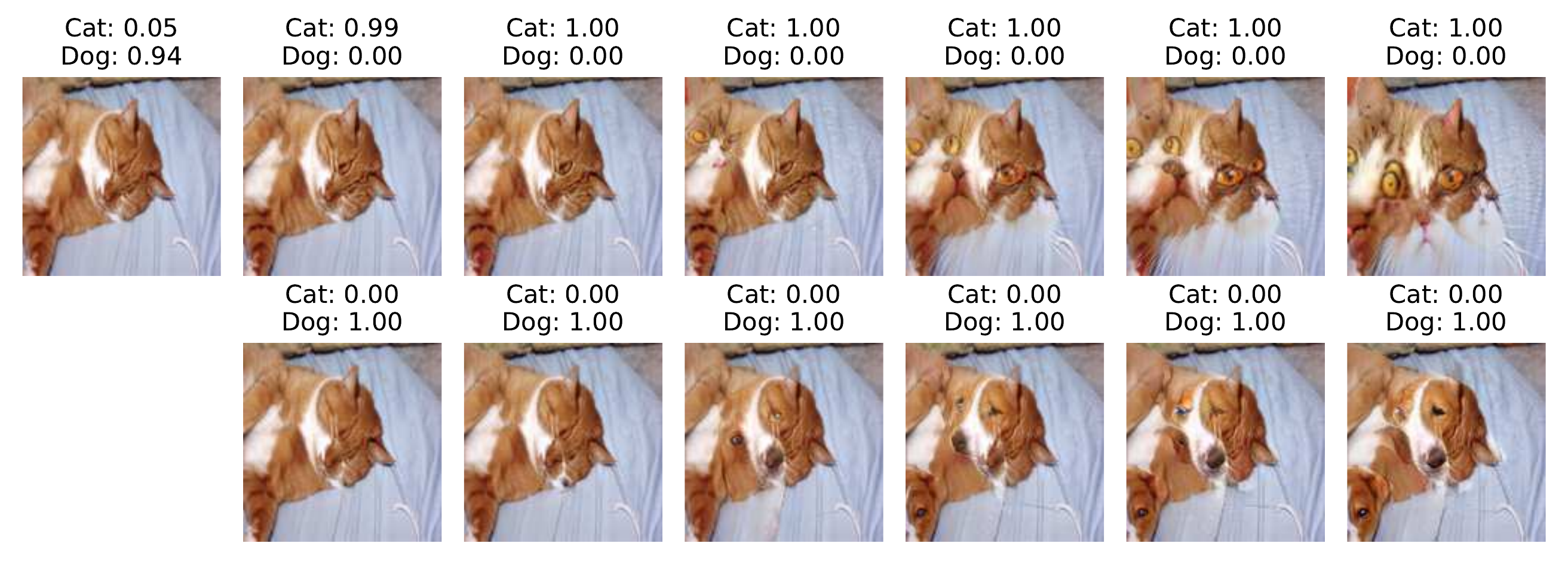}} \\
\end{tabular}	
\caption{\label{fig:vc_imagenet_failure}\textbf{Visual Counterfactuals  - Failure Case for r. ImageNet:} Unusual object locations lead to multiple overlaying
counterfactuals which do not lead to meaningful natural images.
}
\end{figure}

%% file: res/appendix_failure_od_restricted.tex
\begin{figure}[ht!]
\begin{tabular}{p{1cm}x{\breite}x{\breite}x{\breite}x{\breite}x{\breite}x{\breite}x{\breite}x{\breite}}
Model  & Orig.& $\epsilon=3.5$ & $\epsilon=7.0$ & $\epsilon=10.5$ & $\epsilon=14.0$ & $\epsilon=17.5$ & $\epsilon=21.0$\\  
\begin{turn}{90} \hspace{-.9cm} M. AT-3.50 \end{turn}  &  \multicolumn{7}{c}{\includegraphics[width=0.91\textwidth,valign=c]{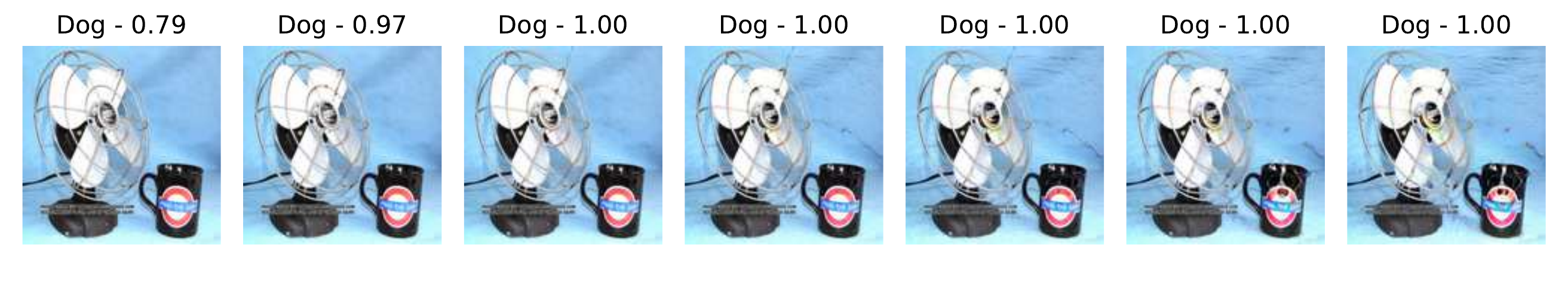}} \\
\hline
\begin{turn}{90} \hspace{-.5cm} AT-3.50 \end{turn}  &  \multicolumn{7}{c}{\includegraphics[width=0.91\textwidth,valign=c]{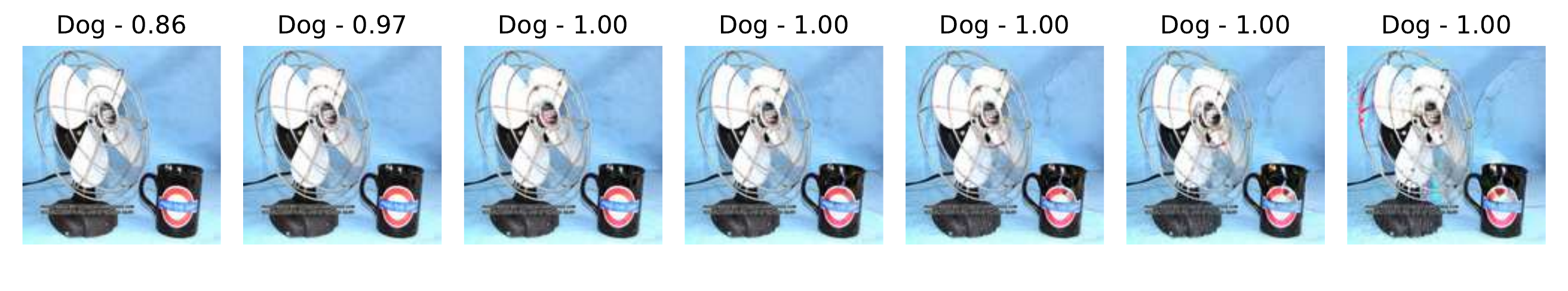}} \\
\hline
\begin{turn}{90} \hspace{-.4cm} R-3.50 \end{turn} & \multicolumn{7}{c}{\includegraphics[width=0.91\textwidth,valign=c]{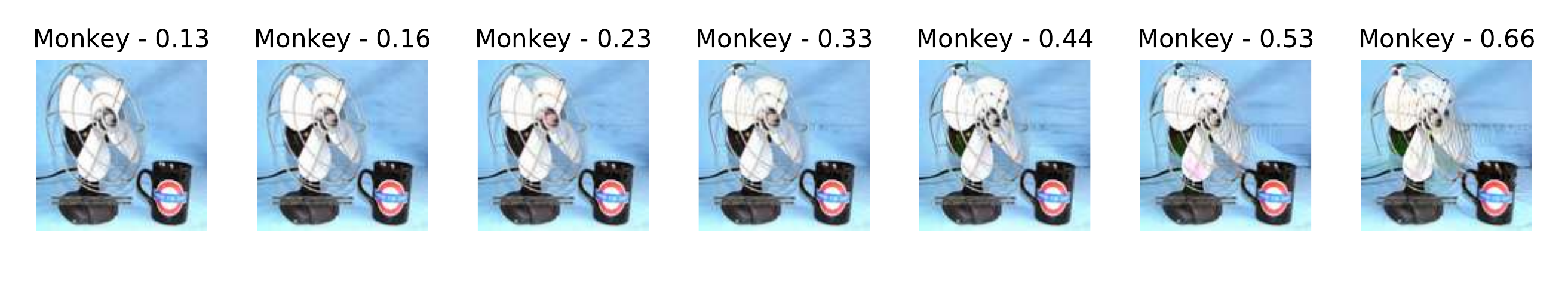}} \\
\hline
\begin{turn}{90} \hspace{-.4cm} R-1.75 \end{turn}  &  \multicolumn{7}{c}{\includegraphics[width=0.91\textwidth,valign=c]{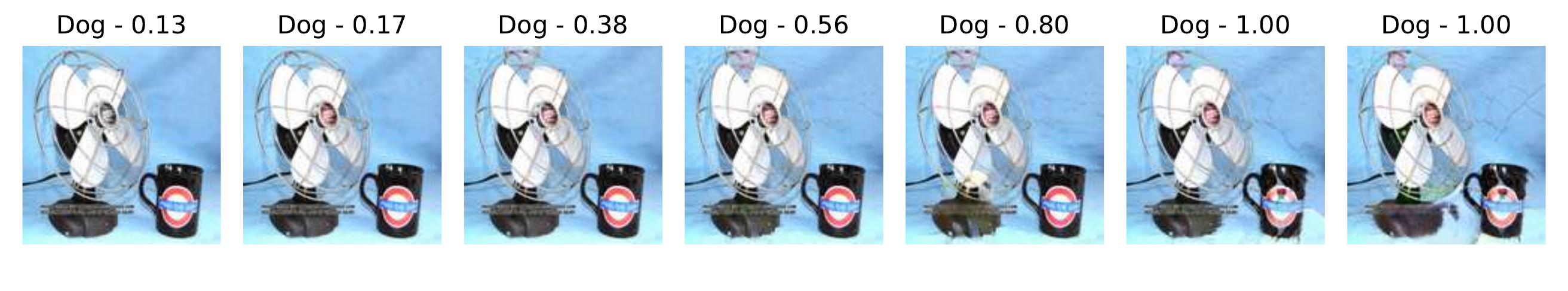}} \\
\end{tabular}	
\vspace{-.5cm}
\caption{\label{fig:od_cifar_fail1}\textbf{Feature Generation on OOD images - Failure Case r. ImageNet:} Overly confident predictions even though little class-specific features have appeared. }

\begin{tabular}{p{1cm}x{\breite}x{\breite}x{\breite}x{\breite}x{\breite}x{\breite}x{\breite}x{\breite}}
\begin{turn}{90} \hspace{-.9cm} M. AT-3.50 \end{turn}  &  \multicolumn{7}{c}{\includegraphics[width=0.91\textwidth,valign=c]{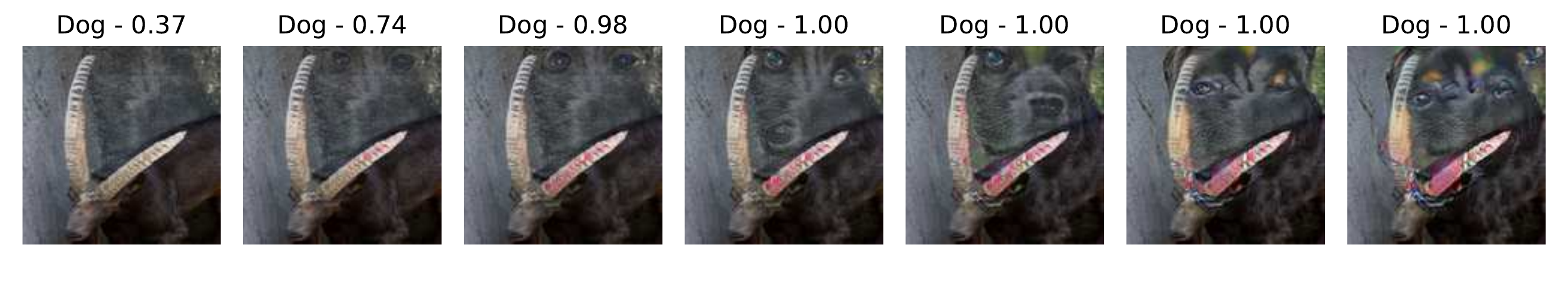}} \\
\hline
\begin{turn}{90} \hspace{-.5cm} AT-3.50 \end{turn}  &  \multicolumn{7}{c}{\includegraphics[width=0.91\textwidth,valign=c]{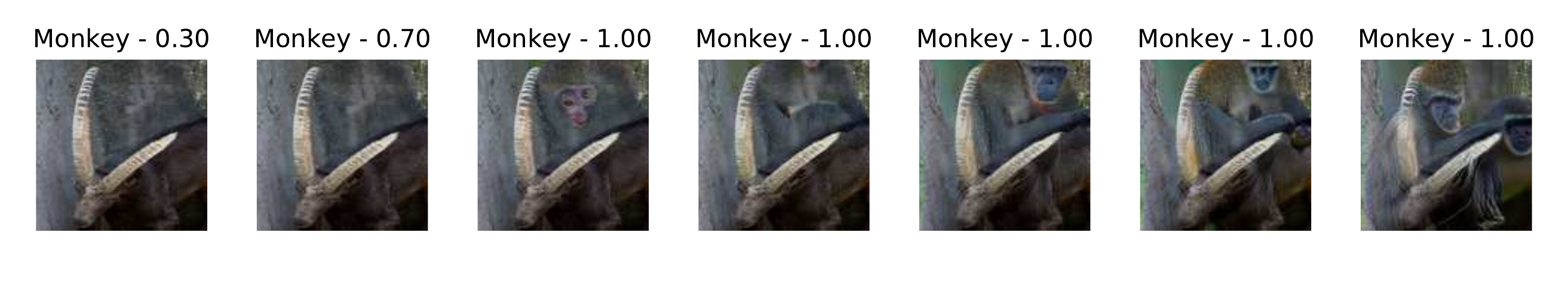}} \\
\hline
\begin{turn}{90} \hspace{-.4cm} R-3.50 \end{turn} & \multicolumn{7}{c}{\includegraphics[width=0.91\textwidth,valign=c]{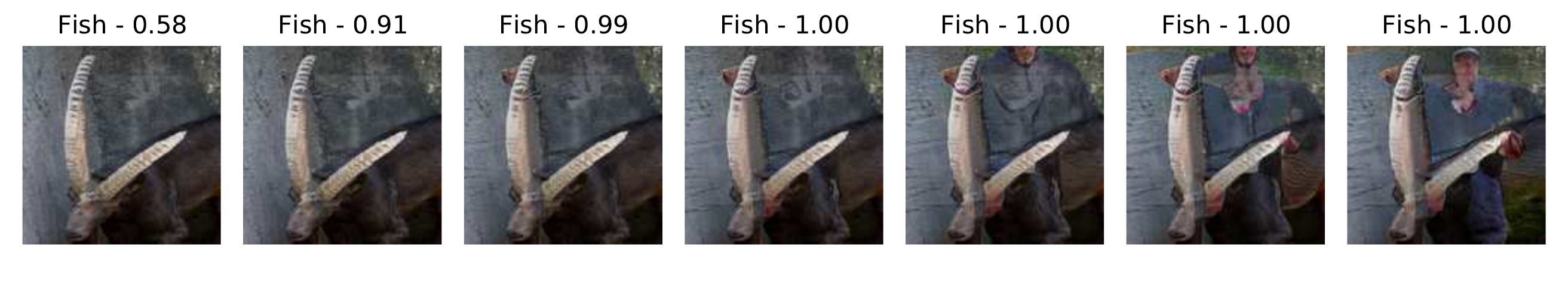}} \\
\hline
\begin{turn}{90} \hspace{-.4cm} R-1.75 \end{turn}  &  \multicolumn{7}{c}{\includegraphics[width=0.91\textwidth,valign=c]{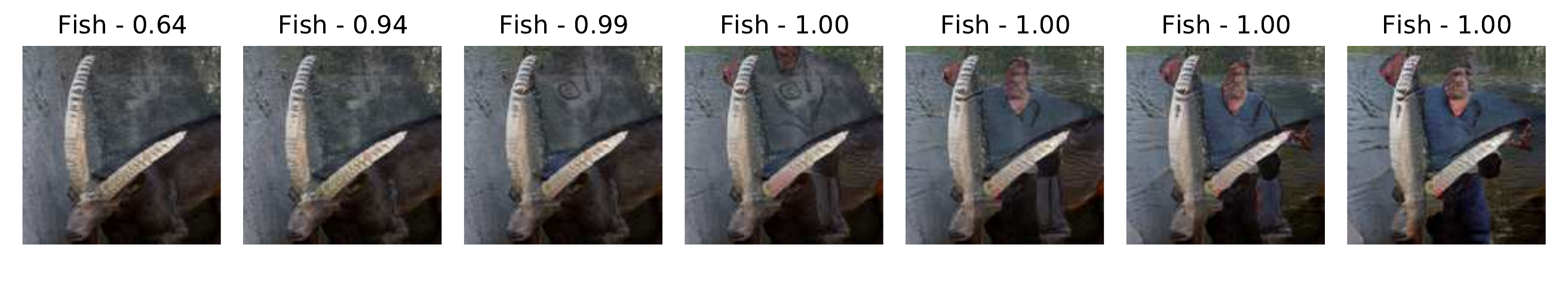}} \\
\end{tabular}	
\vspace{-.5cm}
\caption{\label{fig:od_imagenet_failure}\textbf{Feature Generation on OOD images - Failure Case r. ImageNet:} For the RATIO models which have target fish a human appears as fish training images often show the angler together with the haul. }
\end{figure}

%% file: 5331.bbl
\begin{thebibliography}{10}
\providecommand{\url}[1]{\texttt{#1}}
\providecommand{\urlprefix}{URL }
\providecommand{\doi}[1]{https://doi.org/#1}

\bibitem{AlaEtAl19}
Alayrac, J.B., Uesato, J., Huang, P.S., Fawzi, A., Stanforth, R., Kohli, P.:
  Are labels required for improving adversarial robustness? In: NeurIPS (2019)

\bibitem{andriushchenko2019square}
Andriushchenko, M., Croce, F., Flammarion, N., Hein, M.: Square attack: a
  query-efficient black-box adversarial attack via random search. In: ECCV
  (2020)

\bibitem{AthEtAl2018}
Athalye, A., Carlini, N., Wagner, D.A.: Obfuscated gradients give a false sense
  of security: Circumventing defenses to adversarial examples. In: ICML (2018)

\bibitem{BacEtAl2015}
Bach, S., Binder, A., Montavon, G., Klauschen, F., M{\"u}ller, K.R., Samek, W.:
  On pixel-wise explanations for non-linear classifier decisions by layer-wise
  relevance propagation. PLoS One  \textbf{10}(7),  e0130140 (2015)

\bibitem{BaeEtAl2010}
Baehrens, D., Schroeter, T., Harmeling, S., Kawanabe, M., Hansen, K.,
  M{\"u}ller, K.R.: How to explain individual classification decisions. Journal
  of Machine Learning Research (JMLR)  \textbf{11},  1803--1831 (2010)

\bibitem{BarSelRag2020}
Barocas, S., Selbst, A.D., Raghavan, M.: The hidden assumptions behind
  counterfactual explanations and principal reasons. In: FAT (2020)

\bibitem{bitterwolf20provable}
Bitterwolf, J., Meinke, A., Hein, M.: Provable worst case guarantees for the
  detection of out-of-distribution data. arXiv:2007.08473  (2020)

\bibitem{CarWag2017}
Carlini, N., Wagner, D.: Adversarial examples are not easily detected:
  Bypassing ten detection methods. In: ACM Workshop on Artificial Intelligence
  and Security (2017)

\bibitem{carlini16towards}
Carlini, N., Wagner, D.: Towards evaluating the robustness of neural networks.
  In: 2017 ieee symposium on security and privacy (sp). pp. 39--57. IEEE (2017)

\bibitem{CarEtAl19}
Carmon, Y., Raghunathan, A., Schmidt, L., Duchi, J.C., Liang, P.S.: Unlabeled
  data improves adversarial robustness. In: NeurIPS (2019)

\bibitem{ChaEtAl2019}
Chang, C.H., Creager, E., Goldenberg, A., Duvenaud, D.: Explaining image
  classifiers by counterfactual generation. In: ICLR (2019)

\bibitem{CohenARXIV2019}
Cohen, J.M., Rosenfeld, E., Kolter, J.Z.: Certified adversarial robustness via
  randomized smoothing. In: NeurIPS (2019)

\bibitem{CroEtAl2018}
Croce, F., Andriushchenko, M., Hein, M.: Provable robustness of relu networks
  via maximization of linear regions. In: AISTATS (2019)

\bibitem{CroHei2020}
Croce, F., Hein, M.: Reliable evaluation of adversarial robustness with an
  ensemble of diverse parameter-free attacks. In: ICML (2020)

\bibitem{croce2019minimally}
Croce, F., Hein, M.: Minimally distorted adversarial examples with a fast
  adaptive boundary attack. In: ICML (2020)

\bibitem{cubuk18autoaugment}
Cubuk, E.D., Zoph, B., Mane, D., Vasudevan, V., Le, Q.V.: Autoaugment: Learning
  augmentation strategies from data. In: Proceedings of the IEEE conference on
  computer vision and pattern recognition. pp. 113--123 (2019)

\bibitem{devries17cutout}
DeVries, T., Taylor, G.W.: Improved regularization of convolutional neural
  networks with cutout. arXiv preprint arXiv:1708.04552  (2017)

\bibitem{DongEtAl2017}
Dong, Y., Su, H., Zhu, J., Bao, F.: Towards interpretable deep neural networks
  by leveraging adversarial examples (2017), arXiv preprint, arXiv:1708.05493

\bibitem{robustness}
Engstrom, L., Ilyas, A., Santurkar, S., Tsipras, D.: Robustness (python
  library) (2019), \url{https://github.com/MadryLab/robustness}

\bibitem{GowEtAl18}
Gowal, S., Dvijotham, K., Stanforth, R., Bunel, R., Qin, C., Uesato, J.,
  Arandjelovic, R., Mann, T.A., Kohli, P.: On the effectiveness of interval
  bound propagation for training verifiably robust models (2018), preprint,
  arXiv:1810.12715v3

\bibitem{GoyEtAl2019}
Goyal, Y., Wu, Z., Ernst, J., Batra, D., Parikh, D., Lee, S.: Counterfactual
  visual explanations. In: ICML (2019)

\bibitem{grathwohl2019your}
Grathwohl, W., Wang, K.C., Jacobsen, J.H., Duvenaud, D., Norouzi, M., Swersky,
  K.: Your classifier is secretly an energy based model and you should treat it
  like one. ICLR  (2020)

\bibitem{GuoEtAl2017}
Guo, C., Pleiss, G., Sun, Y., Weinberger, K.: On calibration of modern neural
  networks. In: ICML (2017)

\bibitem{HeiAnd2017}
Hein, M., Andriushchenko, M.: Formal guarantees on the robustness of a
  classifier against adversarial manipulation. In: NeurIPS (2017)

\bibitem{HeiAndBit2019}
Hein, M., Andriushchenko, M., Bitterwolf, J.: Why {ReLU} networks yield
  high-confidence predictions far away from the training data and how to
  mitigate the problem. In: CVPR (2019)

\bibitem{HenEtAl2016}
Hendricks, L.A., Akata, Z., Rohrbach, M., Donahue, J., Schiele, B., Darrell,
  T.: Generating visual explanations. In: ECCV (2016)

\bibitem{HenEtAl2018}
Hendricks, L.A., Hu, R., Darrell, T., Akata, Z.: Grounding visual explanations.
  In: ECCV (2018)

\bibitem{HenGim2017}
Hendrycks, D., Gimpel, K.: A baseline for detecting misclassified and
  out-of-distribution examples in neural networks. In: ICLR (2017)

\bibitem{HenMazDie2019}
Hendrycks, D., Mazeika, M., Dietterich, T.: Deep anomaly detection with outlier
  exposure. In: ICLR (2019)

\bibitem{pmlr-v97-hendrycks19a}
Hendrycks, D., Lee, K., Mazeika, M.: Using pre-training can improve model
  robustness and uncertainty. In: ICML. pp. 2712--2721 (2019)

\bibitem{KatzEtAl2017}
Katz, G., Barrett, C., Dill, D., Julian, K., Kochenderfer, M.: Reluplex: An
  efficient smt solver for verifying deep neural networks. In: CAV (2017)

\bibitem{KrauseStarkDengFei-Fei_3DRR2013}
Krause, J., Stark, M., Deng, J., Fei-Fei, L.: 3d object representations for
  fine-grained categorization. In: 4th International IEEE Workshop on 3D
  Representation and Recognition (3dRR-13). Sydney, Australia (2013)

\bibitem{krizhevsky2009learning}
Krizhevsky, A., Hinton, G.: Learning multiple layers of features from tiny
  images. Tech. rep., Citeseer (2009)

\bibitem{cifar10}
Krizhevsky, A., Hinton, G., et~al.: Learning multiple layers of features from
  tiny images  (2009)

\bibitem{CunBenHin2015}
LeCun, Y., Bengio, Y., Hinton, G.: Deep learning. Nature  \textbf{521} (2015)

\bibitem{lecuyer2018certified}
Lecuyer, M., Atlidakis, V., Geambasu, R., Hsu, D., Jana, S.: Certified
  robustness to adversarial examples with differential privacy. In: IEEE
  Symposium on Security and Privacy (SP) (2019)

\bibitem{LeeEtAl2018}
Lee, K., Lee, H., Lee, K., Shin, J.: Training confidence-calibrated classifiers
  for detecting out-of-distribution samples. In: ICLR (2018)

\bibitem{lee2017cleannet}
Lee, K.H., He, X., Zhang, L., Yang, L.: Cleannet: Transfer learning for
  scalable image classifier training with label noise. In: Proceedings of the
  IEEE Conference on Computer Vision and Pattern Recognition ({CVPR}) (2018)

\bibitem{LeiEtAl2017}
Leibig, C., Allken, V., Ayhan, M.S., Berens, P., Wahl, S.: Leveraging
  uncertainty information from deep neural networks for disease detection.
  Scientific Reports  \textbf{7} (2017)

\bibitem{li2018certified}
Li, B., Chen, C., Wang, W., Carin, L.: Certified adversarial robustness with
  additive noise. In: NeurIPS (2019)

\bibitem{MadEtAl2018}
Madry, A., Makelov, A., Schmidt, L., Tsipras, D., Valdu, A.: Towards deep
  learning models resistant to adversarial attacks. In: ICLR (2018)

\bibitem{maji13fine-grained}
Maji, S., Kannala, J., Rahtu, E., Blaschko, M., Vedaldi, A.: Fine-grained
  visual classification of aircraft. Tech. rep. (2013)

\bibitem{meinke2020towards}
Meinke, A., Hein, M.: Towards neural networks that provably know when they
  don't know. In: ICLR (2020)

\bibitem{Mil2017}
Miller, T.: Explanation in artificial intelligence: Insights from the social
  sciences. Artificial Intelligence  \textbf{267},  1 -- 38 (2019)

\bibitem{MirGehVec2018}
Mirman, M., Gehr, T., Vechev, M.: Differentiable abstract interpretation for
  provably robust neural networks. In: ICML (2018)

\bibitem{MosEtAl18}
Mosbach, M., Andriushchenko, M., Trost, T., Hein, M., Klakow, D.: Logit pairing
  methods can fool gradient-based attacks. In: NeurIPS 2018 Workshop on
  Security in Machine Learning (2018)

\bibitem{najafi2019robustness}
Najafi, A., Maeda, S.i., Koyama, M., Miyato, T.: Robustness to adversarial
  perturbations in learning from incomplete data. In: NeurIPS (2019)

\bibitem{SVHN}
Netzer, Y., Wang, T., Coates, A., Bissacco, A., Wu, B., Ng, A.Y.: Reading
  digits in natural images with unsupervised feature learning. In: NeurIPS
  Workshop on Deep Learning and Unsupervised Feature Learning (2011)

\bibitem{NguYosClu2015}
Nguyen, A., Yosinski, J., Clune, J.: Deep neural networks are easily fooled:
  {H}igh confidence predictions for unrecognizable images. In: CVPR (2015)

\bibitem{Nilsback08}
Nilsback, M.E., Zisserman, A.: Automated flower classification over a large
  number of classes. In: Indian Conference on Computer Vision, Graphics and
  Image Processing (Dec 2008)

\bibitem{ParVit2019}
Álvaro Parafita, Vitrià, J.: Explaining visual models by causal attribution.
  In: ICCV Workshop on XCAI (2019)

\bibitem{recht2018cifar10.1}
Recht, B., Roelofs, R., Schmidt, L., Shankar, V.: Do cifar-10 classifiers
  generalize to cifar-10? (2018), arXiv:1806.00451

\bibitem{rice2020overfitting}
Rice, L., Wong, E., Kolter, J.Z.: Overfitting in adversarially robust deep
  learning. In: ICML (2020)

\bibitem{RonEtAl2019}
Rony, J., Hafemann, L.G., Oliveira, L.S., Ayed, I.B., Sabourin, R., Granger,
  E.: Decoupling direction and norm for efficient gradient-based {L2}
  adversarial attacks and defenses. In: CVPR (2019)

\bibitem{russakovsky2015}
Russakovsky, O., Deng, J., Su, H., Krause, J., Satheesh, S., Ma, S., Huang, Z.,
  Karpathy, A., Khosla, A., Bernstein, M., Berg, A.C., Fei-Fei, L.: {ImageNet
  Large Scale Visual Recognition Challenge}. International Journal of Computer
  Vision (IJCV)  \textbf{115}(3),  211--252 (2015).
  \doi{10.1007/s11263-015-0816-y}

\bibitem{SamEtAl2018}
Samangouei, P., Saeedi, A., Nakagawa, L., Silberman, N.: Explaingan: Model
  explanation via decision boundary crossing transformations. In: ECCV (2018)

\bibitem{santurkar2019computer}
Santurkar, S., Tsipras, D., Tran, B., Ilyas, A., Engstrom, L., Madry, A.:
  Computer vision with a single (robust) classifier. In: NeurIPS (2019)

\bibitem{schmidt2018adversarially}
Schmidt, L., Santurkar, S., Tsipras, D., Talwar, K., Madry, A.: Adversarially
  robust generalization requires more data. In: NeurIPS (2018)

\bibitem{SchEtAl2018}
Schott, L., Rauber, J., Bethge, M., Brendel, W.: Towards the first
  adversarially robust neural network model on mnist. In: ICLR (2019)

\bibitem{sehwag2019better}
Sehwag, V., Bhagoji, A.N., Song, L., Sitawarin, C., Cullina, D., Chiang, M.,
  Mittal, P.: Better the devil you know: An analysis of evasion attacks using
  out-of-distribution adversarial examples. preprint, arXiv:1905.01726  (2019)

\bibitem{StuHeiSch2019}
Stutz, D., Hein, M., Schiele, B.: Disentangling adversarial robustness and
  generalization. In: CVPR (2019)

\bibitem{SzeEtAl2014}
Szegedy, C., Zaremba, W., Sutskever, I., Bruna, J., Erhan, D., Goodfellow, I.,
  Fergus, R.: Intriguing properties of neural networks. In: ICLR. pp.
  2503--2511 (2014)

\bibitem{80mtiny}
Torralba, A., Fergus, R., Freeman, W.T.: 80 million tiny images: A large data
  set for nonparametric object and scene recognition. IEEE transactions on
  pattern analysis and machine intelligence  \textbf{30}(11),  1958--1970
  (2008)

\bibitem{TraBon2019}
Tramèr, F., Boneh, D.: Adversarial training and robustness for multiple
  perturbations. In: NeurIPS (2019)

\bibitem{tsipras2018robustness}
Tsipras, D., Santurkar, S., Engstrom, L., Turner, A., Madry, A.: Robustness may
  be at odds with accuracy. In: ICLR (2019)

\bibitem{stanforth2019labels}
Uesato, J., Alayrac, J.B., Huang, P.S., Stanforth, R., Fawzi, A., Kohli, P.:
  Are labels required for improving adversarial robustness? In: NeurIPS (2019)

\bibitem{WacMitRus2018}
Wachter, S., Mittelstadt, B., Russell, C.: Counterfactual explanations without
  opening the black box: automated decisions and the {GDPR}. Harvard Journal of
  Law and Technology  \textbf{31}(2),  841--887 (2018)

\bibitem{wang2018high}
Wang, T.C., Liu, M.Y., Zhu, J.Y., Tao, A., Kautz, J., Catanzaro, B.:
  High-resolution image synthesis and semantic manipulation with conditional
  gans. In: CVPR (2018)

\bibitem{WonEtAl18}
Wong, E., Schmidt, F., Metzen, J.H., Kolter, J.Z.: Scaling provable adversarial
  defenses. In: NeurIPS (2018)

\bibitem{LSUN}
Yu, F., Seff, A., Zhang, Y., Song, S., Funkhouser, T., Xiao, J.: Lsun:
  Construction of a large-scale image dataset using deep learning with humans
  in the loop (2015), preprint, arXiv:1506.03365v3

\bibitem{ZeiFer2014}
Zeiler, M.D., Fergus, R.: Visualizing and understanding convolutional networks.
  In: ECCV (2014)

\bibitem{ZhaEtAl2019}
Zhang, H., Yu, Y., Jiao, J., Xing, E.P., Ghaoui, L.E., Jordan, M.I.:
  Theoretically principled trade-off between robustness and accuracy. In: ICML
  (2019)

\bibitem{zhu2016generative}
Zhu, J.Y., Kr{\"a}henb{\"u}hl, P., Shechtman, E., Efros, A.A.: Generative
  visual manipulation on the natural image manifold. In: Leibe, B., Matas, J.,
  Sebe, N., Welling, M. (eds.) ECCV (2016)

\end{thebibliography}
